\pdfoutput=1
\documentclass{article}

\usepackage{iclr2027_conference,times}
\iclrfinalcopy
\usepackage{amsmath,amssymb,amsthm,mathtools,mathrsfs}
\usepackage{booktabs,multirow,array,float,longtable}
\usepackage{microtype}
\usepackage{placeins}
\usepackage{graphicx}
\usepackage{xcolor}
\usepackage{url}
\usepackage{hyperref}
\usepackage[capitalize,noabbrev]{cleveref}
\usepackage{enumitem}
\usepackage{aliascnt}

\hypersetup{colorlinks=true,linkcolor=blue!55!black,citecolor=blue!55!black,urlcolor=blue!55!black,pdftitle={From Local Prediction Error to Final Outputs: A Theory of Expert Sharing in Multi-Step Reasoning},pdfauthor={Tiantong Wang, Zhiren Gong, Yixin Wang, Tiantong Wu, Wei Yang Bryan Lim}}

\newtheorem{theorem}{Theorem}
\newaliascnt{proposition}{theorem}
\newtheorem{proposition}[proposition]{Proposition}
\aliascntresetthe{proposition}
\newaliascnt{lemma}{theorem}
\newtheorem{lemma}[lemma]{Lemma}
\aliascntresetthe{lemma}
\newaliascnt{corollary}{theorem}
\newtheorem{corollary}[corollary]{Corollary}
\aliascntresetthe{corollary}
\newaliascnt{definition}{theorem}

\aliascntresetthe{definition}
\newaliascnt{assumption}{theorem}
\newtheorem{assumption}[assumption]{Assumption}
\aliascntresetthe{assumption}
\theoremstyle{remark}
\newaliascnt{remark}{theorem}

\aliascntresetthe{remark}

\crefname{theorem}{Theorem}{Theorems}
\Crefname{theorem}{Theorem}{Theorems}
\crefname{proposition}{Proposition}{Propositions}
\Crefname{proposition}{Proposition}{Propositions}
\crefname{lemma}{Lemma}{Lemmas}
\Crefname{lemma}{Lemma}{Lemmas}
\crefname{corollary}{Corollary}{Corollaries}
\Crefname{corollary}{Corollary}{Corollaries}
\crefname{definition}{Definition}{Definitions}
\Crefname{definition}{Definition}{Definitions}
\crefname{assumption}{Assumption}{Assumptions}
\Crefname{assumption}{Assumption}{Assumptions}
\crefname{remark}{Remark}{Remarks}
\Crefname{remark}{Remark}{Remarks}

\newcommand{\cC}{\mathcal{C}}

\newcommand{\cE}{\mathcal{E}}
\newcommand{\cF}{\mathcal{F}}

\newcommand{\cH}{\mathcal{H}}
\newcommand{\cJ}{\mathcal{J}}

\newcommand{\cS}{\mathcal{S}}
\newcommand{\cT}{\mathcal{T}}
\newcommand{\cX}{\mathcal{X}}
\newcommand{\cY}{\mathcal{Y}}
\newcommand{\cZ}{\mathcal{Z}}
\newcommand{\E}{\mathbb{E}}
\newcommand{\Prb}{\mathbb{P}}
\newcommand{\R}{\mathbb{R}}

\newcommand{\TV}{\operatorname{TV}}
\newcommand{\argmin}{\operatorname*{arg\,min}}

\newcommand{\tr}{\operatorname{tr}}
\newcommand{\norm}[1]{\left\lVert #1\right\rVert}
\newcommand{\abs}[1]{\left\lvert #1\right\rvert}

\newcommand{\Neff}{N^{\mathrm{eff}}}
\newcommand{\BReuse}{\mathsf B^{\mathrm{reuse}}}
\newcommand{\PRG}{\mathsf{PRG}}
\newcommand{\Risk}{\mathcal{R}}

\title{From Local Prediction Error to Final Outputs: A Theory of Expert Sharing in Multi-Step Reasoning}

\author{%
Tiantong Wang\textsuperscript{1},\enspace
Zhiren Gong\textsuperscript{1,2},\enspace
Yixin Wang\textsuperscript{3,1},\enspace
Tiantong Wu\textsuperscript{1},\enspace
Wei Yang Bryan Lim\textsuperscript{1}\\[2pt]
{\normalfont\textsuperscript{1}Nanyang Technological University\quad
\textsuperscript{2}KTH Royal Institute of Technology}\\
{\normalfont\textsuperscript{3}Beijing Institute of Technology}%
}
\date{}

\begin{document}
\maketitle
\lhead{Preprint}
\begin{abstract}
Mixture-of-experts (MoE) models can share experts across layers or computational steps, pooling training information from different contexts. This can reduce local prediction error, but multi-step systems are ultimately evaluated by their final outputs. We study when these local gains survive composition. In specified linear regression models, we derive exact sharing and merging criteria by balancing variance savings against bias from sharing across roles with different target behavior. Under a common fixed linear propagation map, we characterize expected squared final-output risk, retaining error directions and correlations. We construct cases where sharing lowers average local error yet increases final-output error. Within this setting, we also identify when orthogonal full pooling with isotropic noise preserves every local improvement at the final output. For general routed computations, we derive conditional end-to-end error bounds that account for distribution shifts, errors in surrounding components, and changes in expert selection. Separately, we show that correct predictions and route labels alone do not establish that individual experts implement the attributed operations. Controlled experiments realize a strict local-to-final risk reversal. In low-data symbolic settings, trained Transformers exhibit statistically supported teacher-forced one-step gains without statistically resolved free-running final-output gains. Separate intervention tests distinguish successful multi-layer actions from single-expert attribution and generalization to unseen transitions. Together, these results distinguish when sharing improves local estimation, when those gains survive to the final output, and what evidence is needed to attribute reusable operations to individual experts.

\end{abstract}

\section{Introduction}\label{sec:introduction}

A mixture-of-experts (MoE) Transformer uses a router to select among neural modules, called experts, when processing an input representation \citep{shazeer2017outrageously,lepikhin2021gshard,fedus2022switch,du2022glam}. Different layers can maintain separate experts, each with its own parameters, or use a common set of expert parameters across layers, which  is called \emph{expert sharing} \citep{tan2025rexmoe,huang2026unipool,chen2026moue,jaggi2026tying,qiu2026more}. Training information from different layers then shapes the same learned function. For multi-step reasoning, evaluating this choice requires understanding both the accuracy of individual computations and their contribution to the final output.

Choosing which computations should share an expert requires comparing the final-output errors of the resulting shared and separate models. A natural statistical rationale for sharing is that combining information from related contexts can reduce estimation variance, at the cost of bias when their required transformations differ. Multi-task learning provides a foundation for understanding this trade-off \citep{tripuraneni2020theory,tripuraneni2021meta,duan2023adaptive,huang2025molar}. However, even when this trade-off favors sharing in average local prediction error, the corresponding final-output comparison remains a separate question. Shared parameters can change the dependence between errors at different calls, while subsequent computations determine how those errors combine. A criterion based on local error therefore needs an additional connection to the accuracy of the complete computation. The comparison we seek must account jointly for how sharing changes estimation errors and how downstream computation transforms them, identifying when a local preference for shared experts remains justified at the final output. We therefore ask:
\begin{quotation}
\emph{When does expert sharing reduce average local prediction error, \\and under what conditions is this advantage preserved at the final output?}
\end{quotation}
Our analysis shows that the preference can reverse even when both models use the same fixed linear propagation map. Sharing can lower average local prediction error while increasing expected squared final-output error. The reason is that a local improvement can coexist with greater bias in directions amplified downstream, or with a change in how errors across calls reinforce or cancel. We derive exact risk comparisons that retain these effects and characterize the propagation conditions under which every strict local improvement is preserved in a specified sharing model.

Beyond its effect on accuracy, sharing raises a question about what an individual expert computes. Repeated selection of an expert may suggest that it implements an operation common to several contexts. Causal abstraction evaluates such interpretations through interventions on internal computation \citep{geiger2022inducing,geiger2024alignments}. We apply this perspective by changing expert choices while holding the pre-action context fixed and comparing the resulting effects with proposed operations.

\paragraph{Contributions}
\begin{enumerate}[leftmargin=*,itemsep=2pt,topsep=2pt]
    \item \textbf{Exact risk comparisons for expert sharing.}
    In specified linear regression models, we derive necessary and sufficient conditions for a common estimate to outperform separate estimates by balancing variance savings against bias from differences between targets. The same analysis gives exact criteria for merging expert groups and evaluates fixed combinations of separate and common estimates (\cref{thm:exact-frontier,thm:partition-frontier}), with extensions to cross-layer sharing and general estimation geometry.

    \item \textbf{When local improvements survive to the output.}
    Under a common fixed linear propagation map, exact expected squared-risk comparisons yield a strict local-to-final reversal (\cref{prop:local-terminal-reversal}). For nontrivial fixed groupwise averaging and centered estimation noise with equal positive variance in every direction, we characterize the propagation maps that preserve every strict local improvement (\cref{app:uniform-terminal-inheritance}). The condition requires equal positive scaling of squared lengths on the coefficient differences removed by sharing. For general routed computations, we derive conditional error bounds accounting for distribution shifts, changes in expert selection, and errors in surrounding components (\cref{sec:end-to-end}).

    \item \textbf{Separating predictive success from expert-level operations.}
    We construct a model with correct predictions and correct expert labels whose output remains unchanged when a different expert is forced (\cref{prop:route-label-separation}). We then analyze the state information and intervention coverage needed to evaluate operation-level effects, distinguishing individual expert calls from actions spanning several layers (\cref{sec:causal-reuse}).
\end{enumerate}

Additionally, we conduct experiments to reproduce the local risk comparisons and realize a local-to-final reversal. In low-data symbolic tasks, trained Transformers show one-step gains when given correct preceding states, while final-output gains remain statistically unresolved when the models use their own predictions (\cref{fig:exp-local-final-transfer}). Separate intervention and held-out-transition tests examine expert-level attribution and generalization beyond observed transitions. These results provide a basis for evaluating sharing patterns through both local estimation and final-output performance, while identifying the evidence needed to interpret individual experts as executable operations.
\section{Related Work}\label{sec:related-work}
We introduce prior work in expert sharing, statistical pooling, and causal abstraction discussing how local sharing gains change under composition, routing, and intervention-based evaluation.
\paragraph{Expert reuse and routed architectures.}
Classical MoE studies learned gating over specialized predictors \citep{jacobs1991adaptive,jordan1994hierarchical}; sparse Transformers scale conditional computation \citep{shazeer2017outrageously,lepikhin2021gshard,fedus2022switch,du2022glam}.  Recent systems share experts across adjacent layers, global pools, or depth groups \citep{tan2025rexmoe,huang2026unipool,chen2026moue,jaggi2026tying,qiu2026more}.  Expert-choice, soft dispatch, and sequential expert calls use different execution mechanisms \citep{zhou2022expertchoice,puigcerver2023softmoe,wang2025chainexperts}.  Our event-level comparison retains their functional costs; \cref{app:routed-roles,app:architecture-routing} give the architecture representation and cost dictionary.

\paragraph{Statistical sharing and shrinkage.}
Multi-task learning explains gains from shared structure and negative transfer from misspecified relatedness \citep{tripuraneni2020theory,tripuraneni2021meta,duan2023adaptive,huang2025molar}.  MoE theory studies generalization, robustness, expressivity, and task-specialized routing \citep{chowdhury2023patch,zhao2024generalization,puigcerver2022robustness,wang2025expressive,xiang2026taskrouting}.  Our local formulas and known-noise Gaussian plug-in guarantee specialize classical shrinkage \citep{baranchik1970family,stein1981estimation,srinath2018cluster}. The fixed-rule audit uses unbiased risk estimation, and additive transfer uses the standard Wasserstein coupling argument \citep{redko2017theoretical}.  

\paragraph{Reasoning and causal abstraction.}
Chain-of-thought theory studies serial computation through intermediate states \citep{merrill2024expressive,li2024serial}; latent-module work formalizes reusable components in reasoning traces \citep{kong2026reusablemodules}.  Causal abstraction distinguishes behavioral agreement from intervention agreement \citep{geiger2022inducing,geiger2024alignments}.  Applied to route actions, this requires sufficient pre-action states and distinguishable forced effects.  We also analyze randomized action audits via inverse-propensity estimation \citep{horvitz1952generalization}, with a matching lower-bound dependence on per-action coverage.

\section{Routed Roles and Event-Lifted Risk}
\label{sec:framework}
We distinguish call environments from expert blocks and measure their complete-event effects.  This makes tied and separate libraries comparable.
\paragraph{Roles, experts, and the reference.}
Consider one prediction request: the current state is 3 and the next operation is multiplication by two, so the target next state is 6.  In the illustrative two-layer Transformer of \hyperref[app:worked-routed-transformer]{Appendix~\ref*{app:routed-roles}}, both layers process this same request, rather than applying the task operation twice.  At the chosen prediction position, operation $i=2$ is labelled by call type $k=i+1=3$, giving two roles $(1,3)$ and $(2,3)$.

A role $a$ in a finite set $\cJ$ specifies a controlled call environment and its target behavior. Different input numbers can instantiate the same role.  The two layers define distinct roles here because their input sources and surrounding computation differ.  An expert library $\mathcal E_\theta=\{E_{\theta,e}:\mathcal X_e\to\mathcal Y_e\}_{e=1}^K$ contains parameterized functions indexed by $e$.  Separate fitting uses two independent parameter blocks for these roles, and reuse assigns the same block. The reused block may receive different inputs across roles. Its outputs and the roles' target updates may also differ.

In general, a fixed assignment $\alpha_m:\cJ\to[K_m]$ identifies the expert serving each role in comparator $m\in\{\mathrm{reuse},\mathrm{sep}\}$, with $[K_m]=\{1,\ldots,K_m\}$.  Reuse groups $\cJ_e=\{a:\alpha_{\mathrm{reuse}}(a)=e\}$ share functions through compatible interfaces or fixed adapters. The separate model fits a function per role.  In this role-assignment comparison, expert sharing is the common-function use represented by the reuse comparator.  The assignment specifies who should serve a role. The router decides who actually does, without redefining the role.

The \emph{reference} is the declared target computation, not the fitted separate model.  In the example, knowing the target next state is 6 does not determine a layer's target residual.  A local target $f_a^\star$ is specified separately to map hidden vectors to residuals, whereas the symbolic task operator $F_i$ maps task states to task states.  

\paragraph{Complete events and aligned instructions.}
An event is a complete computational block, not necessarily a token or reasoning step.  For $j=0,\ldots,H-1$, states $X_j,\widehat X_j^m\in\cS_j$ retain everything needed for execution, including prefixes, hidden-state arrays, and caches.  A canonical program $\Gamma_j\in\mathfrak G_j^\star$ specifies roles, call order, input selection, and combination.  A fixed implementation map $\iota_j^m:\mathfrak G_j^\star\to\mathfrak G_j^m$ translates it into model-specific instructions.  Conditional on fitted parameters and external randomness,
\begin{equation}
\begin{aligned}
 X_{j+1}&=\Phi_j^\star(X_j;\Gamma_j),\\
 \widehat X_{j+1}^m&=\Phi_j^m(\widehat X_j^m;\widehat\Gamma_j^m),
 &\widehat\Gamma_j^m&=\mathcal R_j^m(\mathcal H_j^m).
\end{aligned}
 \label{eq:rcg-main}
\end{equation}
The maps $\Phi_j^\star$ and $\Phi_j^m$ return complete post-event states in $\cS_{j+1}$, including shared paths and aggregation.  The router uses available history $\mathcal H_j^m$ to choose the \emph{actual} plan $\widehat\Gamma_j^m$, which may differ from the \emph{aligned} plan $\iota_j^m(\Gamma_j)$.    Decisions use only information available when made.  Adaptive internal decisions must be separate events for fixed-sequence replacement, or the whole policy must be one event lift. \cref{app:event-interface} gives the complete history and randomness interface.

\paragraph{Role risk in its call environment.}
Role $a$ has probe-record space $\cZ_a$, distribution $P_a$, and an event lift
\begin{equation}
 \cT_a[f]:\cZ_a\longrightarrow\cS,
 \label{eq:event-lift}
\end{equation}
which inserts expert function $f$ into a specified environment and returns the complete post-event state.  For example, $\cT_a[f](h,u,c)=h+cf(u)$ is a residual update with state $h$, expert input $u$, and scalar gain $c$. Dimensions agree after any fixed adapter.  Let $T_a^\star$ be the target-call counterpart.  For a state metric $d$,
\begin{equation}
 \ell_a(f;z)=d(\cT_a[f](z),T_a^\star(z)),\qquad
 R_a(f)=\E_{Z\sim P_a}\ell_a(f;Z).
 \label{eq:role-risk}
\end{equation}
Both arguments are complete states. The worked example inserts one residual into the full array.  The bounded-loss theorem additionally requires $\ell_a\in[0,1]$ and probe laws, lifts, targets, classes, and permitted expert inputs fixed independently of fitting data, or a valid joint uniform bound.    

\section{The Local Sharing Frontier}
\label{sec:local-theory}
This section mainly discuss that, with call environments fixed, when should roles share a function. The linear model prices pooling and merges by removable variance versus sharing bias. Beyond that, we compare certificates rather than exact risks.

\subsection{An exact linear-operator frontier}
In a specified linear estimation problem, consider $M\ge1$ equally weighted roles with target operators $A_a\in\R^{p\times d}$ and responses $Y_a=X_aA_a^\top+\mathcal E_a$.  Each fixed design $X_a\in\R^{n\times d}$ stacks the training inputs and satisfies $n\ge d$, $X_a^\top X_a=nI_d$, which are not the event states $X_j$.  Designs are held constant as fitting noise varies. Orthogonal columns are a model assumption.  All noise rows are mutually independent across observations and roles, centered, and have covariance $\sigma^2I_p$.  Gaussianity is not required.  Define
\begin{equation}
 \bar A=\frac1M\sum_{a=1}^{M}A_a,
 \qquad B=\frac1M\sum_{a=1}^{M}\norm{A_a-\bar A}_F^2.
 \label{eq:linear-B}
\end{equation}
Thus $B$ measures dispersion among the true role targets.  Ordinary least squares (OLS) gives $\widehat A_a=Y_a^\top X_a/n$ and $\widehat A_{\mathrm{pool}}=M^{-1}\sum_a\widehat A_a$.  For a fixed coefficient $\lambda\in[0,1]$, partial pooling uses
\begin{equation}
 \widehat A_a^{(\lambda)}=(1-\lambda)\widehat A_a+\lambda\widehat A_{\mathrm{pool}}.
 \label{eq:partial-reuse-estimator}
\end{equation}
Separate fitting and full sharing with $\lambda=0$and $1$, and intermediate matrices generally remain role-specific. An independent input has second moment $\E[xx^\top]=I_d$.  The excess prediction risk is
\begin{equation}
 \Risk_\lambda=\frac1M\sum_{a=1}^{M}\E_{\mathcal E,x}
       \norm{(\widehat A_a^{(\lambda)}-A_a)x}_2^2.
 \label{eq:exact-risk-definition}
\end{equation}
It averages fitting noise and test inputs relative to noiseless targets, excluding fresh label noise.  Only under this isotropic test geometry does conditional prediction error equal squared Frobenius parameter error.  This specified squared loss is distinct from the bounded metric loss in \cref{eq:role-risk}.

\begin{theorem}[Exact sharing and partial-pooling risks]
\label{thm:exact-frontier}
Let $D=pd$, $V=D\sigma^2/n$, and $C=V(1-1/M)$.  For the excess risk in \cref{eq:exact-risk-definition},
\begin{align}
 \Risk_{\mathrm{sep}}&=V,\label{eq:exact-sep-risk}\\
 \Risk_{\mathrm{share}}&=B+\frac{V}{M},\label{eq:exact-share-risk}\\
 \Risk_{\lambda}&=V-2C\lambda+(B+C)\lambda^2.\label{eq:exact-partial-risk}
\end{align}
Consequently, full reuse has smaller expected excess prediction risk than separate estimation if and only if
\begin{equation}
 B<C=\frac{pd\sigma^2}{n}\left(1-\frac1M\right).
 \label{eq:exact-threshold}
\end{equation}
If $B+C>0$, the optimal fixed coefficient is $\lambda^\star=C/(B+C)$.
The minimum risk, endpoint comparisons, and degenerate cases are given in
\cref{app:exact-frontier-details}.
\end{theorem}
Here $D$ counts coefficients, $V$ is separate estimation error, and $C=V-V/M$ is the removable variance.  Full pooling pays bias $B$ while averaging independent fitting errors to variance $V/M$.  The oracle coefficient depends on unknown $B$.  Substituting a same-data estimate of $\lambda$ into the fixed-rule identity is invalid. The classical guarantee in \cref{app:classical-plugin} separately requires Gaussian noise, known $q=\sigma^2/n>0$, and $D(M-1)\ge4$.

For allocation, fix a partition $\Pi$ into $K_\Pi$ disjoint nonempty groups before observing fitting noise.  Group $G$ has target mean $\bar A_G=|G|^{-1}\sum_{a\in G}A_a$ and uses the average of its role-wise OLS estimates.
\begin{theorem}[Exact partition and merge frontier]
\label{thm:partition-frontier}
Under the assumptions of \cref{thm:exact-frontier}, the expected role-averaged excess prediction risk of partition $\Pi$ is
\begin{equation}
 \Risk(\Pi)=\frac{1}{M}\left[
 \sum_{G\in\Pi}\sum_{a\in G}\norm{A_a-\bar A_G}_F^2
 {}+K_\Pi V\right].
 \label{eq:partition-risk}
\end{equation}
If two groups $G$ and $H$ are merged, the exact risk change is
\begin{equation}
 \Risk(\Pi_{G\cup H})-\Risk(\Pi)
 =\frac1M\left[
 \frac{|G||H|}{|G|+|H|}\norm{\bar A_G-\bar A_H}_F^2-V
 \right].
 \label{eq:exact-merge-frontier}
\end{equation}
Thus a merge improves expected risk if and only if its between-centroid functional separation is below the variance $V$ saved by removing one independently estimated expert.  Optimizing the number and composition of experts is therefore exactly a function-space clustering objective with penalty $V$ per fitted expert.
\end{theorem}
The variance charge saved by a merge is $V$ before division by $M$, hence $V/M$ in average risk.    Cross-depth patterns and general test geometry are treated in Appendix~\ref{app:linear-sharing-architectures} and \cref{thm:weighted-projection-frontier} in Appendix~\ref{app:assumption-robust}.  

\subsection{A distribution-free comparison}
We now retain the sharing-versus-estimation comparison as a high-probability bound, under explicit sampling and loss conditions.  For fixed candidate class $\cF_e$ and positive evaluation weights $\pi_{e,a}$ summing to one on $\cJ_e$, define
\begin{align}
 R_{\mathrm{sep},e}^*&=\sum_{a\in\cJ_e}\pi_{e,a}\inf_{f\in\cF_e}R_a(f),\\
 R_{\mathrm{share},e}^*&=\inf_{f\in\cF_e}\sum_{a\in\cJ_e}\pi_{e,a}R_a(f),\\
 \BReuse_e&=R_{\mathrm{share},e}^*-R_{\mathrm{sep},e}^*\ge0.
 \label{eq:general-reuse-bias}
\end{align}
The first oracle chooses one function per role, and the second must use a common function, incurring reuse bias.  With $n_a$ mutually independent observations per role, weighted pooling has effective sample size
\begin{equation}
 \Neff_e=\left(\sum_{a\in\cJ_e}\frac{\pi_{e,a}^2}{n_a}\right)^{-1},
 \label{eq:performance-neff}
\end{equation}
which is $Mn$ for balanced roles. With nonempty finite classes, losses in $[0,1]$, fixed objects independent of fitting data, and mutually independent role samples, \cref{thm:finite-class} (Appendix~\ref{app:finite-class-full}) gives shared/separate ERM bounds $R_{\mathrm{sep},e}^*+\BReuse_e+2\epsilon_{\mathrm{sh},e}$ and $R_{\mathrm{sep},e}^*+2\epsilon_{\mathrm{sep},e}$ on one $1-\delta$ fitting-data event. The shared empirical-risk and weighted role-wise deviations use
\( \epsilon_{\mathrm{sh},e}(\delta)=\sqrt{\frac{\log(4|\cF_e|/\delta)}{2\Neff_e}}\) and \(\epsilon_{\mathrm{sep},e}(\delta)=\sum_{a\in\cJ_e}\pi_{e,a}
 \sqrt{\frac{\log(4|\cJ_e||\cF_e|/\delta)}{2n_a}}\), respectively (\cref{eq:finite-radius-shared,eq:finite-radii}), with $|\cF_e|$ counting functions.  Variance interpretations require \cref{prop:hilbert-bias}'s class conditions (Appendix~\ref{app:statistical-proofs}). Approximate ERM and mismatched training weights retain \cref{prop:approximate-weighted-erm}'s terms (Appendix~\ref{app:assumption-robust}). Infinite classes need uniform bounds, and \cref{eq:app-trace-concentration} treats independent traces.

\section{From Local Prediction Error to Final Outputs}
\label{sec:end-to-end}
In this section, we discuss the main argument that lower local prediction error does not necessarily indicate a better final output.  Common fixed linear propagation gives an exact terminal frontier and a reversal. The local squared-risk model in \cref{sec:local-theory} isolates estimation, and a common fixed propagation map then exposes the effect of composition. The metric-error analysis below tracks the conditions needed for a general routed program. The trained-model study in \cref{sec:experiments} measures teacher-forced one-step and free-running final-state errors.

\subsection{Exact terminal geometry and a reversal}
The terminal map determines which local error directions and correlations matter.  Let $\theta$ stack the true role coefficients and $Y=\theta+\zeta$ their estimates, with $\E\zeta=0$ and finite covariance $\Sigma$.  For \emph{every compared rule}, a common matrix $G$, fixed independently of $\zeta$, maps coefficient error to terminal error.  Fix a pooling matrix $P$ independently of $\zeta$, and set $Q=I-P$ and $L_\lambda=I-\lambda Q$, $0\le\lambda\le1$.  Thus $L_0Y=Y$ is separate fitting and $L_1Y=PY$ is full pooling.  For $R_H(\lambda)=\E\|G(L_\lambda Y-\theta)\|_2^2$, \cref{thm:terminal-geometry-frontier} in Appendix~\ref{app:terminal-selection} gives
\begin{equation}
\begin{aligned}
 R_H(\lambda)&=R_H(0)-2a_H\lambda+b_H\lambda^2,\\
 a_H&=\tr(G^\top GQ\Sigma),\\
 b_H&=\|GQ\theta\|_2^2+\tr(G^\top GQ\Sigma Q^\top)\ge0.
\end{aligned}
 \label{eq:main-terminal-frontier}
\end{equation}
The geometry $G^\top G$ weights coefficient errors by their endpoint effects.  The coefficient $b_H$ contains squared bias and noise. $a_H=\E\langle G\zeta,GQ\zeta\rangle$ measures alignment between original noise and the removed component.  It can be negative when pooling destroys beneficial cancellation.  Full pooling therefore improves \emph{actual expected terminal risk} exactly when $2a_H>b_H$.

\paragraph{A local gain can reverse.}
In \cref{prop:local-terminal-reversal} (Appendix~\ref{app:terminal-selection}), three scalar roles have independent centered fitting errors of variance one, target $\theta=(1/2,-1,1/2)^\top$, and propagation $G=(1,-2,1)$.  Uniform full pooling reduces local average risk from $1$ to $5/6$, but raises terminal risk from $6$ to $9$.  The pooled component is annihilated by $G\mathbf1=0$, leaving squared terminal bias $(G\theta)^2=9$.    Partial pooling can still help: the same example has $R_H(\lambda)=6-12\lambda+15\lambda^2$.

Here $G$ comes from the three events in \cref{eq:reversal-reference-events}, starting at $S_0=0$:
\[
 S_1=S_0+\theta_1,\qquad
 S_2=-\tfrac12S_1+\theta_2,\qquad
 S_3=-2S_2+\theta_3.
\]
Each fitted model uses this same recurrence with its fitted coefficients. Expanding it gives the weights $(1,-2,1)$, so the common propagation is fixed by the computation. Equal pooled coefficients cancel through $G\mathbf1=0$, while the target satisfies $G\theta=3$.

Repeated calls can reuse correlated fitted errors despite independent training data (\cref{prop:gramian-amplification,prop:shared-error-coherence}; Appendices~\ref{app:end-to-end-proofs},~\ref{app:assumption-robust}). Appendix~\ref{app:uniform-terminal-inheritance}'s uniform-inheritance result requires orthogonal projection and isotropic noise. A full-law $L^2$ remainder bound yields a same-law true-risk interval (\cref{prop:terminal-remainder-interval}; Appendix~\ref{app:terminal-selection}.3). Joint Taylor and conditional-moment results retain region, moment, and escape assumptions (Appendix~\ref{app:parameter-endogenous}). Support replay recomputes gains, and recorded-gain replay pays an extra discrepancy (Appendix~\ref{app:rank-radius-replay}). 

The uniform-inheritance statement concerns a restricted case: a fixed Euclidean orthogonal pooling projection $P$, $Q=I-P$ with positive rank, centered noise with covariance $qI$, $q>0$, and the same fixed $G$ for both estimates. In that case, \cref{app:uniform-terminal-inheritance} shows that strict local improvement from full pooling implies strict terminal improvement for \emph{every target} $\theta$ exactly when $QG^\top GQ=cQ$ for some $c>0$. Thus $G$ scales squared lengths by the same positive factor on the discarded contrast space. This all-target condition is not necessary for one particular target to benefit.

\subsection{Conditional certificates for general routed programs}
We now bound aligned implementation and routing errors without an exact common linear map.  Use $S_j=X_j$, $\widehat S_j^m=\widehat X_j^m$ from \cref{eq:rcg-main}, now represented in a common metric space $(\cS,d)$.  For $H\ge1$ events write the reference dynamics as
\begin{equation}
 S_{j+1}=T_{j,\Gamma_j}(S_j),
 \label{eq:reference-random-plan}
\end{equation}
where $T_{j,\gamma}=\Phi_j^\star(\cdot;\gamma)$ and $\widehat T^m_{j,\gamma}=\Phi_j^m(\cdot;\gamma)$ use their respective program spaces.  Each reference fixed-program map has a deterministic $L_j$-Lipschitz envelope throughout the compared region, uniformly over admissible programs. This does not assume a hard router is Lipschitz.  Let $A_j=\prod_{r=j+1}^{H-1}L_r$, with empty product one.  At the \emph{same current model state}, define
\begin{align}
 a_j^m&=d\!\left(\widehat T^m_{j,\iota_j^m(\Gamma_j)}(\widehat S_j^m),
                  T_{j,\Gamma_j}(\widehat S_j^m)\right),\\
 r_j^m&=d\!\left(\widehat T^m_{j,\widehat\Gamma_j^m}(\widehat S_j^m),
                  \widehat T^m_{j,\iota_j^m(\Gamma_j)}(\widehat S_j^m)\right).
 \label{eq:plan-local-split}
\end{align}
The first compares fitted and target implementations under aligned instructions. The second changes only the plan at that same state, avoiding a second charge for earlier state error.  With $e_j^m=d(\widehat S_j^m,S_j)$, the triangle inequality gives
\[
 e_{j+1}^m\le L_je_j^m+a_j^m+r_j^m.
\]
 Evaluation expectations condition on fitted models and average fresh trajectories and routing randomness, writing as $\rho_j^m=\E_{\mathrm{eval}}r_j^m$.  The random intended program remains inside the discrepancies.  Role-probe risk controls rollout error only through a further bridge.

\begin{assumption}[Common role-aggregation domination]
\label{ass:role-aggregation}
The role distributions, event-lift losses, target hybrids, and expert classes are fixed independently of the expert-estimation sample.  There are deterministic nonnegative weights $\omega_{j,a}$, also fixed independently of that sample and valid for both comparators, such that, conditional on the fitted models,
\begin{equation}
 \E_{\mathrm{eval}} a_j^m
 \le \tau_j^m+
     \sum_{a\in\cJ}\omega_{j,a}
     R_a\!\left(\widehat f^m_{\alpha_m(a)}\right),
 \qquad \tau_j^m\ge0.
 \label{eq:role-aggregation}
\end{equation}
\end{assumption}
The weights are fixed domination envelopes, not realized gates. Dynamic routing still requires justified occurrence, sensitivity, transfer, and assembly control. Appendix~\ref{app:routed-roles}.4 gives occupancy/transfer sufficient conditions. \cref{prop:additive-occupancy-transfer} (Appendix~\ref{app:assumption-robust}.4) charges additive support mismatch. 

The role risks are measured on fixed probes. In the sufficient construction of Appendix~\ref{app:routed-roles}.4, each $\kappa_{j,q,a}$ for controlled occurrence $q$ of role $a$ bounds both occurrence mass and distribution shift. Summing these bounds yields the common weights $\omega_{j,a}$, rather than empirical call frequencies. The term $\tau_j^m$ retains surrounding-assembly mismatch after target experts replace the controlled calls. In contrast, $\rho_j^m$ averages the effect of changing only the plan at the same current fitted-model state. Within-event effects are accounted for by the declared lifts and weights. $A_j$ propagates the resulting event error through later events. 

For each role set $w_a=\sum_jA_j\omega_{j,a}$, within each reuse group retain $w_a>0$, put $W_e=\sum_{a\in\cJ_e}w_a$, and set $\pi_{e,a}=w_a/W_e$.  Only $W_e>0$ groups enter below.  All oracles, reuse biases, radii, and empirical objectives are recomputed on this active support.  The full convention, including zero-weight roles influencing training, is in \cref{app:terminal-certificate-full}.

\begin{theorem}[Plan-level terminal certificates]
\label{thm:terminal-certificate}
Let initial states agree and let scalar terminal utility $\mathcal U$ be $L_{\mathcal U}$-Lipschitz.  Assume \cref{ass:role-aggregation} and all conditions of \cref{thm:finite-class} on every active group: fixed finite classes, losses in $[0,1]$, fixed probe laws and performance weights independent of fitting data, at least one observation per active role, and mutually independent role observations.  Fit matched performance-weighted ERMs from the same role samples.  For $\delta\in(0,1)$ choose $\delta_e>0$, $\sum_e\delta_e\le\delta$, and use the radii of \cref{eq:finite-radius-shared,eq:finite-radii} at $\delta_e$, with $\epsilon_{\mathrm{reuse},e}=\epsilon_{\mathrm{sh},e}$.  With probability at least $1-\delta$ over fitting data, simultaneously for both models,
\begin{equation}
\begin{aligned}
 \E_{\mathrm{eval}}|\mathcal U(\widehat S_H^m)-\mathcal U(S_H)|&\le L_{\mathcal U}U_m,\\
 U_m&=\sum_j A_j(\tau_j^m+\rho_j^m)\\
 &\quad+\sum_e W_e\bigl[R_{\mathrm{sep},e}^*
       +\mathbf1\{m=\mathrm{reuse}\}\BReuse_e+2\epsilon_{m,e}(\delta_e)\bigr].
\end{aligned}
\label{eq:terminal-certificate-compact}
\end{equation}
\end{theorem}
$U_m$ is a state-error bound, not the utility $\mathcal U$.  The expanded bounds and probability accounting are in \cref{app:terminal-certificate-full}.  Common oracles cancel in the state-certificate gap $\PRG_H=U_{\mathrm{sep}}-U_{\mathrm{reuse}}$:
\begin{equation}
\begin{split}
 \PRG_H
 &=\sum_eW_e\Bigl[
      2\{\epsilon_{\mathrm{sep},e}(\delta_e)
          -\epsilon_{\mathrm{sh},e}(\delta_e)\}
      -\BReuse_e\Bigr]\\
 &\quad-\sum_jA_j\Bigl[(\tau_j^{\mathrm{reuse}}-\tau_j^{\mathrm{sep}})
    +(\rho_j^{\mathrm{reuse}}-\rho_j^{\mathrm{sep}})\Bigr].
\end{split}
\label{eq:prg-corrected}
\end{equation}
Thus $\PRG_H>0$ orders these state-error bounds. Strict utility-bound ordering also requires $L_{\mathcal U}>0$.  It does not order unknown true risks or directly bound discrete answer accuracy.  Model-specific weights prevent this compact cancellation.  Approximate ERM, mismatched training weights, and zero-weight training influences retain their explicit algorithmic-deviation terms (\cref{prop:approximate-weighted-erm}).  Architecture-specific route costs and margin feedback are in \cref{app:architecture-routing,app:endogenous-routing}. Independent actual-performance evaluation is in \cref{prop:paired-real-improvement} (Appendix~\ref{app:assumption-robust}.7).

\section{Why Route Labels Do Not Establish Causal Reuse}
\label{sec:causal-reuse}
The preceding comparisons concern predictive risk and error bounds. In this section, we discuss whether behavior can be attributed to experts implementing executable operations.
Correct expert labels do not always establish operations. We separate labels from intervention fidelity, sufficient state information, and audit coverage.

Let $C$ be the complete executable context and $S=\Psi(C)$ a macro state extracted by a fixed map.  A route action $\mathcal A_i$ specifies a non-anticipating policy and bounded stopping time. Execution produces $\mathfrak T_i(C)$ and macro result $T_i(C)=\Psi(\mathfrak T_i(C))$.  Deterministic statements use greedy execution or a common fixed realization of exogenous randomness. The full policy and stochastic formulations are in \cref{app:causal-proofs,app:extensions-ledger}.  For a candidate operator $F_i:\cS\to\cS$ and fixed pre-action context law $\nu_i$, define
\begin{equation}
 \mu_i=\E_{C\sim\nu_i}d\!\left(T_i(C),F_i(\Psi(C))\right).
 \label{eq:causal-defect-main}
\end{equation}
This is prediction error for an \emph{intervened effect} using the pre-action state.  Small defect also needs non-collapse between actions to support a reusable-operation claim.  A single expert call and a multi-layer compound action are different attribution targets.

\begin{proposition}[Observational alignment does not imply causal reuse]
\label{prop:route-label-separation}
For any finite family of target operators $\{F_i\}_{i=1}^{K}$ with outputs in a common real vector space, there exists a residual routed model with exact task output and route-label accuracy one, while its output under forced expert index $j$ is independent of $j$.  Under a distribution over requested index $I$, state $S$, and forced index $J$, its forced-route fidelity to $F_J$ is exactly
\begin{equation}
 \Prb\!\left(F_I(S)=F_J(S)\right),
 \label{eq:forced-fidelity-construction}
\end{equation}
which equals $1/K$ when $J$ is uniform and independent of $(I,S)$ and the $K$ operator outputs are pairwise distinct at every evaluated state.
\end{proposition}
The shared path computes $F_I(S)$ and every expert contributes zero, while the natural router selects $I$ perfectly.  Forcing $J$ leaves the requested operation $I$ and pre-action prefix unchanged, so the output remains $F_I(S)$.  The $1/K$ specialization requires both independent uniform forcing and pairwise distinct target outputs at every evaluated state.

A compressed state may also omit information needed to predict an action.  \Cref{prop:aliasing} in Appendix~\ref{app:causal-proofs} gives a worst-case diameter obstruction, while \cref{prop:average-aliasing} gives an average squared-loss conditional-variance identity.  The worst pair and probability-weighted average treat rare aliases differently.  The extractor, actions, operators, and alignment must be fixed independently of the audit, and off-policy fidelity needs further control to transfer to on-policy contexts.

Randomized action-wise audits require supported actions and known forcing propensities(\cref{app:causal-audit}).  Their population comparison concerns action-specific $f_A(S)$ versus action-unaware $f(S)$. It equals a local sharing comparison only with matched inputs, classes, losses, and distributions (\cref{app:causal-role-alignment}).  Physical tying alone does not remove action information encoded in the expert input.  These causal checks remain separate from the paired terminal-performance audit.

\section{Experiments: From Local Gains to Final Outputs}
\label{sec:experiments}
We test whether reducing local error improves the final output, first in controlled estimation, and then in trained modern Transformers architecture. The latter compose four operations on $\mathbb Z_{31}$ using four-layer, width-256 causal models trained at horizons 4 and 8. We distinguish teacher-forced one-step error from free-running final-state error.  Methods, full results, and evaluation scope are in \cref{app:experiment-details}.

\begin{figure}[htbp]
\centering
\includegraphics[width=0.495\linewidth]{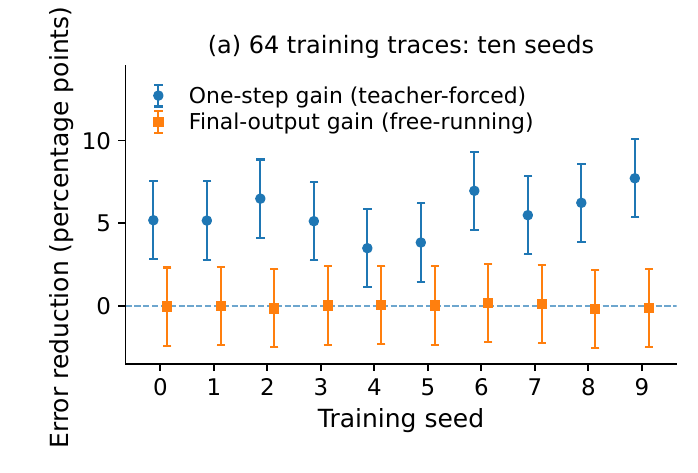}\hfill
\includegraphics[width=0.495\linewidth]{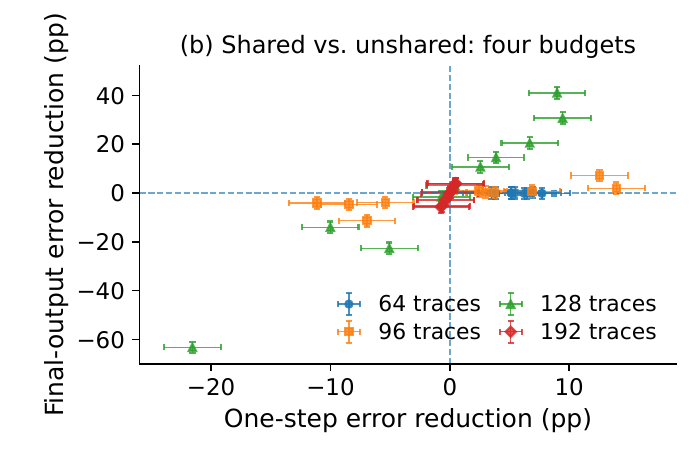}
\caption{\textbf{Local and final gains in trained Transformers.} Positive differences mean lower error for tied experts. (a) At 64 traces, one-step gains are supported, while the final-output difference is not statistically resolved. (b) All forty tying contrasts at 64, 96, 128, and 192 traces. Bars retain the reported simultaneous intervals of the full 240-endpoint follow-up family ($N=32768$, $480$ tails, $\delta=0.05$,  radius $= 2.3657$)}
\label{fig:exp-local-final-transfer}
\end{figure}

\paragraph{One-step gains need not yield detectable final-output gains.}
A training-data sweep compares the same trained model pairs on both endpoints. \Cref{fig:exp-local-final-transfer} shows cross-depth tying versus layer-local experts for all ten seeds at all four budgets of the balanced follow-up. At 64 training traces, tying reduces teacher-forced step error by 3.49--7.72 percentage points, with a positive lower bound in every seed, so the local gains are non-trivial. However, final-output differences range only from $-0.186$ to $+0.192$ points and every interval crosses zero, which shows that no statistically meaningful output gain is observed. Both models remain near chance on the final endpoint in this low-data regime. This is a concrete example, align with our maths abstract construction, which shows that there exists the case when \emph{local improvement without a statistically meaningful final improvement}. Other budgets exhibit larger final changes and seed-dependent directions. 

\section{Conclusion}\label{sec:conclusion}
Expert sharing must be assessed at three levels: local estimation, final-output performance, and intervention effects.  Locally it trades estimation savings against role incompatibility.  Common fixed linear propagation makes the terminal trade-off exact and permits local gains to reverse.  General routed programs require explicit transfer, assembly, and routing control.  Causal reuse additionally requires supported interventions and a sufficient pre-action state.  The experiments make the local-to-final distinction concrete: a controlled program reverses the risk ordering, while trained models exhibit supported one-step gains without resolved final-output gains.

\clearpage
\subsection*{AI Use Statement}
In this work, generative AI tools were used for language editing, code development, and the drafting of candidate mathematical derivations. The authors independently verified the final mathematical content, including assumptions, derivations, propositions, and proofs. The final proofs, citations, code, and experiment provenance were also manually checked or rewritten by the authors. The authors take full responsibility for the final content of this work, including the text, claims, mathematical results, code, and all other materials produced with the assistance of generative AI.
\subsection*{Ethics Statement}
The reported experiments use generated symbolic data and locally trained Transformer models, rather than personal data or a deployed decision system. More efficient conditional computation can reduce resource use, but it can also lower the cost of deploying capable models. Future applications should report both efficiency and capability changes.

\subsection*{Reproducibility Statement}
The assumptions and complete proofs are given in the corresponding theoretical appendices, with a consolidated reading guide in the appendix overview. \Cref{app:experiment-details} specifies the tasks, model architectures, optimization settings, frozen-feature estimators, interventions, and evaluation procedures. Figure captions and \cref{app:new-execution,app:exp-budget} state the sampling units, comparison families, and uncertainty conventions. The supplemented materials preserves the experimental configurations and summaries together with scripts for regenerating the tables and figures.

\clearpage
\bibliography{references}
\bibliographystyle{iclr2027_conference}

\clearpage
\appendix
\section*{Appendix Overview and Reading Paths}
\phantomsection\label{app:overview}
\addcontentsline{toc}{section}{Appendix overview and reading paths}
The appendices separate proofs of the main claims from architectural specializations and evaluation protocols. Readers can use the questions below to choose a path.

\paragraph{Understand the problem settings and argument.}
Start with the \hyperref[app:worked-routed-transformer]{worked Transformer example}, then return to Sections~\ref{sec:local-theory}--\ref{sec:causal-reuse}.  Consult \cref{app:notation} to distinguish exact risk, a certificate, and an audit.
\paragraph{Check a main claim.}
Read \cref{app:exact-frontier-proof} for local identities, \cref{app:statistical-proofs,app:end-to-end-proofs} for statistical and terminal bounds, \cref{app:terminal-selection} for terminal geometry, and \cref{app:causal-proofs} for causal obstructions.  Use \cref{app:operational-assumptions} alongside the full statements.
\paragraph{Choose an extension or evaluation.}
Use H--P selectively for the relevant architecture, geometry, or intervention.  \Cref{app:experiment-details} reports local-to-final experiments, intervention supplements, and evaluation scope. \hyperref[prop:paired-real-improvement]{The paired audit in Appendix~\ref*{app:assumption-robust}.7} addresses actual performance.

\begingroup
\small
\begin{longtable}{@{}p{0.055\textwidth}p{0.83\textwidth}r@{}}
\toprule
 & Question and entry point & Page\\
\midrule
\endfirsthead
\toprule
 & Question and entry point & Page\\
\midrule
\endhead
\hyperref[app:notation]{\ref*{app:notation}} & \hyperref[app:notation]{Symbols and claim types}: what does this term or guarantee mean? & \pageref{app:notation}\\
\hyperref[app:routed-roles]{\ref*{app:routed-roles}} & \hyperref[app:routed-roles]{Roles and event lifts}: how do the definitions fit a concrete model? & \pageref{app:routed-roles}\\
\hyperref[app:statistical-proofs]{\ref*{app:statistical-proofs}} & \hyperref[app:statistical-proofs]{Distribution-free statistics}: what can independent samples guarantee? & \pageref{app:statistical-proofs}\\
\hyperref[app:exact-frontier-proof]{\ref*{app:exact-frontier-proof}} & \hyperref[app:exact-frontier-proof]{Exact local frontiers}: how are pooling risks and selection justified? & \pageref{app:exact-frontier-proof}\\
\hyperref[app:end-to-end-proofs]{\ref*{app:end-to-end-proofs}} & \hyperref[app:end-to-end-proofs]{Terminal certificates}: how do local and routing errors propagate? & \pageref{app:end-to-end-proofs}\\
\hyperref[app:causal-proofs]{\ref*{app:causal-proofs}} & \hyperref[app:causal-proofs]{Causal actions}: why are labels and compressed states insufficient? & \pageref{app:causal-proofs}\\
\hyperref[app:experiment-details]{\ref*{app:experiment-details}} & \hyperref[app:experiment-details]{Local-to-final experiments}: which gains transfer, and what do intervention tests add? & \pageref{app:experiment-details}\\
\hyperref[app:linear-sharing-architectures]{\ref*{app:linear-sharing-architectures}} & \hyperref[app:linear-sharing-architectures]{Unequal data and cross-depth tying}: which effects can be shared? & \pageref{app:linear-sharing-architectures}\\
\hyperref[app:architecture-routing]{\ref*{app:architecture-routing}} & \hyperref[app:architecture-routing]{Routing architectures}: what functional cost does a route change incur? & \pageref{app:architecture-routing}\\
\hyperref[app:endogenous-routing]{\ref*{app:endogenous-routing}} & \hyperref[app:endogenous-routing]{Routing feedback}: when can an earlier error change later support? & \pageref{app:endogenous-routing}\\
\hyperref[app:causal-audit]{\ref*{app:causal-audit}} & \hyperref[app:causal-audit]{Intervention audits}: which action effects have enough coverage? & \pageref{app:causal-audit}\\
\hyperref[app:extensions-ledger]{\ref*{app:extensions-ledger}} & \hyperref[app:extensions-ledger]{Complementary extensions}: what do function corrections and protocols add? & \pageref{app:extensions-ledger}\\
\hyperref[app:assumption-robust]{\ref*{app:assumption-robust}} & \hyperref[app:assumption-robust]{Relaxed assumptions}: what survives design, transfer, or optimization mismatch? & \pageref{app:assumption-robust}\\
\hyperref[app:terminal-selection]{\ref*{app:terminal-selection}} & \hyperref[app:terminal-selection]{Terminal geometry}: when does a local gain reverse or survive uniformly? & \pageref{app:terminal-selection}\\
\hyperref[app:parameter-endogenous]{\ref*{app:parameter-endogenous}} & \hyperref[app:parameter-endogenous]{Nonlinearity and parameters}: how are remainders and selected moments controlled? & \pageref{app:parameter-endogenous}\\
\hyperref[app:rank-radius-replay]{\ref*{app:rank-radius-replay}} & \hyperref[app:rank-radius-replay]{Update budgets and replay}: what do rank, radius, support, and gains constrain? & \pageref{app:rank-radius-replay}\\
\hyperref[app:operational-assumptions]{\ref*{app:operational-assumptions}} & \hyperref[app:operational-assumptions]{Assumption audit}: which assumptions support each result? & \pageref{app:operational-assumptions}\\
\bottomrule
\end{longtable}
\endgroup

\clearpage
\section{Reading Guide: Symbols, Mathematical Terms, and Claim Types}
\label{app:notation}
This guide distinguishes risk comparisons, certificates, and audits, and records the paper's notation. Definitions apply throughout; specialized symbols are also introduced next to their equations, and a reused letter has the scope specified in its section.

\subsection{Three types of guarantees}
A \emph{risk} is an expected loss on a specified data distribution.  A \emph{population oracle} is the best function in a declared class when that distribution is known; an \emph{oracle tuning coefficient} minimizes an expected risk using unknown population quantities.    An \emph{empirical risk} replaces the expectation by a sample average, and an empirical risk minimizer (ERM) minimizes that average.  An \emph{excess prediction risk} here measures error relative to the noiseless target operator, excluding irreducible observation noise that would be common to the compared predictors.

An \emph{exact expected-risk comparison} averages over the specified fitting randomness and is an equality or inequality about that model's true risk.  A \emph{high-probability certificate} says that a bound is valid for all training samples except a set of probability at most $\delta$.  A smaller certificate does not imply smaller actual error: if $r_1\le10$ and $r_2\le20$, either true error can be smaller.  A \emph{held-out audit} estimates the risk difference of already fitted systems on new examples.  \Cref{thm:weighted-projection-frontier} illustrates an exact expected-risk comparison; \cref{thm:terminal-certificate} gives a high-probability certificate; \cref{prop:paired-real-improvement} gives a held-out audit.

\subsection{Index and notation conventions}
Unless stated otherwise, counts are positive integers, sets of roles and experts are finite, all maps and random variables are measurable, and all stated expectations exist.  Probability arguments use standard Borel state and probe spaces (measurable spaces equivalent to Borel subsets of a complete separable metric space); finite sets, Euclidean vectors, and finite token sequences satisfy this condition.  These routine conditions allow conditional distributions and conditional means to be interpreted consistently.

\begin{longtable}{p{0.24\textwidth}p{0.68\textwidth}}
\toprule
Symbol or convention & Meaning and scope \\
\midrule
\endhead
$a\in\cJ$, $e\in[K]$ & A recurring role and an expert parameter-block index.  $[K]=\{1,\ldots,K\}$.  A role includes the call distribution and environment; it need not coincide with a human-assigned semantic label.\\
$m$, $K_m$, $\alpha_m$ & Comparator model, its expert count, and its role-to-expert assignment.  The compact comparison uses $m\in\{\mathrm{reuse},\mathrm{sep}\}$.\\
$\cJ_e$, $M$, $J$ & Roles assigned to one reused expert; the number of roles in a local group; the number of role cells in a stacked model.  A role cell can be a role--layer pair.\\
$j$, $H$, $q$, $Q(\gamma)$ & Event index $j=0,\ldots,H-1$ and horizon $H$; within-event call index $q$ and number of calls in program $\gamma$.  An event need not be one token.\\
$\ell$, $L$ in depth models & Layer index and number of layers.  $L_j$ elsewhere is a Lipschitz constant, while a matrix $L$ in \cref{app:assumption-robust} is a linear sharing rule.\\
$X_j=S_j$, $\widehat X_j^m=\widehat S_j^m$ & Reference and fitted-model pre-event states; $\cS_j$ is their declared stage space.  These are not the regression design matrices $X_a$.\\
$\Gamma_j$, $\gamma$ & Random canonical route program and one fixed program value.  $\mathfrak G_j^\star$ is their space.  Canonical programs use role labels rather than arbitrary expert numbering.\\
$\iota_j^m$, $\widehat\Gamma_j^m$ & Implementation of a canonical program in model $m$, and the actually selected model-specific route.  These can differ.\\
$\Phi_j^m$, $T_{j,\gamma}$ & Complete model-specific event and reference event under program $\gamma$, including dispatch, shared paths, experts, residuals, and aggregation.\\
$z\in\cZ_a$, $P_a$ & A local probe record and its reference distribution.  A record supplies the information needed to execute a controlled call and compare its effect.\\
$\cT_a[f]$, $T_a^\star$ & The event lift with candidate expert $f$ inserted, and its target counterpart.  They have the same output state space even when $f$ returns only a residual vector.\\
$\ell_a$, $R_a$, $\widehat R_a$ & One-example loss, population expected role loss, and sample-average role loss.  $\ell_a$ is a function, not a layer index in this context.\\
$\cF_e$, $\pi_{e,a}$ & Candidate expert functions and within-group evaluation probabilities, fixed independently of expert-fitting data unless selection is controlled.\\
$R^*_{\mathrm{sep},e}$, $R^*_{\mathrm{share},e}$ & Role-specific and shared population oracle risks.  Their difference $\BReuse_e\ge0$ is reuse bias for that class and evaluation distribution.\\
$n_a$, $N$, $\Neff_e$ & Samples from role $a$, total independent sample count, and weighted effective size $(\sum_a\pi_{e,a}^2/n_a)^{-1}$.  Effective size is not the number of expert calls.\\
$\delta$, $\delta_e$, $\epsilon_{\mathrm{sh},e}$, $\epsilon_{\mathrm{sep},e}$ & Allowed failure probabilities and uniform-deviation radii.  Group failure budgets must sum to at most the advertised total.\\
$w_a$, $W_e$, $\omega_{j,a}$ & Terminal role importance, its sum within a group, and event-level occurrence/transfer weight.  The group evaluation weights are $\pi_{e,a}=w_a/W_e$ on positive-weight roles.\\
$a_j^m$, $r_j^m$, $\tau_j^m$, $\rho_j^m$ & Correctly aligned local approximation error, route-induced local error, residual assembly error, and expected route error.  They measure different mechanisms.\\
$A_j$, $U_m$, $\mathcal U$, $L_{\mathcal U}$ & Downstream amplification product, a state-error certificate, terminal utility, and its Lipschitz constant.  $U_m$ is a bound, not utility.\\
$\PRG_H$ & The no-sharing state-bound minus the reuse state-bound under their stated matching assumptions.  Its sign orders those bounds only.\\
$C$, $\Psi$, $F_i$, $T_i$ & Complete executable context, macro-state extractor, proposed macro operator, and the actual macro transition after intervention.  Here $C$ is not the variance saving in the linear model.\\
$\mathcal A_i=(\varpi_i,\sigma_i)$ & A route action: online route policy and bounded stopping time.  $K$ in the causal sections counts actions, which may be compound multi-expert interventions.\\
$\nu_i$, $\mu_i$, $\Xi_i$ & Probe-context law, average intervention defect, and largest action-effect difference between contexts aliased by $\Psi$.\\
$p_a(s)$, $e_a(C)$ & Declared state-based intervention distribution for a population identity, and known action-sampling propensity in an audit.  They are distinct experimental policies.\\
$\beta$, $\beta_j$ & Audit overlap floor (a probability), versus common-state router-logit error bound.  They have different meanings and units.\\
$\gamma_k$, $\Delta_j$, $\zeta_j$ & Top-$k$ support margin, extra functional severity of a support change, and route cost remaining even when supports agree.\\
\bottomrule
\end{longtable}
A hat marks an estimate, fitted object, empirical quantity, or state/route realized by a fitted model; $\Phi_j^m$ and $\mathcal R_j^m$ also belong to the fitted model despite having no hat; $\star$ marks a reference or oracle, specified locally.  $\argmin$ is the set of minimizing arguments; $\inf$ is the greatest lower bound even when no minimizer is attained.  Empty sums are zero and empty products are one.  $\mathbf1\{B\}$ equals one when event $B$ occurs and zero otherwise.  $\operatorname{Law}(X)$ means the probability distribution of $X$, and ``almost surely'' means except on an event of probability zero.

\subsection{Linear-model notation}
For a matrix $A$, $\|A\|_F$ denotes its Frobenius norm and $\|A\|_2$ its operator norm. The common-Hilbert-risk assumption fixes one shared geometry for all role risks. The design Gram matrix $X^\top X$ and the propagation Gramian $P_j^\top P_j$ describe different geometries.

\begin{longtable}{p{0.24\textwidth}p{0.68\textwidth}}
\toprule
Linear-model symbol & Definition \\
\midrule
\endhead
$A_a\in\R^{p\times d}$, $D=pd$ & True input-to-output operator and its scalar coefficient count.\\
$X_a$, $Y_a$, $\mathcal E_a$ & Fixed $n_a\times d$ design matrix, $n_a\times p$ observed responses, and response-noise matrix in $Y_a=X_aA_a^\top+\mathcal E_a$.\\
$\widehat A_a$, $E_a$ & Fitted OLS matrix and its parameter error $E_a=\widehat A_a-A_a$; this $E_a$ is not an expert function $E_{\theta,e}$.\\
$\sigma_a^2$, $v_a$ & Response-noise variance and coefficient-estimation variance $v_a=\sigma_a^2/n_a$ in the orthogonal design.  Equal noise levels are homoscedastic; unequal ones are heteroscedastic.\\
$B$, $V$, $C$ & Balanced squared sharing bias, no-sharing risk $V=pd\sigma^2/n$, and variance saving $C=V(1-1/M)$.\\
$\lambda$, $\lambda^\star$ & Fixed partial-pooling strength and its population-optimal value.  Covariance eigenvalues $\lambda_k$ and resource multipliers $\lambda_{\mathrm{res}}$ are unrelated parameters.\\
$\Pi$, $G,H$, $K_\Pi$ & Partition, two of its disjoint groups, and number of groups.  $H$ here is a group name, not the trace horizon.\\
$B_{\mathrm{role}}$, $B_{\mathrm{depth}}$, $B_{\mathrm{int}}$ & Between-role main-effect, depth main-effect, and residual interaction squared variations in the balanced role--depth array.\\
$\theta$, $Y$, $\zeta$, $\Sigma$ & Stacked true parameters, their noisy base estimate, centered estimation noise, and its full covariance in the general linear observation model.\\
$W$, $L$, $P$, $Q$ & Evaluation weight matrix, fixed linear sharing rule, weighted sharing projector, and discarded-direction projector $Q=I-P$ in \cref{app:assumption-robust}.\\
$\mathbf e_j$, $\xi_j$, $M_j$, $P_j$ & State-error vector, injected local error, linear propagation matrix, and product of later matrices.\\
$b_j$, $\Sigma_{jk}$, $K_e$ & Mean injected error, cross-time covariance, and sum of all terminally propagated effects of one shared expert error.\\
\bottomrule
\end{longtable}

\subsection{Transfer and transport conventions}
The $1$-Wasserstein distance is
\begin{equation}
 W_1(P,Q;d)=\inf_{\gamma\in\Pi(P,Q)}\int d(x,y)\,d\gamma(x,y),
 \label{eq:guide-wasserstein}
\end{equation}
where $\Pi(P,Q)$ is the set of couplings. The finite first-moment condition is $\E d(X,x_0)<\infty$ for one fixed point $x_0$. The coupling notation $\Pi(P,Q)$ is unrelated to a partition $\Pi$. In the discrete route transport of \cref{eq:functional-gate-transport}, costs are distances between expert outputs, not between their labels.

\subsection{Intervention conventions}
The propensity $e_a(C)$ is the probability of actually sampling action $a$ at context $C$, not generally the router's softmax score under deterministic top-$k$ selection. Overlap requires $e_a(C)\ge\beta>0$ on every claimed region/action. Changing the intervention policy or target distribution changes the estimand. A Markov-sufficient macro state makes the transition law depend on the complete context only through that state and the chosen action; otherwise kernels must be defined on the complete context.

\subsection{Computational vocabulary and comparison objects}\label{app:computational-vocabulary}
A role groups specified call environments with a probe law and target; an event contains the calls and surrounding operations being compared. A reasoning step is a task-level operation, with no assumed one-to-one relation to a token, layer, call, or event. The independent statistical unit is the entire independently sampled record, often a complete trace rather than an individual call.

\emph{Support replay} fixes selected indices but recomputes their gains at the replay's own states. \emph{Frozen-gain replay} also fixes recorded gains; natural execution recomputes its selections. Freezing a feature law fixes both the feature map and its evaluation input distribution. 

An estimation-risk frontier averages fitting randomness; a terminal-risk frontier also includes the map carrying errors to the endpoint. An approximation frontier minimizes population discrepancy over a feasible class without charging estimation or optimization.

\subsection{Technical terms used by the extensions}\label{app:extension-vocabulary}
Function-subspace dimension, matrix rank, factor norms, and the norm of an increment $BA-B_0A_0$ are distinct constraints. A fixed core is not optimized, whereas a free core is. A feature second moment is $\E[\varphi\varphi^\top]$, not a centered covariance. Conditioning on a selection cell requires recomputing conditional means and covariances. A full-law risk averages both non-escape and escape outcomes; derivative envelopes must hold throughout the specified region, not only at the expansion point.

\section{Routed Computation Graphs, Roles, and Event Lifts}
\label{app:routed-roles}

\paragraph{Overview.}
This appendix explains what is shared and how its effect is compared inside a complete computation. The worked example makes the roles, assignments, and event lifts concrete; the remaining sections give the precise interfaces and sufficient conditions connecting local probes to rollout errors. 
After the example, Sections~\ref*{app:routed-roles}.1--\ref*{app:routed-roles}.2 define programs and role risks, Sections~\ref*{app:routed-roles}.3--\ref*{app:routed-roles}.4 give the replacement and transfer arguments, and Section~\ref*{app:routed-roles}.5 distinguishes the levels of reuse.

\subsection*{Worked example: two-layer expert reuse}
\phantomsection\label{app:worked-routed-transformer}
\addcontentsline{toc}{subsection}{Worked example: two-layer expert reuse}
\paragraph{One prediction through two layers.}
Consider the request ``current state 3, next operation multiply by two.''  The task in \cref{app:experiment-details} uses $F_2(s)=2s\bmod31$, so the target next state is $F_2(3)=6$.  This is one operation within a supplied program.  To follow the internal computation, use the illustrative two-layer causal Transformer: width 128, four attention heads, and four candidate SwiGLU experts of intermediate width 384 per layer.  Both layers contribute to this one prediction; they do not perform two successive multiplications.  Each receives the position's hidden representation, not the integer 3 as a direct expert input.  The primary trained comparators in \cref{app:new-task} instead have four layers and hidden width 256.  The example uses declared reference functions and probes.

\paragraph{Two roles and their assigned experts.}
For this request, the controlled call type is $k=i+1=3$, since the task operation index is $i=2$.  The chosen prediction position gives role $(1,3)$ in the first layer and role $(2,3)$ in the second.  The separate model uses each layer's own local expert 3---two independent parameter blocks, numbered 3 and 7 when flattened.  The reuse model assigns shared block 3 to both roles.  Its four-expert bank is shared across depths, while routers and normalization remain layer-local; the separate model stores two independent banks.    A different input number can instantiate the same role.

This pair belongs to the full eight-role construction.  Let $k\in\{1,2,3,4\}$ name a controlled call type and $\ell\in\{1,2\}$ its layer.  At symbolic-task prediction positions, $k=i+1$ labels the requested operation $i\in\{0,1,2,3\}$.  The full assignment is
\[
 \cJ=\{(\ell,k):\ell\in\{1,2\},\ k\in\{1,2,3,4\}\},\qquad
 K_{\mathrm{sep}}=8,\quad K_{\mathrm{reuse}}=4,
\]
\[
 \alpha_{\mathrm{sep}}(\ell,k)=4(\ell-1)+k,
 \qquad \alpha_{\mathrm{reuse}}(\ell,k)=k,
 \qquad \cJ_k=\{(1,k),(2,k)\}.
\]
\begin{center}
\begin{tabular}{@{}cccc@{}}
\toprule
Call type & Two distinct roles & Separate blocks & Reused block\\
\midrule
1 & $(1,1),(2,1)$ & $1,5$ & 1\\
2 & $(1,2),(2,2)$ & $2,6$ & 2\\
3 & $(1,3),(2,3)$ & $3,7$ & 3\\
4 & $(1,4),(2,4)$ & $4,8$ & 4\\
\bottomrule
\end{tabular}
\end{center}
A call occurrence belongs to one role.  Paper indices are one-based; code bank positions are $k-1$.  Separate indices 5--8 flatten a layer/index pair, not an eight-way router.  At role $(2,3)$, the reuse assignment specifies block 3, but the natural router may select block 1.  This remains role $(2,3)$, not $(2,1)$.

\paragraph{Reference updates and task outputs.}
The target next state 6 does not uniquely specify a 128-dimensional residual at either layer.  For the local comparison, specify reference weights independently and let $f_a^\star$ be the corresponding same-interface SwiGLU function.  Both fitted models are evaluated against this declared target; the fitted separate model is not the reference.  The local function returns a hidden-state residual, whereas $F_i:\mathbb Z_{31}\to\mathbb Z_{31}$ returns a task state.  These local reference weights are not asserted to form a complete network that outputs 6.  

\paragraph{A residual difference you can calculate.}
For arithmetic only, suppose a valid controlled probe fixes the prefix and all other call contributions.  The base row has first two coordinates $(1,0)$, and the expert output is multiplied by a fixed gain $c=1/2$.  Suppose the candidate residual begins $(3,1)$ and the reference residual begins $(2,1)$, with all remaining output differences zero.  The updated row then begins $(1,0)+\tfrac12(3,1)=(2.5,0.5)$ for the candidate, and $(1,0)+\tfrac12(2,1)=(2,0.5)$ for the reference.  Under the complete-state metric below, this gives a single-probe distance of $0.5$: the expert returns an update, and the comparison measures its effect after insertion into the surrounding state.

Only the first two coordinates of the 128-dimensional row are displayed; all other coordinate differences are zero. These illustrative values give a single-probe state distance of $0.5$.

\paragraph{The complete-state representation.}
Each $E_{\theta,e}:\R^{128}\to\R^{128}$ returns a residual, with parameters
\[
 E_{\theta,e}(u)=D_e[\operatorname{SiLU}(V_eu)\odot(U_eu)],\quad
 U_e,V_e\in\R^{384\times128},\quad D_e\in\R^{128\times384}.
\]
For a fixed-length prefix without dropout, take the event to be one full FFN subblock with residual addition.  Its pre-event state is $X=(p,\mathbf H)$, where $p$ is the prefix and $\mathbf H\in\R^{n\times128}$ is the complete pre-FFN array after attention.  The expert input is $u=\operatorname{LN}(\mathbf H_{t,:})$ at position $t$.  Mask and layer boundary are fixed.  This full-prefix implementation uses no KV-cache path.  Attention, FFN, and readout can be separate events: the horizon $H$ is not automatically the layer count.  $\cS_j$ contains the complete records at boundary $j$; $X_j$ and $\widehat X_j^m$ follow the target and fitted trajectories respectively.

A valid $z\in\cZ_a$ fixes $p,\mathbf H,t,u$, a gain $c$, and the other call contributions.  Let $\mathbf H_{\mathrm{base}}(z)$ retain the residual path and all contributions except the controlled call.  $\operatorname{Insert}_t(v)$ is an $n\times128$ zero array with row $t$ equal to $v^\top$.  The numerical comparison above is an instance of
\[
\begin{aligned}
 \cT_a[f](z)&=(p,\mathbf H_{\mathrm{base}}(z)+\operatorname{Insert}_t(cf(u))),\\
 T_a^\star(z)&=(p,\mathbf H_{\mathrm{base}}(z)+\operatorname{Insert}_t(cf_a^\star(u))).
\end{aligned}
\]
For same-shaped states use
\[
 d((p,\mathbf H),(p',\mathbf H'))=\mathbf1\{p\ne p'\}+\|\mathbf H-\mathbf H'\|_F.
\]
Only one row differs on this probe, so $\ell_a(f;z)=|c|\|f(u)-f_a^\star(u)\|_2$.  Take $P_a$ uniform on a valid probe bank independently fixed before expert fitting; $R_a(f)=\E_{P_a}\ell_a(f;Z)$ averages these complete-state discrepancies.  

\paragraph{Intended and actual execution.}
A canonical $\Gamma_j\in\mathfrak G_j^\star$ names roles, positions, gains or gain rules, and combination for all calls in the event.  $\iota_j^m$ substitutes the assigned model-specific blocks, producing an aligned plan in $\mathfrak G_j^m$.  The actual plan is $\widehat\Gamma_j^m=\mathcal R_j^m(\mathcal H_j^m)$, using only visible prefix, activations, and already revealed randomness.    The execution maps return complete event states.  At one common current state, compare
\[
\begin{array}{ll}
 \Phi_j^m(\widehat X_j^m;\widehat\Gamma_j^m) & \text{actual fitted execution},\\
 \Phi_j^m(\widehat X_j^m;\iota_j^m(\Gamma_j)) & \text{aligned fitted execution},\\
 \Phi_j^\star(\widehat X_j^m;\Gamma_j) & \text{target at the model state}.
\end{array}
\]
The first two isolate routing cost; the last two isolate implementation error.  The reference trajectory itself evaluates $\Phi_j^\star$ at $X_j$, not at $\widehat X_j^m$.

The eight illustrated roles cover controlled prediction-position calls, not all errors in a complete rollout; other calls need their own losses or justified assembly terms.  The Frobenius metric is generally unbounded.    \Cref{app:routed-roles}.2 specifies the fixed-probe conditions, and \cref{sec:frozen-gain-replay} treats fixed-gain replay.  See \cref{app:notation} for component terminology and return to \hyperref[sec:framework]{Section~\ref*{sec:framework}} for the general definitions.

\subsection{Canonical programs and model-specific implementations}
Let
\begin{equation}
 \cE_\theta=\{E_{\theta,e}:\cX_e\rightarrow\cY_e\}_{e=1}^{K}
 \label{eq:app-expert-library}
\end{equation}
be a library of parameter blocks.  A canonical route program $\gamma\in\mathfrak G_j^\star$ is labelled by functional roles rather than by model-specific expert indices.  It contains the controlled calls, their dispatched inputs, mixture weights, call order, shared paths, residual operations, aggregation, and---when needed---a finite non-anticipating policy for internal adaptive choices.  Each model $m$ has an implementation map
\begin{equation}
 \iota_j^m:\mathfrak G_j^\star\longrightarrow\mathfrak G_j^m
 \label{eq:app-plan-implementation-map}
\end{equation}
that replaces every canonical role occurrence by the expert index, tied parameter block, or subgraph used by that comparator.  This map is part of the comparison protocol and aligns canonical roles with each model's executable library.

Conditional on an executable plan and the relevant exogenous randomness, the event map
\begin{equation}
 \Phi_{j}^{m}:\cS_j\times\mathfrak G_j^m\rightarrow\cS_{j+1}
 \label{eq:app-fixed-plan-map}
\end{equation}
is deterministic.  A non-anticipating scheduler produces the actually executed plan $\widehat\Gamma_j^m$ from information available before its corresponding routing decision.  Future random variables may be represented by an exogenous seed in the probability space, but are not made observable to the router merely to obtain a deterministic representation.

\paragraph{Complete interfaces for \cref{eq:rcg-main}.}\label{app:event-interface}
A \emph{state space} $\cS_j$ is the set of admissible complete records immediately before event $j$.  For a Transformer event, a record can include the prefix, the hidden-state array, attention caches, and any earlier values reused by residual connections.  A \emph{state} $X_j\in\cS_j$ is one such record, whereas $X_{j+1}\in\cS_{j+1}$ is the record after the event.  The learned state $\widehat X_j^m$ is a separate trajectory value in the same declared comparison representation.  The spaces $\cX_e,\cY_e$ describe one expert's interface and need not equal the event-state spaces.  If models use different internal representations, the comparison maps and induced event transitions must be specified; equal output dimensions alone do not supply these maps.

The program spaces $\mathfrak G_j^\star$ and $\mathfrak G_j^m$ are sets of admissible instructions.  A lowercase $\gamma$ is a fixed member; $\Gamma_j$ is the possibly random intended program.  A program includes the declared input-selection rules, call order, mixing weights or weight-computation rules, and aggregation.  \emph{Canonical} means that roles are named consistently across models.    The implementation map $\iota_j^m$ is fixed as part of this comparison and may send different roles to the same parameter block.  

Let $\mathscr H_j^m$ be the space of admissible history records at the routing boundary.  Then
\[
 \mathcal H_j^m\in\mathscr H_j^m,
 \qquad \mathcal R_j^m:\mathscr H_j^m\longrightarrow\mathfrak G_j^m,
 \qquad \widehat\Gamma_j^m=\mathcal R_j^m(\mathcal H_j^m).
\]
The history record may contain the router's view of the current state, earlier plans and outputs, and random draws already revealed for this decision.  It need not expose every coordinate of the complete event state.  It is a random record, not itself a sigma-field.  If $\mathscr I_j^m$ is the information allowed before the decision, non-anticipation requires the history and decision to be measurable with respect to $\mathscr I_j^m$.  For an internal adaptive policy this requirement applies separately at each internal decision.  

An \emph{external random tape} is a collection of random draws used to realize all compared executions on one probability space.  Fixing it makes the transitions deterministic as functions of their arguments; only the appropriate revealed portion is visible to a router.  The tape argument is suppressed in $\Phi_j^\star$ and $\Phi_j^m$.  

There are three distinct event outputs at a common model state: actual execution uses $\Phi_j^m(\widehat X_j^m;\widehat\Gamma_j^m)$; aligned execution uses $\Phi_j^m(\widehat X_j^m;\iota_j^m(\Gamma_j))$; the reference comparison uses $\Phi_j^\star(\widehat X_j^m;\Gamma_j)$.  The actual reference trajectory instead evaluates the last map at $X_j$.  These insertions are precisely the split used in \cref{eq:app-path-recursion}.  

In \cref{sec:end-to-end} all compared stage states are represented in a declared common metric space $(\cS,d)$, and the notation is
\[
 S_j=X_j,\quad \widehat S_j^m=\widehat X_j^m,\qquad
 T_{j,\gamma}(s)=\Phi_j^\star(s;\gamma),\quad
 \widehat T^m_{j,\gamma}(s)=\Phi_j^m(s;\gamma).
\]
In the first transition $\gamma\in\mathfrak G_j^\star$; in the second it lies in $\mathfrak G_j^m$.  The semicolon merely separates state and plan arguments.  The intended program may depend on the reference history; recomputing a reference policy on the learned state would define a different comparator.

The abstraction includes the following common cases.
\begin{center}
\small
\begin{tabular}{p{0.26\textwidth}p{0.66\textwidth}}
\toprule
Architecture & Canonical-program representation \\
\midrule
Top-$k$ token routing & One sparse role measure per token; selected calls are combined in parallel. \\
Expert-choice or balanced routing & A batch- or sequence-level role-assignment matrix is chosen jointly. \\
Additive soft gate & A normalized role measure weights expert outputs at a common input. \\
Full Soft MoE dispatch & The program includes dispatch and combine matrices; because slot inputs depend on the assignment, the generic fixed-program bound is used unless extra dispatch sensitivity is controlled. \\
Shared experts & The program contains always-active calls and conditionally selected calls. \\
Global/cross-layer pools & Canonical roles are mapped to a shared model-specific pool at several depths. \\
Tied or recurrent experts & One parameter block implements several canonical occurrences. \\
Sequential/hierarchical experts & A finite expert program is executed; adaptive choices are represented by a non-anticipating policy or split into separate routed events. \\
\bottomrule
\end{tabular}
\end{center}
  It uses their fixed-program transitions, their implementation maps, and the role-level domination condition stated below.  The architecture-specific transport and matching results in \cref{app:architecture-routing} apply only to their stated additive or preselected-sequence specializations.

\subsection{Roles as fixed call environments}
A role $a\in\cJ$ specifies (i) a probe distribution $P_a$ and (ii) a measurable event lift $\cT_a[f]$.  A probe record $z$ contains all information needed to execute the controlled call environment: the pre-event state, canonical program, controlled occurrence, revealed randomness, and the fixed values of the other components in the hybrid comparison.  The lift returns a post-event model state after inserting candidate function $f$ at the controlled occurrence.  The target hybrid $T_a^\star$ has the same codomain.  Therefore
\begin{equation}
 \ell_a(f;z)=d\!\left(\cT_a[f](z),T_a^\star(z)\right)
 \label{eq:app-event-loss}
\end{equation}
is a distance between two event states, even if $f$ itself maps a token representation to an FFN residual.

For the finite-sample theorem, $P_a$, $\cT_a$, $T_a^\star$, and the expert class are fixed independently of the samples used to estimate the experts, and $\ell_a\in[0,1]$.  A frozen surrounding assembly or an independent sample split is sufficient.  If the same observations jointly train the backbone, dispatch, or event lift, one must analyze the joint hypothesis class rather than condition on a data-dependent loss without paying its complexity.  Scaling a bounded distance is valid; post-hoc clipping of an unbounded distance controls only the clipped quantity and cannot silently be substituted into an unclipped terminal bound.

A reuse assignment $\alpha_m:\cJ\to[K_m]$ maps roles to model-specific parameter blocks.  The main compact comparison assumes that the canonical roles, evaluation weights, and expert-estimation classes are common across comparators.  Data-dependent assignments require either an independent selection split or an additional selection-complexity term.

\subsection{A deterministic call-replacement lemma}
Fix a state $s$ and one realized canonical program $\gamma$ with $Q(\gamma)$ controlled call occurrences.  For a model $m$, let $\widetilde T_{j,\gamma}^{m,(q)}(s)$ be the hybrid event output in which the first $q$ canonical occurrences use the fitted expert functions and the remaining occurrences use their target functions, while all non-expert components use the fixed model-$m$ assembly.  Thus $\widetilde T^{m,(Q)}$ is the correctly aligned learned event and $\widetilde T^{m,(0)}$ uses target experts inside the model assembly.  Define
\begin{equation}
 \tau_{j,\gamma}^{m}(s)
 =d\!\left(\widetilde T_{j,\gamma}^{m,(0)}(s),T_{j,\gamma}(s)\right),
 \label{eq:app-assembly-residual}
\end{equation}
which isolates shared-backbone, dispatch, aggregation, or other assembly mismatch.

\begin{lemma}[Call-by-call event replacement]
\label{lem:call-replacement}
For every $s$, $\gamma$, and model $m$,
\begin{equation}
 d\!\left(\widetilde T_{j,\gamma}^{m,(Q)}(s),T_{j,\gamma}(s)\right)
 \le \tau_{j,\gamma}^{m}(s)
 +\sum_{q=1}^{Q(\gamma)}
 d\!\left(\widetilde T_{j,\gamma}^{m,(q)}(s),
          \widetilde T_{j,\gamma}^{m,(q-1)}(s)\right).
 \label{eq:app-call-hybrid}
\end{equation}
Each summand is an event-lift loss for the role of occurrence $q$ under the corresponding hybrid probe record.
\end{lemma}
\begin{proof}
Insert the sequence $\widetilde T^{m,(Q-1)},\ldots,\widetilde T^{m,(0)}$ and apply the triangle inequality.  By construction, adjacent hybrids differ at exactly one controlled occurrence and have outputs in the same post-event state space.  The lemma is a fixed-program hybrid statement.  If later routing decisions adapt to earlier expert outputs, refine those decisions into separate routed events or treat the entire adaptive policy as one indivisible event lift.  Conditioning on one realized trace while replacing an earlier output is valid only when the later decisions are explicitly frozen as part of the intervention; it is not automatically a decomposition of the original adaptive execution.
\end{proof}

\subsection{A rigorous sufficient occupancy-transfer condition}
Condition on the expert-estimation sample and let $Z_{j,q}^m$ be the hybrid probe record of the $q$th controlled occurrence when that occurrence exists.  Assume a deterministic maximum $Q_{\max,j}<\infty$; a summable version gives the same conclusion for an unbounded count.  For each role $a$, define the conditional sub-probability measure
\begin{equation}
 Q_{j,q,a}^m(B)
 =\Prb\!\left(
   Z_{j,q}^m\in B,
   q\le Q(\Gamma_j),
   a(j,\Gamma_j,q)=a
   \mid \mathcal D
 \right),
 \label{eq:app-hybrid-subprobability}
\end{equation}
where $\mathcal D$ denotes the information in the fitted sample.  Formally this is a sigma-field: the collection of events whose truth can be determined from the sample.  Conditioning fixes the fitted parameters, not the fresh evaluation trajectory.  Suppose there are deterministic constants $\kappa_{j,q,a}\ge0$, fixed independently of $\mathcal D$ and valid for both comparators, such that
\begin{equation}
 Q_{j,q,a}^m\ll P_a,
 \qquad
 \left\|\frac{dQ_{j,q,a}^m}{dP_a}\right\|_\infty
 \le \kappa_{j,q,a}
 \quad\text{a.s.}
 \label{eq:app-density-ratio}
\end{equation}
The notation $Q\ll P$ means $P(B)=0$ implies $Q(B)=0$.  The Radon--Nikodym derivative $dQ/dP$ is the weight function satisfying $Q(B)=\int_B(dQ/dP)\,dP$; its $\infty$-norm is the smallest almost-sure upper bound.  Because $Q_{j,q,a}^m$ has total mass at most one (a sub-probability measure), $\kappa_{j,q,a}$ already includes both occurrence mass and distribution shift.  Integrating each hybrid loss in \cref{eq:app-call-hybrid} yields
\begin{equation}
 \E[a_j^m\mid\mathcal D]
 \le \tau_j^m
 +\sum_{a\in\cJ}\omega_{j,a}
   R_a\!\left(\widehat f^m_{\alpha_m(a)}\right),
 \qquad
 \omega_{j,a}=\sum_{q=1}^{Q_{\max,j}}\kappa_{j,q,a},
 \label{eq:app-role-domination-derived}
\end{equation}
where
\begin{equation}
 \tau_j^m
 =\E\!\left[
   \tau_{j,\Gamma_j}^{m}(\widehat S_j^m)
   \mid\mathcal D
 \right].
 \label{eq:app-expected-assembly-residual}
\end{equation}
Indeed, $\int\ell_a(\widehat f;z)\,dQ_{j,q,a}^m(z)\le\kappa_{j,q,a}R_a(\widehat f)$.  This is one sufficient condition for \cref{ass:role-aggregation}.  \Cref{prop:additive-occupancy-transfer} permits support mismatch by retaining an additive discrepancy term.  More refined lifts can absorb within-event suffix sensitivity and give smaller weights.  If only model-specific envelopes $\kappa_{j,q,a}^m$ are credible, each comparator must retain its own terminal weights and full certificate.

\subsection{Hierarchy of reuse claims}\label{app:reuse-meanings}

\begin{enumerate}[leftmargin=*,itemsep=1pt]
\item \emph{Parameter reuse}: identical weights occur in multiple calls.
\item \emph{Statistical reuse}: samples from several roles jointly estimate those weights.
\item \emph{Functional reuse}: one function has low role risk on the stated role distributions.
\item \emph{Causal reuse}: a pre-action route intervention induces a stable macro transition.
\item \emph{Compositional reuse}: that transition remains accurate in previously unseen route sequences or horizons.
\end{enumerate}
Only the first item is an architectural fact.  Each subsequent item requires additional empirical or mathematical conditions.

\paragraph{Partial pooling versus physical tying.}
The interpolation in \cref{eq:partial-reuse-estimator} shares statistical information, but
\[
 \widehat A_a^{(\lambda)}-\widehat A_b^{(\lambda)}
 =(1-\lambda)(\widehat A_a-\widehat A_b).
\]
For $0<\lambda<1$, these matrices need not coincide.  Writing them as a common core plus role-specific residuals is only a reparameterization unless the residual capacity is restricted and counted.  Full pooling and each block of a hard partition do use one common matrix; intermediate shrinkage does not by itself reduce the number of stored role matrices.  Theorems about estimation risk, approximation under an update constraint, and physical storage therefore address different objects.

\paragraph{The expert's information is part of its function class.}
The common operator in the local model receives $x$, not a separate role identifier.  A class of functions $f(x,a)$ can implement role-dependent operations with one parameter block and is not the same class as the shared $f(x)$ in that model.  A hidden feature can also encode role or depth without an explicit identifier.  Thus a fixed feature map, its input law, the permitted conditioning information, and any adapter must be declared before interpreting a sharing bias.

\section{Distribution-Free Statistical Theory}
\label{app:statistical-proofs}

\paragraph{Overview.}
We compare fitting one function for a group with fitting a separate function for each role. Reuse bias measures the population cost of the shared restriction, while concentration controls estimation from independent records. 
Section~\ref*{app:statistical-proofs}.1 interprets reuse bias, Section~\ref*{app:statistical-proofs}.2 proves the finite-class comparison, Section~\ref*{app:statistical-proofs}.3 examines sample allocation, and Section~\ref*{app:statistical-proofs}.4 gives uniform-deviation and trace-level templates.

\subsection{Reuse bias and its functional interpretation}
Fix a group $\cJ_e$ and a class $\cF_e$.  The role distributions, losses, class, weights, and sample counts are fixed independently of the observations used by the ERMs.  For weights $\pi_{e,a}>0$ summing to one, define the shared and separate population oracles as in \cref{eq:general-reuse-bias}.  Nonnegativity of $\BReuse_e$ follows because the shared class is the diagonal subset of the product class $\prod_{a\in\cJ_e}\cF_e$.  The concentration theorem below assumes $\ell_a(f;Z)\in[0,1]$ for every role and candidate; if the surrounding event lift is learned from the same sample, it must be included in the hypothesis class or learned on an independent split.

A useful exact interpretation holds under a common quadratic geometry.
\begin{proposition}[Reuse bias as functional variance]
\label{prop:hilbert-bias}
First take the unrestricted candidate class $\cF_e=\cH$, where $\cH$ is a real Hilbert space, and suppose
\begin{equation}
 R_a(f)=c_a+\norm{f-f_a^\star}_{\cH}^{2},
 \qquad f\in\cH.
 \label{eq:app-hilbert-risk}
\end{equation}
With $\bar f=\sum_a\pi_{e,a}f_a^\star$,
\begin{equation}
 \BReuse_e
 =\sum_a\pi_{e,a}\norm{f_a^\star-\bar f}_{\cH}^{2}
 =\frac12\sum_{a,b}\pi_{e,a}\pi_{e,b}
   \norm{f_a^\star-f_b^\star}_{\cH}^{2}.
 \label{eq:app-hilbert-variance}
\end{equation}
More generally, for any nonempty class $\cF\subseteq\cH$ and risks
$R_a(f)=c_a+\|f-t_a\|_{\cH}^2$, set $\bar t=\sum_a\pi_{e,a}t_a$ and
$d(u,\cF)=\inf_{f\in\cF}\|u-f\|_{\cH}$.  The restricted-class reuse bias is
\begin{equation}
 B_{\cF}
 =\sum_a\pi_{e,a}\|t_a-\bar t\|_{\cH}^2
  +d(\bar t,\cF)^2-\sum_a\pi_{e,a}d(t_a,\cF)^2.
 \label{eq:restricted-hilbert-bias}
\end{equation}
No convexity or attainment of the infima is required.  In particular, if every $t_a\in\cF$ and $\bar t\in\cF$ (or its closure), only the variance term remains.
\end{proposition}
\begin{proof}
Write $f-f_a^\star=(f-\bar f)+(\bar f-f_a^\star)$.  The identity $\sum_a\pi_{e,a}(\bar f-f_a^\star)=0$ cancels the summed cross term, giving
\[
 \sum_a\pi_{e,a}\|f-f_a^\star\|_{\cH}^2
 =\|f-\bar f\|_{\cH}^2+
   \sum_a\pi_{e,a}\|\bar f-f_a^\star\|_{\cH}^2.
\]
Only the first term depends on the shared function, so it is minimized at $f=\bar f$.  The constants $c_a$ cancel when subtracting the separate oracle risk.  Expanding the weighted double sum gives
\[
 \sum_{a,b}\pi_{e,a}\pi_{e,b}\|f_a^\star-f_b^\star\|_{\cH}^2
 =2\left(\sum_a\pi_{e,a}\|f_a^\star\|_{\cH}^2-\|\bar f\|_{\cH}^2\right),
\]
which yields the pairwise form.  For the restricted identity, the same completed-square equality with $t_a$ holds for every $f\in\cF$.  Taking its infimum gives the shared risk $\sum_a\pi_{e,a}c_a+\sum_a\pi_{e,a}\|t_a-\bar t\|^2+d(\bar t,\cF)^2$.  Subtract the separate infima $\sum_a\pi_{e,a}[c_a+d(t_a,\cF)^2]$ to obtain \cref{eq:restricted-hilbert-bias}.
\end{proof}
The proposition motivates functional probes.

\paragraph{A finite-class obstruction to the pure-variance interpretation.}
Let $\cH=\R$, $\cF=\{0,1\}$, $\pi_1=\pi_2=1/2$, and $R_1(f)=f^2$, $R_2(f)=(f-1)^2$.  Losses on the candidate class are in $[0,1]$ and both role targets are realizable.  Nevertheless,
\begin{equation}
 B_{\cF}=\frac12,\qquad
 \sum_a\pi_a(t_a-\bar t)^2=\frac14,\qquad
 d(\bar t,\cF)^2=\frac14.
 \label{eq:restricted-bias-counterexample}
\end{equation}
The missing mean $\bar t=1/2$ accounts for the gap.  A common squared-loss geometry alone therefore does not imply the pure-variance claim.

\subsection{Finite-class comparison and proof}
Now the question is sampling error: how far can every candidate's empirical loss be from its population loss? A simultaneous event controls both fitted comparators.

\label{app:finite-class-full}
\begin{theorem}[Simultaneous finite-class comparison]
\label{thm:finite-class}
Assume the losses from \cref{eq:role-risk} lie in $[0,1]$, $\cF_e$ is finite, $n_a\ge1$, and all role samples are mutually independent.  For $\delta\in(0,1)$, with probability at least $1-\delta$, the shared and role-specific empirical risk minimizers (ERMs) satisfy
\begin{align}
 \sum_a\pi_{e,a}R_a(\widehat f_e)
 &\le R_{\mathrm{sep},e}^*+\BReuse_e+2\epsilon_{\mathrm{sh},e},\\
 \sum_a\pi_{e,a}R_a(\widehat f_a)
 &\le R_{\mathrm{sep},e}^*+2\epsilon_{\mathrm{sep},e},
 \label{eq:finite-class-bounds}
\end{align}
where
\begin{align}
 \epsilon_{\mathrm{sh},e}
 &=\sqrt{\frac{\log(4|\cF_e|/\delta)}{2\Neff_e}},\label{eq:finite-radius-shared}\\
 \epsilon_{\mathrm{sep},e}
 &=\sum_a\pi_{e,a}\sqrt{\frac{\log(4|\cJ_e||\cF_e|/\delta)}{2n_a}}.
 \label{eq:finite-radii}
\end{align}
Thus the displayed shared certificate is smaller whenever $\BReuse_e<2(\epsilon_{\mathrm{sep},e}-\epsilon_{\mathrm{sh},e})$.
\end{theorem}

\paragraph{Proof.}\label{proof:finite-class}
For notational simplicity suppress $e$.  Role $a$ supplies independent observations $Z_{a,1},\ldots,Z_{a,n_a}\sim P_a$, and
\begin{equation}
 \widehat R_a(f)=\frac1{n_a}\sum_{i=1}^{n_a}\ell_a(f;Z_{a,i}),
 \qquad
 \widehat R_{\mathrm{sh}}(f)=\sum_a\pi_a\widehat R_a(f).
 \label{eq:app-empirical-risks}
\end{equation}
For fixed $f$, weighted Hoeffding gives
\begin{equation}
 \Prb\!\left(
 \left|\widehat R_{\mathrm{sh}}(f)-\sum_a\pi_aR_a(f)\right|>t
 \right)
 \le 2\exp\!\left(-2\Neff_e t^2\right),
 \label{eq:app-weighted-hoeffding}
\end{equation}
where $(\Neff_e)^{-1}=\sum_a\pi_a^2/n_a$.  A union bound over $\cF_e$ with failure budget $\delta/2$ yields
\begin{equation}
 \sup_{f\in\cF_e}
 \left|\widehat R_{\mathrm{sh}}(f)-\sum_a\pi_aR_a(f)\right|
 \le \epsilon_{\mathrm{sh},e}.
 \label{eq:app-sh-uniform}
\end{equation}
If $f_{\mathrm{sh}}^\star$ is a shared population minimizer and $\widehat f_e$ is a shared ERM, then
\begin{align}
 \sum_a\pi_aR_a(\widehat f_e)
 &\le \widehat R_{\mathrm{sh}}(\widehat f_e)+\epsilon_{\mathrm{sh},e}\nonumber\\
 &\le \widehat R_{\mathrm{sh}}(f_{\mathrm{sh}}^\star)+\epsilon_{\mathrm{sh},e}\nonumber\\
 &\le R_{\mathrm{share},e}^*+2\epsilon_{\mathrm{sh},e}
 =R_{\mathrm{sep},e}^*+\BReuse_e+2\epsilon_{\mathrm{sh},e}.
 \label{eq:app-sh-erm}
\end{align}
For every role, Hoeffding and a union bound over $|\cJ_e||\cF_e|$ with the remaining failure budget give
\begin{equation}
 \sup_{f\in\cF_e}|\widehat R_a(f)-R_a(f)|
 \le \sqrt{\frac{\log(4|\cJ_e||\cF_e|/\delta)}{2n_a}}.
 \label{eq:app-role-uniform}
\end{equation}
The same ERM argument role by role, followed by multiplication by $\pi_a$ and summation, proves \cref{thm:finite-class}.  The two uniform events hold jointly with probability at least $1-\delta$.

\subsection{What the effective sample size optimizes}
\begin{proposition}[Allocation maximizing the Hoeffding effective sample size]
\label{prop:neff-allocation}
For positive real allocations with $N=\sum_a n_a$,
\begin{equation}
 \Neff_e=\left(\sum_a\frac{\pi_a^2}{n_a}\right)^{-1}\le N,
 \label{eq:app-neff-upper}
\end{equation}
with equality if and only if $n_a=N\pi_a$.
\end{proposition}
\begin{proof}
Cauchy--Schwarz gives
$1=(\sum_a\pi_a)^2\le(\sum_a\pi_a^2/n_a)(\sum_a n_a)=N/\Neff_e$.
Equality requires $\pi_a/\sqrt{n_a}$ to be proportional to $\sqrt{n_a}$.
\end{proof}
This allocation maximizes the particular bounded-loss Hoeffding effective sample size.  A Bernstein analysis can produce variance-aware allocations for unequal variances.

\subsection{Uniform-deviation and trace-level templates}
The finite class is not essential.  If, on one event,
\begin{equation}
 \sup_f\left|\widehat R_{\mathrm{sh}}(f)-R_{\mathrm{sh}}(f)\right|\le u_{\mathrm{sh}},
 \qquad
 \sup_f|\widehat R_a(f)-R_a(f)|\le u_a,
 \label{eq:app-general-uniform}
\end{equation}
then the same proof gives shared excess $2u_{\mathrm{sh}}$ and separate excess $2\sum_a\pi_a u_a$.  Uniform bounds based on Rademacher complexity (fit to random signs) or covering numbers (the size of a finite approximation to a class) may be substituted if they prove exactly \cref{eq:app-general-uniform}.  Algorithm-specific stability, compression, and PAC-Bayes bounds are not automatically uniform-deviation statements; using them requires replacing the learning theorem and matching its particular estimator or randomized predictor.  

Calls inside one autoregressive trace should not be counted as independent samples.  Let $Z_1,\ldots,Z_N$ be independent traces, not necessarily identically distributed, and let $L_r(f)\in[0,1]$ be any aggregate trace loss.  For a finite class $\cF$, with probability at least $1-\delta$,
\begin{equation}
 \sup_{f\in\cF}
 \left|\frac1N\sum_{r=1}^{N}
 \bigl(L_r(f)-\E L_r(f)\bigr)\right|
 \le \sqrt{\frac{\log(2|\cF|/\delta)}{2N}}.
 \label{eq:app-trace-concentration}
\end{equation}
If traces are IID, the centering equals the common population trace risk.  No within-trace independence is needed.

\section{Exact Linear Sharing Frontier}
\label{app:exact-frontier-proof}

\paragraph{Overview.}
We prove the exact pooling and partition risks in the specified linear model. We also separate fixed-rule identities from data-dependent choices: a same-data plug-in can fail, independent validation needs its own guarantee, and the stated high-dimensional Gaussian specialization is classical shrinkage.
Sections~\ref*{app:exact-frontier-proof}.1--\ref*{app:exact-frontier-proof}.3 derive the pooling, merge, and dispersion formulas; Sections~\ref*{app:exact-frontier-proof}.4--\ref*{app:exact-frontier-proof}.6 contrast a failing plug-in, independent validation, and the classical Gaussian shrinkage specialization.

\subsection{Oracle details and proof of the exact sharing frontier}
\label{app:exact-frontier-details}
For $B+C>0$, the optimal fixed coefficient and its minimum risk in
\cref{thm:exact-frontier} are
\begin{equation}
 \lambda^\star=\frac{C}{B+C},
 \qquad
 \Risk_{\lambda^\star}=V-\frac{C^2}{B+C}.
 \label{eq:oracle-lambda}
\end{equation}
If $B=C=0$, every $\lambda\in[0,1]$ gives the same risk $V$.
For $M>1$, $\sigma>0$, and finite $B$, oracle partial pooling is no worse than
either endpoint; it is strictly better than separate estimation and is strictly
better than full reuse whenever $B>0$.

\paragraph{Proof of \cref{thm:exact-frontier}.}\label{proof:exact-frontier}
Write $Y_a=X_aA_a^\top+\mathcal E_a$, where the noise rows have mean zero, covariance $\sigma^2I_p$, and are independent across observations and roles.  Gaussianity is not used in the expected-risk proof.  Orthogonality of the fixed design gives
\begin{equation}
 \widehat A_a=A_a+E_a,
 \qquad
 E_a=\frac1n\mathcal E_a^\top X_a.
 \label{eq:app-ols-error}
\end{equation}
The $pd=D$ entries of $E_a$ are centered and uncorrelated with variance $\sigma^2/n$, and the matrices $E_a$ are independent across roles.  Indeed, for output coordinates $u,v$ and input coordinates $r,t$, $\E[(E_a)_{ur}(E_a)_{vt}]=(\sigma^2/n)\mathbf1\{u=v,r=t\}$.  Hence
\begin{equation}
 \E\norm{E_a}_F^2=\frac{pd\sigma^2}{n}=V.
 \label{eq:app-ols-variance}
\end{equation}
For isotropic test input $x$, $\E_x\norm{(\widehat A-A_a)x}_2^2=\norm{\widehat A-A_a}_F^2$.

Separate estimation therefore has average risk $V$.  Let $\bar E=M^{-1}\sum_aE_a$.  The pooled estimator is $\widehat A_{\mathrm{pool}}=\bar A+\bar E$, so
\begin{align}
 \frac1M\sum_a\E\norm{\widehat A_{\mathrm{pool}}-A_a}_F^2
 &=\frac1M\sum_a\norm{\bar A-A_a}_F^2+\E\norm{\bar E}_F^2\nonumber\\
 &=B+\frac{V}{M},
 \label{eq:app-full-sharing-risk}
\end{align}
where the cross term vanishes and $\E\norm{\bar E}_F^2=V/M$.

For partial reuse,
\begin{equation}
 \widehat A_a^{(\lambda)}-A_a
 =\lambda(\bar A-A_a)+(1-\lambda)E_a+\lambda\bar E.
 \label{eq:app-partial-error}
\end{equation}
The average squared bias is $\lambda^2B$.  For each coefficient, the variance of $(1-\lambda)E_a+\lambda\bar E$ is
\begin{equation}
 \frac{\sigma^2}{n}
 \left[(1-\lambda)^2+\frac{2\lambda(1-\lambda)+\lambda^2}{M}\right].
 \label{eq:app-partial-coeff-var}
\end{equation}
Multiplying by $D$ and simplifying yields
\begin{equation}
 \Risk_\lambda
 =V-2V\left(1-\frac1M\right)\lambda
 +\left[B+V\left(1-\frac1M\right)\right]\lambda^2,
 \label{eq:app-partial-quadratic}
\end{equation}
which is \cref{eq:exact-partial-risk}.  The full-sharing comparison follows from
$\Risk_{\mathrm{share}}-\Risk_{\mathrm{sep}}=B-C$.  Differentiating the strictly convex quadratic when $B+C>0$ gives $\lambda^\star=C/(B+C)$ and the stated minimum.  In the degenerate case $B=C=0$, the quadratic is constant in $\lambda$.  If $M>1$ and $\sigma>0$, then $C>0$, so the derivative at zero is negative.  If additionally $B>0$, then $\lambda^\star<1$ and the interior optimum is strictly below full sharing.

\subsection{Proof of the exact partition and merge frontier}
\label{proof:partition-frontier}
For a group $G\subseteq[M]$, write
\begin{equation}
 \widehat A_G=\frac1{|G|}\sum_{a\in G}\widehat A_a
 =\bar A_G+\bar E_G,
 \qquad
 \E\norm{\bar E_G}_F^2=\frac{V}{|G|}.
 \label{eq:app-group-estimator}
\end{equation}
The total expected risk contributed by roles in $G$ is therefore
\begin{align}
 \sum_{a\in G}\E\norm{\widehat A_G-A_a}_F^2
 &=\sum_{a\in G}\norm{\bar A_G-A_a}_F^2
   +|G|\E\norm{\bar E_G}_F^2\nonumber\\
 &=\sum_{a\in G}\norm{\bar A_G-A_a}_F^2+V.
 \label{eq:app-group-risk}
\end{align}
Summing over the $K_\Pi$ blocks and dividing by $M$ proves \cref{eq:partition-risk}.

For two disjoint groups $G,H$, the Hilbert-space variance decomposition gives
\begin{align}
 &\sum_{a\in G\cup H}\norm{A_a-\bar A_{G\cup H}}_F^2\nonumber\\
 &\quad=\sum_{a\in G}\norm{A_a-\bar A_G}_F^2
 +\sum_{a\in H}\norm{A_a-\bar A_H}_F^2
 +\frac{|G||H|}{|G|+|H|}\norm{\bar A_G-\bar A_H}_F^2.
 \label{eq:app-merge-variance-identity}
\end{align}
Merging increases within-group dispersion by the final term and decreases $K_\Pi V$ by exactly $V$, proving \cref{eq:exact-merge-frontier}.

The merge separation has an exact sampling law under the additional assumption that all noise entries are mutually Gaussian with the stated independent-row covariance and $\sigma>0$.  For two pre-specified, disjoint, nonempty groups, let $m=|G|$, $h=|H|$, $c=mh/(m+h)$, $q=\sigma^2/n$, and define
\begin{equation}
 W_{G,H}=c\norm{\bar A_G-\bar A_H}_F^2,
 \qquad
 \widehat W_{G,H}=c\norm{\widehat A_G-\widehat A_H}_F^2.
 \label{eq:app-merge-statistic}
\end{equation}
Because each coefficient of $\widehat A_G-\widehat A_H$ has variance
$q(1/m+1/h)=q/c$,
\begin{equation}
 \frac{\widehat W_{G,H}}{q}
 \sim \chi'^2_D\!\left(\frac{W_{G,H}}{q}\right),
 \qquad
 \E\widehat W_{G,H}=W_{G,H}+V.
 \label{eq:app-merge-statistic-law}
\end{equation}
Thus $\widehat W_{G,H}-V$ is unbiased for the functional separation governing the merge decision.  Exact noncentral-chi-square confidence bounds can be inverted for a pre-specified pair; searching many candidate merges requires a familywise or validation correction.

\subsection{The heterogeneity is observable up to a known noise correction}
The exact model also gives a direct diagnostic for $B$.
\begin{proposition}[Noisy role dispersion]
\label{prop:observed-dispersion}
Assume $M\ge2$, $\sigma>0$, and independent Gaussian observation noise for the distributional identity below.  Let $q=\sigma^2/n$ and
\begin{equation}
 \widehat B_{\mathrm{raw}}
 =\frac1M\sum_{a=1}^{M}
 \norm{\widehat A_a-\widehat A_{\mathrm{pool}}}_F^2.
 \label{eq:app-braw}
\end{equation}
Then
\begin{equation}
 \E\widehat B_{\mathrm{raw}}=B+C,
 \qquad
 \widehat B_{\mathrm{unb}}=\widehat B_{\mathrm{raw}}-C
 \label{eq:app-b-unbiased}
\end{equation}
 is unbiased for $B$.  More precisely,
\begin{equation}
 \frac{M\widehat B_{\mathrm{raw}}}{q}
 \sim \chi'^2_{D(M-1)}\!\left(\frac{MB}{q}\right).
 \label{eq:app-b-noncentral-chi}
\end{equation}
\end{proposition}
\begin{proof}
For one matrix coefficient, stack the $M$ role estimates and let $P=I_M-\mathbf 1\mathbf 1^\top/M$.  The centered estimate is $P(a+\varepsilon)$ with $\varepsilon\sim\mathcal N(0,qI_M)$.  In an orthonormal basis of the rank-$(M-1)$ image of $P$, its squared norm divided by $q$ is noncentral chi-squared with noncentrality $\norm{Pa}^2/q$.  Summing the independent contributions of all $D$ coefficients gives \cref{eq:app-b-noncentral-chi}; its mean gives \cref{eq:app-b-unbiased} because $Dq(M-1)/M=C$.
\end{proof}
The expectation and unbiasedness formulas need only the second-moment assumptions; the noncentral chi-square law additionally needs Gaussian noise, $q>0$, and $M\ge2$.  When $\sigma=0$, dispersion is observed without noise; when $M=1$, both dispersions and $B,C$ are zero and no chi-square expression is needed.  A noncentral chi-square variable is a sum of squared independent unit-variance normal variables with possibly nonzero means; its degrees of freedom count summands and its noncentrality is the sum of squared means.  Thus the same dispersion that determines negative transfer is estimable from role-specific fits after subtracting estimation noise.  The oracle coefficient can be written $\lambda^\star=C/(B+C)$, suggesting the clipped plug-in rule $\widehat\lambda=1$ when $\widehat B_{\mathrm{raw}}=0$ and $\widehat\lambda=\min\{1,C/\widehat B_{\mathrm{raw}}\}$ otherwise, when the exact assumptions are approximately credible.  It is not justified by substituting $\widehat\lambda$ into the fixed-coefficient risk formula.  The low-dimensional Gaussian counterexample below and the contrast-space result give the dimension-dependent conclusions.  \Cref{app:classical-plugin} supplies the classical known-noise Gaussian guarantee when the contrast dimension $D(M-1)$ is at least four.  

\subsection{A same-data plug-in can worsen risk inside the exact model}
Take $M=2$, $D=1$, $q=1$, and write $Y_i=\theta_i+\varepsilon_i$ with independent $\varepsilon_i\sim\mathcal N(0,1)$.  In the orthonormal contrast coordinate,
$Z=(Y_1-Y_2)/\sqrt2\sim\mathcal N(\mu,1)$, where $\mu=(\theta_1-\theta_2)/\sqrt2$.
Since $C=1/2$ and $\widehat B_{\mathrm{raw}}=Z^2/2$, the plug-in has contrast estimate
\begin{equation}
 h(Z)=\begin{cases}0,&|Z|\le1,\\ Z-1/Z,&|Z|>1.\end{cases}
 \label{eq:plugin-contrast}
\end{equation}
The common-mean coordinate is unchanged.  Put $u(z)=h(z)-z$; this function is continuous and piecewise differentiable.  Gaussian integration by parts, with the boundary terms cancelling at $\pm1$, gives
\begin{align}
 \E(h(Z)-\mu)^2-1
 &=\E\{u(Z)^2+2u'(Z)\}\\
 &=\E\left[(Z^2-2)\mathbf1_{\{|Z|\le1\}}
       +\frac{3}{Z^2}\mathbf1_{\{|Z|>1\}}\right].
 \label{eq:plugin-risk-difference}
\end{align}
For completeness, the integration identity follows from the normal density derivative $\phi'(z-\mu)=-(z-\mu)\phi(z-\mu)$, so $\E[(Z-\mu)u(Z)]=\E u'(Z)$; $u$ and its derivative are integrable here.  At $\mu=6$, Chebyshev's inequality gives $\Prb(4\le Z\le8)\ge3/4$, whereas the normal fourth moment gives $\Prb(|Z|\le1)\le3/5^4$.  Consequently,
\begin{equation}
 \E(h(Z)-6)^2-1
 \ge\frac{3}{64}\frac34-\frac{6}{625}
 =\frac{4089}{160000}>0.
 \label{eq:plugin-analytic-counterexample}
\end{equation}
The average risk of the two roles increases by half this contrast-risk difference.    Unbiased estimation of $B$ in \cref{prop:observed-dispersion} does not imply risk domination after a nonlinear same-data plug-in.

\subsection{Independent validation selects reuse strength with an oracle inequality}
Let $\Lambda\subset[0,1]$ be a finite grid.  Fit $\widehat f_\lambda$ using training data, and evaluate every fitted candidate on an independent weighted validation set with loss in $[0,1]$ and effective size $N_{\mathrm{val}}^{\mathrm{eff}}$.  Let $\widehat\lambda$ minimize validation loss.
Here the weighted validation convention is the one made explicit for finite candidates in \cref{thm:finite-architecture-selection}: conditional on the training information $\mathscr T$, validation records $i=1,\ldots,N$ are independent, and nonnegative weights $v_i$ summing to one are fixed before observing their outcomes. Write $\ell_i(\widehat f_\lambda)\in[0,1]$ for the loss on record $i$. Then $\widehat R(\widehat f_\lambda)=\sum_i v_i\ell_i(\widehat f_\lambda)$, $R(\widehat f_\lambda)=\sum_i v_i\E[\ell_i(\widehat f_\lambda)\mid\mathscr T]$, and $N_{\mathrm{val}}^{\mathrm{eff}}=(\sum_i v_i^2)^{-1}$. The selector minimizes this same weighted empirical risk. Nonidentically distributed records retain this declared weighted target; losses of different candidates on the same record need not be independent.
\begin{proposition}[Validation-selected partial reuse]
\label{prop:validation-lambda}
Conditional on the training sample, with probability at least $1-\delta$ over validation data,
\begin{equation}
 R(\widehat f_{\widehat\lambda})
 \le \min_{\lambda\in\Lambda}R(\widehat f_\lambda)
 +2\sqrt{\frac{\log(2|\Lambda|/\delta)}{2N_{\mathrm{val}}^{\mathrm{eff}}}}.
 \label{eq:app-validation-oracle}
\end{equation}
\end{proposition}
\begin{proof}
Weighted Hoeffding and a union bound give one event on which every validation risk is within the displayed radius of its population risk.  Apply the standard two-deviation ERM argument on that event.
\end{proof}
This result provides a statistically valid way to choose among fitted candidates for the stated bounded validation loss, without estimating $B$ parametrically.  The guarantee evaluates the already fitted candidates using independent validation data.  \Cref{thm:finite-architecture-selection} gives a separate finite-family guarantee under explicit tail assumptions, including a Gaussian quadratic-loss specialization.

\subsection{Classical same-data shrinkage in the contrast space}
\label{app:classical-plugin}
This calculation explains a dimension-specific exception to the preceding warning; it is an application of classical Baranchik--Stein shrinkage \citep{baranchik1970family,stein1981estimation}.  A directly comparable risk identity and Baranchik's sufficient conditions are reproduced in \citet[Section~1, equations~(1.5)--(1.6)]{maruyama2023nonminimaxity}.  Shrinkage toward a fixed subspace, a common mean, or data-selected cluster attractors is also established background \citep{srinath2018cluster}; the calculation here fixes the common-mean subspace.

Retain the balanced fixed-design model of \cref{thm:exact-frontier}, with $M\ge2$, $D=pd$, mutually Gaussian observation noise, and \emph{known} $q=\sigma^2/n>0$.  Stack the $M$ vectorized role estimates as $Y=\theta+\zeta\in\R^{MD}$, so $\zeta\sim\mathcal N(0,qI_{MD})$.  Define
\[
 P=\frac{\mathbf1\mathbf1^\top}{M}\otimes I_D,\qquad Q=I_{MD}-P,
 \qquad k=D(M-1).
\]
Here $P$ replaces every role vector by their common average, and $Q$ retains their differences.  The integer $k$ is the dimension of this \emph{contrast space}, not the number of selected experts.  Choose a fixed matrix $U\in\R^{MD\times k}$ whose columns are an orthonormal basis of that space: $U^\top U=I_k$, $UU^\top=Q$.  Then
\[
 Z=U^\top Y\sim\mathcal N(\mu,qI_k),\qquad \mu=U^\top\theta,
 \qquad M\widehat B_{\mathrm{raw}}=\|Z\|^2,\quad MC=kq.
\]
The same-data coefficient defined after \cref{prop:observed-dispersion} leaves $PY$ unchanged and replaces $Z$ by
\[
 h(Z)=\left(1-\frac{kq}{\|Z\|^2}\right)_+Z,\qquad h(0)=0,
 \quad (t)_+=\max\{t,0\}.
\]
Thus the fitted stacked vector is $PY+Uh(Z)$.  Its role-averaged excess prediction risk satisfies
\[
 \Risk_{\mathrm{plug}}-\Risk_{\mathrm{sep}}
 =\frac1M\bigl(\E\|h(Z)-\mu\|^2-kq\bigr)<0
 \qquad\text{for every fixed $\theta$ when $k\ge4$}.
\]

\begin{proof}[Direct verification of the classical specialization]
Let $u(z)=h(z)-z$ and $s=\|z\|^2$.  The divergence $\operatorname{div}u=\sum_{i=1}^k\partial u_i/\partial z_i$ is the sum of coordinate derivatives.  Away from the sphere $s=kq$,
\[
 \begin{array}{c|c|c}
 &s\le kq&s>kq\\ \hline
 u(z)&-z&-kqz/s\\
 \operatorname{div}u(z)&-k&-kq(k-2)/s\\
 \|u(z)\|^2&s&k^2q^2/s
 \end{array}
\]
The two formulas for $u$ agree on the sphere.  Both pieces have bounded first derivatives on their domains, so $u$ is weakly differentiable (its integration-by-parts derivative has no surface jump term).  It and its derivatives are Gaussian-integrable.  Coordinatewise Gaussian integration by parts gives
\[
 \E[(Z_i-\mu_i)u_i(Z)]=q\E[\partial_i u_i(Z)].
\]
Expanding $h(Z)-\mu=Z-\mu+u(Z)$ and summing therefore gives
\[
 \E\|h(Z)-\mu\|^2-kq
 =\E\left[(s-2kq)\mathbf1_{\{s\le kq\}}
       +\frac{k(4-k)q^2}{s}\mathbf1_{\{s>kq\}}\right].
\]
For $k\ge4$ the outside term is nonpositive, and the inside term is at most $-kq$.  Every nondegenerate Gaussian distribution assigns positive probability to $s\le kq$, so the expectation is strictly negative, also at $k=4$.  Finally $P\zeta$ and $U(h(Z)-\mu)$ lie in orthogonal subspaces for every realization.  The common-mean risk is unchanged, and division by $M$ converts the contrast-risk difference to the original role-average risk.
\end{proof}

\paragraph{Relation to classical shrinkage.}
After standardizing $X=Z/\sqrt q$, the rule is $(1-\phi(w)/w)X$ with $w=\|X\|^2$ and $\phi(w)=\min\{w,k\}$.  In the cited Stein risk identity, substitution gives $w-2k$ on $w\le k$ and $k(4-k)/w$ otherwise.  It also satisfies Baranchik's monotonicity and range conditions $0\le\phi(w)\le2(k-2)$ when $k\ge4$.    The usual positive-part James--Stein rule uses $(k-2)q$, not $kq$, and already dominates the unshrunk estimate for $k\ge3$ \citep{srinath2018cluster}.  That is a different estimator; the threshold four here is not a universal dimension barrier.

The strict improvement is pointwise in the fixed target, not a claimed target-uniform positive improvement amount.    The low-dimensional counterexample in \cref{eq:plugin-analytic-counterexample} remains valid.  The main non-minimaxity result of \citet{maruyama2023nonminimaxity} concerns rules whose shrinkage is exactly zero outside a finite ball; this is not the present rule, whose $\phi(w)$ stays equal to $k$.  Only their introductory classical risk identity is used here.

\section{End-to-End Route-Plan Certificates}
\label{app:end-to-end-proofs}

\paragraph{Overview.}
We track how aligned implementation error and functional routing error propagate to the endpoint. The certificate requires explicit role-to-rollout transfer conditions. A separate linear calculation retains error directions and cross-time covariance rather than replacing them by scalar sensitivity bounds.
Sections~\ref*{app:end-to-end-proofs}.1--\ref*{app:end-to-end-proofs}.2 establish propagation and the conditional certificate; Sections~\ref*{app:end-to-end-proofs}.3--\ref*{app:end-to-end-proofs}.5 address functional route cost, terminal decision margins, and the separate covariance identity.

\subsection{Pathwise propagation with random intended programs}
All random variables are defined on one coupled evaluation probability space.  The intended canonical program $\Gamma_j$ may depend on the reference history and exogenous randomness; it is never moved outside expectation.  The reference and learned recursions are
\begin{equation}
 S_{j+1}=T_{j,\Gamma_j}(S_j),
 \qquad
 \widehat S_{j+1}^m
 =\widehat T^m_{j,\widehat\Gamma_j^m}(\widehat S_j^m).
 \label{eq:app-plan-recursions}
\end{equation}
The implementation $\iota_j^m(\Gamma_j)$ is the model-$m$ executable plan aligned with the intended canonical program.  Assume $T_{j,\gamma}$ is $L_j$-Lipschitz on the reached region uniformly in admissible canonical $\gamma$.  Define $e_j^m=d(\widehat S_j^m,S_j)$.  Inserting the aligned learned transition and the reference map at the learned state gives, pathwise,
\begin{align}
 e_{j+1}^m
 &\le
 d\!\left(\widehat T^m_{j,\widehat\Gamma_j^m}(\widehat S_j^m),
            \widehat T^m_{j,\iota_j^m(\Gamma_j)}(\widehat S_j^m)\right)
 \nonumber\\
 &\quad+
 d\!\left(\widehat T^m_{j,\iota_j^m(\Gamma_j)}(\widehat S_j^m),
            T_{j,\Gamma_j}(\widehat S_j^m)\right)
 +d\!\left(T_{j,\Gamma_j}(\widehat S_j^m),T_{j,\Gamma_j}(S_j)\right)
 \nonumber\\
 &\le r_j^m+a_j^m+L_je_j^m.
 \label{eq:app-path-recursion}
\end{align}
Induction yields
\begin{equation}
 e_H^m
 \le \left(\prod_{r=0}^{H-1}L_r\right)e_0^m
 +\sum_{j=0}^{H-1}A_j(a_j^m+r_j^m),
 \qquad
 A_j=\prod_{r=j+1}^{H-1}L_r.
 \label{eq:app-path-bound}
\end{equation}
No independence between route errors and states is used.  The intended program is coupled to the reference trajectory; the theorem bounds imitation of that coupled reference computation rather than regret against a state-dependent policy recomputed at the learned state.  If sensitivities are random rather than deterministic upper bounds, their products remain inside expectation unless additional moment assumptions justify factorization.

\subsection{Expanded terminal certificate and probability bookkeeping}
\label{app:terminal-certificate-full}
The compact statement of \cref{thm:terminal-certificate} uses the following
complete active-support convention:
\begin{equation}
\begin{aligned}
 w_a&=\sum_{j=0}^{H-1}A_j\omega_{j,a},
 &\cJ_e^+&=\{a\in\cJ_e:w_a>0\},\\
 W_e&=\sum_{a\in\cJ_e^+}w_a,
 &\pi_{e,a}&=\frac{w_a}{W_e}\quad(a\in\cJ_e^+).
\end{aligned}
\label{eq:performance-weights}
\end{equation}
Here $w_a$ is role $a$'s total terminal weight, $\cJ_e^+$ retains its positive-weight roles, $W_e$ is their group total, and $\pi_{e,a}$ gives their relative importance.  Only groups with $W_e>0$ enter the certificate.  For notational economy, $R_{\mathrm{sep},e}^*$, $\BReuse_e$, and the two radii below denote the quantities from \cref{sec:local-theory} recomputed on $\cJ_e^+$ with weights $\pi_{e,a}$, and the empirical objectives use the same active support.  A zero-weight role can be omitted from this comparison; if its data nevertheless influence a shared fitted expert, that influence must be represented by a positive training/certificate weight or by an additional algorithmic-deviation term.  Thus these weights incorporate downstream amplification.

\paragraph{Expanded conditions and bounds.}
Let $\delta\in(0,1)$, let the initial states agree, and let terminal utility (a scalar score) $\mathcal U:\cS\to\R$ be $L_{\mathcal U}$-Lipschitz.  On every active group $\cJ_e^+$ assume all hypotheses of \cref{thm:finite-class}: a fixed finite class, losses in $[0,1]$, at least one observation per active role, and mutually independent role observations under the stated sampling model.  The classes, losses, distributions, and performance weights are fixed independently of these observations.  For each group $e$ with $W_e>0$, choose $\delta_e>0$ with $\sum_e\delta_e\le\delta$, and define $\epsilon_{\mathrm{sh},e}(\delta_e)$ and $\epsilon_{\mathrm{sep},e}(\delta_e)$ by \cref{eq:finite-radius-shared} and \cref{eq:finite-radii}, respectively, with $\delta$ replaced by $\delta_e$.  Assume the group estimators minimize the performance-weighted empirical risks induced by \cref{eq:performance-weights}.  The two comparators are fitted from the same role samples; if they are refitted on independent samples, the confidence allocation must also cover both fitted datasets.  Under \cref{ass:role-aggregation}, with probability at least $1-\delta$ over the expert-estimation samples,
\begin{align}
 \E_{\mathrm{eval}}\abs{\mathcal U(\widehat S_H^{\mathrm{reuse}})-\mathcal U(S_H)}&\le L_{\mathcal U} U_{\mathrm{reuse}},\\
 U_{\mathrm{reuse}}
 &=\sum_jA_j(\tau_j^{\mathrm{reuse}}+\rho_j^{\mathrm{reuse}})
 +\sum_eW_e\bigl(R_{\mathrm{sep},e}^*+\BReuse_e
                  +2\epsilon_{\mathrm{sh},e}(\delta_e)\bigr),
 \label{eq:reuse-certificate}\\
 \E_{\mathrm{eval}}\abs{\mathcal U(\widehat S_H^{\mathrm{sep}})-\mathcal U(S_H)}&\le L_{\mathcal U} U_{\mathrm{sep}},\\
 U_{\mathrm{sep}}
 &=\sum_jA_j(\tau_j^{\mathrm{sep}}+\rho_j^{\mathrm{sep}})
 +\sum_eW_e\bigl(R_{\mathrm{sep},e}^*
                  +2\epsilon_{\mathrm{sep},e}(\delta_e)\bigr).
 \label{eq:sep-certificate}
\end{align}

The symbols $\epsilon_{\mathrm{reuse},e}$ and $\epsilon_{\mathrm{sep},e}$ in
\cref{eq:terminal-certificate-compact} are respectively the shared and separate
radii at the same group budget $\delta_e$.  Its indicator selects the reuse
bias only for the shared model, so substituting either value of $m$ gives
exactly the two expanded bounds above.  The common oracle cancels in
\cref{eq:prg-corrected} only because both comparators use these same groups,
weights, losses, and evaluation populations.  The utility-bound difference is
$L_{\mathcal U}\PRG_H$; strict ordering requires $L_{\mathcal U}>0$.
For $L_{\mathcal U}=0$ both utility bounds vanish.

\paragraph{Proof of \cref{thm:terminal-certificate}.}\label{proof:terminal-certificate}
For each active group choose $\delta_e>0$ with $\sum_e\delta_e\le\delta$ and condition on the simultaneous training event obtained by applying \cref{thm:finite-class} with failure probability $\delta_e$ to group $e$.  Under \cref{ass:role-aggregation}, common initial states, and $\rho_j^m=\E r_j^m$,
\begin{equation}
 \E e_H^m
 \le \sum_jA_j(\tau_j^m+\rho_j^m)
 +\sum_a\left(\sum_jA_j\omega_{j,a}\right)
 R_a\!\left(\widehat f^m_{\alpha_m(a)}\right).
 \label{eq:app-aggregate-roles}
\end{equation}
The coefficient of role $a$ is $w_a$.  Restrict each reuse group to its active support $\cJ_e^+$ from \cref{eq:performance-weights}; then writing $w_a=W_e\pi_{e,a}$ gives
\begin{equation}
 \sum_{a\in\cJ_e^+}w_aR_a(f)
 =W_e\sum_{a\in\cJ_e^+}\pi_{e,a}R_a(f).
 \label{eq:app-group-weight-factor}
\end{equation}
Because the estimators use these fixed performance weights, the shared part of \cref{thm:finite-class} gives the reuse term in \cref{eq:reuse-certificate}; its role-specific part gives \cref{eq:sep-certificate}.  Lipschitz continuity of terminal utility multiplies the state bound by $L_{\mathcal U}$.  Subtracting the two displayed right-hand sides cancels only the genuinely common oracle terms and yields $L_{\mathcal U}\PRG_H$.

If comparator $m$ requires model-specific weights $\omega_{j,a}^{m}$, define $w_a^m=\sum_jA_j\omega_{j,a}^m$ and construct each certificate separately.  Both remain valid, but the oracle and estimation terms generally have different coefficients and the compact expression \cref{eq:prg-corrected} is unavailable.  This is the appropriate treatment when rollout occupancies differ and no common domination envelope is credible.  Likewise, if event lifts or the surrounding assembly are learned from the expert-estimation data, the finite-sample event must be replaced by a uniform bound for the joint class or by sample splitting.

\subsection{Functional route cost}
For a finite canonical plan space and a finite model-specific executable plan space, define
\begin{align}
 C_j^m(\gamma,\gamma')
 &=\Prb(\Gamma_j=\gamma,\widehat\Gamma_j^m=\gamma'),\\
 D_j^m(\gamma,\gamma')
 &=\E\!\left[
 d\!\left(\widehat T^m_{j,\gamma'}(\widehat S_j^m),
          \widehat T^m_{j,\iota_j^m(\gamma)}(\widehat S_j^m)\right)
 \mid \Gamma_j=\gamma,\widehat\Gamma_j^m=\gamma'
 \right],
 \label{eq:app-plan-confusion}
\end{align}
with $D_j^m(\gamma,\gamma')=0$ when the conditioning event has zero probability.  Then
\begin{equation}
 \rho_j^m
 =\sum_{\gamma,\gamma'}
   C_j^m(\gamma,\gamma')D_j^m(\gamma,\gamma').
 \label{eq:app-functional-plan-cost}
\end{equation}
This quantity distinguishes harmless index substitutions from rare but severe functional confusions.  Index accuracy records only the off-alignment mass and is meaningful only after the canonical-to-model alignment has been specified.

For an additive mixture, a direct gate bound is available.
\begin{lemma}[Equal-mass functional gate discrepancy]
\label{lem:gate-discrepancy-revised}
Let $v_1,\ldots,v_K$ lie in a normed vector space and let $g,g'$ have equal total mass.  Then
\begin{equation}
 \norm{\sum_e(g_e-g'_e)v_e}
 \le \frac12\left(\max_{e,e'}\norm{v_e-v_{e'}}\right)\norm{g-g'}_1.
 \label{eq:app-gate-diameter}
\end{equation}
The constant $1/2$ is sharp.
\end{lemma}
\begin{proof}
The positive and negative parts of $g-g'$ have common mass $\norm{g-g'}_1/2$.  Normalize them to two convex combinations of the $v_e$; their distance is at most the diameter.  Equality is attained by two points and disjoint one-hot gates.
\end{proof}
The lemma applies at one common expert input and with the same expert-output collection.  Different states or dispatch-dependent expert inputs require additional state or dispatch sensitivity terms.

\subsection{Terminal decisions with a margin}
Discrete accuracy is not Lipschitz.  It can nevertheless be controlled under a terminal margin when the same deterministic readout $g$ is applied to the reference and learned terminal states.  Let $z=g(S_H)$, suppose $g$ is $L_g$-Lipschitz from $d$ to $\ell_\infty$, and let the reference top-two gap be $\gamma>0$.  For $L_g>0$, the learned argmax is unchanged whenever the pathwise right-hand side of \cref{eq:app-path-bound} is strictly less than $\gamma/(2L_g)$; for $L_g=0$, the readout is constant on the compared region and the conclusion is immediate.  If the learned and reference readout heads differ, an additional common-state readout discrepancy $\norm{\widehat g(\widehat S_H)-g(\widehat S_H)}_\infty$ must also be below the remaining margin budget.  These pathwise robustness statements yield expected-accuracy control when combined with a suitable tail bound.

\subsection{Exact directional amplification}
\begin{proposition}[Exact propagation of correlated errors]
\label{prop:gramian-amplification}
Let $\mathbf e_{j+1}=M_j\mathbf e_j+\xi_j$ in $\R^s$, with $\mathbf e_0=0$, deterministic $s\times s$ matrices $M_j$, and square-integrable errors $\xi_j$.  Define $b_j=\E\xi_j$ and cross-covariances $\Sigma_{jk}=\E[(\xi_j-b_j)(\xi_k-b_k)^\top]$.  With $P_j=M_{H-1}\cdots M_{j+1}$ and $P_{H-1}=I_s$,
\begin{equation}
 \E\|\mathbf e_H\|_2^2
 =\left\|\sum_jP_jb_j\right\|_2^2
 +\sum_{j,k}\tr(P_j\Sigma_{jk}P_k^\top).
 \label{eq:gramian-exact}
\end{equation}
Here $\tr$ sums diagonal matrix entries.  For uncorrelated centered errors only $j=k$ terms remain, weighted by $P_j^\top P_j$, the \emph{Gramian} measuring directional amplification.  Reusing one fitted expert generally creates nonzero cross-time terms, so that special case cannot be assumed from independent training examples.
\end{proposition}

\paragraph{Proof of \cref{prop:gramian-amplification}.}\label{proof:gramian-amplification}
Unrolling $\mathbf e_{j+1}=M_j\mathbf e_j+\xi_j$ gives
\begin{equation}
 \mathbf e_H=\sum_{j=0}^{H-1}P_j\xi_j,
 \qquad P_j=M_{H-1}\cdots M_{j+1}.
 \label{eq:app-linear-unroll}
\end{equation}
The mean is $\sum_jP_jb_j$.  Expanding the covariance without discarding cross-time terms gives
\begin{equation}
 \operatorname{Cov}(\mathbf e_H)
 =\sum_{j,k}P_j\Sigma_{jk}P_k^\top.
 \label{eq:app-linear-cov}
\end{equation}
Use $\E\|X\|_2^2=\|\E X\|_2^2+\tr(\operatorname{Cov}X)$ to obtain \cref{prop:gramian-amplification}.  This identity follows by writing $X=\E X+(X-\E X)$, expanding its squared norm, and observing that the centered cross term has zero expectation.  If all off-diagonal $\Sigma_{jk}$ vanish, the previous diagonal Gramian formula follows.  The double sum is nonnegative as a whole, although individual cross terms may be negative.  If $M_j$ is random and correlated with the errors, the deterministic covariance factorization is not valid: only the pathwise identity $\mathbf e_H=\sum_jP_j\xi_j$ is automatic.  \Cref{prop:shared-error-coherence} shows explicitly how tying parameters generates cross-time terms even in a simple additive model.

\section{Causal Route Actions and Proofs}
\label{app:causal-proofs}

\paragraph{Overview.}
We ask what a forced route actually changes, rather than what its natural label suggests. The counterexamples and risk decompositions distinguish observational agreement from executable reuse and show how an insufficient macro state obstructs prediction of intervention effects. Action coverage is treated separately in \cref{app:causal-audit}.
Sections~\ref*{app:causal-proofs}.1--\ref*{app:causal-proofs}.3 define actions and prove the observational and state-aliasing obstructions; Sections~\ref*{app:causal-proofs}.4--\ref*{app:causal-proofs}.5 cover rollout, reporting requirements, and the average-risk version of the obstruction.

\subsection{Non-anticipating route actions}
A complete micro context $C\in\cC$ contains the prompt, generated prefix, caches, hidden states, router state, and all randomness revealed up to the intervention boundary.  A fixed macro extractor $\Psi:\cC\to\cS$ produces $S=\Psi(C)$.  A route action $\mathcal A_i=(\varpi_i,\sigma_i)$ consists of a non-anticipating policy over future route decisions and a stopping time $1\le\sigma_i\le B_{\mathrm{act}}$.  Every decision made by $\varpi_i$ is measurable with respect to the information available immediately before that decision; neither the policy nor the stopping rule may inspect unrevealed future tokens or a completed post-hoc route trace.

For pathwise statements, execution is greedy or is conditioned on a common realization of exogenous randomness.  Executing $\mathcal A_i$ then yields a deterministic micro transition $\mathfrak T_i(C)$ and macro result $T_i(C)=\Psi(\mathfrak T_i(C))$.  Under stochastic generation, the corresponding object is a transition kernel; the deterministic claims below then apply conditionally, while the Wasserstein formulation in \cref{app:extensions-ledger} controls the induced laws.

For a candidate macro operator $F_i$, fixed before the evaluation sample is observed, the off-policy defect under a fixed probe distribution $\nu_i$ is
\begin{equation}
 \mu_i^{\mathrm{off}}
 =\E_{C\sim\nu_i}d\!\left(T_i(C),F_i(\Psi(C))\right).
 \label{eq:app-off-policy-defect}
\end{equation}
If a controller executes actions on-policy, the corresponding occupancy changes with the action library and controller; an on-policy defect must therefore be evaluated under that induced distribution.  For $K\ge2$, a small defect is meaningful only together with non-collapse between actions, for example
\begin{equation}
 \operatorname{Sep}(\nu)
 =\E_{C\sim\nu}\frac{2}{K(K-1)}
   \sum_{i<i'}d\!\left(T_i(C),T_{i'}(C)\right).
 \label{eq:app-action-separation}
\end{equation}
If $\Psi$, the actions, or the operator class are selected using the same records on which the defect is reported, sample splitting, cross-fitting, or a uniform complexity correction is required.

\subsection{Proof that observational alignment is insufficient}
\begin{proof}[Proof of \cref{prop:route-label-separation}]
Construct a residual routed block whose shared path receives $(S,I)$ and returns $F_I(S)$.  Set every expert contribution identically to zero, and let the natural router output index $I$.  The model output is exactly $F_I(S)$ and natural route-label accuracy is one.  Under an intervention replacing only the routed index by $J$, the shared path is unchanged and every expert contribution remains zero, so the output is still $F_I(S)$ and is independent of $J$.  Forced-route fidelity to the target associated with $J$ is therefore exactly $\Prb(F_I(S)=F_J(S))$.  If $J$ is uniform and independent of $(I,S)$ and the $K$ values $F_1(S),\ldots,F_K(S)$ are pairwise distinct at every evaluated state, equality occurs if and only if $J=I$, giving fidelity $1/K$.
\end{proof}
For comparison, a second model with zero shared contribution and experts $E_i(S)=F_i(S)$ has the same natural outputs and the same labels under route $I$, but a forced route $J$ returns $F_J(S)$.  These two models are observationally equivalent on task outputs and natural route labels and differ under intervention.  The construction is compatible with perfect task performance.  It establishes a non-identifiability result: behavioral and route-label observations alone do not determine where the operation is implemented.

\subsection{State-aliasing obstruction and proof}
\label{app:aliasing-full}
Let $\cC_0$ be a nonempty set of contexts.  Two contexts are \emph{aliased} when they have the same extracted state.  Define the largest difference between their responses to action $i$ by
\begin{equation}
 \Xi_i(\Psi;\cC_0)=
 \sup_{\substack{C,C'\in\cC_0,\ C\ne C'\\\Psi(C)=\Psi(C')}}
 d\!\left(T_i(C),T_i(C')\right),
 \label{eq:aliasing-diameter}
\end{equation}
with value zero when no distinct aliased pair exists.  Thus $\Xi_i$ is the largest possible action-effect discrepancy among contexts that the macro representation cannot distinguish; the supremum need not be attained.
\begin{proposition}[State-aliasing obstruction]
\label{prop:aliasing}
Every deterministic $F_i$ satisfies
\begin{equation}
 \sup_{C\in\cC_0}d\!\left(T_i(C),F_i(\Psi(C))\right)
 \ge \frac12\Xi_i(\Psi;\cC_0).
 \label{eq:aliasing-lower-bound}
\end{equation}
\end{proposition}
\begin{proof}[Proof of \cref{prop:aliasing}]
For any distinct $C,C'$ with $\Psi(C)=\Psi(C')=s$, the triangle inequality gives
\begin{equation}
 d(T_i(C),T_i(C'))
 \le d(T_i(C),F_i(s))+d(F_i(s),T_i(C')).
 \label{eq:app-alias-triangle}
\end{equation}
At least one right-hand term is at least half the left-hand side.  Taking the supremum over feasible distinct aliased pairs proves the claim.  If no such pair exists, the defined diameter is zero and the inequality is trivial.
\end{proof}
This is a worst-case obstruction; rare aliased contexts need not prevent small average defect.  For squared loss, \cref{prop:average-aliasing} instead characterizes the minimum average defect exactly by conditional variance.  Non-collapse, pre-action selectability, and on-policy composition remain additional requirements; definitions and rollout bounds are in \cref{app:causal-proofs}.

\subsection{Causal rollout}
Let $I_j$ be the executed action and let $I_j^\star$ be a fixed or externally coupled reference action sequence on the same probability space.  The reference sequence is held fixed under the coupling; comparing two state-dependent controllers requires an additional policy-stability or selection-mismatch term.  Define
\begin{align}
 \epsilon_j^{\mathrm{abs}}
 &=d\!\left(T_{I_j}(C_j),F_{I_j}(S_j)\right),\\
 \chi_j^{\mathrm{sel}}
 &=d\!\left(F_{I_j}(S_j),F_{I_j^\star}(S_j)\right),
 \label{eq:app-causal-local-errors}
\end{align}
where $S_j=\Psi(C_j)$.  Let $\bar S_0=S_0$ and $\bar S_{j+1}=F_{I_j^\star}(\bar S_j)$, and assume $F_{I_j^\star}$ is $L_j$-Lipschitz on a region containing both trajectories.  Inserting $F_{I_j}(S_j)$ and $F_{I_j^\star}(S_j)$ yields
\begin{equation}
 d(S_{j+1},\bar S_{j+1})
 \le \epsilon_j^{\mathrm{abs}}+\chi_j^{\mathrm{sel}}
 +L_jd(S_j,\bar S_j).
 \label{eq:app-causal-recursion}
\end{equation}
Induction gives
\begin{equation}
 d(S_H,\bar S_H)
 \le \sum_{j=0}^{H-1}
 \left(\prod_{r=j+1}^{H-1}L_r\right)
 (\epsilon_j^{\mathrm{abs}}+\chi_j^{\mathrm{sel}}).
 \label{eq:app-causal-rollout}
\end{equation}
This controls macro-transition mismatch for the coupled reference sequence.  Comparing state-feedback controllers additionally requires control of how the reference policy changes when its input state changes.

\subsection{An exact average-risk form of the aliasing obstruction}
The supremum in \cref{eq:aliasing-lower-bound} is sensitive to rare contexts.  An average-risk conclusion needs their probability, not just their maximum separation.  Fix an action $i$, a probe-context distribution, $S=\Psi(C)$ in a standard Borel space, and a square-integrable actual action outcome $Y_i=T_i(C)$ in a separable real Hilbert space $\cH$.  A stochastic outcome can be included as long as its additional randomness is also averaged.  Define $m_i(S)=\E[Y_i\mid S]$.
\begin{proposition}[Average state-compression obstruction]
\label{prop:average-aliasing}
Among all measurable predictors $F$ with finite second moment,
\begin{equation}
 \inf_F\E\|Y_i-F(S)\|_{\cH}^2
 =\E\|Y_i-m_i(S)\|_{\cH}^2
 =\frac12\E\|Y_i-Y_i'\|_{\cH}^2,
 \label{eq:average-aliasing-identity}
\end{equation}
where, conditional on $S$, $Y_i$ and $Y_i'$ are independent draws from the same conditional action-outcome law.  The optimum is zero exactly when $Y_i=m_i(S)$ almost surely.
\end{proposition}
\begin{proof}
Insert $m_i(S)$ into $Y_i-F(S)$.  Conditional on $S$, the residual $Y_i-m_i(S)$ has mean zero, so its inner product with $m_i(S)-F(S)$ has zero expectation.  The risk equals $\E\|Y_i-m_i(S)\|^2+\E\|m_i(S)-F(S)\|^2$.  The first term is attained by $F=m_i$.  Conditional independence makes the inner product of $Y_i-m_i(S)$ and $Y_i'-m_i(S)$ have zero conditional expectation; expanding their squared difference gives twice the first term.  A nonnegative squared norm has zero mean exactly when it vanishes almost surely.
\end{proof}
For deterministic actions this residual quantifies action-relevant information lost by the state extractor.  With stochastic outcomes it also includes irreducible execution noise; the theorem does not label all conditional variance as compression error.  Restricting $F$ to a neural class adds its approximation error and cannot improve the unrestricted infimum.  Unlike the worst-case diameter, \cref{eq:average-aliasing-identity} weights rare contexts by their probability.

\FloatBarrier
\section{Controlled Experiments: Methods and Results}
\label{app:experiment-details}

\paragraph{Overview.}
This appendix describes three complementary studies: controlled estimators test quantitative risk predictions, trained Transformers compare one-step and final-output performance, and interventions examine the effects attributed to expert calls. The primary experiment family and later extensions retain separate evaluation and confidence scopes. Sections~\ref*{app:experiment-details}.1--\ref*{app:experiment-details}.6 specify the task, models, controlled estimation, trained comparisons, interventions, routing diagnostics, and paired performance evaluation. The remaining sections cover data budgets, transition holdouts, state-space size, state-conditioned variability, and general feature geometry.

\subsection{Shared task, models, and evaluation protocol}
\label{app:new-task}
\label{app:exp-provenance}
The primary experiment family comprises fifteen trained models: three sharing architectures under the task-and-balance objective, and two additional supervision objectives on the tied architecture, each with three training seeds. The tied task-and-balance models are reused as the intervention baseline. Data-budget and generalization studies are described separately with their own settings and evaluation scope.

\paragraph{Task and splits.}
The task composes four operations on a finite state space,
\begin{equation*}
 F_0(s)=s+1,\quad F_1(s)=s+7,\quad F_2(s)=2s,\quad F_3(s)=3s+5\pmod{31},
\end{equation*}
and the stream $\langle\mathrm{bos}\rangle,s_0,z_1,s_1,\ldots,z_H,s_H$, where $s_t=F_{z_t}(s_{t-1})$, and H is the number of operation application. The operation program is externally supplied, not generated by the model. A 37-symbol input vocabulary encodes state $s$ as token $s+2$ and operation $z$ as token $z+33$; a 31-way state head predicts the next state at operation-token positions. Training uses reference prefixes (teacher forcing), whereas free-running evaluation uses greedy state predictions as the subsequent inputs. The first four operations define a family; the protocol hashes their canonical bytes to assign 209 training, 22 development, and 25 test families. All initial states and continuations of one family stay together. Family assignment conditions on the entire four-operation prefix; it does not make those four coordinates independently uniform within a split. This withholds program-prefix families, not every local state--action transition. The strict transition holdout in \cref{app:exp-holdout} is a different training and evaluation protocol.

 The eight artificial estimation roles, three scalar composition roles, and four symbolic task operations belong to distinct experimental settings.

The primary pool contains 16,384 independently sampled trace records, equally split between training horizons 4 and 8; repeated contents are allowed under this sampling law. Three training seeds share this pool. Within a seed, compared models use the same horizon and minibatch-index schedule. Throughout this experimental appendix, H4, H8, H12, and H16 denote operation application number $H=4,8,12,16$, respectively. H8 is the primary free-running evaluation. H4 enumerates all 775 held-out-family traces; H12/H16 are descriptive length-extrapolation tests.

\paragraph{Models and fitting.}
All primary models have four pre-normalized causal Transformer layers, hidden width 256, four attention heads, sinusoidal positions, SwiGLU FFNs of inner width 768, no dropout, and no capacity dropping. Local has four distinct experts per layer; tied reuses one bank of four experts across depth with layer-specific routers and normalization; dense has one unrouted, active-width-matched FFN per layer. Top-1 routing retains the selected softmax gain. Local and tied initialize expert values equally but differ in parameter-object identity. Dense is an active-width control, not an identical initial function.

The training settings below distinguish the fifteen primary architecture/objective runs from the smaller model trained solely to supply frozen features. Follow-up experiments use the changes specified in their respective sections.
\begin{center}
\begin{tabular}{@{}lll@{}}
\toprule
Setting & Primary runs & Feature pilot\\
\midrule
Layers / hidden width / heads & $4/256/4$ & $2/128/4$\\
Routed experts per bank / FFN inner width & $4/768$ & $4/384$\\
Batch size / training records & $64/16{,}384$ & $64/2{,}048$\\
Completed updates / trace presentations & $3{,}000/192{,}000$ & $300/19{,}200$\\
Optimizer / peak learning rate & AdamW / $3\times10^{-4}$ & AdamW / $3\times10^{-4}$\\
Betas / optimizer epsilon & $(0.9,0.95)/10^{-8}$ & $(0.9,0.95)/10^{-8}$\\
Weight decay / gradient-norm clip & $0.01/1$ & $0.01/1$\\
Linear warmup / subsequent schedule & $150$ steps / cosine & $15$ steps / cosine\\
Training autocast / parameters and state & BF16 / FP32 & BF16 / FP32\\
\bottomrule
\end{tabular}
\end{center}
Evaluation uses FP32. Trace presentations count repeated uses of the training pool. Training and neural-model evaluation use NVIDIA RTX A5000 24\,GB GPUs, with one card for the primary experiments and a second for the follow-ups. Controlled linear calculations use float64. Training time, memory, unique parameters, and dispatched computation are reported separately.

\paragraph{Checkpoint selection and development data.}
The primary comparisons use all fifteen checkpoints at the final training update, 3,000. Development monitoring uses a fixed 512-trace set, evaluated every 500 updates. The coefficients in \cref{app:new-causal} specify the three training objectives. Candidate checkpoints, action mappings, evaluation losses, comparison families, and sample sizes are fixed before generating the primary evaluation banks. The mappings are selected using development data as described in \cref{app:new-causal}. Follow-up studies retain their own protocols and are not included retrospectively in the primary comparison family.

\paragraph{Outcomes and independent units.}
Terminal error is the indicator of an incorrect final state; complete-trace error indicates any incorrect predicted state along a free-running trace. A trace can recover from an intermediate error, so these are different endpoints. Teacher-forced step error averages state-prediction mistakes on reference prefixes; it is not the event-lift role loss of \cref{eq:role-risk} and is not a substitute for free-running performance. A complete trace is the independent evaluation unit. Multiple forced actions on one context are averaged before across-context uncertainty is calculated. Across-seed standard deviations describe training variability, not additional audit samples.

\subsection{Controlled local estimation and composition}
\label{app:new-frozen-frontier}
\paragraph{Local estimation gains under controlled sharing.}
Frozen, nonorthogonal Transformer features define a linear residual-estimation problem. All 54 estimation settings agree with their covariance-aware references within five Monte Carlo standard errors; two repetitions on trained features satisfy the same numerical agreement criterion (\cref{app:new-frozen-frontier}). 

\paragraph{Frozen Transformer feature interface.}
The frozen-feature estimation study uses a separately trained two-layer, width-128 Transformer after 300 updates. For each probe trace, one prediction position is selected uniformly. The feature is the normalized input to the expert subblock in the second Transformer layer, after its FFN pre-normalization. Features are extracted in evaluation mode without gradients in FP32, converted to float64, and mapped to 32 dimensions by a fixed random orthonormal $128\times32$ projection. No centering, whitening, or post-hoc direction selection is applied. A controlled residual call uses $T_a[A](z)=h(z)+JAu(z)$ with $J^\top J=I_{16}$, so the squared difference between complete event outputs equals $\|(A-A_a)u(z)\|^2$. Eight artificial statistical roles are defined independently of the symbolic operation, trace content, and layer.

Each role has 64 fixed nonorthogonal design rows. Repeating those rows one, four, or sixteen times, with fresh independent Gaussian label noise of standard deviation 0.1, gives $n=64,256,1024$ independent measurements. The input support does not increase with $n$. Test geometry is the empirical second moment of a separately frozen 8,192-record feature pool. Targets are held constant across the measurement-count sweep and calibrated to three heterogeneity levels $c=0,0.5,2$. Designs, targets, features, surrounding state, and test geometry are fixed before drawing fitting noise. The 32-dimensional test second moment has condition number approximately $1.95\times10^7$; the eight design singular-value ratios range from $1.20\times10^{-4}$ to $1.52\times10^{-4}$. The fitting procedure uses neither ridge regularization nor whitening.

More precisely, the calibration sets $B(c)=cC_{256}$, where $B(c)$ is target sharing bias in this fixed test geometry and $C_{256}$ is the estimation variance removable by pooling at $n=256$. The dimensionless $c$ therefore fixes target heterogeneity relative to one reference variance scale. Writing $V_0(n)$ for separate-estimation risk in this geometry, the reported eight-role construction has $C_n=(7/8)V_0(n)$. Repeated design rows make these variance quantities scale as $1/n$, whereas $B(c)$ stays fixed. Thus $c$ is a target-construction parameter.

Float64 role-wise least squares is solved by reduced QR and followed by the test-geometry projection of \cref{eq:test-weighted-common-operator}. The five fixed coefficients $0,0.25,0.5,0.75,1$ and the population oracle are evaluated on the same noisy observations within each of 128 independent repetitions. Observed risk is computed from the fitted matrices; the analytic reference is stored separately. All 54 estimation settings meet the numerical comparison criterion $|\mathrm{mean}-\mathrm{reference}|\le\max(5\,\mathrm{MCSE},10^{-8})$.Full-pooling ratios cross one between the sampled measurement counts in the nonzero-heterogeneity settings, consistent with the specified reference.

\begingroup
\small
\setlength{\tabcolsep}{4pt}
\begin{longtable}{@{}lrrrrr@{}}
\caption{Frozen-feature estimation endpoints from 128 fitting-noise repetitions. Parentheses are Monte Carlo standard errors (MCSE). Ratios divide the two means.}\label{tab:exp_t1a}\\
\toprule
$c$ & $n$ & Separate (MCSE) & Full (MCSE) & Reference ratio & Observed ratio \\
\midrule
\endfirsthead
\toprule
$c$ & $n$ & Separate (MCSE) & Full (MCSE) & Reference ratio & Observed ratio \\
\midrule
\endhead
0 & 64 & 0.5329 (0.0096) & 0.0664 (0.0012) & 0.1250 & 0.1246 \\
0 & 256 & 0.1355 (0.0022) & 0.0169 (0.0003) & 0.1250 & 0.1250 \\
0 & 1024 & 0.0331 (0.0005) & 0.0042 (0.0001) & 0.1250 & 0.1255 \\
0.5 & 64 & 0.5329 (0.0096) & 0.1240 (0.0012) & 0.2344 & 0.2327 \\
0.5 & 256 & 0.1355 (0.0022) & 0.0745 (0.0003) & 0.5625 & 0.5503 \\
0.5 & 1024 & 0.0331 (0.0005) & 0.0618 (0.0001) & 1.8750 & 1.8644 \\
2 & 64 & 0.5329 (0.0096) & 0.2968 (0.0012) & 0.5625 & 0.5571 \\
2 & 256 & 0.1355 (0.0022) & 0.2474 (0.0003) & 1.8750 & 1.8262 \\
2 & 1024 & 0.0331 (0.0005) & 0.2346 (0.0001) & 7.1250 & 7.0810 \\
\bottomrule
\end{longtable}
\endgroup

\begin{figure}[htbp]
\centering
\includegraphics[width=\linewidth]{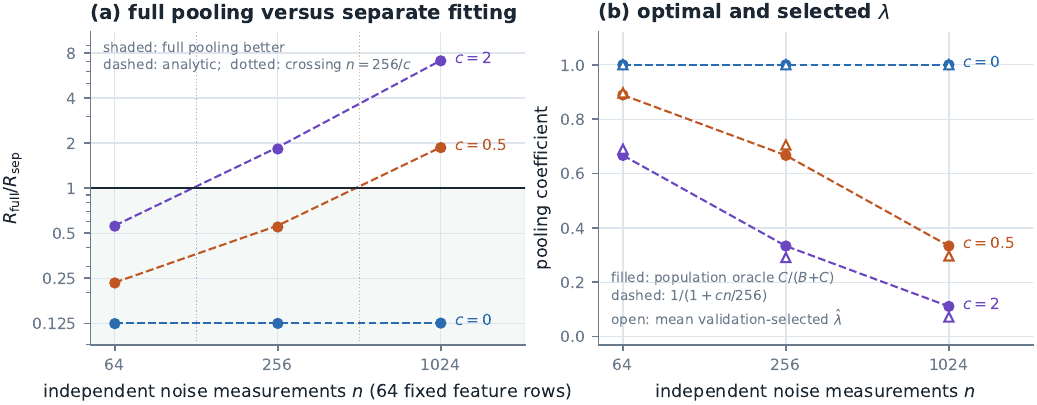}
\caption{Frozen-feature estimation under fixed target heterogeneity. Observed full/separate ratios and the independently selected coefficients are distinct from the analytic lines and population oracle. More measurements of the same feature rows reduce noise without removing fixed target heterogeneity.}
\label{fig:exp-t1a}
\end{figure}

A realizable selector uses 4,096 fresh noisy validation records per fit to choose among the five fixed coefficients. Its risk and coefficient frequencies are reported separately in \cref{tab:exp_t1a_selection}; the oracle target is not made available to selection. The squared Gaussian loss is unbounded, so the bounded primary audit radius is not applied to this estimation study. The trained-feature replication repeats this inserted-estimator design on features from two trained step-3000 checkpoints, with condition numbers about 3,002 and 2,625, respectively; both 54-setting repetitions satisfy the same numerical comparison criterion. 
\begingroup
\small
\setlength{\tabcolsep}{4pt}
\begin{longtable}{@{}rrrrl@{}}
\caption{Validation-selected pooling (4,096 fresh validation records per fit). Frequencies follow the grid 0, 0.25, 0.5, 0.75, 1 and sum to 128; conditional finite-grid excess is computed after selection.}\label{tab:exp_t1a_selection}\\
\toprule
$c$ & $n$ & Selected risk & Excess & Five selection counts \\
\midrule
\endfirsthead
\toprule
$c$ & $n$ & Selected risk & Excess & Five selection counts \\
\midrule
\endhead
0.0 & 64 & 0.0664 & 0.000000 & 0/0/0/0/128 \\
0.5 & 64 & 0.1241 & 0.000782 & 0/0/0/53/75 \\
2.0 & 64 & 0.2245 & 0.000638 & 0/0/31/97/0 \\
0.0 & 256 & 0.0169 & 0.000000 & 0/0/0/0/128 \\
0.5 & 256 & 0.0567 & 0.000169 & 0/0/23/105/0 \\
2.0 & 256 & 0.0976 & 0.000373 & 0/107/21/0/0 \\
0.0 & 1024 & 0.0042 & 0.000000 & 0/0/0/0/128 \\
0.5 & 1024 & 0.0240 & 0.000096 & 0/104/24/0/0 \\
2.0 & 1024 & 0.0329 & 0.000091 & 91/37/0/0/0 \\
\bottomrule
\end{longtable}
\endgroup

\begingroup
\small
\setlength{\tabcolsep}{4pt}
\begin{longtable}{@{}lrrr@{}}
\caption{Trained-feature replication repeats the frozen-estimator protocol using features from two completed step-3000 checkpoints. Counts use the five-MCSE numerical-agreement criterion.}\label{tab:exp_trained_features}\\
\toprule
Feature source & Condition number & Cells & Flagged \\
\midrule
\endfirsthead
\toprule
Feature source & Condition number & Cells & Flagged \\
\midrule
\endhead
trained-local-s0 & 3002 & 54 & 0 \\
trained-tied-s0 & 2625 & 54 & 0 \\
\bottomrule
\end{longtable}
\endgroup

\begin{figure}[htbp]
\centering
\includegraphics[width=\linewidth]{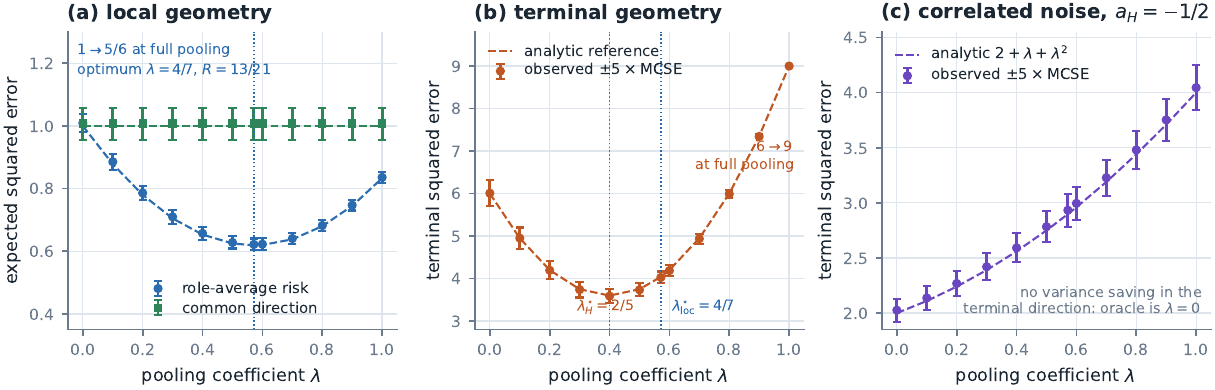}
\caption{\textbf{Controlled local-to-final risk reversal.} The same estimates are evaluated in local and final-output geometries. Markers are reported means with $\pm5\,\mathrm{MCSE}$ over 20,000 noise repetitions; dashed curves are analytic references. Full pooling helps locally but hurts the final output, and the optimal pooling strengths differ. The correlated-noise case isolates a further limit of pooling. Panels (a) and (b) use the same three-role estimates; (c) is a separate two-dimensional model. The common-direction normalization and both constructions are specified in \cref{app:new-frozen-frontier}.}
\label{fig:exp-terminal}
\end{figure}

\paragraph{Local improvements can reverse at the final output.}
The executed three-event construction of \cref{prop:local-terminal-reversal} evaluates the \emph{same fitted coefficients} locally and after composition. Full pooling lowers observed role-average squared error from 1.0090 to 0.8359, while final-output squared error rises from 6.0086 to 9.0000 (\cref{fig:exp-terminal}). The corresponding analytic optima also differ: $\lambda=4/7$ locally versus $2/5$ at the endpoint. Thus a local improvement can accompany worse final output in this controlled program.

\paragraph{Composition of the same estimates.}
The controlled composition study executes the existing three-event construction in \cref{prop:local-terminal-reversal}, with $\theta=(1/2,-1,1/2)^\top$, independent unit-variance coefficient noise, and common-mean pooling. The implementation evaluates both the actual recurrence and its fixed linear terminal map on 20,000 noise repetitions. The role-average, common-direction, and terminal geometries are distinct evaluations of the same estimates. Full pooling changes observed local risk from 1.0090 to 0.8359 and terminal risk from 6.0086 to 9.0000, against references $1\to5/6$ and $6\to9$. The terminal full-pooling output is constant, hence its MCSE is zero. 

In Figure~\ref{fig:exp-terminal}(a), the common-direction curve measures the squared scalar error projection
\[
 \E\left[\left(\frac{\mathbf1^\top(L_\lambda Y-\theta)}{\sqrt3}\right)^2\right],
 \qquad \mathbf1=(1,1,1)^\top.
\]
Its analytic value is 1 at every displayed pooling strength. This is a projection onto a unit direction, rather than role-average risk or the squared error of the arithmetic mean. Panels (a) and (b) evaluate the same three-role estimates. Panel (c) instead uses the separately declared two-dimensional model from Appendix~\ref{app:terminal-selection}.1:
\[
 \theta=0,\quad P=\operatorname{diag}(1,0),\quad G=(1,1),\quad
 \Sigma=\begin{pmatrix}4&-3/2\\-3/2&1\end{pmatrix}.
\]
In that model, suppressing a component removes beneficial noise cancellation, so positive pooling increases terminal risk.
\begingroup
\small
\setlength{\tabcolsep}{4pt}
\begin{longtable}{@{}lrrrr@{}}
\caption{Controlled composition: selected points; 20,000 noise repetitions for each of the two models. Reference and observed values are separate columns. The full-pooling terminal loss is constant, so its MCSE is zero.}\label{tab:exp_t1c}\\
\toprule
Model / geometry & $\lambda$ & Reference & Observed & MCSE \\
\midrule
\endfirsthead
\toprule
Model / geometry & $\lambda$ & Reference & Observed & MCSE \\
\midrule
\endhead
reversal/local & 0.0 & 1.0000 & 1.0090 & 0.00590 \\
reversal/local & 0.4 & 0.6533 & 0.6574 & 0.00406 \\
reversal/local & 1.0 & 0.8333 & 0.8359 & 0.00341 \\
reversal/common & 0.0 & 1.0000 & 1.0077 & 0.01022 \\
reversal/common & 0.4 & 1.0000 & 1.0077 & 0.01022 \\
reversal/common & 1.0 & 1.0000 & 1.0077 & 0.01022 \\
reversal/terminal & 0.0 & 6.0000 & 6.0086 & 0.06031 \\
reversal/terminal & 0.4 & 3.6000 & 3.5875 & 0.03341 \\
reversal/terminal & 1.0 & 9.0000 & 9.0000 & 0.00000 \\
correlated/terminal & 0.0 & 2.0000 & 2.0257 & 0.02060 \\
correlated/terminal & 0.4 & 2.5600 & 2.5917 & 0.02622 \\
correlated/terminal & 1.0 & 4.0000 & 4.0451 & 0.04064 \\
\bottomrule
\end{longtable}
\endgroup

\subsection{Shared, unshared, and dense Transformers}
\label{app:new-trained-reuse}
Nine primary runs compare the three architectures at three seeds, all at the full budget. Tied has 3,433,984 distinct parameters versus 10,511,872 for local, a 67.3\% reduction; dense has 3,429,888. The reported H8 audit contains one erroneous trace across the nine checkpoints, produced by local seed 1. All twelve architecture intervals cover zero. The tested task is saturated at this budget. Matching active FFN width does not establish equal total compute or elapsed time. The per-seed resource quantities are reported separately below.
\begingroup
\small
\setlength{\tabcolsep}{4pt}
\begin{longtable}{@{}lrrrrrr@{}}
\caption{Shared, unshared, and dense Transformers at training update 3000. Errors are percentages on 32,768 H8 traces per checkpoint; training time excludes I/O. Parameter counts are millions; peak is allocated MiB.}\label{tab:exp_t2}\\
\toprule
Model & Seed & Params (M) & Terminal (\%) & Trace (\%) & Step (ms) & Peak \\
\midrule
\endfirsthead
\toprule
Model & Seed & Params (M) & Terminal (\%) & Trace (\%) & Step (ms) & Peak \\
\midrule
\endhead
local & 0 & 10.5119 & 0.0000 & 0.0000 & 31.7 & 278.3 \\
local & 1 & 10.5119 & 0.0031 & 0.0031 & 32.5 & 279.0 \\
local & 2 & 10.5119 & 0.0000 & 0.0000 & 32.5 & 279.3 \\
tied & 0 & 3.4340 & 0.0000 & 0.0000 & 29.9 & 184.8 \\
tied & 1 & 3.4340 & 0.0000 & 0.0000 & 30.5 & 184.5 \\
tied & 2 & 3.4340 & 0.0000 & 0.0000 & 30.2 & 185.1 \\
dense & 0 & 3.4299 & 0.0000 & 0.0000 & 12.7 & 172.1 \\
dense & 1 & 3.4299 & 0.0000 & 0.0000 & 12.4 & 172.1 \\
dense & 2 & 3.4299 & 0.0000 & 0.0000 & 12.9 & 172.1 \\
\bottomrule
\end{longtable}
\endgroup

H4 is also solved on its full finite test population. At H12/H16, terminal error ranges approximately from 0.964 to 0.972 and complete-trace error is at least 0.999 in the supplied primary records. The uniform 31-state guessing error is a scale reference. These trained systems do not demonstrate successful length extrapolation under the reported settings. 
\begingroup
\small
\setlength{\tabcolsep}{4pt}
\begin{longtable}{@{}lrrrrrrr@{}}
\caption{Primary horizon diagnostics (percent error): H4 enumerates all 775 family-held-out traces; H12/H16 use 4,096 traces per checkpoint. Each horizon lists terminal / complete-trace error. }\label{tab:exp_horizons}\\
\toprule
Model & Seed & H4 term. & H4 trace & H12 term. & H12 trace & H16 term. & H16 trace \\
\midrule
\endfirsthead
\toprule
Model & Seed & H4 term. & H4 trace & H12 term. & H12 trace & H16 term. & H16 trace \\
\midrule
\endhead
local & 0 & 0.000 & 0.000 & 96.826 & 99.951 & 96.973 & 100.000 \\
local & 1 & 0.000 & 0.000 & 96.631 & 99.927 & 96.802 & 100.000 \\
local & 2 & 0.000 & 0.000 & 97.046 & 100.000 & 97.119 & 100.000 \\
tied & 0 & 0.000 & 0.000 & 96.899 & 100.000 & 96.655 & 100.000 \\
tied & 1 & 0.000 & 0.000 & 97.046 & 99.902 & 96.851 & 100.000 \\
tied & 2 & 0.000 & 0.000 & 97.192 & 100.000 & 97.070 & 100.000 \\
dense & 0 & 0.000 & 0.000 & 96.387 & 100.000 & 96.704 & 100.000 \\
dense & 1 & 0.000 & 0.000 & 97.095 & 100.000 & 96.753 & 100.000 \\
dense & 2 & 0.000 & 0.000 & 97.070 & 100.000 & 97.217 & 100.000 \\
\bottomrule
\end{longtable}
\endgroup

\subsection{Route labels and executable interventions}
\label{app:new-causal}
\paragraph{Objectives and interventions.}
The three training objectives are cumulative:
\[
\begin{aligned}
 \mathcal L_B&=\mathcal L_{\mathrm{task}}+0.01\mathcal L_{\mathrm{bal}},\\
 \mathcal L_R&=\mathcal L_B+0.1\mathcal L_{\mathrm{route}},\\
 \mathcal L_I&=\mathcal L_R+0.5\mathcal L_{\mathrm{CI}}.
\end{aligned}
\]
The task loss is cross-entropy for the next state. The balancing term is computed over all input tokens, while route supervision is cross-entropy between natural per-layer router logits and the requested operation at state-prediction positions. The natural token/update budget is matched. I also receives a counterfactual forward pass and extra target per training trace, so neither its total computation nor its supervision is matched to R. B is the tied baseline from the architecture comparison, reused rather than retrained: the primary family has fifteen unique runs.

The intervention loss independently draws $t$ uniformly from $\{1,\ldots,H\}$ and operation $i$ uniformly from $\{0,1,2,3\}$ for each training trace. At the corresponding state-prediction position in every primary-model layer, it forces expert $i$ with that layer's detached natural-pass gain; this is a four-layer compound action. The requested operation token and pre-action prefix remain unchanged; the target is $F_i(s_{t-1})$, the chosen operation applied to the pre-action state. Future target tokens and post-action reference states are not inserted. For B, which has no route-label supervision, an operation-to-expert permutation is selected separately at each layer on 4,096 development contexts. All 24 permutations are enumerated; the chosen permutation maximizes natural label agreement, with lexicographic tie-breaking. R/I retain their label-defined identity mapping for the main comparison. Neither final forced fidelity nor test data are used to select the mapping.

At the forced position, let $z_\ell^{\mathrm{nat}}$ denote layer $\ell$'s router logits in the original natural pass. Its recorded gain is the natural winner's probability, $q_\ell^{\mathrm{nat}}=\max_e\operatorname{softmax}(z_\ell^{\mathrm{nat}})_e$. After detachment, this fixed value multiplies the forced expert's output. The convention preserves the original winner's gain rather than looking up the forced expert's original probability or recomputing a gain during intervention. All layers' reference gains come from the same original natural pass. The forced pass still computes its own activations, while the residual and other shared paths retain their declared computation. The operation-to-expert mapping remains development-selected for B and the identity for R/I. This is the frozen-gain convention of \cref{sec:frozen-gain-replay}, distinct from current-gain support replay in \cref{sec:support-replay}.

\paragraph{Evaluation.}
For each of 32,768 independent H8 teacher-prefix contexts, all four actions are forced and averaged. Forced fidelity is the average indicator that the resulting state equals the declared operation's output. Cross-action fidelity excludes the originally requested action. Action separation compares outputs of distinct forced actions and must be read alongside target separation on the same states. The uniform-state, uniformly requested and forced collapsed-control reference is $67/248$, not $1/4$, because several operation outputs coincide; this control distribution is not automatically the distribution of every model-specific probe. Same-index forcing, causal-mask, and explicit-operator controls test implementation rather than learned performance.

The main action is a four-layer compound intervention. Single-layer tests use a separate 2,048-context bank, changing only one call. On-policy diagnostics evaluate each model on its own reached-state distribution. The label-agreement columns use development scores; primary compound fidelity is evaluated on the independent test contexts. These columns therefore describe different evaluation sets.

R and I both have development label agreement 1, but compound fidelities are $0.9467\pm0.0171$ and $1.0000\pm0.0000$, versus $0.2528\pm0.0079$ for B. The I checkpoints have single-layer fidelity $0.3274\pm0.0247$ when averaged over layers. This contrasts attribution scopes; it does not prove that individual experts have no useful partial function.  All these averages are descriptive over three seeds.
\begingroup
\small
\setlength{\tabcolsep}{4pt}
\begin{longtable}{@{}lrrrrrrrr@{}}
\caption{Per-seed primary causal results. Raw/aligned label scores use development data; compound fidelity uses final teacher contexts; L0--L3 are separate single-layer secondary interventions.}\label{tab:exp_t3_perseed}\\
\toprule
Obj. & Seed & Raw & Aligned & Compound & L0 & L1 & L2 & L3 \\
\midrule
\endfirsthead
\toprule
Obj. & Seed & Raw & Aligned & Compound & L0 & L1 & L2 & L3 \\
\midrule
\endhead
B & 0 & 0.2621 & 0.3605 & 0.2567 & 0.267 & 0.267 & 0.266 & 0.267 \\
B & 1 & 0.2051 & 0.3790 & 0.2438 & 0.267 & 0.267 & 0.257 & 0.270 \\
B & 2 & 0.2657 & 0.3297 & 0.2580 & 0.267 & 0.266 & 0.270 & 0.267 \\
R & 0 & 1.0000 & 1.0000 & 0.9632 & 0.279 & 0.267 & 0.293 & 0.269 \\
R & 1 & 1.0000 & 1.0000 & 0.9291 & 0.290 & 0.267 & 0.276 & 0.269 \\
R & 2 & 1.0000 & 1.0000 & 0.9478 & 0.292 & 0.268 & 0.269 & 0.267 \\
I & 0 & 1.0000 & 1.0000 & 1.0000 & 0.401 & 0.267 & 0.269 & 0.267 \\
I & 1 & 1.0000 & 1.0000 & 1.0000 & 0.597 & 0.267 & 0.269 & 0.267 \\
I & 2 & 1.0000 & 1.0000 & 1.0000 & 0.520 & 0.268 & 0.267 & 0.267 \\
\bottomrule
\end{longtable}
\endgroup

\begingroup
\small
\setlength{\tabcolsep}{4pt}
\begin{longtable}{@{}lrrrr@{}}
\caption{On-policy secondary diagnostics, 2,048 contexts per checkpoint. Each model reaches its own context distribution. Mean $\pm$ sample SD over seeds.}\label{tab:exp_onpolicy}\\
\toprule
Objective & Fidelity & Cross-fidelity & Separation & Target sep. \\
\midrule
\endfirsthead
\toprule
Objective & Fidelity & Cross-fidelity & Separation & Target sep. \\
\midrule
\endhead
B & 0.2499 $\pm$ 0.0093 & 0.1033 $\pm$ 0.0095 & 0.5309 $\pm$ 0.0270 & 0.9744 $\pm$ 0.0000 \\
R & 0.9455 $\pm$ 0.0202 & 0.9273 $\pm$ 0.0269 & 0.9658 $\pm$ 0.0015 & 0.9744 $\pm$ 0.0000 \\
I & 1.0000 $\pm$ 0.0000 & 1.0000 $\pm$ 0.0000 & 0.9744 $\pm$ 0.0000 & 0.9744 $\pm$ 0.0000 \\
\bottomrule
\end{longtable}
\endgroup

\paragraph{Intervention effects are a separate question.}
Route-supervised models at 64 traces have label agreement 1 but forced-action fidelity $0.3713\pm0.0267$. At the primary budget, intervention supervision gives compound fidelity 1 without establishing single-expert attribution. Separately holdout-trained models retain seen fidelity 1 but reach only $0.0021\pm0.0037$ on withheld transitions. A separately trained follow-up retains the seen/withheld fidelity gap at moduli 97 and 251 (\cref{app:exp-state-space}). These address causal attribution and coverage; full tables remain in \cref{app:new-causal,app:exp-holdout}.

\subsection{Functional routing, margin, and gain diagnostics}
\label{app:new-diagnostics}
The routing diagnostics use six frozen local/tied checkpoints and 4,096 independent development sites per checkpoint. The nearest, farthest, and random alternative experts are evaluated on one common pre-call input with the same natural gain. The reported local event difference is checked against the residual-output distance. Selecting the nearest alternative makes its local distance smallest by definition; whether it has smaller terminal harm is an empirical question. A logical-site substitution persists through later full-prefix recomputation, while subsequent unforced routes respond naturally. Candidate search is additional diagnostic work.

The measurements show positive distance--harm associations in many accurate primary checkpoints. The lower-budget routing follow-up uses 24 lower-budget checkpoints, 1,024 sites each, and includes repairs as negative loss differences. Associations weaken or change sign in some settings.  Downstream first-disagreement summaries refer to the instrumented operation-token positions, not every token and cache event in the computation.
\begingroup
\small
\setlength{\tabcolsep}{4pt}
\begin{longtable}{@{}llrrrr@{}}
\caption{Functional expert-replacement diagnostics, 4,096 sites per checkpoint. Harm is intervened minus natural terminal error and is not clipped. Distance ordering is by construction; terminal association is empirical. SE is across independent sampled sites as reported.}\label{tab:exp_routing}\\
\toprule
Checkpoint & Alternative & Distance & Harm (pp) & SE (pp) & Spearman \\
\midrule
\endfirsthead
\toprule
Checkpoint & Alternative & Distance & Harm (pp) & SE (pp) & Spearman \\
\midrule
\endhead
local-s0 & nearest & 4.458 & 8.618 & 0.439 & 0.172 \\
local-s0 & random & 4.954 & 13.843 & 0.540 & 0.270 \\
local-s0 & farthest & 5.456 & 19.189 & 0.615 & 0.378 \\
tied-s0 & nearest & 3.576 & 0.781 & 0.138 & 0.013 \\
tied-s0 & random & 3.884 & 2.759 & 0.256 & 0.066 \\
tied-s0 & farthest & 4.190 & 5.029 & 0.342 & 0.115 \\
local-s1 & nearest & 4.024 & 6.396 & 0.382 & 0.149 \\
local-s1 & random & 4.560 & 10.864 & 0.486 & 0.246 \\
local-s1 & farthest & 5.067 & 15.015 & 0.558 & 0.318 \\
tied-s1 & nearest & 2.846 & 1.733 & 0.204 & 0.017 \\
tied-s1 & random & 3.119 & 4.175 & 0.313 & 0.089 \\
tied-s1 & farthest & 3.430 & 6.006 & 0.371 & 0.132 \\
local-s2 & nearest & 4.717 & 7.153 & 0.403 & 0.162 \\
local-s2 & random & 5.218 & 12.695 & 0.520 & 0.266 \\
local-s2 & farthest & 5.699 & 18.579 & 0.608 & 0.365 \\
tied-s2 & nearest & 3.375 & 0.146 & 0.060 & -0.008 \\
tied-s2 & random & 3.727 & 1.709 & 0.203 & 0.041 \\
tied-s2 & farthest & 4.101 & 3.223 & 0.276 & 0.078 \\
\bottomrule
\end{longtable}
\endgroup

\begingroup
\small
\setlength{\tabcolsep}{4pt}
\begin{longtable}{@{}llrrrr@{}}
\caption{Lower-budget expert-replacement diagnostics outside the saturated primary regime. Each checkpoint uses 1,024 sites; selected nearest and farthest summaries are shown. Negative harm represents repairs.}\label{tab:exp_offsaturation}\\
\toprule
Checkpoint & Alternative & Natural error & Harm (pp) & Repair (\%) & Spearman \\
\midrule
\endfirsthead
\toprule
Checkpoint & Alternative & Natural error & Harm (pp) & Repair (\%) & Spearman \\
\midrule
\endhead
EXT-local-B-n128-s0 & nearest & 0.782 & 0.68 & 1.27 & -0.038 \\
EXT-local-B-n128-s0 & farthest & 0.782 & 3.91 & 1.66 & 0.031 \\
EXT-local-B-n128-s1 & nearest & 0.712 & 2.64 & 1.07 & -0.007 \\
EXT-local-B-n128-s1 & farthest & 0.712 & 5.08 & 2.15 & 0.137 \\
EXT-local-B-n128-s2 & nearest & 0.778 & 0.59 & 1.07 & -0.021 \\
EXT-local-B-n128-s2 & farthest & 0.778 & 3.71 & 1.66 & 0.020 \\
EXT-local-B-n192-s0 & nearest & 0.017 & 10.64 & 0.00 & 0.134 \\
EXT-local-B-n192-s0 & farthest & 0.017 & 22.56 & 0.00 & 0.261 \\
EXT-local-B-n192-s1 & nearest & 0.008 & 8.30 & 0.00 & 0.101 \\
EXT-local-B-n192-s1 & farthest & 0.008 & 18.95 & 0.10 & 0.264 \\
EXT-local-B-n192-s2 & nearest & 0.001 & 9.18 & 0.00 & 0.125 \\
EXT-local-B-n192-s2 & farthest & 0.001 & 21.29 & 0.00 & 0.357 \\
EXT-local-B-n256-s0 & nearest & 0.000 & 8.30 & 0.00 & 0.141 \\
EXT-local-B-n256-s0 & farthest & 0.000 & 17.77 & 0.00 & 0.309 \\
EXT-local-B-n256-s1 & nearest & 0.010 & 8.69 & 0.00 & 0.128 \\
EXT-local-B-n256-s1 & farthest & 0.010 & 17.48 & 0.00 & 0.345 \\
EXT-local-B-n256-s2 & nearest & 0.000 & 8.01 & 0.00 & 0.101 \\
EXT-local-B-n256-s2 & farthest & 0.000 & 23.14 & 0.00 & 0.420 \\
EXT-local-B-n96-s0 & nearest & 0.957 & 0.10 & 0.59 & -0.043 \\
EXT-local-B-n96-s0 & farthest & 0.957 & 0.00 & 1.07 & -0.033 \\
EXT-local-B-n96-s1 & nearest & 0.965 & 0.59 & 0.29 & -0.058 \\
EXT-local-B-n96-s1 & farthest & 0.965 & -0.59 & 1.17 & -0.065 \\
EXT-local-B-n96-s2 & nearest & 0.739 & 3.71 & 1.07 & 0.001 \\
EXT-local-B-n96-s2 & farthest & 0.739 & 7.13 & 1.27 & 0.029 \\
EXT-tied-B-n128-s0 & nearest & 0.462 & 1.37 & 1.17 & -0.011 \\
EXT-tied-B-n128-s0 & farthest & 0.462 & 5.47 & 1.37 & 0.045 \\
EXT-tied-B-n128-s1 & nearest & 0.859 & 0.88 & 0.29 & -0.061 \\
EXT-tied-B-n128-s1 & farthest & 0.859 & 0.10 & 1.17 & -0.048 \\
EXT-tied-B-n128-s2 & nearest & 0.510 & 3.22 & 0.68 & 0.008 \\
EXT-tied-B-n128-s2 & farthest & 0.510 & 5.08 & 0.88 & 0.061 \\
EXT-tied-B-n192-s0 & nearest & 0.036 & 1.56 & 0.00 & -0.066 \\
EXT-tied-B-n192-s0 & farthest & 0.036 & 3.81 & 0.10 & -0.000 \\
EXT-tied-B-n192-s1 & nearest & 0.064 & 2.83 & 0.29 & 0.029 \\
EXT-tied-B-n192-s1 & farthest & 0.064 & 6.84 & 0.39 & 0.130 \\
EXT-tied-B-n192-s2 & nearest & 0.002 & 4.10 & 0.00 & 0.025 \\
EXT-tied-B-n192-s2 & farthest & 0.002 & 9.77 & 0.00 & 0.128 \\
EXT-tied-B-n256-s0 & nearest & 0.007 & 2.44 & 0.10 & 0.015 \\
EXT-tied-B-n256-s0 & farthest & 0.007 & 6.45 & 0.10 & 0.114 \\
EXT-tied-B-n256-s1 & nearest & 0.000 & 1.76 & 0.00 & -0.043 \\
EXT-tied-B-n256-s1 & farthest & 0.000 & 7.32 & 0.00 & 0.088 \\
EXT-tied-B-n256-s2 & nearest & 0.000 & 1.76 & 0.00 & -0.056 \\
EXT-tied-B-n256-s2 & farthest & 0.000 & 7.23 & 0.00 & 0.102 \\
EXT-tied-B-n96-s0 & nearest & 0.952 & -0.10 & 1.07 & -0.050 \\
EXT-tied-B-n96-s0 & farthest & 0.952 & -0.29 & 1.07 & -0.052 \\
EXT-tied-B-n96-s1 & nearest & 0.953 & 0.29 & 0.59 & -0.029 \\
EXT-tied-B-n96-s1 & farthest & 0.953 & 0.88 & 0.59 & -0.021 \\
EXT-tied-B-n96-s2 & nearest & 0.850 & 0.98 & 0.39 & -0.026 \\
EXT-tied-B-n96-s2 & farthest & 0.850 & 0.88 & 0.98 & -0.017 \\
\bottomrule
\end{longtable}
\endgroup

The margin-and-gain diagnostic records logits before and after direct score perturbations, common shifts, and hidden-state perturbations. Let $\gamma=z_{(1)}-z_{(2)}$ be the original winner--runner-up logit gap. A direct perturbation lowers the winner by $c\gamma$ and raises the runner-up by $c\gamma$, leaving other logits unchanged; this $c$ is a perturbation scale, not the target-heterogeneity parameter in \cref{app:new-frozen-frontier}. Hidden-state perturbations use $h'=h+\alpha\max\{\|h\|_2,1\}v$, where $v$ is the perturbation-direction vector, and then recompute normalization and routing. The table gives the tested $c$ and $\alpha$ values. With $\epsilon=\|\widetilde z-z\|_\infty$ and $\mathrm{tol}=64\varepsilon_{\mathrm{mach}}(1+\max_e|z_e|)$, where $\varepsilon_{\mathrm{mach}}$ is the floating-point machine precision, a pair is certified if $\gamma-2\epsilon>\mathrm{tol}$, boundary if $|\gamma-2\epsilon|\le\mathrm{tol}$, and condition-not-met otherwise. It evaluates the strict half-margin condition in \cref{lem:topk-margin-stability} from the actual scores with a floating-point tolerance. Across the measured perturbation pairs, 150,095 satisfy the strict condition with slack exceeding the floating-point tolerance, with zero natural-support violations among them. Boundary cases are retained.  The same-support gain measurement also reports nonzero coefficient changes, so support stability is not update equality. This is a sampled local diagnostic, not the trajectory-level conditions or a non-vacuous numerical certificate for the general routed bound.
\begingroup
\small
\setlength{\tabcolsep}{4pt}
\begin{longtable}{@{}lrrrrrr@{}}
\caption{Router-margin and gain summaries pooled over six checkpoints. A certified violation is a natural support change despite the strict half-margin condition with numerical tolerance. Failure of the sufficient condition does not require a support change.}\label{tab:exp_margin}\\
\toprule
Perturbation & Pairs & Certified & Boundary & Not met & Violations & Changed/not met \\
\midrule
\endfirsthead
\toprule
Perturbation & Pairs & Certified & Boundary & Not met & Violations & Changed/not met \\
\midrule
\endhead
direct\_c=0 & 24576 & 24574 & 2 & 0 & 0 & 0 \\
direct\_c=0.24 & 24576 & 24573 & 3 & 0 & 0 & 0 \\
direct\_c=0.49 & 24576 & 24515 & 61 & 0 & 0 & 0 \\
direct\_c=0.5 & 24576 & 0 & 24576 & 0 & 0 & 0 \\
direct\_c=0.51 & 24576 & 0 & 61 & 24515 & 0 & 24515 \\
direct\_c=1.0 & 24576 & 0 & 2 & 24574 & 0 & 24574 \\
common\_logit\_shift & 24576 & 4224 & 0 & 20352 & 0 & 0 \\
hidden\_alpha=0.0001 & 24576 & 24557 & 4 & 15 & 0 & 5 \\
hidden\_alpha=0.001 & 24576 & 24409 & 3 & 164 & 0 & 32 \\
hidden\_alpha=0.01 & 24576 & 23243 & 0 & 1333 & 0 & 255 \\
\bottomrule
\end{longtable}
\endgroup

\subsection{Independent paired performance evaluation}
\label{app:new-execution}
The primary evaluation comprises eighteen paired comparisons: tied/local and dense/local at each of three seeds under terminal and complete-trace error (twelve), plus R/B and I/R under $1-\mathrm{Fid}$ (six). Loss differences are reference minus candidate. We apply both directions of \cref{prop:paired-real-improvement}, with thirty-six tails, total failure allowance 0.05, and 32,768 independent traces/contexts per checkpoint. The resulting two-sided radius is 2.0039 percentage points.

Architecture and intervention comparisons use two independent evaluation banks, each containing 32,768 held-out-family H8 traces. Within each comparison, candidate and reference predictions are paired on the same records. The resulting intervals cross zero for all twelve architecture comparisons and favor the candidate for all six supervision comparisons. These conclusions concern fixed checkpoints on the declared evaluation distributions. Four forced actions are averaged within each context before across-context uncertainty is computed; neither those actions nor the three training seeds increase the independent sample count. The fixed-family guarantee requires candidates to be fixed before the independent evaluation data are inspected.
\begingroup
\small
\setlength{\tabcolsep}{4pt}
\begin{longtable}{@{}lrlrrr@{}}
\caption{All 18 primary paired comparisons. Positive differences favor the first model; numbers are percentage points. Every interval has radius 2.0039 pp (36 tails, $\delta=0.05$, $N=32768$).}\label{tab:exp_audit}\\
\toprule
Candidate/ref. & Seed & Loss & Difference & Lower & Upper \\
\midrule
\endfirsthead
\toprule
Candidate/ref. & Seed & Loss & Difference & Lower & Upper \\
\midrule
\endhead
tied/local & 0 & terminal & 0.0000 & -2.0039 & 2.0039 \\
tied/local & 0 & trace & 0.0000 & -2.0039 & 2.0039 \\
dense/local & 0 & terminal & 0.0000 & -2.0039 & 2.0039 \\
dense/local & 0 & trace & 0.0000 & -2.0039 & 2.0039 \\
R/B & 0 & 1--Fid & 70.6573 & 68.6534 & 72.6613 \\
I/R & 0 & 1--Fid & 3.6774 & 1.6735 & 5.6813 \\
tied/local & 1 & terminal & 0.0031 & -2.0009 & 2.0070 \\
tied/local & 1 & trace & 0.0031 & -2.0009 & 2.0070 \\
dense/local & 1 & terminal & 0.0031 & -2.0009 & 2.0070 \\
dense/local & 1 & trace & 0.0031 & -2.0009 & 2.0070 \\
R/B & 1 & 1--Fid & 68.5387 & 66.5348 & 70.5426 \\
I/R & 1 & 1--Fid & 7.0854 & 5.0815 & 9.0893 \\
tied/local & 2 & terminal & 0.0000 & -2.0039 & 2.0039 \\
tied/local & 2 & trace & 0.0000 & -2.0039 & 2.0039 \\
dense/local & 2 & terminal & 0.0000 & -2.0039 & 2.0039 \\
dense/local & 2 & trace & 0.0000 & -2.0039 & 2.0039 \\
R/B & 2 & 1--Fid & 68.9774 & 66.9734 & 70.9813 \\
I/R & 2 & 1--Fid & 5.2246 & 3.2207 & 7.2285 \\
\bottomrule
\end{longtable}
\endgroup

\subsection{Local-to-final gains across data budgets and seeds}
\label{app:exp-budget}
The training-data sweep varies the training-record budget over 17 values from 16 to 16,384, using the same 3,000-update policy. The original three seeds give 153 architecture configurations. The original interval families stay separate from the ten-seed follow-up at budgets 64, 96, 128, and 192. 

A configuration specifies an architecture, budget, and seed; a contrast instead specifies a model pair, budget, and seed. Here the three architectures also yield three pairs: tied/local, dense/local, and dense/tied. Thus both counts happen to equal $17\times3\times3=153$ in the original sweep. Each contrast has one local and one terminal endpoint, giving 306 endpoint comparisons and 612 one-sided tails for two-sided simultaneous intervals.

\paragraph{Budget- and seed-dependent comparisons.}
The data do not satisfy dense $\le$ tied $\le$ local at every budget. The original three-seed means already have exceptions at 24, 192, and 256; at 192, teacher-forced step errors are approximately 0 (dense), 0.00109 (local), and 0.00496 (tied).  At 64 traces the three original tied/local step improvements are 5.18, 5.16, and 6.49 percentage points, whereas terminal differences are approximately $-0.043,-0.006,-0.143$ points with intervals covering zero. An established local prediction gain therefore does not imply a detectable gain on that terminal endpoint. 
\begingroup
\small
\setlength{\tabcolsep}{4pt}
\begin{longtable}{@{}rrrr@{}}
\caption{Original three-seed architecture data-budget sweep: teacher-forced step error on the reported H8 bank (153 configuration summaries). Values are mean $\pm$ sample SD over three training seeds.}\label{tab:exp_ladder}\\
\toprule
Training traces & Local & Tied & Dense \\
\midrule
\endfirsthead
\toprule
Training traces & Local & Tied & Dense \\
\midrule
\endhead
16 & 0.9467 $\pm$ 0.0034 & 0.9436 $\pm$ 0.0032 & 0.9389 $\pm$ 0.0065 \\
24 & 0.9322 $\pm$ 0.0008 & 0.9245 $\pm$ 0.0039 & 0.9247 $\pm$ 0.0110 \\
32 & 0.9167 $\pm$ 0.0035 & 0.8999 $\pm$ 0.0087 & 0.8917 $\pm$ 0.0130 \\
48 & 0.8651 $\pm$ 0.0107 & 0.8224 $\pm$ 0.0239 & 0.8160 $\pm$ 0.0524 \\
64 & 0.7959 $\pm$ 0.0220 & 0.7398 $\pm$ 0.0293 & 0.7033 $\pm$ 0.0488 \\
96 & 0.4133 $\pm$ 0.2122 & 0.3670 $\pm$ 0.1186 & 0.1377 $\pm$ 0.0459 \\
128 & 0.1638 $\pm$ 0.0119 & 0.1433 $\pm$ 0.0950 & 0.0179 $\pm$ 0.0186 \\
192 & 0.0011 $\pm$ 0.0009 & 0.0050 $\pm$ 0.0043 & 0.0000 $\pm$ 0.0000 \\
256 & 0.0004 $\pm$ 0.0006 & 0.0004 $\pm$ 0.0006 & 0.0000 $\pm$ 0.0000 \\
384 & 0.0002 $\pm$ 0.0002 & 0.0000 $\pm$ 0.0000 & 0.0000 $\pm$ 0.0000 \\
512 & 0.0001 $\pm$ 0.0002 & 0.0000 $\pm$ 0.0000 & 0.0000 $\pm$ 0.0000 \\
768 & 0.0000 $\pm$ 0.0000 & 0.0000 $\pm$ 0.0000 & 0.0000 $\pm$ 0.0000 \\
1024 & 0.0000 $\pm$ 0.0000 & 0.0000 $\pm$ 0.0000 & 0.0000 $\pm$ 0.0000 \\
2048 & 0.0000 $\pm$ 0.0000 & 0.0000 $\pm$ 0.0000 & 0.0000 $\pm$ 0.0000 \\
4096 & 0.0000 $\pm$ 0.0000 & 0.0000 $\pm$ 0.0000 & 0.0000 $\pm$ 0.0000 \\
8192 & 0.0000 $\pm$ 0.0000 & 0.0000 $\pm$ 0.0000 & 0.0000 $\pm$ 0.0000 \\
16384 & 0.0000 $\pm$ 0.0000 & 0.0000 $\pm$ 0.0000 & 0.0000 $\pm$ 0.0000 \\
\bottomrule
\end{longtable}
\endgroup

\begingroup
\small
\setlength{\tabcolsep}{4pt}
\begin{longtable}{@{}rrrr@{}}
\caption{Free-running terminal error for exactly the same original three-seed architecture checkpoints and H8 bank as the preceding step-error table. Values are mean $\pm$ sample SD.}\label{tab:exp_ladder_terminal}\\
\toprule
Training traces & Local & Tied & Dense \\
\midrule
\endfirsthead
\toprule
Training traces & Local & Tied & Dense \\
\midrule
\endhead
16 & 0.9671 $\pm$ 0.0014 & 0.9678 $\pm$ 0.0008 & 0.9668 $\pm$ 0.0011 \\
24 & 0.9665 $\pm$ 0.0006 & 0.9666 $\pm$ 0.0004 & 0.9673 $\pm$ 0.0010 \\
32 & 0.9680 $\pm$ 0.0005 & 0.9671 $\pm$ 0.0009 & 0.9679 $\pm$ 0.0006 \\
48 & 0.9667 $\pm$ 0.0009 & 0.9673 $\pm$ 0.0012 & 0.9684 $\pm$ 0.0003 \\
64 & 0.9679 $\pm$ 0.0006 & 0.9685 $\pm$ 0.0001 & 0.9677 $\pm$ 0.0008 \\
96 & 0.8883 $\pm$ 0.1295 & 0.9174 $\pm$ 0.0565 & 0.6444 $\pm$ 0.1314 \\
128 & 0.7580 $\pm$ 0.0342 & 0.6342 $\pm$ 0.2061 & 0.1304 $\pm$ 0.1300 \\
192 & 0.0086 $\pm$ 0.0072 & 0.0387 $\pm$ 0.0334 & 0.0000 $\pm$ 0.0000 \\
256 & 0.0030 $\pm$ 0.0049 & 0.0030 $\pm$ 0.0049 & 0.0000 $\pm$ 0.0000 \\
384 & 0.0019 $\pm$ 0.0017 & 0.0000 $\pm$ 0.0000 & 0.0000 $\pm$ 0.0000 \\
512 & 0.0011 $\pm$ 0.0019 & 0.0000 $\pm$ 0.0000 & 0.0000 $\pm$ 0.0000 \\
768 & 0.0000 $\pm$ 0.0000 & 0.0000 $\pm$ 0.0000 & 0.0000 $\pm$ 0.0000 \\
1024 & 0.0000 $\pm$ 0.0000 & 0.0000 $\pm$ 0.0000 & 0.0000 $\pm$ 0.0000 \\
2048 & 0.0000 $\pm$ 0.0000 & 0.0000 $\pm$ 0.0000 & 0.0000 $\pm$ 0.0000 \\
4096 & 0.0000 $\pm$ 0.0000 & 0.0000 $\pm$ 0.0000 & 0.0000 $\pm$ 0.0000 \\
8192 & 0.0000 $\pm$ 0.0000 & 0.0000 $\pm$ 0.0000 & 0.0000 $\pm$ 0.0000 \\
16384 & 0.0000 $\pm$ 0.0001 & 0.0000 $\pm$ 0.0000 & 0.0000 $\pm$ 0.0000 \\
\bottomrule
\end{longtable}
\endgroup

\begingroup
\small
\setlength{\tabcolsep}{4pt}
\begin{longtable}{@{}rrrrrrr@{}}
\caption{Tied versus local at 64 training traces, original architecture-sweep family. Values and bounds are percentage points. Differences are local minus tied error. The intervals retain the full 153-pair, 306-endpoint, 612-tail sweep budget ($N=32768$, $\delta=0.05$; radius $2.39685$ percentage points). }\label{tab:exp_local_terminal}\\
\toprule
Seed & Step diff. & Step lower & Step upper & Term. diff. & Term. lower & Term. upper \\
\midrule
\endfirsthead
\toprule
Seed & Step diff. & Step lower & Step upper & Term. diff. & Term. lower & Term. upper \\
\midrule
\endhead
0 & 5.183 & 2.786 & 7.580 & -0.043 & -2.440 & 2.354 \\
1 & 5.162 & 2.765 & 7.559 & -0.006 & -2.403 & 2.391 \\
2 & 6.488 & 4.092 & 8.885 & -0.143 & -2.540 & 2.253 \\
\bottomrule
\end{longtable}
\endgroup

In the balanced ten-seed follow-up at 96 traces, the local tied/local contrast is supported in opposite directions: six seeds favor tied and four favor local. At 64, all ten favor tied on the local endpoint. This motivates reporting seeds. The 153 original local/terminal configuration contrasts and 120 follow-up contrasts overlap in 36 configurations, leaving 237 unique configurations and 84 newly added contrasts. Their original confidence budgets are not combined into a new 237-comparison family. 

The follow-up uses four budgets (64, 96, 128, 192), seeds $0$--$9$, and the same three model pairs: $4\times10\times3=120$ contrasts, or 240 endpoints and 480 one-sided tails for two-sided intervals. Figure~\ref{fig:exp-local-final-transfer} displays the 40 tied/local contrasts and their two endpoints while retaining that full-family interval budget. 
\begingroup
\small
\setlength{\tabcolsep}{4pt}
\begin{longtable}{@{}rlrrc@{}}
\caption{Balanced ten-seed architecture follow-up; means and sample SD are computed from run summaries. Last column counts lower / higher / inconclusive local risk versus local under the separate follow-up intervals.}\label{tab:exp_seed}\\
\toprule
Traces & Model & Step error & Terminal error & Lower/higher/unclear \\
\midrule
\endfirsthead
\toprule
Traces & Model & Step error & Terminal error & Lower/higher/unclear \\
\midrule
\endhead
64 & local & 0.7684 $\pm$ 0.0233 & 0.9678 $\pm$ 0.0009 & -- \\
64 & tied & 0.7127 $\pm$ 0.0271 & 0.9679 $\pm$ 0.0005 & 10/0/0 \\
64 & dense & 0.6499 $\pm$ 0.0530 & 0.9676 $\pm$ 0.0006 & 10/0/0 \\
96 & local & 0.3736 $\pm$ 0.1210 & 0.9129 $\pm$ 0.0677 & -- \\
96 & tied & 0.3630 $\pm$ 0.0753 & 0.9256 $\pm$ 0.0355 & 6/4/0 \\
96 & dense & 0.1583 $\pm$ 0.0518 & 0.6899 $\pm$ 0.1248 & 9/0/1 \\
128 & local & 0.1047 $\pm$ 0.0516 & 0.5478 $\pm$ 0.2017 & -- \\
128 & tied & 0.1100 $\pm$ 0.0779 & 0.5286 $\pm$ 0.2134 & 5/3/2 \\
128 & dense & 0.0114 $\pm$ 0.0112 & 0.0854 $\pm$ 0.0797 & 10/0/0 \\
192 & local & 0.0031 $\pm$ 0.0026 & 0.0247 $\pm$ 0.0209 & -- \\
192 & tied & 0.0038 $\pm$ 0.0029 & 0.0298 $\pm$ 0.0229 & 0/0/10 \\
192 & dense & 0.0001 $\pm$ 0.0003 & 0.0012 $\pm$ 0.0020 & 0/0/10 \\
\bottomrule
\end{longtable}
\endgroup

\paragraph{Paired one-step and final-output gains.}
Figure~\ref{fig:exp-local-final-transfer} uses all forty tied/local contrasts in the balanced follow-up: all ten seeds at all four budgets (64, 96, 128, 192). Its 64-trace panel repeats ten of those contrasts. Every bar retains its original 240-endpoint, 480-tail family interval with $\delta=0.05$ and 32,768 independent traces per checkpoint.

Across those ten 64-trace seeds, one-step improvements range from 3.4931 to 7.7190 percentage points, with ten positive lower bounds. Final-output differences range from $-0.1862$ to $+0.1923$ points, with ten intervals crossing zero. Both models have near-chance final predictions in that low-data region, so the result illustrates a lack of statistically resolved endpoint improvement. Other displayed budgets show larger final changes and seed-dependent directions. No displayed comparison establishes opposite signs on both endpoints. This empirical observable uses reference prefixes locally and model-generated prefixes finally.

The one-step metric uses reference prefixes, whereas the final-state and complete-trace metrics use free-running predictions. The independent-step reference used for complete-trace error concerns failure anywhere along a trace; it is not a final-state error model. Agreement with that reference does not establish independence of actual prediction errors or exclude propagation and recovery.

\paragraph{Supervision at low data.}
The supervision-budget sweep includes ten budgets (64, 96, 128, 192, 256, 512, 1,024, 2,048, 4,096, 16,384), three objectives, and the three original training seeds, yielding ninety configuration summaries. The B models reuse tied checkpoints from the architecture sweep. Additional seed comparisons retain their separate evaluation families. At 64 traces, R retains raw route agreement 1 while forced fidelity is $0.3713\pm0.0267$. Labels and action effects therefore have different empirical behavior. The extra intervention targets also constitute extra supervision, so observed I improvements are not free at matched total training cost.
\begingroup
\small
\setlength{\tabcolsep}{4pt}
\begin{longtable}{@{}rlrrrr@{}}
\caption{Supervision-budget sweep over ten training-data budgets, three objectives, and the three original seeds. The baseline reuses tied checkpoints from the architecture sweep. Causal evaluation uses 16,384 contexts.}\label{tab:exp_supervision}\\
\toprule
Traces & Obj. & Terminal error & Raw labels & Forced fidelity & Cross-fidelity \\
\midrule
\endfirsthead
\toprule
Traces & Obj. & Terminal error & Raw labels & Forced fidelity & Cross-fidelity \\
\midrule
\endhead
64 & B & 0.9685 $\pm$ 0.0001 & 0.2443 $\pm$ 0.0528 & 0.1724 $\pm$ 0.0108 & 0.1442 $\pm$ 0.0067 \\
64 & R & 0.9642 $\pm$ 0.0006 & 1.0000 $\pm$ 0.0000 & 0.3713 $\pm$ 0.0267 & 0.3406 $\pm$ 0.0298 \\
64 & I & 0.1257 $\pm$ 0.0454 & 1.0000 $\pm$ 0.0000 & 0.9943 $\pm$ 0.0014 & 0.9978 $\pm$ 0.0006 \\
96 & B & 0.9174 $\pm$ 0.0565 & 0.2154 $\pm$ 0.0581 & 0.3097 $\pm$ 0.0293 & 0.2102 $\pm$ 0.0134 \\
96 & R & 0.5103 $\pm$ 0.1574 & 1.0000 $\pm$ 0.0000 & 0.7963 $\pm$ 0.0260 & 0.7606 $\pm$ 0.0351 \\
96 & I & 0.0215 $\pm$ 0.0098 & 1.0000 $\pm$ 0.0000 & 0.9991 $\pm$ 0.0002 & 0.9997 $\pm$ 0.0005 \\
128 & B & 0.6342 $\pm$ 0.2061 & 0.2530 $\pm$ 0.0417 & 0.3531 $\pm$ 0.0671 & 0.2115 $\pm$ 0.0629 \\
128 & R & 0.0636 $\pm$ 0.0842 & 1.0000 $\pm$ 0.0000 & 0.7909 $\pm$ 0.0713 & 0.7239 $\pm$ 0.0974 \\
128 & I & 0.0024 $\pm$ 0.0014 & 1.0000 $\pm$ 0.0000 & 0.9999 $\pm$ 0.0001 & 1.0000 $\pm$ 0.0000 \\
192 & B & 0.0387 $\pm$ 0.0334 & 0.2424 $\pm$ 0.0483 & 0.2918 $\pm$ 0.0204 & 0.1218 $\pm$ 0.0047 \\
192 & R & 0.0056 $\pm$ 0.0039 & 1.0000 $\pm$ 0.0000 & 0.8779 $\pm$ 0.0307 & 0.8373 $\pm$ 0.0407 \\
192 & I & 0.0008 $\pm$ 0.0012 & 1.0000 $\pm$ 0.0000 & 1.0000 $\pm$ 0.0000 & 1.0000 $\pm$ 0.0000 \\
256 & B & 0.0030 $\pm$ 0.0049 & 0.2576 $\pm$ 0.0180 & 0.2943 $\pm$ 0.0052 & 0.1393 $\pm$ 0.0172 \\
256 & R & 0.0013 $\pm$ 0.0023 & 1.0000 $\pm$ 0.0000 & 0.8120 $\pm$ 0.0292 & 0.7494 $\pm$ 0.0388 \\
256 & I & 0.0000 $\pm$ 0.0000 & 1.0000 $\pm$ 0.0000 & 1.0000 $\pm$ 0.0000 & 1.0000 $\pm$ 0.0000 \\
512 & B & 0.0000 $\pm$ 0.0000 & 0.2312 $\pm$ 0.0172 & 0.2832 $\pm$ 0.0394 & 0.1312 $\pm$ 0.0330 \\
512 & R & 0.0000 $\pm$ 0.0000 & 1.0000 $\pm$ 0.0000 & 0.9441 $\pm$ 0.0119 & 0.9254 $\pm$ 0.0158 \\
512 & I & 0.0000 $\pm$ 0.0000 & 1.0000 $\pm$ 0.0000 & 1.0000 $\pm$ 0.0000 & 1.0000 $\pm$ 0.0000 \\
1024 & B & 0.0000 $\pm$ 0.0000 & 0.2436 $\pm$ 0.0435 & 0.2796 $\pm$ 0.0242 & 0.1325 $\pm$ 0.0267 \\
1024 & R & 0.0000 $\pm$ 0.0000 & 1.0000 $\pm$ 0.0000 & 0.9557 $\pm$ 0.0243 & 0.9409 $\pm$ 0.0324 \\
1024 & I & 0.0000 $\pm$ 0.0000 & 1.0000 $\pm$ 0.0000 & 1.0000 $\pm$ 0.0000 & 1.0000 $\pm$ 0.0000 \\
2048 & B & 0.0000 $\pm$ 0.0000 & 0.2625 $\pm$ 0.0498 & 0.2704 $\pm$ 0.0184 & 0.1261 $\pm$ 0.0289 \\
2048 & R & 0.0000 $\pm$ 0.0000 & 1.0000 $\pm$ 0.0000 & 0.9563 $\pm$ 0.0042 & 0.9418 $\pm$ 0.0056 \\
2048 & I & 0.0001 $\pm$ 0.0002 & 1.0000 $\pm$ 0.0000 & 1.0000 $\pm$ 0.0000 & 1.0000 $\pm$ 0.0000 \\
4096 & B & 0.0000 $\pm$ 0.0000 & 0.2445 $\pm$ 0.0307 & 0.2615 $\pm$ 0.0294 & 0.1119 $\pm$ 0.0280 \\
4096 & R & 0.0000 $\pm$ 0.0000 & 1.0000 $\pm$ 0.0000 & 0.9671 $\pm$ 0.0090 & 0.9561 $\pm$ 0.0120 \\
4096 & I & 0.0000 $\pm$ 0.0000 & 1.0000 $\pm$ 0.0000 & 1.0000 $\pm$ 0.0000 & 1.0000 $\pm$ 0.0000 \\
16384 & B & 0.0000 $\pm$ 0.0000 & 0.2465 $\pm$ 0.0372 & 0.2539 $\pm$ 0.0281 & 0.1052 $\pm$ 0.0226 \\
16384 & R & 0.0000 $\pm$ 0.0000 & 1.0000 $\pm$ 0.0000 & 0.9457 $\pm$ 0.0159 & 0.9276 $\pm$ 0.0212 \\
16384 & I & 0.0000 $\pm$ 0.0000 & 1.0000 $\pm$ 0.0000 & 1.0000 $\pm$ 0.0000 & 1.0000 $\pm$ 0.0000 \\
\bottomrule
\end{longtable}
\endgroup

\subsection{Strict state--action holdout and capacity checks}
\label{app:exp-holdout}
The transition-holdout study uses separate B/R/I trainings with a declared set of sixteen of the 124 state--action pairs withheld. Ordinary training traces containing excluded transitions are rejected, and such pairs are excluded from intervention-training targets; input and output state symbols remain available elsewhere. Thus this is stronger than family holdout, but it is still one finite symbolic task and one selected holdout set.

The fixed withheld pairs, written as (pre-action state, operation ID) with IDs $0$--$3$, are
\[
\begin{gathered}
 (1,1),\ (3,0),\ (3,3),\ (6,1),\ (7,2),\ (8,3),\ (10,0),\ (10,3),\\
 (11,3),\ (13,2),\ (18,2),\ (24,0),\ (25,0),\ (25,1),\ (28,3),\ (29,0).
\end{gathered}
\]
The same set is used across this B/R/I holdout family, with at most two operations excluded at any state. The per-state withholding study in \cref{app:exp-state-space} uses a different design.

Evaluation partitions forced cells by the pre-state and forced action. Evaluation includes 6,987 eligible withheld contexts containing 8,593 withheld action cells; seen statistics use 16,384 contexts and 56,943 seen cells. Context averaging is retained. Natural test traces are separately partitioned into 21,919 using at least one withheld transition and 10,849 clean traces.

The holdout-trained I models have seen fidelity 1 and withheld fidelity $0.0021\pm0.0037$ over three seeds. Natural terminal error on traces using a withheld transition is about 0.984, while clean-trace error is zero. Separately trained full-table I controls have fidelity 1 on both evaluation subsets. These are different checkpoints. The observations show that coverage-conditioned compound success did not generalize in this setting. They are consistent with memorization of seen transitions, but do not uniquely identify the model's internal implementation or negate measured in-distribution intervention effects.
\begingroup
\small
\setlength{\tabcolsep}{4pt}
\begin{longtable}{@{}lrrrrrr@{}}
\caption{Unseen-transition evaluation: separate trainings excluding the 16 declared state--action pairs. Seen/withheld fidelity is context-averaged within the relevant action subset. Withheld SE is the reported across-context estimate; clean/uses columns partition natural test traces. }\label{tab:exp_holdout}\\
\toprule
Obj. & Seed & Seen fid. & Withheld fid. & Withheld SE & Clean term. & Uses term. \\
\midrule
\endfirsthead
\toprule
Obj. & Seed & Seen fid. & Withheld fid. & Withheld SE & Clean term. & Uses term. \\
\midrule
\endhead
B & 0 & 0.26882 & 0.07528 & 0.00316 & 0.00000 & 0.98120 \\
B & 1 & 0.15763 & 0.05167 & 0.00252 & 0.00000 & 0.97226 \\
B & 2 & 0.31621 & 0.04687 & 0.00246 & 0.00000 & 0.98362 \\
R & 0 & 0.93353 & 0.02218 & 0.00176 & 0.00000 & 0.98298 \\
R & 1 & 0.96587 & 0.02290 & 0.00179 & 0.00000 & 0.98262 \\
R & 2 & 0.90390 & 0.00000 & 0.00000 & 0.00000 & 0.98435 \\
I & 0 & 1.00000 & 0.00000 & 0.00000 & 0.00000 & 0.98202 \\
I & 1 & 1.00000 & 0.00644 & 0.00096 & 0.00000 & 0.98540 \\
I & 2 & 1.00000 & 0.00000 & 0.00000 & 0.00000 & 0.98558 \\
\bottomrule
\end{longtable}
\endgroup

\begingroup
\small
\setlength{\tabcolsep}{4pt}
\begin{longtable}{@{}lrrrr@{}}
\caption{Holdout-trained models versus separately trained full-table controls, evaluated on the same subset labels. Mean $\pm$ sample SD over seeds. }\label{tab:exp_holdout_control}\\
\toprule
Obj. & Holdout: seen & Holdout: withheld & Full table: seen & Full table: subset \\
\midrule
\endfirsthead
\toprule
Obj. & Holdout: seen & Holdout: withheld & Full table: seen & Full table: subset \\
\midrule
\endhead
B & 0.2476 $\pm$ 0.0814 & 0.0579 $\pm$ 0.0152 & 0.1879 $\pm$ 0.0129 & 0.2484 $\pm$ 0.0708 \\
R & 0.9344 $\pm$ 0.0310 & 0.0150 $\pm$ 0.0130 & 0.9451 $\pm$ 0.0168 & 0.9493 $\pm$ 0.0165 \\
I & 1.0000 $\pm$ 0.0000 & 0.0021 $\pm$ 0.0037 & 1.0000 $\pm$ 0.0000 & 1.0000 $\pm$ 0.0000 \\
\bottomrule
\end{longtable}
\endgroup

The capacity-and-holdout-size follow-up varies width, depth, training-pool size, training steps, and nested holdout size. Width 256 to 512 is a doubling, though parameter count grows by roughly four. The combined setting has eight width-512 layers, 65,536 training traces, and 12,000 steps. These tested modifications do not yield uniformly high held-out fidelity, but all-run upper bounds must retain exceptions: \texttt{wide-s0} is 0.02834, \texttt{long-s1} 0.01174, the four-pair holdout at seed~2 is 0.13128, and the eight-pair holdout at seed~2 is 0.01236. Seen fidelity is not exactly one in every run either. All 24 follow-up summaries are shown below. 
\begingroup
\small
\setlength{\tabcolsep}{4pt}
\begin{longtable}{@{}lrrrrrr@{}}
\caption{All 24 capacity and holdout-size run summaries, including favorable exceptions and non-perfect seen fidelity. The base 16-pair results remain part of the base transition-holdout study. }\label{tab:exp_robustness}\\
\toprule
Setting & Seed & Hole & Params (M) & Seen fidelity & Withheld fidelity & Uses error \\
\midrule
\endfirsthead
\toprule
Setting & Seed & Hole & Params (M) & Seen fidelity & Withheld fidelity & Uses error \\
\midrule
\endhead
all & 0 & 16 & 17.89 & 1.00000 & 0.00114 & 0.9875 \\
all & 1 & 16 & 17.89 & 1.00000 & 0.00000 & 0.9794 \\
all & 2 & 16 & 17.89 & 1.00000 & 0.00057 & 0.9916 \\
base & 0 & 32 & 3.43 & 0.99472 & 0.00249 & 0.9774 \\
base & 0 & 4 & 3.43 & 1.00000 & 0.00000 & 0.9990 \\
base & 0 & 8 & 3.43 & 1.00000 & 0.00000 & 0.9935 \\
base & 1 & 32 & 3.43 & 0.99750 & 0.00000 & 0.9755 \\
base & 1 & 4 & 3.43 & 1.00000 & 0.00000 & 0.9977 \\
base & 1 & 8 & 3.43 & 1.00000 & 0.00000 & 0.9925 \\
base & 2 & 32 & 3.43 & 0.99872 & 0.00045 & 0.9736 \\
base & 2 & 4 & 3.43 & 1.00000 & 0.13128 & 0.9987 \\
base & 2 & 8 & 3.43 & 1.00000 & 0.01236 & 0.9949 \\
data & 0 & 16 & 3.43 & 1.00000 & 0.00000 & 0.9814 \\
data & 1 & 16 & 3.43 & 1.00000 & 0.00744 & 0.9848 \\
data & 2 & 16 & 3.43 & 1.00000 & 0.00143 & 0.9867 \\
deep & 0 & 16 & 4.49 & 1.00000 & 0.00000 & 0.9832 \\
deep & 1 & 16 & 4.49 & 1.00000 & 0.00587 & 0.9888 \\
deep & 2 & 16 & 4.49 & 0.99998 & 0.00243 & 0.9870 \\
long & 0 & 16 & 3.43 & 1.00000 & 0.00000 & 0.9789 \\
long & 1 & 16 & 3.43 & 1.00000 & 0.01174 & 0.9810 \\
long & 2 & 16 & 3.43 & 1.00000 & 0.00000 & 0.9821 \\
wide & 0 & 16 & 13.68 & 1.00000 & 0.02834 & 0.9812 \\
wide & 1 & 16 & 13.68 & 1.00000 & 0.00157 & 0.9858 \\
wide & 2 & 16 & 13.68 & 1.00000 & 0.00215 & 0.9872 \\
\bottomrule
\end{longtable}
\endgroup

\subsubsection{State--action generalization across state-space sizes}
\label{app:exp-state-space}
\paragraph{Question and protocol.}
The follow-up tests whether the seen/withheld fidelity gap is confined to the original 31-state setting. It applies the same four operations modulo $p=31,97,251$, giving transition tables with 124, 388, and 1,004 entries. At every state, zero, one, or two of the four operations are withheld, corresponding to 0\%, 25\%, or 50\% of the table. This per-state design differs from the preceding sixteen-pair holdout. The grid contains three training seeds for each of nine settings, giving 27 trained configurations.

The follow-up retains the four-layer, width-256 tied architecture and intervention-supervised objective, with modulus-specific state tokens and a $p$-way output head. Ordinary training traces containing a withheld transition are rejected, and counterfactual targets are restricted to allowed transitions. The no-holdout arm is separately trained at each modulus; it checks task learnability under full coverage.

Table~\ref{tab:exp-statespace-settings} gives the training-pool sizes and update budgets. These budgets target approximately comparable nominal exposure; total compute is not exactly matched, and exposure is nominal rather than a measured post-rejection count for each transition. Parameter counts also change with the vocabulary and output head. 
\begin{table}[htbp]
\centering
\small
\setlength{\tabcolsep}{3pt}
\caption{State-space follow-up: actual reported budgets and evaluation subset sizes. Each row has three training seeds.}
\label{tab:exp-statespace-settings}
\begin{tabular}{@{}rrrrrrrrr@{}}
\toprule
$p$ & $4p$ & Withheld & Pairs & Train traces & Updates & Parameters & $N_{\rm clean}$ & $N_{\rm use}$ \\
\midrule
31 & 124 & 0\% & 0 & 20,666 & 3,000 & 3,433,984 & 16384 & 0 \\
31 & 124 & 25\% & 31 & 15,500 & 3,000 & 3,433,984 & 1705 & 14679 \\
31 & 124 & 50\% & 62 & 10,334 & 3,000 & 3,433,984 & 54 & 16330 \\
97 & 388 & 0\% & 0 & 64,666 & 9,000 & 3,467,776 & 16384 & 0 \\
97 & 388 & 25\% & 97 & 48,500 & 9,000 & 3,467,776 & 1618 & 14766 \\
97 & 388 & 50\% & 194 & 32,334 & 9,000 & 3,467,776 & 72 & 16312 \\
251 & 1004 & 0\% & 0 & 167,334 & 24,000 & 3,546,624 & 16384 & 0 \\
251 & 1004 & 25\% & 251 & 125,500 & 24,000 & 3,546,624 & 1655 & 14729 \\
251 & 1004 & 50\% & 502 & 83,666 & 24,000 & 3,546,624 & 70 & 16314 \\
\bottomrule
\end{tabular}
\par\smallskip
\begin{minipage}{\linewidth}\small $N_{\rm clean}$ and $N_{\rm use}$ are counts per seed in a 16,384-trace task bank; they are constant across the three seeds of each setting. The latter subset uses at least one withheld transition. \end{minipage}
\end{table}

\paragraph{Evaluation and results.}
Forced-action evaluation uses 16,384 contexts per run and averages the applicable actions within a context before aggregation. At 25\% and 50\% withholding, every state has respectively one or two excluded actions, giving 16,384 and 32,768 withheld action cells, but still only 16,384 contexts. In the no-holdout control, withheld fidelity is undefined, not zero. Natural task evaluation separately partitions 16,384 traces into clean traces and those using at least one withheld transition; their counts must not be interchanged with the causal-context counts.

All nine full-coverage control checkpoints have observed fidelity one. Across the eighteen holdout-trained checkpoints, withheld fidelity is at most 0.00647, while seen fidelity is at least 0.97763. Table~\ref{tab:exp-statespace-results} reports every setting, and Table~\ref{tab:exp-statespace-perseed} retains every seed. Observed clean-subset terminal error is zero in all 27 evaluated configurations, but at 50\% withholding the clean subsets contain only 54, 72, and 70 traces for the three moduli. 
\begin{table}[htbp]
\centering
\small
\setlength{\tabcolsep}{3pt}
\caption{State-space follow-up: forced-action fidelity and free-running terminal error (mean $\pm$ sample SD across three seeds). All values are proportions.}
\label{tab:exp-statespace-results}
\begin{tabular}{@{}rrccc@{}}
\toprule
$p$ & Withheld & Withheld fidelity & Seen fidelity & Error on traces using withheld pairs \\
\midrule
31 & 0\% & --- & $1.00000\pm0.00000$ & --- \\
31 & 25\% & $0.00224\pm0.00367$ & $1.00000\pm0.00000$ & $0.97693\pm0.00346$ \\
31 & 50\% & $0.00212\pm0.00229$ & $0.98133\pm0.00556$ & $0.96928\pm0.00138$ \\
97 & 0\% & --- & $1.00000\pm0.00000$ & --- \\
97 & 25\% & $0.00006\pm0.00011$ & $1.00000\pm0.00000$ & $0.99199\pm0.00076$ \\
97 & 50\% & $0.00119\pm0.00104$ & $0.99887\pm0.00028$ & $0.99023\pm0.00085$ \\
251 & 0\% & --- & $1.00000\pm0.00000$ & --- \\
251 & 25\% & $0.00014\pm0.00025$ & $1.00000\pm0.00000$ & $0.99647\pm0.00072$ \\
251 & 50\% & $0.00078\pm0.00104$ & $0.99964\pm0.00051$ & $0.99577\pm0.00022$ \\
\bottomrule
\end{tabular}
\par\smallskip
\begin{minipage}{\linewidth}\small All nine full-coverage control checkpoints have observed fidelity one. Clean-subset terminal error is zero in all 27 runs, but the subset counts in Table~\ref{tab:exp-statespace-settings} must be retained. Dashes denote an empty subset. These are descriptive three-seed summaries.\end{minipage}
\end{table}

\begin{table}[htbp]
\centering
\small
\setlength{\tabcolsep}{3pt}
\caption{All 27 state-space follow-up run summaries. Fidelity and error columns are proportions.}
\label{tab:exp-statespace-perseed}
\begin{tabular}{@{}rrr rrrrrr@{}}
\toprule
$p$ & Withheld & Seed & Withheld fid. & Seen fid. & $N_{\rm clean}$ & Clean error & $N_{\rm use}$ & Error on use \\
\midrule
31 & 0\% & 0 & --- & 1.000000 & 16384 & 0.000000 & 0 & --- \\
31 & 0\% & 1 & --- & 1.000000 & 16384 & 0.000000 & 0 & --- \\
31 & 0\% & 2 & --- & 1.000000 & 16384 & 0.000000 & 0 & --- \\
31 & 25\% & 0 & 0.000000 & 1.000000 & 1705 & 0.000000 & 14679 & 0.980857 \\
31 & 25\% & 1 & 0.006470 & 1.000000 & 1705 & 0.000000 & 14679 & 0.974317 \\
31 & 25\% & 2 & 0.000244 & 1.000000 & 1705 & 0.000000 & 14679 & 0.975611 \\
31 & 50\% & 0 & 0.001434 & 0.978638 & 54 & 0.000000 & 16330 & 0.968279 \\
31 & 50\% & 1 & 0.004669 & 0.977631 & 54 & 0.000000 & 16330 & 0.970851 \\
31 & 50\% & 2 & 0.000244 & 0.987732 & 54 & 0.000000 & 16330 & 0.968708 \\
97 & 0\% & 0 & --- & 1.000000 & 16384 & 0.000000 & 0 & --- \\
97 & 0\% & 1 & --- & 1.000000 & 16384 & 0.000000 & 0 & --- \\
97 & 0\% & 2 & --- & 1.000000 & 16384 & 0.000000 & 0 & --- \\
97 & 25\% & 0 & 0.000000 & 1.000000 & 1618 & 0.000000 & 14766 & 0.991128 \\
97 & 25\% & 1 & 0.000183 & 1.000000 & 1618 & 0.000000 & 14766 & 0.992280 \\
97 & 25\% & 2 & 0.000000 & 1.000000 & 1618 & 0.000000 & 14766 & 0.992550 \\
97 & 50\% & 0 & 0.000488 & 0.999176 & 72 & 0.000000 & 16312 & 0.991172 \\
97 & 50\% & 1 & 0.002380 & 0.998627 & 72 & 0.000000 & 16312 & 0.989517 \\
97 & 50\% & 2 & 0.000702 & 0.998810 & 72 & 0.000000 & 16312 & 0.990007 \\
251 & 0\% & 0 & --- & 1.000000 & 16384 & 0.000000 & 0 & --- \\
251 & 0\% & 1 & --- & 1.000000 & 16384 & 0.000000 & 0 & --- \\
251 & 0\% & 2 & --- & 1.000000 & 16384 & 0.000000 & 0 & --- \\
251 & 25\% & 0 & 0.000000 & 1.000000 & 1655 & 0.000000 & 14729 & 0.997148 \\
251 & 25\% & 1 & 0.000427 & 1.000000 & 1655 & 0.000000 & 14729 & 0.996537 \\
251 & 25\% & 2 & 0.000000 & 1.000000 & 1655 & 0.000000 & 14729 & 0.995723 \\
251 & 50\% & 0 & 0.000275 & 0.999054 & 70 & 0.000000 & 16314 & 0.995954 \\
251 & 50\% & 1 & 0.001984 & 1.000000 & 70 & 0.000000 & 16314 & 0.995525 \\
251 & 50\% & 2 & 0.000092 & 0.999878 & 70 & 0.000000 & 16314 & 0.995832 \\
\bottomrule
\end{tabular}
\par\smallskip
\begin{minipage}{\linewidth}\small Withheld actions are averaged within each eligible context before aggregation; at 25\% and 50\% withholding there are 16,384 contexts, with one and two withheld actions per context, respectively.  Empty subsets are shown as dashes.\end{minipage}
\end{table}

\paragraph{Coverage-dependent generalization.}
The enlarged state spaces preserve a substantial coverage gap in this finite task family under the reported training recipe. Uniform-state guessing at $1/p$ serves as a scale reference. The displayed means and seed standard deviations are descriptive summaries over three training seeds; the primary eighteen-pair audit retains its own confidence budget.
\FloatBarrier

\subsection{State compression and general-design supplements}
\label{app:exp-supplements}
\paragraph{State-conditioned intervention variability.}
The study groups intervention outputs by the declared pre-action state and action, using a one-hot outcome representation across 124 cells. The table reports the centered outcome-variance statistic, cell-weighted fidelity, and fraction of empirically deterministic cells for 63 checkpoints. These are different descriptive quantities. In particular, the corresponding classification-loss analysis uses a conditional mode; it is not the squared-loss conditional-mean identity of \cref{prop:average-aliasing}.

The centered and unbiased-pair variance summaries differ by up to 0.00159009, so they are not treated as interchangeable finite-sample estimators. We report the centered statistic descriptively for each fixed checkpoint and state extractor, measuring conditional outcome variability.
\begingroup
\small
\setlength{\tabcolsep}{4pt}
\begin{longtable}{@{}rlrrr@{}}
\caption{State-conditioned intervention variability for 63 fixed checkpoints and the specified state extractor. The centered one-hot outcome statistic and cell-weighted fidelity measure different losses. }\label{tab:exp_compression}\\
\toprule
Traces & Obj. & Centered variance & Cell-weighted fidelity & Deterministic fraction \\
\midrule
\endfirsthead
\toprule
Traces & Obj. & Centered variance & Cell-weighted fidelity & Deterministic fraction \\
\midrule
\endhead
64 & B & 0.8221 $\pm$ 0.0183 & 0.1724 $\pm$ 0.0108 & 0.0000 $\pm$ 0.0000 \\
64 & R & 0.7370 $\pm$ 0.0205 & 0.3713 $\pm$ 0.0267 & 0.0000 $\pm$ 0.0000 \\
64 & I & 0.0107 $\pm$ 0.0024 & 0.9943 $\pm$ 0.0014 & 0.7366 $\pm$ 0.0123 \\
96 & B & 0.6418 $\pm$ 0.0159 & 0.3097 $\pm$ 0.0293 & 0.0000 $\pm$ 0.0000 \\
96 & R & 0.2941 $\pm$ 0.0388 & 0.7963 $\pm$ 0.0260 & 0.0430 $\pm$ 0.0123 \\
96 & I & 0.0017 $\pm$ 0.0003 & 0.9991 $\pm$ 0.0002 & 0.9194 $\pm$ 0.0689 \\
128 & B & 0.6299 $\pm$ 0.0487 & 0.3531 $\pm$ 0.0671 & 0.0027 $\pm$ 0.0047 \\
128 & R & 0.3465 $\pm$ 0.1069 & 0.7909 $\pm$ 0.0713 & 0.0027 $\pm$ 0.0047 \\
128 & I & 0.0002 $\pm$ 0.0002 & 0.9999 $\pm$ 0.0001 & 0.9812 $\pm$ 0.0123 \\
192 & B & 0.6719 $\pm$ 0.0184 & 0.2918 $\pm$ 0.0204 & 0.0108 $\pm$ 0.0186 \\
192 & R & 0.2158 $\pm$ 0.0498 & 0.8779 $\pm$ 0.0307 & 0.0269 $\pm$ 0.0259 \\
192 & I & 0.0001 $\pm$ 0.0001 & 1.0000 $\pm$ 0.0000 & 0.9866 $\pm$ 0.0123 \\
256 & B & 0.6706 $\pm$ 0.0182 & 0.2943 $\pm$ 0.0052 & 0.0081 $\pm$ 0.0081 \\
256 & R & 0.3215 $\pm$ 0.0436 & 0.8120 $\pm$ 0.0292 & 0.0054 $\pm$ 0.0093 \\
256 & I & 0.0000 $\pm$ 0.0000 & 1.0000 $\pm$ 0.0000 & 1.0000 $\pm$ 0.0000 \\
512 & B & 0.6688 $\pm$ 0.0153 & 0.2832 $\pm$ 0.0394 & 0.0054 $\pm$ 0.0047 \\
512 & R & 0.1046 $\pm$ 0.0222 & 0.9441 $\pm$ 0.0119 & 0.1317 $\pm$ 0.1266 \\
512 & I & 0.0000 $\pm$ 0.0000 & 1.0000 $\pm$ 0.0000 & 1.0000 $\pm$ 0.0000 \\
16384 & B & 0.6834 $\pm$ 0.0145 & 0.2537 $\pm$ 0.0075 & 0.0027 $\pm$ 0.0047 \\
16384 & R & 0.1013 $\pm$ 0.0289 & 0.9457 $\pm$ 0.0159 & 0.2016 $\pm$ 0.1290 \\
16384 & I & 0.0000 $\pm$ 0.0000 & 1.0000 $\pm$ 0.0000 & 1.0000 $\pm$ 0.0000 \\
\bottomrule
\end{longtable}
\endgroup

\paragraph{Pooling under general feature geometry.}
This supplement uses 512 noise repetitions on the frozen nonorthogonal designs. The 54 partition cells and 54 merge cells are interpreted using the existing covariance-aware fixed-estimator framework in \cref{thm:general-smoother-risk,thm:weighted-projection-frontier}. The empirical risks satisfy the numerical comparison criterion, but the roles have unequal estimation variances. Applying a single global equal-variance price outside the balanced assumptions yields four merge-sign mismatches. 
\begingroup
\small
\setlength{\tabcolsep}{4pt}
\begin{longtable}{@{}rrlrrrr@{}}
\caption{General-design merging: four sign discrepancies obtained by applying a global equal-variance proxy outside its assumptions. The covariance-aware reference and observed mean agree in sign.  Groups $i/j$ denotes the merge $G=\{i\}$, $H=\{j\}$ using zero-based role indices $0,\ldots,7$; $0/4$ and $2/6$ therefore identify two singleton-group pairs, not group sizes.}\label{tab:exp_merge}\\
\toprule
$c$ & $n$ & Groups & Global proxy & Covariance ref. & Observed & MCSE \\
\midrule
\endfirsthead
\toprule
$c$ & $n$ & Groups & Global proxy & Covariance ref. & Observed & MCSE \\
\midrule
\endhead
2.0 & 64 & 0/4 & -0.03990 & 0.00545 & 0.00543 & 0.00007 \\
2.0 & 64 & 2/6 & -0.03990 & 0.00385 & 0.00377 & 0.00008 \\
0.5 & 256 & 0/4 & -0.00998 & 0.00136 & 0.00135 & 0.00002 \\
0.5 & 256 & 2/6 & -0.00997 & 0.00096 & 0.00094 & 0.00002 \\
\bottomrule
\end{longtable}
\endgroup

The nine plug-in entries show lower observed risk than the separate-risk reference in this particular simulation.  Contrast dimension $k\ge4$ is not sufficient for the classical guarantee in \cref{app:classical-plugin}: the spherical Gaussian contrast-noise and matching loss assumptions also matter and are not automatically established by the anisotropic design. 
\begingroup
\small
\setlength{\tabcolsep}{4pt}
\begin{longtable}{@{}rrrrrr@{}}
\caption{General-design same-data shrinkage: Monte Carlo risk versus an analytic separate-risk reference, not a paired significance test. }\label{tab:exp_plugin}\\
\toprule
$c$ & $n$ & Separate ref. & Plug-in mean & MCSE & Savings (\%) \\
\midrule
\endfirsthead
\toprule
$c$ & $n$ & Separate ref. & Plug-in mean & MCSE & Savings (\%) \\
\midrule
\endhead
0.0 & 64 & 0.5268 & 0.0738 & 0.00110 & 86.0 \\
0.5 & 64 & 0.5268 & 0.1268 & 0.00101 & 75.9 \\
2.0 & 64 & 0.5268 & 0.2292 & 0.00115 & 56.5 \\
0.0 & 256 & 0.1317 & 0.0180 & 0.00026 & 86.4 \\
0.5 & 256 & 0.1317 & 0.0569 & 0.00027 & 56.8 \\
2.0 & 256 & 0.1317 & 0.0939 & 0.00054 & 28.7 \\
0.0 & 1024 & 0.0329 & 0.0046 & 0.00008 & 86.0 \\
0.5 & 1024 & 0.0329 & 0.0237 & 0.00014 & 28.1 \\
2.0 & 1024 & 0.0329 & 0.0300 & 0.00022 & 8.9 \\
\bottomrule
\end{longtable}
\endgroup

\FloatBarrier

\section{General Linear Sharing Subspaces and Cross-Depth Architectures}
\label{app:linear-sharing-architectures}

\paragraph{Overview.}
We extend the local linear comparison to unequal role weights, sample counts, and noise levels, and to fixed cross-depth sharing patterns. The formulas separate the approximation cost of tying from its estimation cost, explaining which role, depth, and interaction effects a declared linear architecture retains. Fixed global-pool assignments and depth-conditioned operator subspaces are instances of this projection comparison.
Section~\ref*{app:linear-sharing-architectures}.1 treats unequal roles and sampling, Section~\ref*{app:linear-sharing-architectures}.2 gives the projection formulation, Section~\ref*{app:linear-sharing-architectures}.3 instantiates cross-depth and multi-axis patterns, and Section~\ref*{app:linear-sharing-architectures}.4 collects the proofs.

\subsection{Unequal roles and heteroscedastic noise}
For role $a\in[M]$, let $A_a\in\R^{p\times d}$, assume $n_a\ge d$, and let the fixed design satisfy
\begin{equation}
 X_a^\top X_a=n_a I_d.
 \label{eq:hetero-design}
\end{equation}
The response matrix is $Y_a=X_aA_a^\top+\mathcal E_a$, where noise rows are mutually independent across observations and roles, centered, and have covariance $\sigma_a^2I_p$.  Different $\sigma_a^2$ are called \emph{heteroscedastic} noise: some roles have noisier targets than others.  Gaussianity is not required for any expectation identity in this section.  Let $\pi_a>0$ with $\sum_a\pi_a=1$, put $D=pd$ and $v_a=\sigma_a^2/n_a$, and define
\begin{align}
 \bar A_\pi&=\sum_a\pi_aA_a,
 &B_\pi&=\sum_a\pi_a\norm{A_a-\bar A_\pi}_F^2,
 \label{eq:hetero-bias}\\
 V_{\mathrm{sep}}&=D\sum_a\pi_av_a,
 &V_{\mathrm{pool}}&=D\sum_a\pi_a^2v_a,
 &C_\pi&=V_{\mathrm{sep}}-V_{\mathrm{pool}}.
 \label{eq:hetero-variance-terms}
\end{align}
The weighted pooled empirical objective is $\sum_a(\pi_a/n_a)\norm{Y_a-X_aA^\top}_F^2$, whose minimizer is
\begin{equation}
 \widehat A_\pi=\sum_a\pi_a\widehat A_a,
 \qquad
 \widehat A_a=Y_a^\top X_a/n_a.
 \label{eq:hetero-weighted-pool}
\end{equation}
For a common shrinkage coefficient $\lambda\in[0,1]$, define $\widehat A_a^{(\lambda)}=(1-\lambda)\widehat A_a+\lambda\widehat A_\pi$.  Its role-weighted excess prediction risk is
\begin{equation}
 \Risk_\lambda
 =\sum_a\pi_a\E_{\mathcal E,x}
   \norm{(\widehat A_a^{(\lambda)}-A_a)x}_2^2,
 \qquad \E[xx^\top]=I_d.
 \label{eq:hetero-risk-definition}
\end{equation}

\begin{theorem}[Exact weighted and heteroscedastic reuse frontier]
\label{thm:hetero-frontier}
For the role-weighted excess prediction risk in \cref{eq:hetero-risk-definition},
\begin{equation}
 \Risk_\lambda
 =V_{\mathrm{sep}}-2C_\pi\lambda+(B_\pi+C_\pi)\lambda^2.
 \label{eq:hetero-partial-risk}
\end{equation}
In particular,
\begin{equation}
 \Risk_{\mathrm{sep}}=V_{\mathrm{sep}},
 \qquad
 \Risk_{\mathrm{share}}=B_\pi+V_{\mathrm{pool}}.
 \label{eq:hetero-endpoints}
\end{equation}
Full reuse is better than separate estimation if and only if
\begin{equation}
 B_\pi<C_\pi
 =D\sum_a\frac{\pi_a(1-\pi_a)\sigma_a^2}{n_a}.
 \label{eq:hetero-exact-threshold}
\end{equation}
When $C_\pi>0$, the optimal common shrinkage strength is
\begin{equation}
 \lambda_\pi^\star=\frac{C_\pi}{B_\pi+C_\pi},
 \qquad
 \Risk_{\lambda_\pi^\star}
 =V_{\mathrm{sep}}-\frac{C_\pi^2}{B_\pi+C_\pi}.
 \label{eq:hetero-oracle-lambda}
\end{equation}
\end{theorem}

If $C_\pi=0<B_\pi$, the optimal coefficient is zero; if $C_\pi=B_\pi=0$, every coefficient is optimal.  The quadratic form is retained, with the sharing bias $B_\pi$, separate-estimation risk $V_{\mathrm{sep}}$, and removable variance $C_\pi$ given by their weighted, heteroscedastic counterparts.  A role with large evaluation mass but very few or very noisy samples contributes strongly to both the separate risk and the potential value of pooling.  Conversely, raw role multiplicity can be large while $C_\pi$ is small when almost all mass lies on one role.

The theorem also distinguishes two sample-allocation problems that are sometimes conflated.
\begin{corollary}[Architecture-dependent optimal sampling]
\label{cor:architecture-dependent-sampling}
Fix the idealized continuous variance-allocation problem $\sum_an_a=N$ and assume $\sigma_a>0$.  The formulas below apply when the displayed unconstrained allocations satisfy the feasibility requirements of the orthogonal designs, including any lower bounds such as $n_a\ge d$; otherwise the corresponding constrained KKT solution is used.  The allocation minimizing the pooled estimation variance is
\begin{equation}
 n_a^{\mathrm{pool}}
 =N\frac{\pi_a\sigma_a}{\sum_b\pi_b\sigma_b},
 \qquad
 \min V_{\mathrm{pool}}
 =\frac{D}{N}\left(\sum_a\pi_a\sigma_a\right)^2,
 \label{eq:pool-optimal-allocation}
\end{equation}
whereas the allocation minimizing separate-estimation risk is
\begin{equation}
 n_a^{\mathrm{sep}}
 =N\frac{\sqrt{\pi_a}\sigma_a}{\sum_b\sqrt{\pi_b}\sigma_b},
 \qquad
 \min V_{\mathrm{sep}}
 =\frac{D}{N}\left(\sum_a\sqrt{\pi_a}\sigma_a\right)^2.
 \label{eq:sep-optimal-allocation}
\end{equation}
If each architecture may choose its own allocation, full sharing has lower optimized risk exactly when
\begin{equation}
 B_\pi
 <\frac{D}{N}
 \left[
 \left(\sum_a\sqrt{\pi_a}\sigma_a\right)^2
 -\left(\sum_a\pi_a\sigma_a\right)^2
 \right].
 \label{eq:optimized-allocation-threshold}
\end{equation}
\end{corollary}
For completeness, minimizing $\sum_ac_a/n_a$ with $c_a>0$, $n_a\ge d$, and $\sum_an_a=N>Md$ gives $n_a=\max\{d,\sqrt{c_a/t}\}$, where $t>0$ is chosen to meet the budget.  The Karush--Kuhn--Tucker (KKT) conditions are the stationarity and complementary-slackness conditions for this constrained convex problem: a coordinate above its floor has derivative $-c_a/n_a^2+t=0$; a coordinate at the floor need not satisfy that equality.  If $N=Md$, every $n_a=d$.  Integer allocations require a separate discrete solution.

This result concerns data acquisition or replay allocation.  

\subsection{Any linear sharing architecture is a projection problem}
A large class of tying and conditioning rules can be described by a linear subspace of admissible role-wise functions.  The following theorem gives an exact bias--dimension decomposition.

\begin{theorem}[Exact risk of a linear sharing subspace]
\label{thm:linear-sharing-subspace}
Let $J$ role cells have true parameter vectors $\theta_1,\ldots,\theta_J\in\R^D$.  Suppose their unconstrained stacked estimator satisfies
\begin{equation}
 \widehat\theta=\theta+\zeta,
 \qquad
 \E\zeta=0,
 \qquad
 \operatorname{Cov}(\zeta)=vI_{JD}.
 \label{eq:subspace-observation}
\end{equation}
Here stacking means concatenating the $J$ length-$D$ vectors in $\R^{JD}$, and $\operatorname{Cov}$ denotes covariance.  The subspace must be chosen before observing this fitting noise.  Let $\mathcal U\subseteq\R^{JD}$ be a fixed linear sharing subspace, let $P_{\mathcal U}$ be its Euclidean orthogonal projector, and fit $\widehat\theta_{\mathcal U}=P_{\mathcal U}\widehat\theta$.  Define the role-averaged squared parameter error by $\Risk(\mathcal U)=J^{-1}\E\norm{\widehat\theta_{\mathcal U}-\theta}_2^2$.  Then
\begin{equation}
 \Risk(\mathcal U)
 =\frac1J\norm{(I-P_{\mathcal U})\theta}_2^2
 +\frac{v}{J}\dim(\mathcal U).
 \label{eq:subspace-risk}
\end{equation}
\end{theorem}
The first term is exactly the squared approximation cost imposed by the fixed subspace; the second charges one isotropic-noise degree of freedom for every fitted scalar parameter.  The partition result in \cref{thm:partition-frontier} is the special case in which $\mathcal U$ contains arrays constant inside every expert group.  More generally, tied layers, shared cores, additive depth embeddings, and fixed low-dimensional modulation bases correspond to different choices of $\mathcal U$.

\subsection{An exact cross-depth tying frontier}
We now instantiate \cref{thm:linear-sharing-subspace} for architectures that reuse experts across depth.  Index a semantic role by $a\in[M]$ and depth by $\ell\in[L]$.  The true local operator is $A_{a\ell}\in\R^{p\times d}$, each cell has an orthogonal design with $n\ge d$ samples and homoscedastic noise $\sigma^2$, with noise independent across cells, and
\begin{equation}
 V=\frac{pd\sigma^2}{n}
 \label{eq:cross-depth-cell-variance}
\end{equation}
would be the expected risk of one cell-specific OLS fit.  Define
\begin{align}
 \bar A&=\frac1{ML}\sum_{a,\ell}A_{a\ell},
 &\bar A_{a\cdot}&=\frac1L\sum_\ell A_{a\ell},
 &\bar A_{\cdot\ell}&=\frac1M\sum_aA_{a\ell},
 \label{eq:cross-depth-means}\\
 B_{\mathrm{role}}&=\frac1M\sum_a\norm{\bar A_{a\cdot}-\bar A}_F^2,
 &B_{\mathrm{depth}}&=\frac1L\sum_\ell\norm{\bar A_{\cdot\ell}-\bar A}_F^2,
 \label{eq:cross-depth-main-effects}\\
 R_{a\ell}^{\mathrm{int}}
 &=A_{a\ell}-\bar A_{a\cdot}-\bar A_{\cdot\ell}+\bar A,
 \nonumber\\
 B_{\mathrm{int}}&=\frac1{ML}\sum_{a,\ell}
 \norm{R_{a\ell}^{\mathrm{int}}}_F^2.
 \label{eq:cross-depth-interaction}
\end{align}
The dot in a subscript means averaging over that index.  The role effect changes $a$ only; the depth effect changes $\ell$ only.  The interaction $R_{a\ell}^{\mathrm{int}}$ is what remains when the two effects cannot explain a cell jointly.  Under the balanced Frobenius inner product, these components are orthogonal (their pairwise inner products sum to zero).  We use this as an algebraic two-way analysis of variance (ANOVA) decomposition.

\begin{theorem}[Exact risks for cross-layer reuse patterns]
\label{thm:cross-depth-frontier}
Under the balanced cell model above, the exact role--cell-averaged excess prediction risks are those in \cref{tab:cross-depth-risks}.
\begin{table}[H]
\centering\small
\caption{Exact cross-depth sharing risks.  ``Role-tied'' means that the same semantic expert is reused at every depth; ``additive depth-conditioned'' allows a semantic expert main effect plus a depth main effect but no role--depth interaction.}
\label{tab:cross-depth-risks}
\setlength{\tabcolsep}{3.8pt}
\begin{tabular}{p{0.25\textwidth}p{0.25\textwidth}p{0.14\textwidth}p{0.25\textwidth}}
\toprule
Architecture subspace & Approximation term & Dimension & Exact risk \\
\midrule
Cell-specific layer-local experts & $0$ & $MLD$ & $V$ \\
One globally tied expert & $B_{\mathrm{role}}+B_{\mathrm{depth}}+B_{\mathrm{int}}$ & $D$ & $B_{\mathrm{role}}+B_{\mathrm{depth}}+B_{\mathrm{int}}+V/(ML)$ \\
Role-tied across depth & $B_{\mathrm{depth}}+B_{\mathrm{int}}$ & $MD$ & $B_{\mathrm{depth}}+B_{\mathrm{int}}+V/L$ \\
One role-shared expert per depth & $B_{\mathrm{role}}+B_{\mathrm{int}}$ & $LD$ & $B_{\mathrm{role}}+B_{\mathrm{int}}+V/M$ \\
Additive depth-conditioned reuse & $B_{\mathrm{int}}$ & $(M+L-1)D$ & $B_{\mathrm{int}}+V(M+L-1)/(ML)$ \\
\bottomrule
\end{tabular}
\end{table}
Consequently:
\begin{align}
 \text{role-tied beats cell-specific}
 &\iff B_{\mathrm{depth}}+B_{\mathrm{int}}
 <V\left(1-\frac1L\right),
 \label{eq:role-tied-threshold}\\
 \text{additive beats role-tied}
 &\iff B_{\mathrm{depth}}
 >V\frac{L-1}{ML},
 \label{eq:depth-conditioning-threshold}\\
 \text{additive beats cell-specific}
 &\iff B_{\mathrm{int}}
 <V\frac{(M-1)(L-1)}{ML}.
 \label{eq:additive-vs-separate-threshold}
\end{align}
\end{theorem}

These formulas separate three reasons that cross-layer reuse may fail.  A globally tied pool pays semantic-role variation, depth variation, and their interaction.  Tying the same semantic expert across depth removes the role main effect but still pays depth shift and role--depth interaction.  A depth-conditioned shared expert removes both main effects; only the interaction remains.  Within this additive operator model, depth conditioning is useful precisely when the removed depth effect exceeds the variance cost of its additional $(L-1)D$ scalar degrees of freedom.  Interpreting a nonlinear depth embedding through this subspace requires an explicit realization or local-linearization argument.  If $B_{\mathrm{int}}$ is large, neither global tying nor additive depth conditioning is sufficient, and layer-specific residual capacity is statistically justified.

The same calculation extends to several architectural axes.
\begin{corollary}[Multi-axis additive reuse]
\label{cor:multi-axis-additive}
Suppose role cells are indexed by $r$ factors of sizes $m_1,\ldots,m_r$, with $J=\prod_{k=1}^rm_k$ balanced cells.  Let $\mathcal U_{\mathrm{add}}$ contain a grand mean plus one zero-sum main effect for every factor, and define the normalized interaction residual
\begin{equation}
 B_{\ge2}=\frac1J\norm{(I-P_{\mathcal U_{\mathrm{add}}})\theta}_2^2.
 \label{eq:multi-axis-residual-definition}
\end{equation}
It contains all interactions of order at least two.  Then
\begin{equation}
 \Risk(\mathcal U_{\mathrm{add}})
 =B_{\ge2}
 +\frac{V}{J}\left[1+\sum_{k=1}^r(m_k-1)\right].
 \label{eq:multi-axis-additive-risk}
\end{equation}
\end{corollary}
This covers, for example, semantic role $\times$ layer $\times$ modality or role $\times$ reasoning position $\times$ tool state.  For the declared linear subspaces, adding a factor-specific correction is worthwhile only when the removed component exceeds its estimation price.  \Cref{cor:nested-sharing-frontier} gives the covariance-weighted version without balanced isotropic noise.

\subsection{Proofs}
\begin{proof}[Proof of \cref{thm:hetero-frontier}]
Orthogonality gives $\widehat A_a=A_a+E_a$, with centered, uncorrelated entries of variance $v_a$ and independent error matrices across roles.  Let $\bar E_\pi=\sum_b\pi_bE_b$.  Then
\begin{equation}
 \widehat A_a^{(\lambda)}-A_a
 =\lambda(\bar A_\pi-A_a)+(1-\lambda)E_a+\lambda\bar E_\pi.
 \label{eq:hetero-error-decomposition}
\end{equation}
The role-averaged squared bias is $\lambda^2B_\pi$.  For each scalar coefficient, the weighted variance is
\begin{align}
 &\sum_a\pi_a\operatorname{Var}\!\bigl((1-\lambda)E_a+\lambda\bar E_\pi\bigr)\nonumber\\
 &\quad=(1-\lambda)^2\sum_a\pi_av_a
 +(2\lambda-\lambda^2)\sum_a\pi_a^2v_a.
 \label{eq:hetero-variance-calculation}
\end{align}
Multiplying by $D$ and adding the bias gives \cref{eq:hetero-partial-risk}.  Setting $\lambda=0,1$, subtracting the endpoint risks, and minimizing the convex quadratic prove the remaining statements.
\end{proof}

\begin{proof}[Proof of \cref{cor:architecture-dependent-sampling}]
Minimizing $\sum_a c_a/n_a$ subject to $\sum_an_a=N$ gives $n_a\propto\sqrt{c_a}$ by Lagrange multipliers, or equivalently by Cauchy--Schwarz.  For pooled estimation, $c_a=\pi_a^2\sigma_a^2$, yielding $n_a\propto\pi_a\sigma_a$ and the first minimum.  For separate estimation, $c_a=\pi_a\sigma_a^2$, yielding $n_a\propto\sqrt{\pi_a}\sigma_a$ and the second.  Comparing the two optimized endpoint risks gives \cref{eq:optimized-allocation-threshold}.
\end{proof}

\begin{proof}[Proof of \cref{thm:linear-sharing-subspace}]
Write $P=P_{\mathcal U}$.  Because $P\theta-\theta=-(I-P)\theta$ is orthogonal to $P\zeta$ and $\E\zeta=0$,
\begin{equation}
 \E\norm{P\widehat\theta-\theta}_2^2
 =\norm{(I-P)\theta}_2^2+\E\norm{P\zeta}_2^2.
 \label{eq:subspace-bias-variance}
\end{equation}
Moreover,
\begin{equation}
 \E\norm{P\zeta}_2^2
 =\tr(P\,vI\,P^\top)
 =v\tr(P)
 =v\dim(\mathcal U).
 \label{eq:subspace-noise-trace}
\end{equation}
Divide by $J$.
\end{proof}

\begin{proof}[Proof of \cref{thm:cross-depth-frontier}]
Stack the $ML$ operator vectors and apply \cref{thm:linear-sharing-subspace} with scalar noise variance $v=\sigma^2/n$ and block dimension $D=pd$.  The cell-specific, globally tied, role-tied, depth-specific, and additive subspaces have dimensions $MLD$, $D$, $MD$, $LD$, and $(M+L-1)D$, respectively.  Orthogonal two-way ANOVA gives
\begin{equation}
 \frac1{ML}\sum_{a,\ell}\norm{A_{a\ell}-\bar A}_F^2
 =B_{\mathrm{role}}+B_{\mathrm{depth}}+B_{\mathrm{int}},
 \label{eq:two-way-anova}
\end{equation}
while projection onto role means leaves $B_{\mathrm{depth}}+B_{\mathrm{int}}$, projection onto depth means leaves $B_{\mathrm{role}}+B_{\mathrm{int}}$, and projection onto the additive subspace leaves $B_{\mathrm{int}}$.  Substituting each residual and dimension into \cref{eq:subspace-risk} gives the table.  Pairwise subtraction gives \cref{eq:role-tied-threshold,eq:depth-conditioning-threshold,eq:additive-vs-separate-threshold}.
\end{proof}

\begin{proof}[Proof of \cref{cor:multi-axis-additive}]
The additive subspace has one grand-mean block and $(m_k-1)$ independent main-effect blocks for axis $k$, hence dimension $D[1+\sum_k(m_k-1)]$.  Its orthogonal residual consists of interactions of order at least two.  Apply \cref{thm:linear-sharing-subspace}.
\end{proof}

\section{Architecture-Specific Functional Routing Bounds}
\label{app:architecture-routing}

\paragraph{Overview.}
A routing change matters through the event output it changes, not just expert-index disagreement. We bound the generic cost $\rho_j$ from \cref{eq:app-functional-plan-cost} for additive soft and uniform top-$k$ gates, capacity-limited row-separable assignments, additive hierarchies, and preselected chains. Full soft dispatch also changes expert inputs and requires the additional term in \cref{thm:full-soft-dispatch}. Each bound applies only to its stated interface. The transport notation follows \citet{peyre2019computational}; the inequalities are proved directly.
Sections~\ref*{app:architecture-routing}.1--\ref*{app:architecture-routing}.5 treat the individual routing interfaces; Section~\ref*{app:architecture-routing}.6 summarizes their cost terms, and Section~\ref*{app:architecture-routing}.7 supplies the remaining proofs.

\subsection{Additive soft and top-k gates as transport over expert functions}
Fix a common expert input $u$ and let $v_e(u)$ be the routed residual produced by expert $e$ in a normed vector space.  For normalized gates $g,h\in\Delta_K:=\{w\in\R^K:w_e\ge0,\ \sum_ew_e=1\}$, define the discrete coupling set
\begin{equation}
 \Pi(g,h)=\left\{\gamma\in\R_+^{K\times K}:
 \sum_{e'}\gamma_{ee'}=g_e,
 \ \sum_e\gamma_{ee'}=h_{e'}\right\},
 \label{eq:gate-coupling-set}
\end{equation}
and the input-dependent functional transport cost
\begin{equation}
 \mathsf W_u(g,h)
 =\min_{\gamma\in\Pi(g,h)}
 \sum_{e,e'}\gamma_{ee'}c_u(e,e'),
 \qquad
 c_u(e,e')=\norm{v_e(u)-v_{e'}(u)}.
 \label{eq:functional-gate-transport}
\end{equation}

\begin{theorem}[Functional transport bound for additive routing]
\label{thm:functional-transport-gate}
Let
\begin{equation}
 M(u,g)=B(u)+\sum_{e=1}^Kg_ev_e(u),
 \label{eq:additive-route-output}
\end{equation}
where $B(u)$ contains every dense or always-active shared path and where changing the gate does not change $u$ or the collection $\{v_e(u)\}$.  Then
\begin{equation}
 \norm{M(u,g)-M(u,h)}
 \le \mathsf W_u(g,h)
 \le \frac{D_u}{2}\norm{g-h}_1,
 \qquad
 D_u=\max_{e,e'}c_u(e,e').
 \label{eq:transport-gate-bound}
\end{equation}
The first inequality can be strictly sharper than the diameter--total-variation bound when most mismatched mass moves between functionally similar experts.
\end{theorem}

The shared term $B(u)$ cancels exactly.  Thus an always-active expert can improve the event while contributing nothing to the discrepancy between two conditional gates at the same input.  Conversely, a dominant shared path can make natural task accuracy insensitive to the routed residual, which is one mechanism behind the causal non-identifiability result in \cref{prop:route-label-separation}.

\paragraph{Named architecture instances.}
Mixtral combines two selected SwiGLU experts using a softmax over the selected logits \citep{jiang2024mixtral}; this is weighted top-$2$, not the equal-$1/2$ special case below. DeepSeekMoE combines always-active shared experts with weighted routed experts \citep{dai2024deepseekmoe}, corresponding to $B(u)$ plus the conditional sum in \cref{eq:additive-route-output}. The actual gate normalization and selected mass must be retained; sub-probability gates use the null-atom augmentation below. Shared paths cancel only between two gates of the same fixed model at the same input, not between separately trained models. 

For uniform top-k routing, the transport cost becomes a matching problem.
\begin{corollary}[Uniform top-k functional matching]
\label{cor:topk-matching}
Let $S=\{s_1,\ldots,s_k\}$ and $T=\{t_1,\ldots,t_k\}$ be two $k$-element expert sets, and let both gates assign mass $1/k$ to every selected expert.  Then
\begin{equation}
 \mathsf W_u(g_S,g_T)
 =\frac1k\min_{\pi\in\mathfrak S_k}
 \sum_{r=1}^k
 \norm{v_{s_r}(u)-v_{t_{\pi(r)}}(u)}.
 \label{eq:topk-matching-cost}
\end{equation}
Consequently, two disjoint top-k supports can have small functional route cost, while replacing one selected expert can be severe if the substitute is functionally distant.
\end{corollary}
Here $\mathfrak S_k$ is the set of the $k!$ permutations of $\{1,\ldots,k\}$, so the minimum matches each selected expert to one expert in the other set.  The formula is invariant to enumeration.  For nonuniform within-support weights, use the general transport problem in \cref{eq:functional-gate-transport}; support overlap alone does not determine the route cost.

\subsection{Capacity drops and row-separable expert-choice assignment}
Token-choice and expert-choice routers may drop, pad, or reroute mass under capacity constraints \citep{zhou2022expertchoice,lewis2021base}.  A zero-residual ``drop'' atom makes the associated functional cost explicit.  For a sub-probability gate $g\in\R_+^K$ with $\sum_eg_e\le1$, introduce atom $0$ with $v_0(u)=0$ and augmented mass
\begin{equation}
 \widetilde g_0=1-\sum_{e=1}^Kg_e,
 \qquad
 \widetilde g_e=g_e\quad(e\ge1).
 \label{eq:null-expert-augmentation}
\end{equation}
Applying \cref{thm:functional-transport-gate} to $\widetilde g$ and $\widetilde h$ charges
\begin{equation}
 c_u(e,0)=\norm{v_e(u)}
 \label{eq:null-expert-cost}
\end{equation}
for dropping the contribution of expert $e$.  Hence a drop-rate statistic is insufficient unless it is weighted by the magnitude of the dropped residual.

For $T$ token or slot states $u_t$, suppose an assignment matrix acts row-separably after the $u_t$ are fixed, so row $t$ produces $M(u_t,g_t)$ as in \cref{eq:additive-route-output}.  If matrices $G,H$ induce row gates $g_t,h_t$, then under the sum norm on the batch output,
\begin{equation}
 \sum_{t=1}^T\norm{M(u_t,g_t)-M(u_t,h_t)}
 \le\sum_{t=1}^T\mathsf W_{u_t}(\widetilde g_t,\widetilde h_t).
 \label{eq:batch-assignment-transport}
\end{equation}
The same statement divided by $T$ holds for the average norm, while a maximum norm is controlled by the maximum row cost.  Joint capacity constraints may couple how the rows are selected, but no row-independence assumption is used once the two assignment matrices are fixed.  Dispatch mechanisms that change expert-slot inputs use the generic fixed-plan discrepancy; \cref{thm:full-soft-dispatch} gives a bound when dispatch sensitivity is controlled.

\subsection{Full soft dispatch: expert inputs may also change}
Full Soft MoE constructs weighted token combinations as expert inputs before combining their outputs \citep{puigcerver2023softmoe}.  The common-input gate bound alone is therefore insufficient.  Let $x_1,\ldots,x_T\in\R^d$ be fixed token states and let there be $Q$ \emph{slots}, meaning $Q$ intermediate vectors sent to designated expert calls.  A dispatch matrix $D\in\R_+^{T\times Q}$ has column sums one; a combine matrix $C\in\R_+^{T\times Q}$ has row sums one.  With fixed slot-to-expert assignment $e(q)$, define
\begin{equation}
 u_q(D)=\sum_tD_{tq}x_t,\quad
 v_q(D)=E_{e(q)}(u_q(D)),\quad
 y_t(D,C)=\sum_qC_{tq}v_q(D).
 \label{eq:full-soft-dispatch-map}
\end{equation}
Let $(D',C')$ be another pair with the same normalizations and the same expert functions.  Expert $E_{e(q)}$ is assumed $L_q$-Lipschitz on the convex hull of the token states, using the respective input and output norms.  The convex hull consists of all their nonnegative weighted averages with total weight one.

Let $\mathsf W_D^{\mathrm{out}}$ be discrete transport over slot outputs $\{v_q(D)\}$, with ground cost $\|v_q(D)-v_{q'}(D)\|$.  Let $\mathsf W_X^{\mathrm{in}}$ be transport over the token states, with cost $\|x_t-x_{t'}\|$.  Define $\bar C'_q=T^{-1}\sum_tC'_{tq}$, the average output-token weight assigned to slot $q$.

\begin{theorem}[Dispatch-plus-combine discrepancy for full soft routing]
\label{thm:full-soft-dispatch}
Under the normalization and Lipschitz assumptions above,
\begin{align}
 \frac1T\sum_t\|y_t(D,C)-y_t(D',C')\|
 &\le\frac1T\sum_t
       \mathsf W_D^{\mathrm{out}}(C_{t,:},C'_{t,:})\nonumber\\
 &\quad+\sum_q\bar C'_q L_q
       \mathsf W_X^{\mathrm{in}}(D_{:,q},D'_{:,q}).
 \label{eq:full-soft-dispatch-bound}
\end{align}
The reverse insertion gives another valid bound using output geometry at $D'$ and average combine weights from $C$; their minimum is valid.  If $D_X=\max_{t,t'}\|x_t-x_{t'}\|$, each input transport term is at most $D_X\|D_{:,q}-D'_{:,q}\|_1/2$.
\end{theorem}
\begin{proof}
Insert $y_t(D,C')$.  The first difference keeps slot outputs fixed and is bounded by output transport.  The second difference is at most
$\sum_qC'_{tq}\|E_{e(q)}(u_q(D))-E_{e(q)}(u_q(D'))\|$.
Use the expert's Lipschitz bound and the barycenter transport bound
$\|u_q(D)-u_q(D')\|\le\mathsf W_X^{\mathrm{in}}(D_{:,q},D'_{:,q})$.
Average over $t$.  Reversing the insertion proves the alternative; the diameter inequality follows from \cref{thm:functional-transport-gate} applied to token vectors.
\end{proof}
This extends the routing analysis beyond the common-expert-input assumption.  If dispatch is unchanged, the second term vanishes.  If the slot-to-expert assignment or the token states also change, an additional expert-substitution or input-state discrepancy must be included.  The theorem bounds the route component of a fixed trained event, not the learned expert approximation error already charged elsewhere.

\subsection{Hierarchical additive routing separates coarse and fine errors}
Consider a two-level additive hierarchy with groups $r\in[G]$, within-group experts $e\in[K_r]$, outer gate $p\in\Delta_G$, and conditional inner gates $q_r\in\Delta_{K_r}$.  All vectors below are evaluated at the same fixed input.  Define
\begin{equation}
 m_r(q_r)=\sum_eq_{r,e}v_{r,e},
 \qquad
 M(p,q)=\sum_rp_rm_r(q_r).
 \label{eq:hierarchical-output}
\end{equation}

\begin{proposition}[Hierarchical functional route decomposition]
\label{prop:hierarchical-route-decomposition}
For two hierarchical routes $(p,q)$ and $(p',q')$,
\begin{equation}
 \norm{M(p,q)-M(p',q')}
 \le
 \mathsf W^{\mathrm{out}}_{q}(p,p')
 +\sum_{r=1}^Gp'_r\mathsf W^{\mathrm{in}}_r(q_r,q'_r),
 \label{eq:hierarchical-route-bound}
\end{equation}
where the outer transport uses
\begin{equation}
 c_q^{\mathrm{out}}(r,r')
 =\norm{m_r(q_r)-m_{r'}(q_{r'})},
 \label{eq:hierarchical-outer-cost}
\end{equation}
and the inner transport in group $r$ uses
\begin{equation}
 c_r^{\mathrm{in}}(e,e')=\norm{v_{r,e}-v_{r,e'}}.
 \label{eq:hierarchical-inner-cost}
\end{equation}
The reverse insertion gives a second valid bound with $(p,q)$ and $(p',q')$ interchanged; their minimum may be used.
\end{proposition}
The first term measures a coarse group mistake, and the second measures fine selection errors after the outer mass is fixed.  A single flat route-accuracy number cannot reveal which level consumes the reuse gain.

\subsection{Preselected sequential expert chains have within-event amplification}
Sequential architectures call experts one after another inside one routed event \citep{wang2025chainexperts}.  The theorem here concerns two preselected, equal-length sequences.  If a later expert choice adapts to an earlier expert output, each adaptive choice must instead be made a separate routed event or the entire adaptive policy must be compared through the generic plan-level distance.

Let stage $s\in[Q]$ map metric space $(\cX_{s-1},d_{s-1})$ to $(\cX_s,d_s)$ through $F_{s,e}$.  Consider
\begin{equation}
 \gamma=(e_1,\ldots,e_Q),
 \qquad
 \gamma'=(e'_1,\ldots,e'_Q),
 \label{eq:sequential-routes}
\end{equation}
and suppose every suffix map $F_{r,e_r}$ from the first sequence is $L_r$-Lipschitz on the reached hybrid region.  Let
\begin{equation}
 x'_{s-1}=F_{s-1,e'_{s-1}}\circ\cdots\circ F_{1,e'_1}(x),
 \qquad
 A^{\mathrm{in}}_s=\prod_{r=s+1}^{Q}L_r,
 \label{eq:sequential-prefix-and-weight}
\end{equation}
with the empty prefix and product interpreted as the identity and one.

\begin{theorem}[Sequential route-plan discrepancy]
\label{thm:sequential-route-discrepancy}
The output maps $F_\gamma=F_{Q,e_Q}\circ\cdots\circ F_{1,e_1}$ and $F_{\gamma'}$ satisfy
\begin{equation}
 d_Q(F_\gamma(x),F_{\gamma'}(x))
 \le\sum_{s=1}^{Q}A_s^{\mathrm{in}}
 d_s\!\left(F_{s,e_s}(x'_{s-1}),F_{s,e'_s}(x'_{s-1})\right).
 \label{eq:sequential-route-bound}
\end{equation}
\end{theorem}
Thus one early substitution can have a large cost in an expansive chain, while several index differences can be harmless when the substituted experts agree on the reached hybrid states.  Variable-length chains can be embedded in a common maximum length by padding a terminated chain with an explicit identity/no-op transition on the terminal state.  A zero map is generally not valid padding because it changes the state.

Combining \cref{thm:sequential-route-discrepancy} with the outer amplification $A_j$ in \cref{thm:terminal-certificate} produces two scales of sensitivity: $A_s^{\mathrm{in}}$ measures propagation inside one routed event, while $A_j$ measures propagation through later events.  For a call at within-event position $s$ of global event $j$, the natural terminal importance is their product.

\subsection{Architecture-to-cost dictionary}
\begin{table}[H]
\centering\small
\caption{Architecture-specific quantities that upper-bound the generic functional routing term under their stated structural assumptions.}
\label{tab:architecture-cost-dictionary}
\setlength{\tabcolsep}{3.5pt}
\begin{tabular}{p{0.24\textwidth}p{0.60\textwidth}}
\toprule
Architecture & Functional route-cost proxy \\
\midrule
Additive soft gate & $\mathsf W_u(g,h)$ over expert outputs evaluated at one common input \\
Uniform top-k & Minimum bipartite matching cost in \cref{eq:topk-matching-cost} \\
Capacity-limited additive routing & Augmented transport through a zero-residual atom, charging $\norm{v_e(u)}$ for dropped mass \\
Row-separable expert-choice assignment & Sum, average, or maximum of row-wise transport costs after token inputs are fixed \\
Full soft dispatch & Output transport plus Lipschitz-weighted input-dispatch transport in \cref{eq:full-soft-dispatch-bound} \\
Hierarchical additive MoE & Coarse outer transport plus conditional within-group transport \\
Preselected sequential experts & Suffix-amplified per-position substitutions in \cref{eq:sequential-route-bound} \\
Shared-expert residual block & Same conditional cost after exact cancellation of route-independent paths \\
General dispatch/adaptive program & Direct plan-level transition distance $r_j$; additional dispatch or policy sensitivity is required for a finer bound \\
\bottomrule
\end{tabular}
\end{table}
These are upper bounds on transition discrepancy.  They can be estimated from counterfactual forward passes while preserving the active-expert budget.

\subsection{Proofs}
\begin{proof}[Proof of \cref{thm:functional-transport-gate}]
For any $\gamma\in\Pi(g,h)$,
\begin{equation}
 \sum_e(g_e-h_e)v_e
 =\sum_{e,e'}\gamma_{ee'}(v_e-v_{e'}).
 \label{eq:coupling-vector-identity}
\end{equation}
The triangle inequality gives
\begin{equation}
 \norm{\sum_e(g_e-h_e)v_e}
 \le\sum_{e,e'}\gamma_{ee'}c_u(e,e').
 \label{eq:coupling-vector-bound}
\end{equation}
Minimizing over couplings proves the first inequality.  There exists a coupling that leaves the common mass $\min(g_e,h_e)$ on the diagonal and transports only total mass $\TV(g,h)=\norm{g-h}_1/2$ off the diagonal.  Every moved unit costs at most $D_u$, proving the second inequality.
\end{proof}

\begin{proof}[Proof of \cref{cor:topk-matching}]
After restricting to the two $k$-point uniform measures and multiplying a coupling by $k$, the feasible set consists of nonnegative matrices whose row sums and column sums all equal one.  This is the \emph{doubly stochastic polytope}: a bounded set described by linear equality and inequality constraints.  An extreme point cannot be a nontrivial convex combination of two other feasible points.  Such points are permutation matrices (one entry equal to one per row and column), and a linear objective attains a minimum at one of them; this is the standard doubly stochastic matrix characterization \citep{peyre2019computational}.  Dividing by $k$ gives \cref{eq:topk-matching-cost}.  Shared atoms simply admit zero-cost matched pairs.
\end{proof}

\begin{proof}[Proof of \cref{prop:hierarchical-route-decomposition}]
Insert $M(p',q)$:
\begin{equation}
 \norm{M(p,q)-M(p',q')}
 \le\norm{M(p,q)-M(p',q)}
 +\norm{M(p',q)-M(p',q')}.
 \label{eq:hierarchical-insertion}
\end{equation}
Apply \cref{thm:functional-transport-gate} to the outer mixtures of vectors $m_r(q_r)$ for the first term.  For the second term, use the triangle inequality over groups and apply the same theorem inside each group.  The reverse insertion is identical with the two routes exchanged.
\end{proof}

\begin{proof}[Proof of \cref{thm:sequential-route-discrepancy}]
For $s=0,\ldots,Q$, let $H_s$ use the prefix $(e'_1,\ldots,e'_s)$ from $\gamma'$ and the suffix $(e_{s+1},\ldots,e_Q)$ from $\gamma$.  Thus $H_0=F_\gamma$ and $H_Q=F_{\gamma'}$.  Adjacent hybrids $H_{s-1}$ and $H_s$ receive the same state $x'_{s-1}$ at stage $s$, differ only between $F_{s,e_s}$ and $F_{s,e'_s}$, and share the suffix $(e_{s+1},\ldots,e_Q)$.  That suffix is $A_s^{\mathrm{in}}$-Lipschitz from $d_s$ to $d_Q$.  The triangle inequality over $H_0,\ldots,H_Q$ proves \cref{eq:sequential-route-bound}.
\end{proof}

\section{Endogenous Routing Feedback and Margin-Stable Traces}
\label{app:endogenous-routing}

\paragraph{Overview.}
An early state error can change a later route and thereby create more error. We give sufficient support-stability conditions and probabilistic bounds for this feedback, retaining both router margins and expert-output geometry. Conditional margin control and weaker marginal control yield different guarantees.
Sections~\ref*{app:endogenous-routing}.1--\ref*{app:endogenous-routing}.2 treat soft-routing sensitivity and hard-support stability; Sections~\ref*{app:endogenous-routing}.3--\ref*{app:endogenous-routing}.4 distinguish conditional from marginal margin bounds, followed by implications and proofs in Sections~\ref*{app:endogenous-routing}.5--\ref*{app:endogenous-routing}.6.

\subsection{Soft additive routing produces an additional Lipschitz term}
Consider an additive routed event
\begin{equation}
 F(s)=M(s,g(s))
 =B(s)+\sum_{e=1}^Kg_e(s)v_e(s),
 \label{eq:closed-loop-soft-event}
\end{equation}
with $g(s)\in\Delta_K$.  The output is in a normed vector space and its distance is the norm of a difference; when one metric $d$ is used on both sides, this compatibility is assumed.  Otherwise use separate input metric and output norm.  Assume that for every fixed gate $h$,
\begin{equation}
 d(M(s,h),M(s',h))\le L_xd(s,s').
 \label{eq:fixed-gate-state-lipschitz}
\end{equation}
At state $s'$, define the functional transport cost $\mathsf W_{s'}$ as in \cref{eq:functional-gate-transport}, using the collection $\{v_e(s')\}$ at the common input $s'$.

\begin{proposition}[Closed-loop sensitivity of an additive soft MoE event]
\label{prop:soft-router-closed-loop}
If the router is functionally transport-Lipschitz,
\begin{equation}
 \mathsf W_{s'}(g(s),g(s'))\le L_Wd(s,s'),
 \label{eq:functional-router-lipschitz}
\end{equation}
then
\begin{equation}
 d(F(s),F(s'))\le(L_x+L_W)d(s,s').
 \label{eq:soft-closed-loop-lipschitz}
\end{equation}
In particular, if $\norm{g(s)-g(s')}_1\le L_gd(s,s')$ and the expert-output diameter at the reached states is uniformly at most $D_{\max}$, one may take
\begin{equation}
 L_W\le\frac12D_{\max}L_g.
 \label{eq:soft-router-diameter-lipschitz}
\end{equation}
\end{proposition}
Thus soft routing adds a sensitivity proportional to both gate movement and functional expert separation.  This result applies to the additive common-input form; a dispatch mechanism whose gate changes expert inputs requires an additional dispatch-sensitivity term.

\subsection{Hard top-k support is locally stable inside a margin tube}
Assume the reference and learned router coordinates have been placed in one fixed canonical role alignment (or already share the same expert coordinates).  Let $z:\cS\to\R^K$ be reference router logits, fix $1\le k<K$, and let $\mathcal I_k(z)$ denote the indices of the $k$ largest coordinates under a fixed tie rule.  Write $z_{(1)}\ge\cdots\ge z_{(K)}$ and define the top-k support margin
\begin{equation}
 \gamma_k(s)=z_{(k)}(s)-z_{(k+1)}(s),
 \label{eq:topk-margin}
\end{equation}
with $k=1$ giving the ordinary winner margin.

\begin{lemma}[Top-$k$ support-margin stability]
\label{lem:topk-margin-stability}
If another logit vector $\widetilde z$ satisfies
\begin{equation}
 \norm{\widetilde z-z(s)}_\infty<\frac12\gamma_k(s),
 \label{eq:topk-margin-condition}
\end{equation}
then $\mathcal I_k(\widetilde z)=\mathcal I_k(z(s))$.  More generally, if a learned router $\widehat z$ satisfies
\begin{equation}
 \norm{\widehat z(s')-z(s')}_\infty\le\beta,
 \qquad
 \norm{z(s')-z(s)}_\infty\le L_Rd(s,s'),
 \label{eq:router-perturbation-decomposition}
\end{equation}
then the supports agree whenever
\begin{equation}
 \gamma_k(s)>2\bigl(\beta+L_Rd(s,s')\bigr).
 \label{eq:route-margin-tube}
\end{equation}
\end{lemma}
For unit-gain top-$1$ or uniform hard top-k, support agreement determines the entire gate.  For top-k followed by nonuniform normalized weights, the lemma controls only the support; any remaining within-support weight discrepancy can be bounded by the transport result in \cref{thm:functional-transport-gate}.  Top-$1$ with a selected softmax gain also retains this weight variation; \cref{prop:explicit-block-envelopes} differentiates that gain explicitly.

The margin tube yields a pathwise no-support-cascade condition.  Let the reference support at event $j$ be $\mathcal I_k(z_j(S_j))$ and the learned support be $\mathcal I_k(\widehat z_j(\widehat S_j))$.  Assume the reference logits are $L_{R,j}$-Lipschitz, and that
\begin{equation}
 \norm{\widehat z_j(s)-z_j(s)}_\infty\le\beta_j
 \label{eq:pathwise-router-model-gap}
\end{equation}
holds on the reached tube.  Assume every reference transition with its complete route (support and any route weights) frozen is $L_j$-Lipschitz, uniformly over the reference routes reached.  Finally, suppose that whenever the learned and reference supports agree, the same-state event discrepancy--including model approximation and, for nonuniform top-k, any within-support weight difference--is at most $\eta_j$.  Define
\begin{equation}
 \bar e_0=d(\widehat S_0,S_0),
 \qquad
 \bar e_{j+1}=L_j\bar e_j+\eta_j.
 \label{eq:no-cascade-envelope}
\end{equation}

\begin{theorem}[Margin-certified absence of support cascades]
\label{thm:no-route-cascade}
If, along the reference trace,
\begin{equation}
 \gamma_{k,j}(S_j)
 >2\bigl(\beta_j+L_{R,j}\bar e_j\bigr)
 \qquad\text{for every }j,
 \label{eq:no-cascade-margin-condition}
\end{equation}
then the learned and reference top-k supports agree at every event and
\begin{equation}
 d(\widehat S_j,S_j)\le\bar e_j
 \qquad\text{for all }j.
 \label{eq:no-cascade-error-bound}
\end{equation}
For unit-gain top-$1$ and uniform hard top-k, the corresponding discrete route sequence is unchanged, so no support-change routing penalty is incurred.  For nonuniform gates, including selected-softmax-gain top-$1$, same-support weight error remains inside $\eta_j$.
\end{theorem}
The condition is sufficient rather than necessary.  It identifies a support-stability basin: local approximation or within-support weight error may be nonzero, but it cannot change future selected supports while the accumulated state deviation remains inside every margin tube.

\subsection{A margin-distribution bound for expected support changes}
Uniform positive margins are often unrealistic.  An expected guarantee follows from a conditional anti-concentration condition.  Let
\begin{equation}
 E_j=d(\widehat S_j,S_j),
 \qquad
 B_j=\norm{\widehat z_j(\widehat S_j)-z_j(\widehat S_j)}_\infty,
 \qquad
 U_j=B_j+L_{R,j}E_j.
 \label{eq:expected-router-perturbation}
\end{equation}
By \cref{lem:topk-margin-stability}, a support mismatch implies $\gamma_{k,j}(S_j)\le2U_j$.  Assume that there exists a version of the conditional margin CDF such that, almost surely, for every $t\ge0$,
\begin{equation}
 \Prb\!\left(\gamma_{k,j}(S_j)\le t\mid U_j\right)
 \le \min\{1,c_jt^{\alpha_j}\}
 \label{eq:conditional-margin-anticoncentration}
\end{equation}
for constants $c_j>0$ and $\alpha_j>0$.  This conditional statement is essential: an unconditional margin histogram does not generally control a random, dependent threshold $2U_j$.

Let $M_j$ be the indicator of a support mismatch.  Let $r_j\ge0$ denote the full route-induced transition discrepancy at event $j$, and assume the pointwise decomposition
\begin{equation}
 r_j\le\zeta_j+\Delta_jM_j,
 \qquad \zeta_j\ge0.
 \label{eq:support-route-decomposition}
\end{equation}
Here $\zeta_j$ bounds route discrepancy that can remain even when supports agree, such as nonuniform within-support weight changes; for unit-gain top-$1$ or uniform top-k, one may take $\zeta_j=0$.  The deterministic constant $\Delta_j\ge0$ bounds the additional severity associated with a support change.

\begin{theorem}[Margin-mediated expected support-change cost]
\label{thm:margin-mediated-routing}
Under the preceding conditions, with $\rho_j=\E r_j$,
\begin{equation}
 \rho_j
 \le\E\zeta_j
 +\Delta_j\E\min\left\{1,c_j(2U_j)^{\alpha_j}\right\}.
 \label{eq:margin-route-cost}
\end{equation}
For $\alpha_j=1$ and finite first moments, the simpler (possibly looser) bound holds without restricting the range of $U_j$:
\begin{equation}
 \rho_j
 \le\E\zeta_j+2c_j\Delta_j
 \left(\E B_j+L_{R,j}\E E_j\right).
 \label{eq:linear-margin-route-cost}
\end{equation}
Consequently, if a pathwise local recursion has the form
\begin{equation}
 E_{j+1}\le a_j+L_jE_j+r_j,
 \label{eq:expected-cascade-recursion}
\end{equation}
then
\begin{equation}
 \E E_{j+1}
 \le \E a_j+\E\zeta_j+2c_j\Delta_j\E B_j
 +\left(L_j+2c_j\Delta_jL_{R,j}\right)\E E_j.
 \label{eq:effective-margin-amplification}
\end{equation}
\end{theorem}
The extra factor $2c_j\Delta_jL_{R,j}$ is an endogenous support-routing amplification term.  It is large when margins concentrate near zero, logits are state-sensitive, or support substitutions are functionally severe.  For $\alpha_j\ne1$, \cref{eq:margin-route-cost} yields a nonlinear moment recursion involving $\E U_j^{\alpha_j}$ (or the clipped moment appearing explicitly in the theorem).

\subsection{A weaker marginal-margin guarantee under arbitrary dependence}
Conditioning on $U_j$ in \cref{eq:conditional-margin-anticoncentration} can be restrictive because $U_j$ itself depends on the trajectory and router margin.  A marginal margin assumption still gives a guarantee, but generally a slower one.  Suppress $j$ and let $\gamma\ge0$ be a support margin, $U\ge0$ the logit perturbation bound, and $M$ the support-mismatch indicator.  Suppose $M\le\mathbf1\{\gamma\le2U\}$, $\Prb(\gamma\le t)\le ct^\alpha$ for all $t\ge0$, and $m_r=\E U^r<\infty$ for some moment order $r>0$.  The constants $c>0,\alpha>0$ control how much margin probability lies near zero; this is called a margin anti-concentration condition.

\begin{theorem}[Dependence-robust marginal-margin bound]
\label{thm:marginal-margin-routing}
No independence between $\gamma$ and $U$ is required for
\begin{equation}
 \Prb(M=1)\le
 \min\left\{1,\inf_{t>0}\left[c(2t)^\alpha+\frac{m_r}{t^r}\right]\right\}.
 \label{eq:marginal-margin-split}
\end{equation}
A sharper explicit bound, not merely the optimization of the preceding split, is
\begin{equation}
 \Prb(M=1)\le
 \min\{1,K_{\alpha,r}(2^\alpha c)^{r/(\alpha+r)}
                     m_r^{\alpha/(\alpha+r)}\},
 \label{eq:marginal-margin-optimized}
\end{equation}
where $K_{\alpha,r}=((\alpha+r)/\alpha)^{\alpha/(\alpha+r)}$.
For $m_r=0$, the mismatch probability is zero.  If the route loss satisfies $r_{\mathrm{route}}\le\zeta+\Delta M$, multiply the mismatch bound by $\Delta$ and add $\E\zeta$ to obtain a functional-cost bound.
\end{theorem}
\begin{proof}
For every fixed threshold $t>0$, a mismatch implies either $\gamma\le2t$ or $U>t$.  A union bound and Markov's inequality give \cref{eq:marginal-margin-split}.  For the sharper bound, let $E=\{M=1\}$ and $p=\Prb(E)$.  On $E$, $U\ge\gamma/2$.  The tail-integral identity for a nonnegative random variable and $\Prb(E,\gamma>t)\ge p-ct^\alpha$ imply, for $p>0$,
\begin{align}
 m_r&\ge 2^{-r}\E[\gamma^r\mathbf1_E]\\
 &\ge 2^{-r}\int_0^{(p/c)^{1/\alpha}}
       rt^{r-1}(p-ct^\alpha)\,dt\\
 &=2^{-r}\frac{\alpha}{\alpha+r}
       c^{-r/\alpha}p^{1+r/\alpha}.
 \label{eq:marginal-margin-tail-integral}
\end{align}
Rearranging yields \cref{eq:marginal-margin-optimized}; the case $p=0$ is immediate.  If $m_r=0$, then $U=0$ almost surely and $\Prb(\gamma=0)=0$ by the margin assumption.  No conditioning or independence is used.  For example, $\alpha=r=c=1$ yields $2\sqrt{m_1}$ rather than the looser $2\sqrt{2m_1}$ from optimizing the split.
\end{proof}

\begin{proposition}[Sharp exponent and limiting constant under marginal control]
\label{prop:marginal-margin-sharpness}
For every $\alpha,r,c>0$, every $\varepsilon>0$, and every $0<b\le c^{-1/\alpha}$, there is a two-expert top-$1$ example with margin CDF $\Prb(\gamma\le t)=\min\{1,ct^\alpha\}$ and strict support changes such that
\begin{equation}
 \Prb(M=1)=
 \left[\frac{\alpha+r}{\alpha}
       \left(\frac{2}{1+\varepsilon}\right)^r
       c^{r/\alpha}\E U^r\right]^{\alpha/(\alpha+r)}.
 \label{eq:marginal-margin-lower-example}
\end{equation}
Letting $b\downarrow0$ shows that the moment exponent cannot improve.  Letting $\varepsilon\downarrow0$ shows that the multiplicative constant in the nontrivial part of \cref{eq:marginal-margin-optimized} cannot be uniformly reduced under the stated assumptions.
\end{proposition}
\begin{proof}
Use logits $z=(\gamma/2,-\gamma/2)$.  On $\{\gamma\le b\}$ replace them by $\widetilde z=(-\varepsilon\gamma/2,\varepsilon\gamma/2)$; otherwise leave them unchanged.  The reference and perturbed winners differ strictly except on the zero-probability event $\gamma=0$.  Thus
\[
 U=\frac{1+\varepsilon}{2}\gamma\mathbf1_{\{\gamma\le b\}},\qquad
 \Prb(M=1)=cb^\alpha,\qquad
 \E U^r=\left(\frac{1+\varepsilon}{2}\right)^r
          \frac{c\alpha}{\alpha+r}b^{\alpha+r}.
\]
Eliminating $b$ proves the formula.  Choose $cb^\alpha<1$ to avoid saturation and then let $\varepsilon\downarrow0$ for limiting sharpness.  Positive $\varepsilon$ avoids relying on a favorable tie rule.
\end{proof}
The perturbation in this example uses only currently available logits, not future information.  Small-margin states are precisely the ones receiving error.  This explains why an unconditional margin histogram cannot justify the faster conditional bound.  For a first-moment state certificate and $r=1$, the robust dependence on the error moment has exponent $\alpha/(\alpha+1)$, not $\alpha$.

\subsection{Implications for expert reuse}
Cross-layer and global pools can increase the number of candidate experts, which can alter router margins. They may also place functionally redundant experts near one another, reducing $\Delta_j$ or the transport cost of a support change.  Therefore candidate-pool size alone does not determine route stability.  The relevant measurements are jointly:
\begin{enumerate}[leftmargin=*,itemsep=1pt]
\item the top-k support-margin distribution under the reached occupancy.
\item the state-to-logit sensitivity $L_{R,j}$ or an empirical local analogue.
\item same-support weight discrepancy $\zeta_j$ when weights are not uniform.
\item the functional geometry of candidate expert outputs.
\item the position-dependent downstream amplification.
\end{enumerate}
These quantities turn the otherwise exogenous $\rho_j$ into a measurable closed-loop diagnostic.  They also explain why route consistency can deteriorate with reasoning length even when the one-step expert estimators remain unchanged.

\subsection{Proofs}
\begin{proof}[Proof of \cref{prop:soft-router-closed-loop}]
Insert $M(s',g(s))$:
\begin{align}
 d(F(s),F(s'))
 &\le d(M(s,g(s)),M(s',g(s)))\nonumber\\
 &\quad+d(M(s',g(s)),M(s',g(s'))).
 \label{eq:soft-router-insertion}
\end{align}
The first term is at most $L_xd(s,s')$ by \cref{eq:fixed-gate-state-lipschitz}; the second is at most $\mathsf W_{s'}(g(s),g(s'))$ by \cref{thm:functional-transport-gate}.  This proves \cref{eq:soft-closed-loop-lipschitz}.  The diameter specialization follows from the second inequality in \cref{eq:transport-gate-bound}.
\end{proof}

\begin{proof}[Proof of \cref{lem:topk-margin-stability}]
Every selected logit can decrease by at most $\norm{\widetilde z-z(s)}_\infty$, and every unselected logit can increase by at most the same amount.  If this perturbation is below $\gamma_k(s)/2$, the smallest selected coordinate remains strictly larger than the largest unselected coordinate.  For the second statement, the triangle inequality bounds $\norm{\widehat z(s')-z(s)}_\infty$ by $\beta+L_Rd(s,s')$.
\end{proof}

\begin{proof}[Proof of \cref{thm:no-route-cascade}]
Proceed by induction.  At $j=0$, the error is at most $\bar e_0$.  If $d(\widehat S_j,S_j)\le\bar e_j$, \cref{eq:no-cascade-margin-condition} and \cref{lem:topk-margin-stability} imply identical supports.  By the assumed same-support event-discrepancy bound and the Lipschitz property of the reference fixed-support transition,
\begin{equation}
 d(\widehat S_{j+1},S_{j+1})
 \le\eta_j+L_jd(\widehat S_j,S_j)
 \le\bar e_{j+1}.
 \label{eq:no-cascade-induction}
\end{equation}
Thus support agreement and the error envelope hold at every event.
\end{proof}

\begin{proof}[Proof of \cref{thm:margin-mediated-routing}]
A support mismatch implies $\gamma_{k,j}(S_j)\le2U_j$.  Conditioning on $U_j$ and applying \cref{eq:conditional-margin-anticoncentration} at the random threshold $2U_j$ gives
\begin{equation}
 \Prb(M_j=1)
 \le\E\min\{1,c_j(2U_j)^{\alpha_j}\}.
 \label{eq:mismatch-probability-margin}
\end{equation}
Taking expectations in \cref{eq:support-route-decomposition} proves \cref{eq:margin-route-cost}.  For $\alpha_j=1$, use $\min\{1,2c_jU_j\}\le2c_jU_j$, substitute $U_j=B_j+L_{R,j}E_j$ and use linearity of expectation.  Taking expectations in \cref{eq:expected-cascade-recursion} and inserting the resulting route-cost bound proves \cref{eq:effective-margin-amplification}.
\end{proof}

\section{Counterfactual Compatibility and the Sample Complexity of Route Audits}
\label{app:causal-audit}

\paragraph{Overview.}
We separate a population comparison of action effects from the finite-sample task of auditing them. Under a declared randomized intervention law, counterfactual variance measures incompatibility of action means; absent overlap, unsupported effects are not identified. Independent records and positive forcing probabilities support action-specific concentration.  
Sections~\ref*{app:causal-audit}.1--\ref*{app:causal-audit}.2 establish the population comparison and the overlap obstruction; Sections~\ref*{app:causal-audit}.3--\ref*{app:causal-audit}.5 give audit bounds and budget allocation, with remaining proofs in Section~\ref*{app:causal-audit}.6.

\subsection{Reuse bias as counterfactual transition variance}
Let $S$ be a pre-action macro state in a standard Borel space and let $A\in[K]$ be drawn from a declared randomized intervention distribution $p_a(S)=\Prb(A=a\mid S)$.  Let $Y^a$ be the potential post-action transition target for action $a$ in a separable real Hilbert space $\cH$.  Assume every $Y^a$ is Bochner measurable and square-integrable, conditional randomization $A\perp(Y^1,\ldots,Y^K)\mid S$, and consistency $Y=Y^A$.  Define the Bochner conditional mean
\begin{equation}
 m_a(s)=\E[Y^a\mid S=s].
 \label{eq:counterfactual-conditional-mean}
\end{equation}
Values of $m_a$ outside the support of $p_a(S)$ are immaterial to the displayed risks.  The evaluation experiment draws $A$ and observes $Y^A$.  Compare action-specific predictors $f_A(S)$ with one shared predictor $f(S)$ under squared loss.

\begin{theorem}[Counterfactual decomposition of reuse bias]
\label{thm:counterfactual-reuse-bias}
Under the conditions above, allow all measurable predictors with finite weighted second moment.  The action-specific and shared Bayes predictors are
\begin{equation}
 f_a^\star(s)=m_a(s),
 \qquad
 f_{\mathrm{share}}^\star(s)=\bar m(s):=\sum_ap_a(s)m_a(s)
 \label{eq:counterfactual-bayes-predictors}
\end{equation}
on the corresponding supports.  Their oracle-risk difference is
\begin{align}
 \BReuse_{\mathrm{cf}}
 &=\E\norm{m_A(S)-\bar m(S)}_{\cH}^2
 \label{eq:counterfactual-bias-variance}\\
 &=\frac12\E_S\sum_{a,b}p_a(S)p_b(S)
 \norm{m_a(S)-m_b(S)}_{\cH}^2.
 \label{eq:counterfactual-pairwise-bias}
\end{align}
\end{theorem}
This is the action-level analogue of \cref{prop:hilbert-bias}.  It separates irreducible transition uncertainty, which appears in both oracle risks, from incompatibility among conditional action means, which is paid only by the shared operator.  The intervention distribution is part of the estimand: changing which actions are compared, or where they are forced, changes $\BReuse_{\mathrm{cf}}$.

The identity clarifies the relation between functional and causal reuse.  Small observational disagreement among naturally selected experts does not imply small \cref{eq:counterfactual-pairwise-bias}; the latter compares actions at common pre-action states.  Conversely, distinct indices can have negligible causal sharing bias if their forced conditional transition means coincide on the stated intervention occupancy.  The theorem is a population identity.  A plug-in estimate of $\BReuse_{\mathrm{cf}}$ must estimate the conditional means $m_a$ and evaluate their squared differences on independent or cross-fitted records, or else control the complexity and overfitting of the learned functions.

\paragraph{When the local and causal comparisons are the same.}\label{app:causal-role-alignment}
For a supported action set, define $\pi_a=\Prb(A=a)>0$ and the local population risk
\[
 R_a(f)=\E[\|Y^a-f(S)\|_{\cH}^2\mid A=a].
\]
Use the same predictors of $S$, the same squared loss, and these weights in the local shared/separate oracle definitions.  Consistency and the law of total expectation give $\sum_a\pi_aR_a(f)=\E\|Y^A-f(S)\|_{\cH}^2$; allowing a separate $f_a$ gives the corresponding action-specific objective.  Their population bias is then exactly the bias in \cref{thm:counterfactual-reuse-bias}.  This alignment uses the action-conditional state laws, which need not equal the marginal law of $S$.  

Without this matched experiment and class, the two biases are different quantities.  In particular, tying a physical expert does not remove action information encoded in its input.  A statistical role such as operation--depth pair $(i,\ell)$ identifies a call environment; a route action specifies an intervention on executable calls.  Equating those indices requires an explicit map, not merely identical names.  

\subsection{Without overlap, unsupported route effects are not identifiable}
Let an observational or intervention policy select action $a$ with propensity $e_a(C)=\Prb(A=a\mid C)$.  Suppose there is a measurable set $\mathcal B$ with $\Prb(C\in\mathcal B)>0$ on which $e_a(C)=0$.

\begin{proposition}[No-overlap non-identifiability]
\label{prop:no-overlap-nonidentifiability}
Without structural restrictions linking unsupported and supported action effects, the distribution of $T_a(C)$ on $\mathcal B$ is not identified from data generated by the policy whenever the model class permits two different effects there.  A defect or utility depending nontrivially on those differences is likewise not identified; a constant summary need not be affected.  In a nondegenerate binary-output model, for example, there exist two action libraries with identical observed-data distributions but different values of
\begin{equation}
 \E\!\left[d\!\left(T_a(C),F_a(\Psi(C))\right)\mathbf 1\{C\in\mathcal B\}\right].
 \label{eq:unsupported-defect}
\end{equation}
\end{proposition}
The conclusion is not repaired by more samples from the same policy.  It requires either a structural extrapolation assumption or interventions that assign positive probability to the unsupported action--context region.

\subsection{A finite-sample audit of fixed action-specific defects}
The macro extractor $\Psi$, action definitions, candidate operators $F_a$, and bounded losses in this subsection are fixed independently of the audit records.  They may be learned on a separate fitting split and then conditioned upon.  Cross-fitting requires a separate conditional bound for each held-out fold and a union bound over folds; if the fitted operators differ across folds, the resulting target is their average fold-specific risk, not automatically the risk of a final refitted model.  Treating all cross-fitted scores as IID conditional on every fold fit is not justified.  Reusing the same audit observations to select these objects requires a uniform complexity correction.

Suppose the context--potential-outcome records are IID across $i=1,\ldots,N$.  Conditional on context $C_i$, an action can have stochastic output $T_a(C_i;U_i)$, where $U_i$ is unrevealed execution randomness.  Define the bounded potential losses and their conditional means by
\begin{equation}
 L_{i,a}=d\!\left(T_a(C_i;U_i),F_a(\Psi(C_i))\right)\in[0,1],\quad h_a(c)=\E[L_{i,a}\mid C_i=c].
 \label{eq:potential-intervention-loss}
\end{equation}
Conditional on $C_i$, randomize $A_i$ independently of its potential-loss vector with known propensity $e_a(C_i)$, independently across records, and observe $L_i=L_{i,A_i}$.  Deterministic execution is the special case without $U_i$.  Let
\begin{equation}
 \mu_a=\E L_{i,a}=\E h_a(C_i)
 \label{eq:action-defect-target}
\end{equation}
under the probe-context distribution.  The Horvitz--Thompson estimator is \citep{horvitz1952generalization}
\begin{equation}
 \widehat\mu_a
 =\frac1N\sum_{i=1}^N
 \frac{\mathbf 1\{A_i=a\}L_i}{e_a(C_i)}.
 \label{eq:ipw-route-audit}
\end{equation}

\begin{theorem}[Simultaneous concentration of a randomized route audit]
\label{thm:ipw-route-audit}
Assume the intervention policy has overlap
\begin{equation}
 e_a(C)\ge\beta>0
 \quad\text{a.s. for every claimed action }a\in[K],\qquad 0<\beta\le1/K.
 \label{eq:route-audit-overlap}
\end{equation}
Then each $\widehat\mu_a$ is unbiased, and with probability at least $1-\delta$,
\begin{equation}
 \max_{a\in[K]}\abs{\widehat\mu_a-\mu_a}
 \le
 \sqrt{\frac{2\log(2K/\delta)}{N\beta}}
 +\frac{\log(2K/\delta)}{3N\beta}.
 \label{eq:ipw-route-audit-bound}
\end{equation}
Consequently, $N\beta$ is the leading effective sample size for a simultaneous claim about the least-covered action.  Under uniform random forcing, $\beta=1/K$, so estimating all $K$ action defects to fixed accuracy requires order $K$ audit records, up to the usual accuracy and logarithmic factors.
\end{theorem}

The theorem applies to a loss evaluated after one expert call, a compound multi-layer route action, or a bounded stopping-time macro action, provided the action and loss are defined before the audit records are observed.  If an action changes active compute, the comparator should include matched-compute null or substitute actions so that the audit isolates functionality rather than budget.

For fixed nonnegative terminal weights $w_a$, the same simultaneous event implies the conservative bound
\begin{equation}
 \abs{\sum_aw_a\widehat\mu_a-\sum_aw_a\mu_a}
 \le\left(\sum_aw_a\right)
 \left[
 \sqrt{\frac{2\log(2K/\delta)}{N\beta}}
 +\frac{\log(2K/\delta)}{3N\beta}
 \right].
 \label{eq:weighted-audit-bound}
\end{equation}
This is useful when every action-specific defect must also be reported, but it is not the optimal rate for every aggregate.  In particular, for the behavior-policy-weighted target
\begin{equation}
 \mu_e=\E\sum_ae_a(C)h_a(C)=\E L_i,
 \label{eq:behavior-weighted-audit-target}
\end{equation}
the direct estimator $\widehat\mu_e=N^{-1}\sum_iL_i$ satisfies, by Hoeffding,
\begin{equation}
 \abs{\widehat\mu_e-\mu_e}
 \le\sqrt{\frac{\log(2/\delta)}{2N}}
 \label{eq:behavior-weighted-audit-bound}
\end{equation}
with probability at least $1-\delta$, without a $1/\beta$ factor.  Thus $N\beta$ is the relevant size for uniform action-specific or least-covered-action claims, not a universal sample size for all policy averages.  A different target policy $q_a(C)$ can be evaluated by weighting with $q_a(C)/e_a(C)$, whose rate depends on the resulting importance-weight envelope or variance.

\subsection{The per-action coverage dependence is minimax unavoidable}
The $1/\sqrt{N\beta}$ term is not merely an artifact of inverse-propensity weighting.

\begin{proposition}[Per-action coverage lower bound]
\label{prop:route-audit-lower-bound}
There is a universal constant $c>0$ such that, for any estimator $\widetilde\mu_a$ based on $N$ randomized records with constant propensity $e_a=\beta$, one can choose a bounded binary intervention-outcome distribution for which
\begin{equation}
 \E\abs{\widetilde\mu_a-\mu_a}
 \ge c\min\left\{1,\frac1{\sqrt{N\beta}}\right\}.
 \label{eq:route-audit-minimax-lower}
\end{equation}
\end{proposition}
Thus an action that is forced only rarely cannot be audited precisely merely because the total token count is large.    The proof is a two-point Le Cam argument \citep{tsybakov2009nonparametric}.

\subsection{Allocating a finite intervention budget}
Uniform forcing is simple but is not generally variance optimal for a performance-weighted aggregate.  Suppose the target is $\mu_w=\sum_aw_a\mu_a$, constant action propensities $e_a\ge0$ sum to one, and
\begin{equation}
 s_a=\E L_{i,a}^2.
 \label{eq:action-second-moment}
\end{equation}
For every action with $w_a^2s_a>0$, require $e_a>0$.  The unbiased single-record estimator is $w_{A_i}L_i/e_{A_i}$, with second moment $\sum_aw_a^2s_a/e_a$ over the active set.

\begin{proposition}[Variance-optimal route-intervention allocation]
\label{prop:optimal-intervention-allocation}
Assume $w_a\ge0$ and not all $w_a\sqrt{s_a}$ vanish.  Let $\mathcal A_+=\{a:w_a\sqrt{s_a}>0\}$.  Ignoring positivity requirements for actions outside $\mathcal A_+$ and lower-propensity constraints, the action distribution minimizing the second moment of the weighted estimator is
\begin{equation}
 e_a^\star
 =\frac{w_a\sqrt{s_a}}{\sum_{b\in\mathcal A_+}w_b\sqrt{s_b}}
 \quad(a\in\mathcal A_+),
 \qquad e_a^\star=0\quad(a\notin\mathcal A_+),
 \label{eq:optimal-audit-propensity}
\end{equation}
with minimum
\begin{equation}
 \left(\sum_{a\in\mathcal A_+}w_a\sqrt{s_a}\right)^2.
 \label{eq:optimal-audit-second-moment}
\end{equation}
\end{proposition}
A pilot sample can estimate $s_a$, after which an independent second phase can use \cref{eq:optimal-audit-propensity}.  If action-specific claims are required for every action, or if future targets may assign them positive weight, impose a floor $e_a\ge\beta$ with $K\beta\le1$; if $K\beta<1$ and some $w_a\sqrt{s_a}>0$, the constrained solution is $e_a=\max\{\beta,w_a\sqrt{s_a}/t\}$, where $t>0$ is chosen so that $\sum_ae_a=1$.  If $K\beta=1$, the unique feasible allocation is $e_a=\beta$.  These formulas follow from the Lagrange stationarity condition on coordinates above the floor; when all second-moment contributions vanish, every feasible allocation is optimal.  Actions with zero contribution to the declared aggregate may receive zero mass only when no claim about their individual effects is made.

\subsection{Proofs}
\begin{proof}[Proof of \cref{thm:counterfactual-reuse-bias}]
For a supported action $a$, conditional randomization gives $\E[Y^a\mid S,A=a]=m_a(S)$.  The Hilbert-space Pythagorean identity for Bochner conditional expectation gives
\begin{equation}
 \E\!\left[\norm{Y^a-f(S)}^2\mid S,A=a\right]
 =\E\!\left[\norm{Y^a-m_a(S)}^2\mid S\right]
 +\norm{m_a(S)-f(S)}^2.
 \label{eq:conditional-bayes-decomposition}
\end{equation}
Action-specific prediction sets $f_a=m_a$ and leaves only the first term.  The shared problem minimizes $\sum_ap_a(S)\norm{m_a(S)-f(S)}^2$ pointwise, whose minimizer is $\bar m(S)$.  Subtracting the common irreducible term yields \cref{eq:counterfactual-bias-variance}.  Applying the weighted Hilbert variance identity conditionally on $S$ gives \cref{eq:counterfactual-pairwise-bias}.
\end{proof}

\begin{proof}[Proof of \cref{prop:no-overlap-nonidentifiability}]
Construct two action libraries that agree for every action--context pair with positive observation probability and differ only in the value of $T_a(C)$ for $C\in\mathcal B$.  Because action $a$ has zero probability on $\mathcal B$, the changed transition is never observed, so the induced observed-data distributions are identical.  Choosing the two unsupported transitions to have different distances from $F_a(\Psi(C))$ changes \cref{eq:unsupported-defect}.  Therefore the estimand is not identified without additional structural restrictions.
\end{proof}

\begin{proof}[Proof of \cref{thm:ipw-route-audit}]
For fixed $a$, let
\begin{equation}
 X_{i,a}=\frac{\mathbf 1\{A_i=a\}L_i}{e_a(C_i)}.
 \label{eq:ipw-summand}
\end{equation}
Conditional randomization gives $\E[X_{i,a}\mid C_i]=h_a(C_i)$, hence $\E X_{i,a}=\mu_a$.  Moreover $0\le X_{i,a}\le1/\beta$ and
\begin{equation}
 \E X_{i,a}^2
 \le\E\frac{\mathbf 1\{A_i=a\}}{e_a(C_i)^2}
 =\E\frac1{e_a(C_i)}
 \le\frac1\beta.
 \label{eq:ipw-variance-bound}
\end{equation}
Thus $\operatorname{Var}(X_{i,a})\le1/\beta$, and $|X_{i,a}-\mu_a|\le1/\beta$.  To make the Bernstein constant explicit, put $Z_a=\widehat\mu_a-\mu_a$.  Expanding the exponential moment of each centered bounded summand and using $k!\ge2\cdot3^{k-2}$ for integers $k\ge2$ gives
\[
 \log\E e^{tZ_a}\le
 \frac{t^2/(N\beta)}{2(1-|t|/(3N\beta))},
 \qquad |t|<3N\beta.
\]
This is an exponential-moment bound for both signs of $Z_a$.  The Chernoff argument (apply Markov's inequality to $e^{tZ_a}$ and optimize $t$) gives
$\Prb\{|Z_a|>\sqrt{2x/(N\beta)}+x/(3N\beta)\}\le2e^{-x}$.
Take $x=\log(2K/\delta)$ and union-bound over actions.
\end{proof}

\begin{proof}[Proof of \cref{prop:route-audit-lower-bound}]
It suffices to consider a context-free stochastic binary outcome, allowed by the potential-loss model above.  Under hypotheses $P_-$ and $P_+$, action $a$ is selected with probability $\beta$, and conditional on selection its outcome is Bernoulli with means $1/2-\varepsilon$ and $1/2+\varepsilon$, respectively; every other observation has the same law.  For $\varepsilon\le1/4$, the one-record Kullback--Leibler divergence is at most $16\beta\varepsilon^2$, so the $N$-record divergence is at most $16N\beta\varepsilon^2$.  Choosing $\varepsilon=c_0\min\{1,(N\beta)^{-1/2}\}$ with a sufficiently small universal $c_0\le1/4$ keeps total variation bounded away from one by Pinsker's inequality.  Le Cam's two-point inequality then lower-bounds the maximum expected absolute error by a universal constant times the parameter separation, proving \cref{eq:route-audit-minimax-lower}.
\end{proof}

\begin{proof}[Proof of \cref{prop:optimal-intervention-allocation}]
Restrict to $\mathcal A_+$.  Cauchy--Schwarz gives
\begin{equation}
 \left(\sum_{a\in\mathcal A_+}w_a\sqrt{s_a}\right)^2
 =\left(\sum_{a\in\mathcal A_+}
 \frac{w_a\sqrt{s_a}}{\sqrt{e_a}}\sqrt{e_a}\right)^2
 \le\left(\sum_{a\in\mathcal A_+}\frac{w_a^2s_a}{e_a}\right)
       \left(\sum_{a\in\mathcal A_+}e_a\right).
 \label{eq:audit-allocation-cauchy}
\end{equation}
At an optimum no mass is assigned outside $\mathcal A_+$, so the second factor is one.  Equality holds exactly when $e_a\propto w_a\sqrt{s_a}$ on $\mathcal A_+$, yielding the stated allocation and minimum.
\end{proof}

\section{Extensions, Assumption Ledger, and Research Protocol}
\label{app:extensions-ledger}

\paragraph{Overview.}
This appendix collects complementary extensions and proposed uses of the framework: functional corrections, grouping selection, stochastic macro transitions, and resource-adjusted scores. 
Sections~\ref*{app:extensions-ledger}.1--\ref*{app:extensions-ledger}.3 cover functional corrections, grouping, and stochastic dynamics; Sections~\ref*{app:extensions-ledger}.4--\ref*{app:extensions-ledger}.5 distinguish resource preferences and assumptions.

\subsection{Partial functional reuse beyond a scalar shrinkage coefficient}
The exact linear theorem uses one scalar $\lambda$.  A function-space shared core plus an $r$-dimensional role correction gives a complementary interpolation.
\begin{proposition}[Residual bias of a rank-$r$ functional correction]
\label{prop:rank-r-functional}
Under the unrestricted Hilbert quadratic setting $\cF_e=\cH$ of \cref{prop:hilbert-bias}, assume finite-dimensional $\cH$, fix an integer $0\le r\le\dim(\cH)$, and define
\begin{equation}
 \Sigma=\sum_a\pi_a(f_a^\star-\bar f)\otimes(f_a^\star-\bar f)
 \label{eq:app-between-role-cov}
\end{equation}
where $(u\otimes u)v=u\langle u,v\rangle$ and the eigenvalues are $\lambda_1\ge\lambda_2\ge\cdots\ge0$.  They measure between-role variation along orthogonal function directions, not estimation-noise variance.  Among role predictors $f_a=\bar f+Uz_a$ with $U:\R^r\to\cH$, $U^\top U=I_r$, and role-specific coordinates $z_a\in\R^r$, sharing an $r$-dimensional correction subspace, the minimum average excess risk is
\begin{equation}
 (\BReuse_e)^{(r)}=\sum_{k>r}\lambda_k.
 \label{eq:app-rank-r-tail}
\end{equation}
\end{proposition}
\begin{proof}
For fixed subspace $U$, the optimal role correction is the orthogonal projection of $f_a^\star-\bar f$ onto $U$.  The average residual is $\tr((I-P_U)\Sigma)$.  The rank-$r$ projector maximizing captured trace spans the top $r$ eigenvectors, with the usual convention that the tail is zero once $r$ reaches the rank.
\end{proof}
This is a function-space statement.  Connecting it to a rank-$r$ LoRA parameterization requires a separate realizability assumption about the network Jacobian or adapter class, and adding estimation variance for the learned role coordinates is necessary for a complete finite-sample adapter comparison.

For a concrete obstruction, take linear maps on $\R^2$, input second moment $I_2$, two equally weighted role targets $f_\pm(x)=\pm I_2x$, and a fixed zero shared core.  Their centered variation spans one function direction, so a one-dimensional functional correction is exact.  But either target matrix has two unit singular values: a matrix-rank-one update to this fixed core has squared Frobenius residual at least one.  Thus function-subspace dimension one does not imply parameter-update rank one.  

\Cref{sec:adapter-realizability,sec:factor-tangent} gives exact frozen-feature adapter tails, a two-role free-core result, and initialization-dependent tangent images.  

\subsection{Data-dependent grouping}
The fixed-rule identities condition on a reuse assignment independent of fitting noise.  Adaptive grouping therefore needs a separate selection argument.

In the exact linear model, \cref{thm:partition-frontier} supplies a population objective without an ad hoc clustering metric: minimize within-group Frobenius dispersion plus $V$ per fitted expert.  Its pairwise merge statistic has the exact law in \cref{eq:app-merge-statistic-law}.  Outside that model, the following validation-based construction retains the same logic without assuming that raw parameter distance equals functional heterogeneity.

First, fit a finite candidate family $\mathcal A$ of groupings using training data and select among those fitted candidates on an independent validation split.  Conditional on training, if validation loss is bounded and has effective size $N_{\mathrm{val}}^{\mathrm{eff}}$, validation ERM satisfies the analogue of \cref{eq:app-validation-oracle} with $|\Lambda|$ replaced by $|\mathcal A|$.  Second, if grouping and expert fitting use the same observations, uniform convergence must cover the union of all grouped hypothesis classes; a finite grouping family contributes an additional $\log|\mathcal A|$ term.    An untouched final test is still required.  \Cref{thm:finite-architecture-selection} provides a finite-candidate oracle inequality for explicit loss-tail conditions, including a sufficient Gaussian quadratic model.  \Cref{eq:plugin-analytic-counterexample} illustrates the risk of a same-data dispersion plug-in.

A certificate-guided grouping objective may use
\begin{equation}
 \widehat U(\alpha)
 =\sum_e\widehat W_e
  \left(\widehat B_e+2\widehat\epsilon_{\mathrm{sh},e}\right)
 +\sum_j\widehat A_j(\widehat\tau_j+\widehat\rho_j)
 +\lambda_{\mathrm{res}}\operatorname{Cost}(\alpha),
 \label{eq:app-group-objective}
\end{equation}
where the common role-specific oracle term has been omitted only when it is invariant across the compared partitions.  Every fitted functional probe and plug-in quantity used to choose $\alpha$ must be computed on training/validation data or by cross-fitting, and final evaluation is performed once on a held-out test set.

\subsection{Stochastic macro dynamics}
When the macro state is Markov-sufficient, deterministic transitions can be replaced by Markov kernels on a Polish (complete separable) metric space $(\cS,d)$ with finite first moments.  Let $W_d$ denote the $1$-Wasserstein distance with ground metric $d$, and let $Q_j$ be the actual kernel and $\bar Q_j$ a reference kernel satisfying
\begin{equation}
 \sup_s W_d(Q_j(s,\cdot),\bar Q_j(s,\cdot))\le\eta_j,
 \qquad
 W_d(\bar Q_j(s,\cdot),\bar Q_j(s',\cdot))\le L_jd(s,s').
 \label{eq:app-wasserstein-assumptions}
\end{equation}
For distributions $\xi_{j+1}=\xi_jQ_j$ and $\bar\xi_{j+1}=\bar\xi_j\bar Q_j$, conditional coupling and the triangle inequality give
\begin{equation}
 W_d(\xi_{j+1},\bar\xi_{j+1})
 \le \eta_j+L_jW_d(\xi_j,\bar\xi_j),
 \label{eq:app-wasserstein-recursion}
\end{equation}
which telescopes exactly as \cref{eq:app-path-bound}.  Standard kernel measurability is assumed.  If the one-step discrepancy is state-dependent rather than uniformly bounded, its expectation under $\xi_j$ replaces $\eta_j$.  Without macro-state sufficiency, kernels must be defined on the full micro context before being pushed forward through $\Psi$.

\subsection{Resource-adjusted reuse}
We evaluate statistical gains alongside serving costs.  Define every cost difference as
\begin{equation}
 \Delta C_\bullet=C_\bullet^{\mathrm{reuse}}-C_\bullet^{\mathrm{sep}},
 \label{eq:app-resource-difference-sign}
\end{equation}
so a positive value means that reuse is more costly.  A deployment score may retain system costs separately:
\begin{equation}
 \PRG_H^{\mathrm{resource}}
 =\PRG_H
 -\lambda_{\mathrm{mem}}\Delta C_{\mathrm{memory}}
 -\lambda_{\mathrm{comm}}\Delta C_{\mathrm{communication}}
 -\lambda_{\mathrm{lat}}\Delta C_{\mathrm{latency}}
 -\lambda_{\mathrm{flop}}\Delta C_{\mathrm{flop}}.
 \label{eq:app-resource-prg}
\end{equation}
The nonnegative coefficients convert the different resource units into the chosen score units and encode deployment preferences rather than statistical constants, and the raw cost changes should be reported alongside the scalarized score.

\subsection{Assumption dependencies}
\Cref{app:operational-assumptions} audits the assumptions of each core result and extension, including the exact-model restrictions, estimator independence, transfer conditions, route semantics, and causal coverage.  It distinguishes mathematical sufficiency from empirical plausibility.  The covariance and marginal-margin extensions in \cref{app:assumption-robust,app:endogenous-routing} are used when the stronger uncorrelated-error or conditional-margin assumptions are not credible.

\section{Stronger Results under Weaker or More Explicit Assumptions}
\label{app:assumption-robust}

\paragraph{Overview.}
We identify what remains valid when the simplest assumptions are relaxed. The extensions retain general test geometry and covariance, optimization and transfer costs, and escape events. The final subsection separately explains how held-out paired evaluation can support actual performance improvement. Each extension specifies the object, randomness, and fixed quantities.
Sections~\ref*{app:assumption-robust}.1--\ref*{app:assumption-robust}.2 develop the general linear geometry; Sections~\ref*{app:assumption-robust}.3--\ref*{app:assumption-robust}.6 retain optimization, transfer, repeated-error, and escape costs, before the paired audit in Section~\ref*{app:assumption-robust}.7.

\subsection{General designs, correlated estimation errors, and test geometry}
The design Gram describes which coefficients can be estimated; the test geometry describes which errors matter at evaluation. They play different roles, even when both are built from features.

The orthogonal design $X_a^\top X_a=n_aI_d$ is useful for isolating a scalar variance price, but it is not needed for an exact linear risk calculation.  Let $J$ role cells have parameter vectors $\theta_a\in\R^D$, stacked into $\theta\in\R^{JD}$.  Suppose an unbiased base estimator has the form
\begin{equation}
 Y=\theta+\zeta,\qquad \E\zeta=0,\qquad
 \E[\zeta\zeta^\top]=\Sigma\succeq0.
 \label{eq:robust-sequence-model}
\end{equation}
Here $Y$ denotes the stacked parameter estimates, not the earlier response matrix $Y_a$.  The covariance $\Sigma$ may have nonzero cross-role blocks.  Fix a symmetric positive-semidefinite matrix $W\in\R^{JD\times JD}$, and define $\|u\|_W^2=u^\top Wu$.  This is a squared seminorm: directions in the null space of $W$ do not affect the chosen evaluation loss.  A fixed matrix $L\in\R^{JD\times JD}$ specifies a \emph{linear sharing rule} $\widetilde\theta_L=LY$.  The risk is
\begin{equation}
 \Risk_W(L)=\E\|LY-\theta\|_W^2.
 \label{eq:general-smoother-risk-definition}
\end{equation}
``Fixed'' means chosen independently of $\zeta$, or conditioned on an independent pilot sample.  No Gaussian distribution is assumed.

\begin{theorem}[Exact covariance-aware risk of a fixed linear sharing rule]
\label{thm:general-smoother-risk}
Under \cref{eq:robust-sequence-model},
\begin{equation}
 \Risk_W(L)=\|(L-I)\theta\|_W^2+\tr(WL\Sigma L^\top).
 \label{eq:general-smoother-risk}
\end{equation}
If $\Sigma$ is known, then
\begin{equation}
 \widehat{\Risk}_W(L)=\|(L-I)Y\|_W^2
       +2\tr(WL\Sigma)-\tr(W\Sigma)
 \label{eq:general-unbiased-risk}
\end{equation}
is unbiased for the risk of this fixed $L$.
\end{theorem}
\begin{proof}
Write $LY-\theta=(L-I)\theta+L\zeta$.  Its deterministic--centered cross term has zero expectation, and $\E[\zeta^\top L^\top WL\zeta]=\tr(WL\Sigma L^\top)$, proving \cref{eq:general-smoother-risk}.  Also
\[
 \E\|(L-I)Y\|_W^2
 =\|(L-I)\theta\|_W^2+
   \tr\bigl(W(L-I)\Sigma(L-I)^\top\bigr).
\]
Expanding the last trace and using symmetry of $W,\Sigma$ leaves the risk minus $2\tr(WL\Sigma)+\tr(W\Sigma)$, proving unbiasedness.
\end{proof}
This is the fixed-linear-estimator case of unbiased risk estimation, with an elementary second-moment proof \citep{stein1981estimation}.  Unbiasedness for each pre-specified $L$ does not imply unbiasedness of the minimum over data-selected rules.  Selecting a grouping on these scores requires independent evaluation or uniform selection control.  Replacing $\Sigma$ by an estimate also adds an estimation problem.  The unbiased score may be negative in a sample even though its expected risk is nonnegative.

\paragraph{Instantiation for nonorthogonal operator regression.}
For each role, let $X_a\in\R^{n_a\times d}$ have full column rank, define its design Gram matrix $G_a=X_a^\top X_a\succ0$, and observe $Y_a=X_aA_a^\top+\mathcal E_a$.  Assume the noise rows are centered, independent, and have output covariance $\Omega_a\in\R^{p\times p}$.  Then
\begin{equation}
 \widehat A_a=Y_a^\top X_aG_a^{-1},\qquad
 \operatorname{Cov}(\operatorname{vec}(\widehat A_a-A_a))
       =G_a^{-1}\otimes\Omega_a.
 \label{eq:general-design-ols}
\end{equation}
The operation $\operatorname{vec}$ stacks a matrix's columns; here $\otimes$ is the Kronecker product, which forms a block matrix by multiplying each entry of its first argument by the second matrix.  Independence across roles makes the stacked covariance block diagonal; correlated role fits instead require their actual cross-covariances.

Let the independent test input in role $a$ have second moment $T_a=\E[x_ax_a^\top]\succeq0$, and let evaluation probabilities $\pi_a>0$ sum to one.  Since
\begin{equation}
 \E_{x_a}\|B x_a\|_2^2
 =\tr(BT_aB^\top)
 =\operatorname{vec}(B)^\top(T_a\otimes I_p)\operatorname{vec}(B),
 \label{eq:test-geometry-identity}
\end{equation}
choose
\begin{equation}
 W=\operatorname{blockdiag}_{a=1}^{J}
       \bigl(\pi_a(T_a\otimes I_p)\bigr).
 \label{eq:general-test-weight}
\end{equation}
Then \cref{eq:general-smoother-risk} is exactly the test prediction risk.  This accommodates anisotropic representations, unequal test frequencies, and ill-conditioned but full-rank designs.  A nearly singular $G_a$ increases estimation variance through $G_a^{-1}$; role count alone does not describe this difficulty.  For rank-deficient designs, OLS is not an unbiased estimator of all coordinates without an identifiability restriction, so \cref{eq:robust-sequence-model} cannot simply be assumed.

\subsection{The exact pooling frontier survives arbitrary covariance}
For this subsection suppose $W\succ0$, so it defines an inner product $\langle u,v\rangle_W=u^\top Wv$.  A fixed sharing subspace $\mathcal V\subseteq\R^{JD}$ consists of all parameter arrays allowed by a tying rule.  If the columns of a full-column-rank matrix $B_{\mathcal V}$ form a basis for that subspace, its weighted orthogonal projector is
\begin{equation}
 P=B_{\mathcal V}(B_{\mathcal V}^\top WB_{\mathcal V})^{-1}
       B_{\mathcal V}^\top W.
 \label{eq:weighted-projector-definition}
\end{equation}
For the zero subspace define $P=0$.  In either case $P^2=P$, $P^\top W=WP$, and $Pu$ is the closest admissible vector to $u$ in the $W$-norm.  It need not be symmetric in ordinary Euclidean coordinates.  Let $Q=I-P$ and define
\begin{equation}
 B_W=\|Q\theta\|_W^2,\quad
 V_0=\tr(W\Sigma),\quad
 V_P=\tr(WP\Sigma P^\top),\quad
 C_W=\tr(WQ\Sigma Q^\top)\ge0.
 \label{eq:weighted-pooling-quantities}
\end{equation}
The bias $B_W$ is the population cost of the sharing restriction, $V_0$ is the no-sharing estimation risk, and $C_W$ is the noise variance removed by that restriction.

\begin{theorem}[Covariance- and geometry-aware partial pooling]
\label{thm:weighted-projection-frontier}
For $L_\lambda=I-\lambda Q$, $0\le\lambda\le1$, under \cref{eq:robust-sequence-model} and $W\succ0$,
\begin{equation}
\begin{aligned}
 V_0&=V_P+C_W,\\
 \Risk_W(L_\lambda)&=V_0-2C_W\lambda+(B_W+C_W)\lambda^2.
\end{aligned}
 \label{eq:weighted-projection-frontier}
\end{equation}
Full projection improves expected risk over $L=I$ exactly when $B_W<C_W$.  If $B_W+C_W>0$, the optimal fixed coefficient is $\lambda^\star=C_W/(B_W+C_W)$; if both vanish, every coefficient has the same risk.  No Gaussianity, isotropic covariance, or independence across role estimates is required.
\end{theorem}
\begin{proof}
Weighted orthogonality gives $P^\top WQ=0$.  Consequently the noise components $P\zeta$ and $Q\zeta$ are orthogonal in the evaluation geometry for every realization, even if they are statistically dependent.  Therefore $V_0=V_P+C_W$.  The error decomposes as
\[
 L_\lambda Y-\theta=-\lambda Q\theta+P\zeta+(1-\lambda)Q\zeta.
\]
The bias--noise cross term has expectation zero, while the two projected noise components are pointwise orthogonal.  Thus the risk is $\lambda^2B_W+V_P+(1-\lambda)^2C_W$.  Expansion and minimization prove the claims.
\end{proof}
This strengthens the scalar frontier.  If every role estimate has exactly the same random error and the sharing subspace contains all common-role arrays, then $Q\zeta=0$ and $C_W=0$: averaging does not remove that common error.  Full sharing then only adds bias.  Conversely, covariance concentrated in directions excluded by the sharing subspace makes pooling especially useful.

For common-operator sharing in \cref{eq:general-test-weight}, the fitted shared matrix is
\begin{equation}
 \widehat A_{\mathrm{test}}
 =\left(\sum_a\pi_a\widehat A_aT_a\right)
   \left(\sum_a\pi_aT_a\right)^{-1},
 \label{eq:test-weighted-common-operator}
\end{equation}
provided all $T_a\succ0$.  This is the weighted projection estimator, not necessarily ordinary pooled training OLS.  Pooled OLS weights roles by their training design Grams; \cref{eq:test-weighted-common-operator} weights them by the declared test geometry.  

\begin{corollary}[Exact value of enlarging a sharing architecture]
\label{cor:nested-sharing-frontier}
Let $\mathcal U\subseteq\mathcal V$ be two fixed sharing subspaces, with weighted projectors $P_{\mathcal U},P_{\mathcal V}$.  Define $Q_{\mathrm{new}}=P_{\mathcal V}-P_{\mathcal U}$.  Then
\begin{equation}
 \Risk_W(P_{\mathcal U})-\Risk_W(P_{\mathcal V})
 =\|Q_{\mathrm{new}}\theta\|_W^2
  -\tr(WQ_{\mathrm{new}}\Sigma Q_{\mathrm{new}}^\top).
 \label{eq:nested-sharing-frontier}
\end{equation}
Thus an additional depth effect, role correction, or fixed adapter direction is useful exactly when its recovered signal exceeds its covariance-weighted estimation cost.
\end{corollary}
\begin{proof}
Because the subspaces are nested, $Q_{\mathrm{new}}$ projects onto the directions in $\mathcal V$ weighted-orthogonal to $\mathcal U$.  Both bias and noise split into weighted-orthogonal components.  Enlarging the subspace removes squared bias $\|Q_{\mathrm{new}}\theta\|_W^2$ and adds variance $\tr(WQ_{\mathrm{new}}\Sigma Q_{\mathrm{new}}^\top)$.
\end{proof}
The earlier dimension penalty is recovered when $W=I/J$ and $\Sigma=vI$.  In general, equal parameter counts do not imply equal estimation costs.  Learned subspaces or ranks must still be selected independently or analyzed with a selection penalty.

\subsection{Approximate training and mismatched role weights}
  Fix one group and the evaluation weight vector $\pi$, allowing zero entries provided the empirical losses are defined for every role used in training.  Let $\nu$ be the normalized weight vector used by the training algorithm.  Define
\begin{equation}
 \widehat R_\pi(f)=\sum_a\pi_a\widehat R_a(f),\quad
 \widehat R_\nu(f)=\sum_a\nu_a\widehat R_a(f),\quad
 \TV(\pi,\nu)=\tfrac12\sum_a|\pi_a-\nu_a|.
 \label{eq:training-weight-objectives}
\end{equation}
Assume $0\le\ell_a\le1$ and $\widehat f\in\cF$ satisfies
$\widehat R_\nu(\widehat f)\le\inf_{f\in\cF}\widehat R_\nu(f)+\eta_{\mathrm{opt}}$, where $\eta_{\mathrm{opt}}\ge0$ is an empirical optimization gap.  The gap may be computationally unknown and is retained explicitly in the bound.

\begin{proposition}[Approximate ERM with explicit weight mismatch]
\label{prop:approximate-weighted-erm}
On any event with $\sup_f|R_\pi(f)-\widehat R_\pi(f)|\le u_\pi$,
\begin{equation}
 R_\pi(\widehat f)\le\inf_fR_\pi(f)+2u_\pi
       +\eta_{\mathrm{opt}}+2\TV(\pi,\nu).
 \label{eq:approx-erm-mismatch-bound}
\end{equation}
The training weights $\nu$ may themselves be data dependent: the bound uses their realized distance from the fixed evaluation weights, while the uniform event is over the fixed class and fixed $\pi$.
\end{proposition}
\begin{proof}
For every $f$, bounded losses imply
$|\widehat R_\pi(f)-\widehat R_\nu(f)|\le\TV(\pi,\nu)$.
Move from $R_\pi(\widehat f)$ to its empirical value, then to the training objective, apply approximate minimization, and reverse the two changes at a population-optimal candidate (or an arbitrarily close one).  The two uniform deviations contribute $2u_\pi$ and the two weight changes contribute $2\TV(\pi,\nu)$.
\end{proof}
In \cref{thm:terminal-certificate}, the shared certificate therefore acquires
$\sum_eW_e[\eta_{\mathrm{opt},e}+2\TV(\pi_e,\nu_e)]$.
If separate fits have gaps $\eta_a$, their certificate acquires $\sum_aw_a\eta_a$.  The corrected bound difference is the original $\PRG_H$ plus the separate correction minus the shared correction.  Zero-evaluation-weight roles can influence training through $\nu$ and are now explicitly charged rather than discarded.  For task losses unrelated to the event-lift loss, an optimization-gap or calibration bound remains necessary.

\subsection{Transfer without identical distribution supports}
A bounded density ratio is a strong sufficient condition: it fails if rollout visits states outside the probe distribution's support.  Even on overlapping supports, two shifted Gaussians with equal variance generally have an unbounded density ratio.  The following replacement needs no absolute continuity.

Use the conditional occurrence measures $Q_{j,q,a}^m$ from \cref{eq:app-hybrid-subprobability}.  Let $s_{j,q,a}^m=Q_{j,q,a}^m(\cZ_a)$ be the probability that occurrence $q$ exists and has role $a$.  When $s_{j,q,a}^m>0$, let $\nu_{j,q,a}^m=Q_{j,q,a}^m/s_{j,q,a}^m$ be its conditional probe distribution.  Define a discrepancy over the actual loss class by
\begin{equation}
 \mathfrak d_a(\nu,P_a)
 =\sup_{f\in\cF_a}
   \left|\E_{Z\sim\nu}\ell_a(f;Z)-\E_{Z\sim P_a}\ell_a(f;Z)\right|.
 \label{eq:loss-class-discrepancy}
\end{equation}
The class $\cF_a$ is the applicable fixed expert class for role $a$.  This discrepancy measures changes in task loss, not changes in arbitrary features; for losses in $[0,1]$ it is at most one.

\begin{proposition}[Additive occupancy transfer]
\label{prop:additive-occupancy-transfer}
Assume the call-replacement lemma applies and choose deterministic occurrence bounds
$s_{j,q,a}^m\le\bar s_{j,q,a}$, valid for both comparators independently of expert-fitting data.  Then \cref{ass:role-aggregation} holds with
\begin{equation}
\begin{aligned}
 \omega_{j,a}&=\sum_q\bar s_{j,q,a},\\
 \widetilde\tau_j^m&=\tau_j^m+
 \sum_{q,a}s_{j,q,a}^m
       \mathfrak d_a(\nu_{j,q,a}^m,P_a).
\end{aligned}
 \label{eq:additive-shift-certificate}
\end{equation}
Zero-mass occurrences contribute zero.  If the probe space has metric $d_{\cZ_a}$ and every $\ell_a(f;\cdot)$ is uniformly $L_a^{\mathrm{probe}}$-Lipschitz, then, for distributions with finite first moments,
\begin{equation}
 \mathfrak d_a(\nu,P_a)
 \le L_a^{\mathrm{probe}}W_1(\nu,P_a;d_{\cZ_a}).
 \label{eq:wasserstein-probe-transfer}
\end{equation}
For bounded losses, the minimum of this expression and one is also valid.
\end{proposition}
\begin{proof}
For the fitted function, conditional on its fitting data,
\[
 \int\ell_a(\widehat f;z)\,dQ_{j,q,a}^m(z)
 =s_{j,q,a}^m\E_{\nu_{j,q,a}^m}\ell_a(\widehat f;Z)
 \le\bar s_{j,q,a}R_a(\widehat f)
   +s_{j,q,a}^m\mathfrak d_a(\nu_{j,q,a}^m,P_a).
\]
Nonnegativity of $R_a$ justifies replacing $s$ by $\bar s$ only in its coefficient.  Sum the call losses.  For \cref{eq:wasserstein-probe-transfer}, couple $Z\sim\nu$ and $Z'\sim P_a$.  For every $f$ the expectation difference is at most $L_a^{\mathrm{probe}}\E d_{\cZ_a}(Z,Z')$.  Minimize over couplings and take the supremum over $f$.
\end{proof}
The coupling argument is the standard mechanism behind Wasserstein transfer bounds \citep{redko2017theoretical}; here it is applied to role-conditioned event probes and their occurrence masses.  Small transport distance is useful only when the loss class has controlled smoothness.    The additive shift term must be bounded or estimated independently, and its uncertainty reported.

\subsection{Repeated experts create coherent, rather than fresh, errors}
The off-diagonal terms in \cref{eq:gramian-exact} have a direct interpretation when a fitted parameter is reused.  Consider the exact additive-error model
\begin{equation}
 \xi_j=b_j+B_j\delta_{e_j}+\epsilon_j.
 \label{eq:shared-error-model}
\end{equation}
Here the expert-index sequence $e_j$ and matrices $M_j,B_j$ are deterministic.  Each $\delta_e\in\R^{d_e}$ is a centered random fitting error drawn once per expert and reused at every occurrence of that expert, with covariance $V_e$.  The $s\times d_{e_j}$ matrix $B_j$ maps parameter error into a local state perturbation.  The errors $\epsilon_j\in\R^s$ are centered with covariances $R_j$, pairwise uncorrelated across time and uncorrelated with all $\delta_e$; distinct $\delta_e$ are uncorrelated.  All second moments are finite.  Define
\begin{equation}
 K_e=\sum_{j:e_j=e}P_jB_j.
 \label{eq:reuse-coherence-matrix}
\end{equation}
Thus $K_e$ adds all uses of expert $e$ after mapping them into terminal-state coordinates.

\begin{proposition}[Exact covariance cost of repeated expert use]
\label{prop:shared-error-coherence}
For the linear recursion in \cref{prop:gramian-amplification} and \cref{eq:shared-error-model},
\begin{equation}
 \E\|\mathbf e_H\|_2^2
 =\left\|\sum_jP_jb_j\right\|_2^2
   +\sum_e\tr(K_eV_eK_e^\top)
   +\sum_j\tr(P_jR_jP_j^\top).
 \label{eq:reuse-coherence-risk}
\end{equation}
\end{proposition}
\begin{proof}
Unroll the recursion and collect repeated occurrences of the same random vector:
$\mathbf e_H=\sum_jP_jb_j+\sum_eK_e\delta_e+\sum_jP_j\epsilon_j$.
The specified cross-covariances vanish between these collected components.  Take their covariance and then its trace.
\end{proof}
For the scalar example $M_j=B_j=1$, $b_j=\epsilon_j=0$, one expert used $H$ times with fitting-error variance $v$ gives $\E e_H^2=H^2v$.  Fresh independent errors with the same variance at each call instead give $Hv$.  Oppositely signed propagated effects may cancel, so reuse does not always increase the covariance cost.  The relevant quantity is $K_e$, not merely occurrence count.

Equation~\eqref{eq:shared-error-model} is exact for additive residual perturbations.  For a nonlinear expert it may be a first-order linearization; applying it then requires a separate remainder bound.  Also, conditioning on a route sequence selected using $\delta_e$ generally changes the error means and covariances.  The deterministic/exogenous-route specialization must not be justified by conditioning on arbitrary endogenous routes and retaining the original $V_e$.

\subsection{Local stability regions with an explicit escape cost}
Uniform control on every possible hidden state can be much stronger than needed.  Let $\Omega$ be the event that the reference, learned, and required hybrid states remain in specified deterministic regions where the local recursion and a utility Lipschitz constant $L_{\mathcal U}$ are valid.  Assume common initial states and that terminal utility is globally bounded in an interval of width $R_{\mathcal U}$.  Let $A_j$ be the products of the local reference Lipschitz envelopes.  Conditional on the fitted models, suppose
\[
 \E\left[\mathbf1_{\Omega}\sum_j A_j(a_j+r_j)\right]\le U_{\mathrm{loc}},
 \qquad \Prb(\Omega^c)\le p_{\mathrm{exit}}.
\]
Here $U_{\mathrm{loc}}$ controls only paths remaining inside the region, and $p_{\mathrm{exit}}$ is the probability of leaving it.
\begin{proposition}[Localized utility certificate]
\label{prop:localized-certificate}
Under these conditions,
\begin{equation}
 \E|\mathcal U(\widehat S_H)-\mathcal U(S_H)|
 \le L_{\mathcal U}U_{\mathrm{loc}}+R_{\mathcal U}p_{\mathrm{exit}}.
 \label{eq:localized-utility-certificate}
\end{equation}
\end{proposition}
\begin{proof}
On $\Omega$, apply the pathwise recursion and the local utility bound.  On $\Omega^c$, use the global utility range.  Split the expectation into these two events.
\end{proof}
The indicator-restricted probe measures can be treated by the same nonnegative-loss aggregation argument as other sub-probability measures.  No independence of $\Omega$ and local errors is required.  A local analysis is informative only when the escape probability is also controlled.

\subsection{Expected accuracy and actual improvement are separate claims}
A Lipschitz state certificate does not directly bound discrete answer accuracy.  Let the same readout $g:\cS\to\R^K$ be $L_g$-Lipschitz from $d$ to $\ell_\infty$, let $E=d(\widehat S_H,S_H)$, and let $\gamma\ge0$ be the reference top-two logit gap.  For any deterministic $t>0$,
\begin{equation}
 \Prb(\text{different predicted answers})
 \le \Prb(\gamma\le2L_gt)+\Prb(E>t).
 \label{eq:expected-answer-margin}
\end{equation}
This follows by splitting on $E\le t$ and applying the pathwise margin test.  It needs no independence between the margin and state error.  A moment bound gives $\Prb(E>t)\le\E E^r/t^r$ for any $r>0$.  A model-specific readout adds its common-state logit discrepancy to $L_gE$.  This controls disagreement with the reference; correctness relative to ground truth additionally requires the reference's own accuracy.

One can instead test actual improvement directly.  Fix $K_{\mathrm{cand}}$ fitted candidate systems and a separate comparator before observing an independent IID audit set of size $N$.  All losses $\ell_m(Z)$ and $\ell_{\mathrm{sep}}(Z)$ lie in $[0,1]$ and are measured on the same example $Z$.  Define the paired differences and their means by
\begin{equation}
 D_{i,m}=\ell_{\mathrm{sep}}(Z_i)-\ell_m(Z_i),\quad
 \widehat\Delta_m=\frac1N\sum_iD_{i,m},\quad
 \Delta_m=\E D_{i,m}.
 \label{eq:paired-terminal-audit}
\end{equation}
Positive $\Delta_m$ means lower true risk for candidate $m$ under this audit distribution.

\begin{proposition}[A held-out certificate of actual performance improvement]
\label{prop:paired-real-improvement}
Conditional on fitted systems, with probability at least $1-\delta$,
\begin{equation}
 \Delta_m\ge\widehat\Delta_m-
 \sqrt{\frac{2\log(K_{\mathrm{cand}}/\delta)}{N}}
 \qquad\text{simultaneously for all }m.
 \label{eq:paired-true-improvement-bound}
\end{equation}
Thus exceeding this threshold certifies improvement of a fitted system's true audit-distribution risk, not just improvement of an upper bound.
\end{proposition}
\begin{proof}
Each paired difference lies in $[-1,1]$.  One-sided Hoeffding gives
$\Prb(\widehat\Delta_m-\Delta_m>t)\le\exp(-Nt^2/2)$.
Set the probability to $\delta/K_{\mathrm{cand}}$ and take a union bound.  Pairing requires no independence between the two models' losses on one example; independence is across audit records.
\end{proof}
The same simultaneous event permits selection among this pre-fitted finite candidate family.  Repeatedly generating new candidates in response to the audit results invalidates this guarantee without further selection control.  For unbounded squared loss, the bounded Hoeffding radius cannot be used; a justified tail model or a separately declared bounded evaluation loss is needed.  This audit and $\PRG_H$ answer different questions: the latter explains a mechanism under modeling assumptions, whereas the former tests actual performance on a specified population.

\section{Terminal-Risk Frontiers and Valid Adaptive Selection}
\label{app:terminal-selection}

\paragraph{Overview.}
We connect local sharing to a common terminal geometry, exhibit a reversal of local improvement, and characterize a restricted setting in which every local full-pooling gain survives. Independent selection and nonlinear approximation require separate guarantees.  Fixed matrices may instead be conditioned on an independent pilot, provided the stated conditional moments hold; this does not permit treating a same-data selected rule as fixed.
Section~\ref*{app:terminal-selection}.1 derives the terminal frontier and reversal, Section~\ref*{app:terminal-selection}.2 treats independent finite-family selection, Section~\ref*{app:terminal-selection}.3 retains the nonlinear remainder, and Section~\ref*{app:terminal-selection}.4 characterizes uniform preservation in its stated submodel.

\subsection{The terminal geometry can change the preferred amount of sharing}
Let $Y=\theta+\zeta\in\R^p$ collect unbiased estimates of $p$ role coefficients, with fixed target $\theta$, $\E\zeta=0$, and $\E[\zeta\zeta^\top]=\Sigma\succeq0$.  A fixed linear sharing rule $L$ estimates $\theta$ by $LY$.  Suppose the terminal error of every compared rule is exactly
\begin{equation}
 e_H(L)=G(LY-\theta),\qquad G\in\R^{s\times p}.
 \label{eq:terminal-linear-estimation-map}
\end{equation}
The same propagation matrix $G$ is used for all comparators and is independent of $\zeta$.  For example, in \cref{prop:gramian-amplification}, if each local injection is $\xi_j=H_j(LY-\theta)$ with fixed $H_j$, then $G=\sum_jP_jH_j$.  This sums repeated uses before calculating their covariance.  Define $W_H=G^\top G$; it is positive semidefinite and may be singular, so $\|x\|_{W_H}^2=x^\top W_Hx$ is a squared seminorm rather than necessarily a squared norm.

\begin{theorem}[Exact sharing frontier in terminal geometry]
\label{thm:terminal-geometry-frontier}
Under \cref{eq:terminal-linear-estimation-map}, for every fixed $L$,
\begin{equation}
 R_H(L):=\E\|e_H(L)\|_2^2
 =\|(L-I)\theta\|_{W_H}^2+\tr(W_HL\Sigma L^\top).
 \label{eq:terminal-linear-risk}
\end{equation}
Fix a pooling matrix $P$ independently of $\zeta$, put $Q=I-P$, and let $L_\lambda=I-\lambda Q$ for $0\le\lambda\le1$.  Neither Euclidean orthogonality nor orthogonality in $W_H$ is presumed.  Set
\begin{equation}
 \begin{aligned}
 a_H&=\tr(W_HQ\Sigma),\\
 b_H&=\|Q\theta\|_{W_H}^2+\tr(W_HQ\Sigma Q^\top)\ge0.
 \end{aligned}
 \label{eq:terminal-frontier-coefficients}
\end{equation}
Then
\begin{equation}
 R_H(L_\lambda)=R_H(I)-2a_H\lambda+b_H\lambda^2.
 \label{eq:terminal-risk-quadratic}
\end{equation}
Full pooling strictly improves terminal risk if and only if $2a_H>b_H$.  If $b_H>0$, an optimal coefficient is
\begin{equation}
 \lambda_H^\star=\min\{1,\max\{0,a_H/b_H\}\}.
 \label{eq:terminal-oracle-lambda}
\end{equation}
If $b_H=0$, necessarily $a_H=0$ and all coefficients have the same risk.
\end{theorem}
\begin{proof}
Expand $LY-\theta=(L-I)\theta+L\zeta$.  The centered cross term vanishes and $\E\|GL\zeta\|^2=\tr(W_HL\Sigma L^\top)$, proving \cref{eq:terminal-linear-risk} even when $W_H$ is singular.  Substitution of $L_\lambda$ gives a deterministic bias $-\lambda GQ\theta$ and variance
\[
 \E\|G\zeta-\lambda GQ\zeta\|^2
 =\E\|G\zeta\|^2-2\lambda\E\langle G\zeta,GQ\zeta\rangle
       +\lambda^2\E\|GQ\zeta\|^2.
\]
The middle expectation equals $a_H$ by the trace identity and symmetry of $W_H,\Sigma$.  This proves the quadratic and the strict full-pooling comparison.  Minimize the convex quadratic on $[0,1]$ when $b_H>0$.  When $b_H=0$, both $GQ\theta=0$ and $GQ\zeta=0$ almost surely; hence the middle expectation is zero as well.
\end{proof}

\paragraph{Why the linear term need not be a nonnegative variance saving.}
If $P$ is a projection and $P^\top W_HQ=0$, then $a_H=\tr(W_HQ\Sigma Q^\top)\ge0$: its two projected output components are pointwise orthogonal.  General pooling need not satisfy this condition.  For example, take $\theta=0$, $P=\operatorname{diag}(1,0)$, $G=(1,1)$, and
\[
 \Sigma=\begin{pmatrix}4&-3/2\\-3/2&1\end{pmatrix}\succ0.
\]
Then $a_H=-1/2$, $b_H=1$, and $R_H(L_\lambda)=2+\lambda+\lambda^2$.  Suppressing a component removes beneficial cancellation and hurts every positive sharing strength.  

\begin{proposition}[Local pooling improvement can reverse at the terminal state]
\label{prop:local-terminal-reversal}
There is a three-role linear model in which full pooling reduces average local mean squared error from $1$ to $5/6$ but increases terminal mean squared error from $6$ to $9$.  The reversal persists if local evaluation and pooling use weights proportional to the magnitudes of the terminal propagation coefficients.
\end{proposition}
\begin{proof}
Let $Y_a=\theta_a+\varepsilon_a$ with independent centered errors of variance one and
$\theta=(1/2,-1,1/2)^\top$.  Local separate risk is one.  For uniform pooling, $\bar\theta=0$, the average squared role discrepancy is $1/2$, and the pooled variance is $1/3$, giving $5/6$.

Take the reference events, from the common initial state $S_0=0$, to be
\begin{equation}
 S_1=S_0+\theta_1,\qquad
 S_2=-\tfrac12S_1+\theta_2,\qquad
 S_3=-2S_2+\theta_3.
 \label{eq:reversal-reference-events}
\end{equation}
Each learned model uses the same events with its fitted coefficients.  Thus $G=(1,-2,1)$, separate terminal risk is $\|G\|_2^2=6$, and $G\mathbf1=0$, $G\theta=3$.  Full pooling uses $\bar Y\mathbf1$ and has the constant error $G(\bar Y\mathbf1-\theta)=-3$, hence risk nine.  For partial uniform pooling,
\begin{equation}
 R_H(L_\lambda)=6(1-\lambda)^2+9\lambda^2
 =6-12\lambda+15\lambda^2,
 \label{eq:reversal-partial-risk}
\end{equation}
which is minimized at $\lambda=2/5$ with risk $18/5$.

For the alternative weights $\pi=(1/4,1/2,1/4)$, use the weighted pool $\sum_a\pi_aY_a$.  Its weighted target mean is $-1/4$, weighted squared bias is $9/16$, and variance is $\sum_a\pi_a^2=3/8$.  Weighted local risk is therefore $15/16<1$.  Any common scalar pool is still annihilated by $G\mathbf1=0$, so its terminal risk remains nine.
\end{proof}
  Here the terminal-optimal partial rule substantially outperforms both endpoints.  

\subsection{Selection of finitely many groups, subspaces, or sharing strengths}
Let $\mathscr A$ be a finite set of $K\ge1$ fitted candidates, each fixed conditional on a training/pilot sigma-field $\mathscr T$.  Candidates can have different partitions, subspaces, or sharing strengths.  An independent validation sample has records indexed by $i=1,\ldots,N$, independent conditional on $\mathscr T$.  Losses of different candidates on the same record may be dependent.  Fix nonnegative weights $w_i$ summing to one, measurable with respect to $\mathscr T$ and thus fixed before validation, and define
\[
 \widehat R_m=\sum_iw_i\ell_{i,m},\qquad
 R_m=\sum_iw_i\E[\ell_{i,m}\mid\mathscr T],\qquad m\in\mathscr A.
\]
The risk is for this declared validation target; nonidentically distributed records do not define an unspecified IID population.  The candidates must not be created adaptively from these same validation outcomes.

\begin{theorem}[Independent finite-family selection with explicit loss tails]
\label{thm:finite-architecture-selection}
Assume the centered loss $X_{i,m}=\ell_{i,m}-\E[\ell_{i,m}\mid\mathscr T]$ obeys, conditional on $\mathscr T$,
\begin{equation}
 \log\E[e^{tX_{i,m}}\mid\mathscr T]
 \le\frac{v_{i,m}t^2}{2(1-b_{i,m}|t|)}
 \quad\text{for }|t|<1/b_{i,m},
 \label{eq:selection-tail-assumption}
\end{equation}
where $v_{i,m},b_{i,m}\ge0$ are fixed conditional on $\mathscr T$; when $b_{i,m}=0$ the domain is all real $t$.  For $\delta\in(0,1)$ set
\begin{equation}
 \begin{aligned}
 x&=\log(2K/\delta),&
 V_m&=\sum_iw_i^2v_{i,m},& B_m&=\max_i w_i b_{i,m},\\
 r_m&=\sqrt{2V_mx}+B_mx.
 \end{aligned}
 \label{eq:selection-confidence-radii}
\end{equation}
With conditional probability at least $1-\delta$, simultaneously for every candidate,
$|\widehat R_m-R_m|\le r_m$.  On this event the penalized selector
$\widehat m\in\argmin_m(\widehat R_m+r_m)$ satisfies
\begin{equation}
 R_{\widehat m}\le\min_{m\in\mathscr A}(R_m+2r_m).
 \label{eq:finite-selection-oracle}
\end{equation}
An unpenalized empirical minimizer instead satisfies
$R_{\widehat m}\le\min_mR_m+2\max_mr_m$.
\end{theorem}
\begin{proof}
Conditional independence and \cref{eq:selection-tail-assumption} imply that the log moment generating function of $\sum_iw_iX_{i,m}$ is at most $V_mt^2/[2(1-B_m|t|)]$.  For $V_m>0$ and the upper tail, use Chernoff's inequality at
$t=\sqrt{2x/V_m}/(1+B_m\sqrt{2x/V_m})$ to get probability at most $e^{-x}$ above $\sqrt{2V_mx}+B_mx$.  Apply the same argument to the negative sum.  If $V_m=0$, all positive-weight centered losses have zero variance and the claim is immediate.  A union bound over both tails of the $K$ candidates gives the simultaneous event.  On that event, for every $m$,
\[
 R_{\widehat m}\le\widehat R_{\widehat m}+r_{\widehat m}
 \le\widehat R_m+r_m\le R_m+2r_m.
\]
For the raw minimizer the same argument uses $r_{\widehat m}+r_m\le2\max_mr_m$.
\end{proof}

The tail hypothesis is often called a two-sided sub-gamma condition: $v$ controls the quadratic fluctuation scale and $b$ the range over which the exponential moment is controlled.  This tail condition is an additional assumption beyond covariance information.  For bounded losses in $[0,1]$, Hoeffding's lemma gives $v_{i,m}=1/4$, $b_{i,m}=0$, recovering \cref{prop:validation-lambda}.  To implement the penalized selector, valid radii or deterministic upper bounds on them must be available; replacing unknown parameters by estimates from the same validation outcomes needs a new guarantee.

\paragraph{A sufficient condition for unbounded quadratic validation loss.}
Conditional on training and fixed validation inputs (include those inputs in the conditioning information), suppose
$\ell_{i,m}=\|\mu_{i,m}+\epsilon_i\|_2^2$, where $\mu_{i,m}$ is fixed and the validation errors are independent $\epsilon_i\sim\mathcal N(0,\Omega_i)$.  Errors may be shared across candidates at the same record.  Then \cref{eq:selection-tail-assumption} holds with
\begin{equation}
 v_{i,m}=2\tr(\Omega_i^2)+4\mu_{i,m}^\top\Omega_i\mu_{i,m},
 \qquad b_{i,m}=2\|\Omega_i\|_{\mathrm{op}}.
 \label{eq:quadratic-validation-tail}
\end{equation}
Here $\|\Omega_i\|_{\mathrm{op}}$ is its largest eigenvalue.  To verify this, diagonalize $\Omega_i$ with eigenvalues $\omega_j\ge0$ and corresponding coordinates $\mu_j$.  The centered quadratic log moment generating function is
\[
 \sum_j\left[-\tfrac12\log(1-2t\omega_j)-t\omega_j
       +\frac{2t^2\omega_j\mu_j^2}{1-2t\omega_j}\right].
\]
For $|2t\omega_j|<1$, the power series for the logarithm bounds the first two terms by $t^2\omega_j^2/(1-2|t|\|\Omega_i\|_{\mathrm{op}})$; the last term is bounded with the same denominator.  Summing gives \cref{eq:quadratic-validation-tail}.  Null eigenvalues cause no difficulty because their centered contributions are zero.  Known $\Omega_i$ and a bound $\|\mu_{i,m}\|\le M_{i,m}$ permit the conservative substitution $\mu_{i,m}^\top\Omega_i\mu_{i,m}\le\|\Omega_i\|_{\mathrm{op}}M_{i,m}^2$.  Without such information these are theoretical, not automatically computable, radii.  

\paragraph{Independent selection and refitting give a different exact identity.}
If a pilot selects a rule $L=L(\mathscr T)$ and a fresh fitting error satisfies $\E[\zeta\mid\mathscr T]=0$, $\operatorname{Cov}(\zeta\mid\mathscr T)=\Sigma$, then conditioning and averaging \cref{eq:general-smoother-risk} yields the average of the fixed-rule risks.  The averaging uses fresh fitting errors after pilot selection.  Independence of the final audit also remains necessary for \cref{prop:paired-real-improvement}.

\subsection{Terminal risk with controlled nonlinear remainders}
The following result converts a controlled first-order remainder into an interval for actual terminal risk.
\begin{proposition}[Terminal risk intervals with a controlled remainder]
\label{prop:terminal-remainder-interval}
Let the actual terminal error be $e=e_{\mathrm{lin}}+r$ under the same probability law, with square-integrable vectors, $R_{\mathrm{lin}}=\E\|e_{\mathrm{lin}}\|_2^2$, and $\E\|r\|_2^2\le\varepsilon^2$.  No independence is required.  Then
\begin{equation}
 \bigl(\max\{0,\sqrt{R_{\mathrm{lin}}}-\varepsilon\}\bigr)^2
 \le\E\|e\|_2^2
 \le(\sqrt{R_{\mathrm{lin}}}+\varepsilon)^2.
 \label{eq:terminal-remainder-risk-interval}
\end{equation}
In particular, a reuse model has strictly smaller actual risk than a separate model if the reuse upper endpoint is smaller than the separate lower endpoint, using valid remainder bounds for both.
\end{proposition}
\begin{proof}
Use the triangle and reverse-triangle inequalities for the $L^2$ norm of random vectors: $\|e\|_{L^2}\le\|e_{\mathrm{lin}}\|_{L^2}+\|r\|_{L^2}$ and $\|e\|_{L^2}\ge\|e_{\mathrm{lin}}\|_{L^2}-\|r\|_{L^2}$.  Nonnegativity and squaring give the endpoints; strict endpoint separation gives the comparison.
\end{proof}

\paragraph{Two diagnostics against automatic extrapolation.}
For the scalar nonlinear map $e(\delta)=\delta^2$ at $\delta=0$, the linearization is zero, but $\delta\sim\mathcal N(0,v)$ gives true squared error $\E\delta^4=3v^2$.  Mere knowledge of the covariance does not bound a general fourth-order remainder.  In an endogenous linear-looking map, take $\delta\sim\mathcal N(0,1)$ and $G(\delta)=\mathbf1_{\{|\delta|>1\}}$.  If $\phi,\Phi$ denote the standard normal density and CDF, respectively,
\[
 \E\bigl[(G(\delta)\delta)^2\bigr]
 =2[\phi(1)+1-\Phi(1)]
 >2[1-\Phi(1)]=\E[G(\delta)^2]\E\delta^2.
\]
The inequality follows by integration by parts over $|\delta|>1$.  Averaging a fixed-$G$ formula while retaining unconditional noise covariance is therefore invalid when the route selected $G$ from the same error.  A valid analysis must use the resulting conditional means and covariances or a pathwise remainder bound.  For nonlinear neural traces, specifying the expansion point, reached region, derivatives, noise moments, and exit contribution is still a separate modeling and estimation task.

\Cref{app:parameter-endogenous} now supplies explicit joint-derivative remainder envelopes and conditional route moments for specified models.

\subsection{A corollary: when every local full-pooling gain survives}
\label{app:uniform-terminal-inheritance}
The reversal in \cref{prop:local-terminal-reversal} raises a uniform question: which fixed propagation geometries preserve a strict local full-pooling improvement for \emph{every} possible target?  The following is a structural corollary of \cref{thm:terminal-geometry-frontier}, proved by an elementary spectral argument.    The role of loss geometry in shrinkage is established in, for example, \citet[Section~5]{hansen2016efficient}.  Matrix-risk ordering, which controls all linear combinations, is also distinct from merely lowering its trace \citep[Section~1]{matsuda2021matrix}.  

Let $Y=\theta+\zeta\in\R^p$, with $\E\zeta=0$ and $\operatorname{Cov}(\zeta)=qI_p$, $q>0$.  Let $P=P^\top=P^2$ be a fixed Euclidean orthogonal pooling projection, $Q=I_p-P$, and $k=\operatorname{rank}(Q)>0$.  The separate and fully pooled estimates are $Y$ and $PY$.  Let the same fixed matrix $G\in\R^{s\times p}$ propagate either estimate's error, as in \cref{eq:terminal-linear-estimation-map}.  Define the pooled-minus-separate risk differences
\[
 \begin{aligned}
 \Delta_{\mathrm{loc}}(\theta)
 &=\E\|PY-\theta\|^2-\E\|Y-\theta\|^2,\\
 \Delta_H(\theta)
 &=\E\|G(PY-\theta)\|^2-\E\|G(Y-\theta)\|^2,
 \qquad A=QG^\top GQ.
 \end{aligned}
\]
These are unnormalized risks; dividing local risk by a positive role count leaves its sign unchanged.  The matrix $A$ is positive semidefinite, vanishes on $\operatorname{range}(P)$, and describes amplification restricted to the contrast space $\operatorname{range}(Q)$.  Only the stated second moments, not Gaussianity, are required.

\paragraph{Corollary of \cref{thm:terminal-geometry-frontier}.}
For these fixed $P,G,q$,
\[
 \left[\ \forall\theta\in\R^p:\quad
 \Delta_{\mathrm{loc}}(\theta)<0\ \Longrightarrow\ \Delta_H(\theta)<0\ \right]
 \quad\Longleftrightarrow\quad
 A=cQ\quad\text{for some }c>0.
\]
Equivalently, $G$ must be a positive scaled isometry on the discarded contrast space: $\|Gv\|^2=c\|v\|^2$ for every $v\in\operatorname{range}(Q)$.  This defines the term \emph{scaled isometry}; it imposes no such requirement on all of $\R^p$.

\begin{proof}
Since $P^2=P=P^\top$ and the noise is isotropic, the fixed-rule risk identities give
\[
 \Delta_{\mathrm{loc}}(\theta)=\|Q\theta\|^2-qk,
 \qquad
 \Delta_H(\theta)=\theta^\top A\theta-q\tr(A).
\]
For the second equality, $\tr(G^\top GP)-\tr(G^\top G)=-\tr(G^\top GQ)=-\tr(A)$.  Thus no additional assumption $P^\top G^\top GQ=0$ is needed.  If $A=cQ$ with $c>0$, then $\Delta_H=c\Delta_{\mathrm{loc}}$ for every target, proving sufficiency.

For necessity, if $A=0$, the target $\theta=0$ has $\Delta_{\mathrm{loc}}=-qk<0$ but $\Delta_H=0$, contradicting the strict implication.  Otherwise let $\lambda_{\max}>0$ be the largest eigenvalue of $A$ restricted to its $k$-dimensional contrast space.  If this restriction is not a scalar multiple of the identity, its largest eigenvalue exceeds the average:
\[
 \lambda_{\max}>\frac{\tr(A)}{k}.
\]
Choose a unit eigenvector $v\in\operatorname{range}(Q)$ for $\lambda_{\max}$ and a scalar $t$ satisfying
\[
 \frac{q\tr(A)}{\lambda_{\max}}<t<qk.
\]
The interval is nonempty.  At $\theta=\sqrt t\,v$, the local difference is $t-qk<0$, whereas the terminal difference is $t\lambda_{\max}-q\tr(A)>0$.  This strict reversal contradicts the uniform implication.  Hence the restriction must be $cI_k$ with $c>0$, equivalent to $A=cQ$ on the full space.
\end{proof}

\paragraph{Consequences of the all-target condition.}
The corollary generalizes the isolated reversal to every nonscalar contrast geometry and identifies exactly when no target can reverse a local gain.  At $k=1$, any nonzero restriction is automatically scalar.  The cases $k=0$ and $q=0$ are excluded because no strict local full-pooling gain is possible; the implication would then be vacuous.  The case $A=0$ preserves only equality of terminal risks, not strict improvement.  A single target may still benefit when $A\ne cQ$, so this is not a necessary condition for one particular model to benefit.  It compares full projection with separate estimation, not all possible partial-pooling coefficients or same-data adaptive rules.  An error-dependent $G$, a different $G$ for each comparator, anisotropic noise, or nonlinear/naturally rerouted execution requires its own analysis.

\section{Explicit Nonlinear, Endogenous-Route, and Adapter Specializations}
\label{app:parameter-endogenous}

\paragraph{Overview.}
We make three interfaces explicit: how derivatives control a nonlinear remainder, how route selection changes error moments, and which functional corrections a parameterization can realize. These resolve the conditional modeling questions in \cref{prop:terminal-remainder-interval,thm:terminal-geometry-frontier,prop:rank-r-functional}. Every specialization fixes its reference before taking expectations. Calculations use real arithmetic; floating-point error, a different architecture, or a changed input distribution needs its own accounting.
Sections~\ref*{app:parameter-endogenous}.1--\ref*{app:parameter-endogenous}.2 bound nonlinear remainders and block derivatives; Sections~\ref*{app:parameter-endogenous}.3--\ref*{app:parameter-endogenous}.4 analyze route-selected moments; Sections~\ref*{app:parameter-endogenous}.5--\ref*{app:parameter-endogenous}.6 distinguish adapter realizability from its initialization-dependent tangent image.

\subsection{A finite-horizon remainder from joint state--parameter derivatives}
\label{sec:joint-taylor-remainder}

Use Euclidean norms on states and parameters and the Frobenius norm when a
matrix is viewed as a parameter vector.  Let $d\in\R^p$ denote one parameter
perturbation, reused at every occurrence of the relevant parameters; it is not
redrawn at successive events.  Conditional on a fixed input and reference
parameters, define
\[
 s_{j+1}=F_j(s_j,0),\qquad
 \widehat s_{j+1}=F_j(\widehat s_j,d),\qquad \widehat s_0=s_0.
\]
Here $F_j$ is a twice continuously differentiable event, or a smooth extension of
one specified hard-routing branch.  A branch extension evaluates the chosen
expert and its prescribed smooth gain even at states where it would not win.
Actual hard-routed execution equals this extension only on a separately
certified route-agreement event.

For $M_j=D_sF_j(s_j,0)$ and $B_j=D_dF_j(s_j,0)$ define the linear error
\begin{equation}
 z_0=0,\quad z_{j+1}=M_jz_j+B_jd,\quad
 G_0=0,\quad G_{j+1}=M_jG_j+B_j,
 \qquad z_j=G_jd.
 \label{eq:joint-linear-recursion}
\end{equation}
The symbol $D^2F_j$ is a bilinear second derivative, with norm
$\sup_{\|v\|,\|w\|\le1}\|D^2F_j[v,w]\|$, on the product space of
state and parameter directions.  

\begin{theorem}[Localized finite-horizon Taylor remainder]
\label{thm:joint-taylor-remainder}
Fix deterministic constants $L_j,K_j,h_j\ge0$ that bound, respectively,
$\|D_sF_j\|$, $\|D_dF_j\|$, and $\|D^2F_j\|$ on specified regions.
Let $\Omega$ be an event on which actual execution uses the specified branches
and the segment from $(s_j,0)$ to $(\widehat s_j,d)$ lies in these regions
for every $j$.  Define deterministic sequences
\begin{equation}
 c_0=b_0=0,\qquad
 c_{j+1}=L_jc_j+K_j,\qquad
 b_{j+1}=\|M_j\|_2 b_j+\tfrac12h_j(c_j^2+1).
 \label{eq:remainder-envelope-recursion}
\end{equation}
Then, on $\Omega$, for every event,
\begin{equation}
 \|\widehat s_j-s_j\|_2\le c_j\|d\|_2,\qquad
 \|\widehat s_j-s_j-z_j\|_2\le b_j\|d\|_2^2.
 \label{eq:joint-taylor-pathwise}
\end{equation}
Write $e=\widehat s_H-s_H$ and $r=e-G_Hd$ under the full probability law.
If $q>2$, $\Prb(\Omega^c)\le p_{\rm exit}$, and
$\|e\|_{L^q}+\|G_Hd\|_{L^q}\le T_q<\infty$, then
\begin{equation}
 \E\|r\|_2^2
 \le b_H^2\E[\|d\|_2^4\mathbf1_\Omega]
       +T_q^2p_{\rm exit}^{\,1-2/q}.
 \label{eq:remainder-with-exit}
\end{equation}
Thus the square root of this right-hand side is a valid $\varepsilon$ in
\cref{prop:terminal-remainder-interval}.  If $\Prb(\Omega)=1$, only a finite
fourth moment is needed and the exit term is absent.
\end{theorem}
\begin{proof}
On the segment in the statement, the integral first-derivative formula gives
$\|\widehat s_{j+1}-s_{j+1}\|\le L_j\|\widehat s_j-s_j\|+K_j\|d\|$.
Induction proves the first bound.  The integral second-order Taylor formula at
$(s_j,0)$ has a remainder $t_j$ bounded by
$\frac12h_j(\|\widehat s_j-s_j\|^2+\|d\|^2)$.
Subtracting \cref{eq:joint-linear-recursion} gives
$r_{j+1}=M_jr_j+t_j$.  The first bound and induction give the stated $b_j$.
For the full-law bound, split $\E\|r\|^2$ over $\Omega$ and its complement.
H\"older's inequality gives
$\E[\|r\|^2\mathbf1_{\Omega^c}]
\le\|r\|_{L^q}^2\Prb(\Omega^c)^{1-2/q}$;
the $L^q$ triangle inequality bounds $\|r\|_{L^q}$ by $T_q$.
No independence between $d$, $\Omega$, $e$, and $r$ is used.
\end{proof}

\paragraph{Moments, small-noise scaling, and an obstruction.}
If the hypotheses hold globally and $d\sim\mathcal N(\mu,\Sigma)$, then
\begin{equation}
 \E\|d\|^4=(\|\mu\|^2+\tr\Sigma)^2
               +2\tr(\Sigma^2)+4\mu^\top\Sigma\mu.
 \label{eq:gaussian-fourth-norm}
\end{equation}
To verify it, write $d=\mu+Z$, diagonalize $\Sigma$, use independent centered
normal coordinates with fourth moments $3\sigma_i^4$, and note that odd terms
vanish.  In particular, for $d=\sigma Z$ and constants $b_H$ independent of
$\sigma$, the remainder norm is $O(\sigma^2)$ if $\E\|Z\|^4<\infty$.
For centered $Z$, the linear risk is $\sigma^2\tr(G_H\operatorname{Cov}(Z)G_H^\top)$;
an $o(1)$ relative-risk error additionally requires this coefficient to be
strictly positive.  A zero linear risk provides no relative guarantee.
Covariance alone cannot replace the fourth moment: let $d_N$ equal
$\sqrt N,-\sqrt N,0$ with probabilities $1/(2N),1/(2N),1-1/N$.
Then $\E d_N^2=1$ but $\E d_N^4=N$.  The map $F(d)=d^2$ has constant
Hessian two and unbounded squared remainder along this family.
A small escape probability alone is likewise insufficient for unbounded state
loss: an error of size $p^{-1}$ on an event of probability $p$ has risk $p^{-1}$.
A bounded \emph{utility} does not bound this state-space remainder.

\paragraph{From one prefix to an input population.}
The statement can be applied conditional on a random input $X$ without assuming
that the parameter error is independent of $X$.  If its conditional premises
hold with measurable constants and a finite right-hand side, then
\begin{equation}
 \E\|r\|^2\le\E_X\!\left[
 b_H(X)^2\E[\|d\|^4\mathbf1_\Omega\mid X]
 +T_q(X)^2p_{\rm exit}(X)^{1-2/q}\right].
 \label{eq:input-integrated-remainder}
\end{equation}
This is the tower property applied to \cref{eq:remainder-with-exit}.  The required integrability can fail if small margins
or large derivative constants occur on the input population.

\subsection{Explicit constants for the normalized gated blocks in the preparation code}
\label{sec:explicit-block-constants}

We now specify a concrete specialization.  Freeze the weights of attention, normalization,
router, the up/gate matrices, and any depth correction (not their activations); perturb only expert
\emph{down} matrices.  This matches the mathematical form of the corresponding
subblock in \texttt{experiments/transformer\_core.py}.  Joint changes to the router/up/gate/normalizer or finite-precision execution require new terms. For one token let
\begin{equation}
 \begin{split}
 N(x)&=\gamma\odot\frac{\Pi x}{(\|\Pi x\|^2/d_x+\epsilon_N)^{1/2}}+\beta,
       \quad \Pi=I-\mathbf1\mathbf1^\top/d_x,\quad\epsilon_N>0,\\
 h(u)&=\operatorname{SiLU}(Vu)\odot(Uu),\qquad
 g_e(u)=\operatorname{softmax}(Ru)_e,\\
 F_e(x,\Delta)&=x+(D_e+\Delta)\,[g_e(N(x))h(N(x))]+A N(x).
 \end{split}
 \label{eq:explicit-normalized-expert}
\end{equation}
Here $d_x$ is state width, $\gamma,\beta$ are fixed LayerNorm scale and bias,
$U,V$ are expert up and gate matrices, $D_e$ is its down matrix, $R$ the router
matrix, and $A$ an optional fixed depth-correction matrix (zero if absent).
The entrywise product is $\odot$, and
$\operatorname{SiLU}(t)=t/(1+e^{-t})$.
The gain is the selected softmax probability, not a unit gate; support
agreement does not freeze it.  A deliberately frozen gain in $[0,1]$ is also
allowed by setting its derivatives below to zero.

\begin{proposition}[Explicit smooth-branch derivative envelopes]
\label{prop:explicit-block-envelopes}
Put $g=\|\gamma\|_\infty$, and define
\begin{equation}
 \begin{aligned}
 R_N&=g\sqrt{d_x}+\|\beta\|_2,&
 L_N&=g/\sqrt{\epsilon_N},&
 H_N&=6g/(\sqrt{d_x}\epsilon_N),\\
 a&=\|U\|_2,&v&=\|V\|_2,&T&=vR_N,\\
 s_1&=1+T/4,&s_2&=1/2+T/4.
 \end{aligned}
 \label{eq:normalization-activation-constants}
\end{equation}
Valid bounds for $h$ and its first two derivatives on $\|u\|\le R_N$ are
\begin{equation}
 H_0=avR_N^2,\qquad
 H_1=avR_N(s_1+1),\qquad
 H_2=aR_Ns_2v^2+2s_1av.
 \label{eq:gated-feature-constants}
\end{equation}
Let $r_R=\|R\|_2$, $g_1=2r_RL_N$, and
$g_2=6r_R^2L_N^2+2r_RH_N$.  For $k_e(x)=g_e(N(x))h(N(x))$, take
\begin{equation}
 \begin{split}
 k_0&=H_0,\qquad k_1=H_1L_N+g_1H_0,\\
 k_2&=H_2L_N^2+H_1H_N+2g_1H_1L_N+g_2H_0.
 \end{split}
 \label{eq:gain-feature-constants}
\end{equation}
Uniformly for all states and $\|\Delta\|_F\le t$, the constants
\begin{equation}
 \begin{split}
 L&=1+(\|D_e\|_2+t)k_1+\|A\|_2L_N,\qquad K=k_0,\\
 h_F&=(\|D_e\|_2+t)k_2+\|A\|_2H_N+2k_1
 \end{split}
 \label{eq:explicit-event-LKH}
\end{equation}
bound $\|D_xF_e\|$, $\|D_\Delta F_e\|$, and its joint second derivative.
For $n$ tokens with rowwise normalization and a fixed support per row, use
Frobenius state norms, maximize these per-expert constants over the selected
bank, and replace $K$ by $\sqrt n\max_e k_{0,e}$.  No extra factor $n$
is needed in $L$ or $h_F$, even when the down-matrix perturbations are shared
across tokens, provided their stacked Frobenius norm is at most $t$.
\end{proposition}
\begin{proof}
Write $z=\Pi x$ and $q=\|z\|^2/d_x+\epsilon_N$.  Apart from the fixed diagonal
scale, the derivative of $z/\sqrt q$ is
$q^{-1/2}\Pi-q^{-3/2}zz^\top/d_x$, whose norm is at most
$\epsilon_N^{-1/2}$.  Its second derivative in directions $u,w$ is
\[
 -\frac{\Pi u\langle z,\Pi w\rangle+\Pi w\langle z,\Pi u\rangle
           +z\langle\Pi u,\Pi w\rangle}{d_xq^{3/2}}
 +\frac{3z\langle z,\Pi u\rangle\langle z,\Pi w\rangle}{d_x^2q^{5/2}}.
\]
Since $\|z\|\le\sqrt{d_xq}$, its norm for unit directions is at most
$6/(\sqrt{d_x}\epsilon_N)$.  This also proves the stated bound on $N$.
If $p(t)=(1+e^{-t})^{-1}$, then
$\operatorname{SiLU}'=p+tp(1-p)$ and
$\operatorname{SiLU}''=2p(1-p)+tp(1-p)(1-2p)$.
The inequalities $p(1-p)\le1/4$ and $|t|\le T$ give $s_1,s_2$.
Twice applying the product rule to $h$ gives \cref{eq:gated-feature-constants}.
For a softmax coordinate, the logit gradient is $p_e(e_e-p)$ and its Hessian
is $p_e[(e_e-p)(e_e-p)^\top-(\operatorname{diag}(p)-pp^\top)]$.
Their norms are bounded by two and six, respectively.  Composition with
$RN$ and the product rule prove \cref{eq:gain-feature-constants}.
Finally $D_\Delta^2F_e=0$, each state--parameter cross derivative has norm at
most $k_1$, and the state Hessian is bounded by
$(\|D_e\|+t)k_2+\|A\|H_N$.  Sum these three terms on the joint product space.
For rowwise maps, the sum of products of squared row-direction norms is at
most the product of their sums.  The same inequality applies when a repeated
parameter direction is selected from the stacked perturbation.  The parameter
first derivative has at most $n$ rows of norm $k_0\|\Delta\|_F$, giving only
the stated $\sqrt n$ factor in $K$.
\end{proof}

\paragraph{Closing the surrounding fixed-prefix attention computation.}
We complete the fixed-prefix Transformer analysis with an explicit, conservative bound for attention in the preparation architecture.  Fix prefix length $n$, head width $d_h$, all
weights, and a fixed causal mask with at least one allowed key per row.  Softmax
is taken only over allowed logits.  With normalized input $Z$ satisfying
$\|Z\|_F\le Z_0=\sqrt n R_N$, one attention head is
\[
 A_h(Z)=P_h(Z)\,ZW_{v,h}^\top,\qquad
 P_h(Z)=\operatorname{softmax}_{\rm rows}
  ((ZW_{q,h}^\top)(ZW_{k,h}^\top)^\top/\sqrt{d_h}).
\]
Set $q_h=\|W_{q,h}\|_2$, $k_h=\|W_{k,h}\|_2$,
$v_h=\|W_{v,h}\|_2$, and
\begin{equation}
 \begin{aligned}
 s_{1,h}&=2q_hk_hZ_0/\sqrt{d_h},&
 s_{2,h}&=2q_hk_h/\sqrt{d_h},\\
 p_{1,h}&=2s_{1,h},&p_{2,h}&=6s_{1,h}^2+2s_{2,h},\\
 a_{1,h}&=p_{1,h}v_hZ_0+\sqrt n\,v_h,&
 a_{2,h}&=p_{2,h}v_hZ_0+2p_{1,h}v_h.
 \end{aligned}
 \label{eq:attention-derivative-constants}
\end{equation}
For the concatenated heads followed by output matrix $W_o$, take
$a_i=\|W_o\|_2(\sum_h a_{i,h}^2)^{1/2}$, $i=1,2$.
An attention-plus-residual event $X\mapsto X+A(N(X))$ therefore has
\begin{equation}
 L_{\rm att}=1+a_1L_N,\qquad K_{\rm att}=0,\qquad
 h_{\rm att}=a_2L_N^2+a_1H_N.
 \label{eq:attention-LKH}
\end{equation}
For completeness, these bounds follow by differentiating the bilinear score
map: its first and second derivatives have norms at most $s_{1,h},s_{2,h}$.
The row-softmax first and second derivative norms are at most two and six.
For the latter, its bilinear form is
$p\odot[(u-\E_pu)(w-\E_pw)-\operatorname{Cov}_p(u,w)]$;
its Euclidean norm is at most its $\ell_1$ norm, which is at most
$5\|u\|_2\|w\|_2\le6\|u\|_2\|w\|_2$.
Combine the rows by the squared-row inequality used above, and use
$\|P_h\|_2\le\|P_h\|_F\le\sqrt n$.
Two product rules give $a_{1,h},a_{2,h}$; concatenation and composition with
LayerNorm give \cref{eq:attention-LKH}.
The terminal LayerNorm followed by a fixed linear head similarly has
$L=\|W_{\rm head}\|_2L_N$, $K=0$, $h=\|W_{\rm head}\|_2H_N$.
Alternating these events with \cref{eq:explicit-event-LKH} covers a full
\emph{fixed-prefix forward pass} with the specified down-weight perturbations.

\paragraph{A noncircular route and region check.}
Choose a perturbation radius $t$ first, compute the uniform constants above
and $c_j$ by \cref{eq:remainder-envelope-recursion}, and evaluate reference
router margins before examining perturbed executions.  At each FFN event,
let $\gamma_j$ be the minimum reference top-one logit gap over its tokens.
The sufficient tests
\begin{equation}
 \|d\|\le t,\qquad
 \gamma_j>2\|R_j\|_2L_{N,j}c_jt\quad\hbox{for every FFN event}
 \label{eq:explicit-branch-tube}
\end{equation}
ensure route agreement by induction and \cref{lem:topk-margin-stability}.
The derivative bounds are global in the state, so no additional state-region
escape occurs in this particular normalized architecture.  The test controls
support only; the varying gain has already been included in $k_1,k_2$.
If the tests hold and $d$ is bounded by $t$ almost surely, set
$p_{\rm exit}=0$.  For an unbounded perturbation law, use at least
$\Prb(\|d\|>t)$ in the exit accounting rather than discard it.
For a fixed input with all strict margins, a positive sufficiently small $t$
exists because the finitely many constants and recursions are continuous and
finite at $t=0$.  The admissible radius can depend on the input.
Small $\epsilon_N$, large weights, long prefixes, or depth can make these
conservative constants uninformative.

\paragraph{An explicit tail envelope for this specialization.}
There is also a global, route-independent growth bound for the down-only
perturbation model at a fixed prefix.  A normalized attention residual is
uniformly bounded by its fixed weights and $Z_0$; a routed FFN residual is
bounded by $\sqrt n\max_e[(\|D_e\|_2+\|d\|)H_{0,e}+\|A\|_2R_N]$.
Summing residual magnitudes through finitely many blocks gives
$\|\widehat X-X\|_F\le C_0+C_1\|d\|$ with explicit nonnegative constants
obtained from these displayed sums for the perturbed and reference passes.
If the endpoint is the final normalized logits, the sharper bound
$\|e\|_F\le2\|W_{\rm head}\|_2\sqrt n R_{N,\rm final}$ is available.
Consequently a finite $q$-th parameter moment supplies a finite $T_q$ in
\cref{eq:remainder-with-exit}; for Gaussian perturbations one may use $q=4$
and \cref{eq:gaussian-fourth-norm}.  The actual weights, margins, covariance,
and target prefix distribution are still needed to obtain a numerical
certificate for a trained model.    \Cref{thm:support-replay-bridge,prop:frozen-gain-replay} distinguish current-gain support replay from the preparation code's frozen-gain intervention.

\subsection{Conditional moments under endogenous routing}
\label{sec:endogenous-conditional-moments}

Let $\xi\in\R^p$ be a square-integrable error vector containing all reused
parameter or local-injection errors.  Let $\mathscr R$ be a sigma-field that
makes the actual coefficient matrix $G$ and offset $b$ measurable; both may be
selected using $\xi$.  It may have to contain more than the expert indices:
within-support gains, states, and Jacobians can still vary at a fixed index.
Assume $b+G\xi$ and the conditional-moment expressions below are integrable
at the indicated squared orders.  Finite route families with bounded $G,b$
and $\xi\in L^2$ are sufficient.

\begin{theorem}[Exact risk with route-selected conditional moments]
\label{thm:endogenous-conditional-risk}
For $\mu_{\mathscr R}=\E[\xi\mid\mathscr R]$ and
$\Sigma_{\mathscr R}=\operatorname{Cov}(\xi\mid\mathscr R)$, an exactly
route-affine error $e=b+G\xi$ obeys
\begin{equation}
 \E\|e\|_2^2
 =\E\left[\|b+G\mu_{\mathscr R}\|_2^2
                +\tr(G\Sigma_{\mathscr R}G^\top)\right].
 \label{eq:endogenous-conditional-risk}
\end{equation}
For a finite route variable $T$, with constant $G_t,b_t$ on each positive-mass
cell, this is the sum of these conditional risks weighted by $p_t=\Prb(T=t)$.
Neither the conditional means nor the conditional cross-covariances may be
replaced by their unconditional values without an additional argument.

A quadratic sharing frontier remains possible under \emph{one fixed endogenous
route map}.  Specifically, let $Q$ and $\theta$ be deterministic, and fix the map $T=T(\xi)$
and cell matrices $G_t$ independently of $\lambda$.  Suppose all compared
errors are exactly
$e_\lambda=G_T[\xi-\lambda Q(\theta+\xi)]$.
Then
\begin{equation}
 R(\lambda)=R(0)-2a\lambda+b_2\lambda^2,\quad
 a=\E\langle G_T\xi,G_TQ(\theta+\xi)\rangle,\quad
 b_2=\E\|G_TQ(\theta+\xi)\|^2\ge0.
 \label{eq:endogenous-fixed-partition-frontier}
\end{equation}
With $W_t=G_t^\top G_t$, $\mu_t=\E[\xi\mid T=t]$,
$\Sigma_t=\operatorname{Cov}(\xi\mid T=t)$, the coefficients equal
\begin{equation}
 \begin{split}
 a&=\sum_t p_t\{\mu_t^\top W_tQ(\theta+\mu_t)+\tr(W_tQ\Sigma_t)\},\\
 b_2&=\sum_t p_t\{\|Q(\theta+\mu_t)\|_{W_t}^2
                          +\tr(W_tQ\Sigma_tQ^\top)\}.
 \end{split}
 \label{eq:endogenous-frontier-coefficients}
\end{equation}
Full sharing improves this risk exactly when $2a>b_2$; the minimizer on
$[0,1]$ is $\min\{1,\max\{0,a/b_2\}\}$ if $b_2>0$.
If $b_2=0$, then $a=0$ and all coefficients have equal risk.
\end{theorem}
\begin{proof}
Write $\xi=\mu_{\mathscr R}+\eta$ with
$\E[\eta\mid\mathscr R]=0$.  Conditional on $\mathscr R$, expand the squared
norm; the cross term vanishes and the remaining quadratic noise term is
$\tr(G\Sigma_{\mathscr R}G^\top)$.  Take expectations.
For the sharing statement set $X=G_T\xi$ and $Y=G_TQ(\theta+\xi)$.
Expand $\E\|X-\lambda Y\|^2$ and then condition each coefficient on $T$;
this gives \cref{eq:endogenous-frontier-coefficients}.
Minimize the resulting quadratic.  If $\E\|Y\|^2=0$, then $Y=0$ almost surely
and $\E\langle X,Y\rangle=0$.
\end{proof}

This is a conditional second-moment identity.  It also applies to fixed linear propagations within a route cell after
stacking all local errors, so it retains cross-time dependence.  If the cell map
or coefficients change with $\lambda$, every candidate must use its own
conditional law; \cref{eq:endogenous-fixed-partition-frontier} is no longer
one common quadratic.  For a nonlinear route cell the identity applies to its
linear part only, with the remainder treated by
\cref{thm:joint-taylor-remainder,prop:terminal-remainder-interval}.

\paragraph{General fixed Gaussian route cells via their probabilities.}
There is a useful identity beyond one half-space.  Let $\Sigma\succ0$ be fixed,
let $\mathcal C_t$ be a fixed Borel selection region, and write
$p_t(\mu)=\Pr_{\xi\sim\mathcal N(\mu,\Sigma)}(\xi\in\mathcal C_t)>0$.
When differentiating, hold the region and covariance fixed.  Then
\begin{equation}
 \mu_t=\mu+\Sigma\nabla_\mu\log p_t(\mu),\qquad
 \Sigma_t=\Sigma+\Sigma\nabla_\mu^2\log p_t(\mu)\Sigma.
 \label{eq:gaussian-cell-score-moments}
\end{equation}
To prove this, differentiate the Gaussian density inside its integral on
$\mathcal C_t$.  The first density derivative is
$\Sigma^{-1}(\xi-\mu)$ times the density, and the second is
$[\Sigma^{-1}(\xi-\mu)(\xi-\mu)^\top\Sigma^{-1}-\Sigma^{-1}]$ times that
density.  Divide the integrated expressions by $p_t$ and subtract the outer
product of the conditional mean displacement.  Differentiation is justified
by locally integrable polynomial-times-Gaussian envelopes even if the region's
boundary is not smooth.  For singular covariance, apply the identity in
whitened coordinates on its fixed affine support, varying the mean only within
that support; ambient mean derivatives need not exist across moving supports.
This handles fixed intersections or nonlinear regions analytically in terms of
$p_t$.  Regions depending on the differentiated mean require boundary terms;
regions or propagation coefficients changing across sharing candidates require
separate calculations.  

\subsection{A complete two-expert Gaussian threshold calculation}
\label{sec:gaussian-threshold-route}

\begin{proposition}[Half-space route probabilities, means, and covariances]
\label{prop:gaussian-threshold-moments}
Let $\xi\sim\mathcal N(\mu,\Sigma)$ in $\R^p$, allowing singular $\Sigma$,
and select route $+$ when $a^\top\xi>\tau$, route $-$ otherwise.
Assume first that $s^2=a^\top\Sigma a>0$, and put
\[
 \begin{gathered}
 \kappa=(\tau-a^\top\mu)/s,\quad h=\Sigma a/s,\\
 p_+=1-\Phi(\kappa),\quad p_-=\Phi(\kappa),\quad
 m_+=\phi(\kappa)/p_+,\quad m_-=\phi(\kappa)/p_-.
 \end{gathered}
\]
The functions $\phi,\Phi$ are the standard normal density and CDF.
For finite $\tau$, both probabilities are positive and
\begin{equation}
 \begin{aligned}
 \mu_+&=\mu+hm_+,&
 \Sigma_+&=\Sigma+hh^\top(\kappa m_+-m_+^2),\\
 \mu_-&=\mu-hm_-,&
 \Sigma_-&=\Sigma-hh^\top(\kappa m_-+m_-^2).
 \end{aligned}
 \label{eq:gaussian-threshold-conditional}
\end{equation}
For any fixed offsets $b_\pm$ and matrices $G_\pm$, these moments and
\cref{eq:endogenous-conditional-risk} give the exact risk of
$e=b_T+G_T\xi$.
If $s=0$, the score is constant almost surely: use its single realized branch
(with the stated tie convention), and the original $\mu,\Sigma$ on that branch;
there is no division by $s$ or conditioning on a zero-mass cell.
\end{proposition}
\begin{proof}
Let $Z=(a^\top\xi-a^\top\mu)/s$ and $\eta=\xi-\mu-hZ$.
Then $Z\sim\mathcal N(0,1)$ and $\eta$ is independent of $Z$, including in the
singular Gaussian case, because their cross-covariance is zero.
Moreover $\operatorname{Cov}(\eta)=\Sigma-hh^\top\succeq0$.
Integration by parts using $\phi'(z)=-z\phi(z)$ gives
\[
 \int_\kappa^\infty z\phi(z)\,dz=\phi(\kappa),\qquad
 \int_\kappa^\infty z^2\phi(z)\,dz=p_++\kappa\phi(\kappa).
\]
Subtract from the unconditional first and second moments to obtain the lower
half-space formulas.  Substitute in $\xi=\mu+hZ+\eta$ and subtract the
conditional means' outer products to obtain \cref{eq:gaussian-threshold-conditional}.
The variance-zero score is deterministic and is handled directly.
\end{proof}

\paragraph{An equivalent form avoiding division by small route probabilities.}
The unnormalized moments of the upper cell are
\begin{equation}
 \begin{split}
 u_+&=\E[\xi\mathbf1_{T=+}]=p_+\mu+h\phi(\kappa),\\
 V_+&=\E[\xi\xi^\top\mathbf1_{T=+}]\\
    &=p_+(\Sigma+\mu\mu^\top)
       +\phi(\kappa)(\mu h^\top+h\mu^\top+\kappa hh^\top).
 \end{split}
 \label{eq:threshold-unnormalized-moments}
\end{equation}
The lower cell has $u_-=p_-\mu-h\phi(\kappa)$ and the corresponding negative
sign in the correction to $p_-(\Sigma+\mu\mu^\top)$.
The exact risk is therefore
\begin{equation}
 \sum_{t\in\{+,-\}}
 \left[p_t\|b_t\|^2+2b_t^\top G_tu_t+\tr(G_tV_tG_t^\top)\right].
 \label{eq:threshold-raw-risk}
\end{equation}
These formulas follow from the same two integrals; they avoid inverse small
probabilities, although extreme-tail cancellation still requires numerically
stable CDF/survival functions and appropriate precision.  They specialize classical truncated-normal moment calculations \citep{manjunath2021truncated}.

\paragraph{Why the mean and cross-covariance both matter.}
For $\xi\sim\mathcal N(0,1)$, $T=\operatorname{sign}(\xi)$, and $G_T=T$,
the error is $e=|\xi|$.  Its mean is $\sqrt{2/\pi}$, variance is $1-2/\pi$,
and squared risk is one.  Keeping only the conditional variance loses the
conditional-mean contribution $2/\pi$.
For an independent standard bivariate normal $(\xi_1,\xi_2)$, select $+$ when
$\xi_1+\xi_2>\sqrt2$.  In the notation above, $\kappa=1$ and
$h=(1,1)^\top/\sqrt2$, so the off-diagonal conditional covariance is
$(m_+-m_+^2)/2<0$.  It is zero before selection.  Thus even independently fitted
errors need not be independent or uncorrelated after routing.

\paragraph{Recomputing the route can destroy the quadratic frontier.}
Let $\xi\sim\mathcal N(0,1)$, $a_\lambda=1-\lambda$, and use two branches
whose error is $e_\lambda=a_\lambda\xi$ when
$|a_\lambda\xi|>\tau>0$ and zero otherwise.
For $0\le\lambda<1$,
\begin{equation}
 R(\lambda)=2a_\lambda^2
 \left[\frac{\tau}{a_\lambda}\phi\!\left(\frac{\tau}{a_\lambda}\right)
       +1-\Phi\!\left(\frac{\tau}{a_\lambda}\right)\right],\qquad R(1)=0.
 \label{eq:changing-route-nonquadratic}
\end{equation}
The same integral as above proves this formula.  It is not a quadratic in
$\lambda$: as $\lambda\uparrow1$ it is smaller than every positive power of
$1-\lambda$, yet it is positive at zero.  A nonzero quadratic cannot have that
property.  This is a controlled example of route-dependent propagation.

\subsection{Which functional corrections can a low-rank parameter update realize?}
\label{sec:adapter-realizability}

LoRA represents a change to a specified frozen linear weight by a matrix
product $BA$ \citep{hu2021lora}.  Its \emph{matrix rank} is at most the factor
width.  This differs from the dimension of the span of an arbitrary collection
of functions in \cref{prop:rank-r-functional}.  To make the comparison precise,
fix a feature map $\varphi(x)\in\R^k$, its input law, and its uncentered second
moment $C=\E[\varphi(x)\varphi(x)^\top]\succeq0$.  Let $W_0\in\R^{d\times k}$
be the fixed core and $W_a^\star$ a role's linear target on these features.
Write $K_a=W_a^\star-W_0$.  Only updates to this output weight are allowed;
changing the feature map changes the problem.

\begin{theorem}[Exact adapter approximation bias in feature and output geometry]
\label{thm:feature-weighted-low-rank}
Let $S\succeq0$ be a fixed output-loss matrix and let $r$ be an integer with
$0\le r\le\min(d,k)$.
Define $T_a=S^{1/2}K_aC^{1/2}$.  Then
\begin{equation}
 \inf_{\operatorname{rank}(\Delta_a)\le r}
  \E\|(K_a-\Delta_a)\varphi(x)\|_S^2
   =\sum_{i>r}\sigma_i(T_a)^2.
 \label{eq:weighted-adapter-tail}
\end{equation}
Here $\sigma_i$ are decreasing singular values, $S^{1/2},C^{1/2}$ are the
symmetric positive-semidefinite square roots, and $\|y\|_S^2=y^\top Sy$.
Both $S$ and $C$ may be singular.  If $T_{a,r}$ is a truncated singular-value
decomposition, a realizing update is
\begin{equation}
 \Delta_a=(S^{1/2})^\dagger T_{a,r}(C^{1/2})^\dagger,
 \label{eq:adapter-pseudoinverse-lift}
\end{equation}
where $\dagger$ denotes the Moore--Penrose inverse (invert the nonzero singular
values and leave zero ones zero).  The optimal risk is zero exactly when
$\operatorname{rank}(T_a)\le r$.  For independently chosen role adapters,
the weighted average optimum is the sum of \cref{eq:weighted-adapter-tail}
with the declared role weights; this is an approximation result, not an
estimation or optimization guarantee.
\end{theorem}
\begin{proof}
The objective equals $\|T_a-S^{1/2}\Delta_aC^{1/2}\|_F^2$.
For any matrix $Z$ of rank at most $r$, let $P_Z$ project onto its column space.
Then $\|T_a-Z\|_F^2\ge\|(I-P_Z)T_a\|_F^2$.
Diagonalizing $T_aT_a^\top$ shows
$\tr(P_ZT_aT_a^\top)\le\sum_{i\le r}\sigma_i(T_a)^2$:
in an eigenbasis the diagonal weights of a rank-at-most-$r$ orthogonal
projector lie in $[0,1]$ and sum to at most $r$.
This proves the lower bound.  The left and right singular directions of
$T_a$ with nonzero singular values lie in the ranges of $S^{1/2}$ and
$C^{1/2}$, respectively.  Therefore the pseudoinverse lift has rank at most
$r$ and satisfies $S^{1/2}\Delta_aC^{1/2}=T_{a,r}$, attaining the bound.
Every such rank-at-most-$r$ update can be factored as $BA$ by its singular-value
decomposition, padding the factors with zeros when necessary.
\end{proof}

The covariance must be a \emph{feature second moment}, not its centered
covariance when features have nonzero mean.  Null directions of $C$ are
unobserved under this law; they need not remain null after distribution shift.
Likewise $S$ null directions are invisible to this loss, not necessarily to
another output metric or a downstream computation.

The infimum has no update-norm or factor-norm constraint.  A pseudoinverse lift
can have large norm on poorly excited feature directions.  A bounded-radius
adapter problem, including a radius needed for a downstream Taylor bound,
therefore need not attain this unconstrained tail value.  \Cref{thm:rank-radius-frontier,prop:total-radius-allocation} give the radius-constrained specializations and count each distinct updated matrix once.

\paragraph{Targets outside the linear feature span.}
For an arbitrary square-integrable desired correction $d_\star(x)$, let
$B=\E[d_\star\varphi^\top]$, $K=BC^\dagger$, and
$u=d_\star-K\varphi$.  A vector in the null space of $C$ is orthogonal to
$\varphi$ almost surely, so $B=BC^\dagger C$ and $\E[u\varphi^\top]=0$.
Consequently, for any fixed $S\succeq0$,
\begin{equation}
 \inf_{\operatorname{rank}(\Delta)\le r}\E\|d_\star-\Delta\varphi\|_S^2
 =\E\|u\|_S^2+\sum_{i>r}\sigma_i(S^{1/2}KC^{1/2})^2.
 \label{eq:misspecified-feature-adapter-tail}
\end{equation}
This follows by expanding the square, eliminating the cross term using the
orthogonality just proved, and applying \cref{thm:feature-weighted-low-rank}.
The feature-misspecification residual is retained for targets outside the feature span.

\paragraph{A shared factor is a different architecture.}
With Euclidean output loss, role-specific feature second moments $C_a$, and
one common column factor $U\in\R^{d\times r}$, $U^\top U=I_r$, write
$\Delta_a=UZ_a$.  For fixed $U$ the best value is
$\sum_a\pi_a\|(I-UU^\top)K_aC_a^{1/2}\|_F^2$, attained by $Z_a=U^\top K_a$.
Optimizing $U$ gives
\begin{equation}
 \inf_{U,Z_a}\sum_a\pi_a\E\|(K_a-UZ_a)\varphi_a\|^2
 =\sum_{i>r}\lambda_i\!\left(\sum_a\pi_a K_aC_aK_a^\top\right).
 \label{eq:shared-column-adapter-tail}
\end{equation}
This follows by the same trace maximization as in the proof above.
These eigenvectors are output directions; they are not the function-space
eigenvectors in \cref{prop:rank-r-functional}, and the constraint is stronger
than separate rank-$r$ updates.  For example, two targets $\operatorname{diag}(1,0)$
and $\operatorname{diag}(0,1)$ under $C=I_2$ and zero core admit separate rank-one
updates, but a common one-dimensional output factor has average residual $1/2$.

\paragraph{Two concrete insertion points.}
For the preparation code's \texttt{tied\_depth} term, take
$\varphi(x)=N(x)$ and $\Delta=\texttt{depth\_up}\,\texttt{depth\_down}$.
It is a parallel residual correction, not a modification of an existing expert
weight.  For a LoRA update to the down weight of a frozen gated expert, take
$\varphi(x)=\operatorname{SiLU}(VN(x))\odot UN(x)$, multiplied by its fixed
reference gain when that gain is included in the output loss.
A state-dependent but frozen gain function can be absorbed into $\varphi$;
a route-index indicator can likewise be included if the evaluation averages
calls where that expert is not selected.  The surrounding input map, gain
function, and selection law must remain fixed across the compared adapters.
These are exact \emph{local} feature maps.  An adapter may change the inputs or
routes of later blocks, for which the preceding Taylor or conditional-route
analysis is still required.

\begin{proposition}[Two roles with a freely learned common core]
\label{prop:free-core-two-role}
For two targets $W_1^\star,W_2^\star$, a common feature second moment $C$, and
weights $\pi_1,\pi_2>0$ with $\pi_1+\pi_2=1$, and integer $r\ge0$,
suppose the common core $W_0$
is an arbitrary matrix rather than fixed.  Under Euclidean output loss,
\begin{equation}
 \inf_{W_0,\,\operatorname{rank}(\Delta_i)\le r}
  \sum_{i=1}^2\pi_i\|(W_i^\star-W_0-\Delta_i)C^{1/2}\|_F^2
 =\pi_1\pi_2\sum_{j>2r}
      \sigma_j((W_1^\star-W_2^\star)C^{1/2})^2.
 \label{eq:free-core-two-role-tail}
\end{equation}
In particular, exact realization on the feature law is possible exactly when
the displayed difference has rank at most $2r$.
\end{proposition}
\begin{proof}
For fixed updates, completing the square over the free core gives
$\pi_1\pi_2\|(W_1^\star-W_2^\star-\Delta_1+\Delta_2)C^{1/2}\|_F^2$.
The difference of two rank-at-most-$r$ matrices has rank at most $2r$, yielding
the lower bound by \cref{thm:feature-weighted-low-rank}.
Conversely, split a best rank-at-most-$2r$ truncation $Z$ of the whitened target
difference into two groups of at most $r$ singular terms, $Z=Z_1+Z_2$.
Take $\Delta_1=Z_1(C^{1/2})^\dagger$ and
$\Delta_2=-Z_2(C^{1/2})^\dagger$.  Their whitened difference is exactly $Z$;
choose $W_0=\pi_1(W_1^\star-\Delta_1)+\pi_2(W_2^\star-\Delta_2)$.
This attains the lower bound, including singular $C$ and $2r$ exceeding the
matrix dimensions.
\end{proof}

For the earlier targets $\pm I_2$ with $C=I_2$, the fixed zero core leaves
residual one for each rank-one adapter.  A free core
$W_0=\operatorname{diag}(1,-1)$ instead allows
$\Delta_+=\operatorname{diag}(0,2)$ and
$\Delta_-=\operatorname{diag}(-2,0)$, both rank one, and zero residual.
Thus the fixed-core counterexample cannot be used as a lower bound for jointly
learned cores.  The two-role formula does not extend to an arbitrary number of
roles by checking pairwise ranks: simultaneous realizability adds compatibility
constraints.  For example, take three targets $0,I_2,2I_2$, $C=I_2$, and
$r=1$.  Every pairwise difference has rank at most $2r$.  Exact simultaneous
realization would require $\det W_0=0$ and $\det(I_2-W_0)=0$, hence
$\operatorname{tr}W_0=1$.  But then
$\det(2I_2-W_0)=4-2\operatorname{tr}W_0+\det W_0=2$, a contradiction.
This is an obstruction to exact realization, not a computed positive lower
bound on the three-role infimum.  

\subsection{Initialization-dependent low-rank tangent images}
\label{sec:factor-tangent}

\begin{proposition}[Initialization-dependent tangent image and its missing term]
\label{prop:factor-tangent-image}
For the feature map above with Euclidean output loss, parameterize an update
by $BA$.  At factors $(B_0,A_0)$ its exact change is
\begin{equation}
 (B_0+\dot B)(A_0+\dot A)-B_0A_0
     =B_0\dot A+\dot B A_0+\dot B\dot A.
 \label{eq:factor-exact-increment}
\end{equation}
Thus the first-order parameter-to-function image is
$\{(B_0\dot A+\dot B A_0)\varphi\}$, not all rank-at-most-$r$ corrections.
For $B_0=0$, put $V_0=A_0C^{1/2}$ and let $P_{V_0}$ be the orthogonal
projector onto its row space.  If $K\varphi$ is the desired functional change,
then
\begin{equation}
 \inf_{\dot B}\E\|(K-\dot B A_0)\varphi\|^2
    =\|KC^{1/2}(I-P_{V_0})\|_F^2,
 \label{eq:tangent-image-residual}
\end{equation}
attained by $\dot B=KC^{1/2}V_0^\dagger$.
The omitted term in \cref{eq:factor-exact-increment} obeys
\begin{equation}
 \|\dot B\dot A\varphi\|_{L^2}
 \le \|C^{1/2}\|_2\|\dot B\|_F\|\dot A\|_F
 \le\tfrac12\|C^{1/2}\|_2(\|\dot B\|_F^2+\|\dot A\|_F^2).
 \label{eq:factor-quadratic-remainder}
\end{equation}
\end{proposition}
\begin{proof}
Multiply the two factors and group linear and quadratic terms.  When $B_0=0$,
the matrix approximation problem is $\inf_{\dot B}\|KC^{1/2}-\dot B V_0\|_F^2$.
Each row of $\dot B V_0$ lies in the row space of $V_0$, and every row projection
is attained by the displayed pseudoinverse, proving the equality.
Finally use submultiplicativity of Frobenius and operator norms on
$\dot B\dot A C^{1/2}$, followed by $2ab\le a^2+b^2$.
\end{proof}

At the preparation code's zero \texttt{depth\_up} initialization, the initial
functional tangent therefore only uses the row space of the initialized
\texttt{depth\_down} after feature whitening.  It is generally narrower than
the class obtained by training both factors.  If both factors are zero, the
first derivative is zero but nonzero finite updates remain possible through
$\dot B\dot A$.  More concretely, with $r=1$, $C=I_2$,
$A_0=(1,0)$, $B_0=0$, and target $K=\operatorname{diag}(0,1)$, the tangent
residual is one while a finite rank-one update using row factor $(0,1)$ is
exact.  A tangent obstruction is not a global impossibility theorem.

\paragraph{A rigorous nonlinear parameter-to-function fallback.}
For a general insertion point, fix a Hilbert space of functions on a fixed
probe law and suppose $\mathcal F(W)$ is twice differentiable on the parameter
ball $\|\Delta\|_F\le t$, with a uniform second-derivative norm at most $H$.
Let $J=D\mathcal F(W_0)$, let $d_\star$ be a target functional correction, and
let $\mathcal S_r(t)=\{\Delta:\operatorname{rank}(\Delta)\le r,\|\Delta\|_F\le t\}$.
Define the constrained image distance
\[
 b_r(t)=\inf_{\Delta\in\mathcal S_r(t)}\|d_\star-J\Delta\|_{\mathcal H},
 \qquad \eta=Ht^2/2.
\]
Uniform Taylor remainder and the triangle inequalities give
\begin{equation}
 (\max\{0,b_r(t)-\eta\})^2
 \le\inf_{\Delta\in\mathcal S_r(t)}
       \|d_\star-[\mathcal F(W_0+\Delta)-\mathcal F(W_0)]\|_{\mathcal H}^2
 \le(b_r(t)+\eta)^2.
 \label{eq:nonlinear-adapter-image-interval}
\end{equation}
Indeed, apply the pointwise norm inequalities for every feasible $\Delta$ and
then take infima; an approximating sequence suffices if an infimum is not
attained.  The image $J(\mathcal S_r(t))$ is generally not a linear subspace,
so projecting onto the full range of $J$ only supplies a relaxation, not a
realizability certificate.  Smoothness into $L^2$ requires a uniform
square-integrable derivative envelope on the declared input law; pointwise
differentiability alone does not supply it.  For hard routing one must first
certify a common smooth branch or separately pay for route changes.

\paragraph{Applying the conditional bounds.}
  A numerical nonlinear bound still needs a specified checkpoint,
reference input population, perturbed weights, perturbation law, and meaningful
margin/tail bounds.  A multistep endogenous route generally conditions on an
intersection of nonlinear decisions, rather than the single half-space solved
here; selected indices alone do not fix gains or Jacobians.  The radius extension in \cref{app:rank-radius-replay} fixes the local feature law and core, constrains the actual output-matrix update, and separates approximation from terminal performance.

\section{Rank-Constrained Updates and Precisely Defined Branch Replay}
\label{app:rank-radius-replay}

\paragraph{Overview.}
We address two limits of a local approximation argument: the update must fit a budget on actual parameter changes, and fixed-support execution must be distinguished from natural routing. The guarantees separately track matrix rank, total radius, support disagreement, and the extra cost of replaying recorded gains, for fixed features/cores and finite fixed-prefix computations. The matrix arguments extend classical low-rank least-squares geometry \citep{eckart1936approximation}; the routing arguments are pathwise couplings, not assumptions of route independence. 
Sections~\ref*{app:rank-radius-replay}.1--\ref*{app:rank-radius-replay}.2 handle one-matrix and total update budgets; Section~\ref*{app:rank-radius-replay}.3 compares natural execution with current-gain support replay, and Section~\ref*{app:rank-radius-replay}.4 accounts for freezing the gains as well.

\subsection{One matrix: an exact isotropic radius frontier and semidefinite bounds}
\label{sec:rank-radius}

Fix a core $W_0\in\R^{d\times k}$ and target $W^\star$, and set
$K=W^\star-W_0$.  The feature map $\varphi$, its probability law, and
$C=\E[\varphi\varphi^\top]\succeq0$ remain fixed as the update changes.
For output geometry $S\succeq0$, integer $0\le r\le\min(d,k)$ and finite
$t\ge0$, define
\begin{equation}
 \mathcal E_{r,t}(K;S,C)
 =\min_{\substack{\operatorname{rank}(\Delta)\le r\\\|\Delta\|_F\le t}}
       \|S^{1/2}(K-\Delta)C^{1/2}\|_F^2.
 \label{eq:rank-radius-objective}
\end{equation}
This equals $\min\E\|(K-\Delta)\varphi\|_S^2$.
The constraint is on the actual matrix update, not on a factor's norm or on a
training gradient.  The feasible set is nonempty and compact: rank at most $r$
is closed (all $(r+1)$-minors vanish), and the Frobenius ball is closed and
bounded.  Hence the continuous objective attains its minimum, including for
singular $S,C$.  Write $(x)_+=\max\{x,0\}$ and let $T_r$ denote a truncated
SVD retaining at most $r$ singular terms of $T$; $T_0=0$.

\begin{theorem}[Rank and matrix-radius approximation]
\label{thm:rank-radius-frontier}
In isotropic geometry, put
$q_r=\|K_r\|_F=(\sum_{i\le r}\sigma_i(K)^2)^{1/2}$.
Then
\begin{equation}
 \mathcal E_{r,t}(K;I,I)
 =\sum_{i>r}\sigma_i(K)^2+(q_r-t)_+^2.
 \label{eq:isotropic-rank-radius}
\end{equation}
For $q_r>0$, an attaining update is
$\Delta^\star=\min\{1,t/q_r\}K_r$; for $q_r=0$, use $\Delta^\star=0$.

For arbitrary positive-semidefinite $S,C$, set
\begin{equation}
 \begin{gathered}
 T=S^{1/2}KC^{1/2},\quad q_r=\|T_r\|_F,\quad
 E_r=\sum_{i>r}\sigma_i(T)^2,\quad
 \kappa=\|S^{1/2}\|_2\|C^{1/2}\|_2,\\
 \Delta_0=(S^{1/2})^\dagger T_r(C^{1/2})^\dagger,\qquad
 \alpha=\begin{cases}
 \min\{1,t/\|\Delta_0\|_F\},&\Delta_0\ne0,\\
 1,&\Delta_0=0.
 \end{cases}
 \end{gathered}
 \label{eq:psd-radius-definitions}
\end{equation}
Then
\begin{equation}
 E_r+(q_r-\kappa t)_+^2
 \ \le\ \mathcal E_{r,t}(K;S,C)
 \ \le\ E_r+(1-\alpha)^2q_r^2.
 \label{eq:psd-rank-radius-bounds}
\end{equation}
The right-hand value is attained by the feasible candidate
$\alpha\Delta_0$, but need not be the minimum.  If
$t\ge\|\Delta_0\|_F$, the two bounds equal $E_r$.
These conventions cover $r=0$, $t=0$, a zero target, null feature/output
geometries, and tied singular values without dividing by zero.
\end{theorem}
\begin{proof}
For any feasible $\Delta$ let $P$ project onto its column space and put
$u=\|\Delta\|_F$.  Since $P\Delta=\Delta$,
$\langle K,\Delta\rangle_F=\langle PK,\Delta\rangle_F\le u\|PK\|_F$.
In an eigenbasis of $KK^\top$, the diagonal entries of $P$ lie in $[0,1]$
and sum to at most $r$.  It follows that
$\|PK\|_F^2=\tr(PKK^\top)\le q_r^2$.
Consequently
\[
 \|K-\Delta\|_F^2\ge\|K\|_F^2+u^2-2uq_r
 =\sum_{i>r}\sigma_i(K)^2+(u-q_r)^2.
\]
Minimize over $0\le u\le t$.  A scaled $K_r$ has the optimal norm and attains
the inner-product bound; the zero case is immediate.

For the general lower bound, $Z=S^{1/2}\Delta C^{1/2}$ has rank at most $r$
and norm at most $\kappa t$.  Minimizing over all such $Z$ relaxes the actual
image constraint, so the isotropic formula applied to $T$ gives the left-hand
side of \cref{eq:psd-rank-radius-bounds}.  Every nonzero left/right singular
direction of $T$ lies in the range of $S^{1/2}/C^{1/2}$, respectively.
Thus $S^{1/2}\Delta_0C^{1/2}=T_r$ and
$\operatorname{rank}(\Delta_0)\le r$, also when these matrices are singular.
Scaling by $\alpha$ makes the update feasible.  Orthogonality of $T-T_r$ and
$T_r$ gives
$\|T-\alpha T_r\|_F^2=E_r+(1-\alpha)^2q_r^2$.
If $t\ge\|\Delta_0\|_F$, this equals $E_r$ and
$q_r=\|S^{1/2}\Delta_0 C^{1/2}\|_F\le\kappa t$, proving the final claim.
\end{proof}

\paragraph{A suboptimality example in semidefinite geometry.}
Take $S=I_2$, $C=\operatorname{diag}(10^{-2},1)$,
$K=\operatorname{diag}(20,1)$, and $r=t=1$.
Then $T=\operatorname{diag}(2,1)$ and
$\Delta_0=\operatorname{diag}(20,0)$.  The scaled candidate
$\alpha\Delta_0=\operatorname{diag}(1,0)$ has loss
$(2-0.1)^2+1=4.61$.  The feasible rank-one update
$\operatorname{diag}(0,1)$ has loss $4$.  Thus scaled transformed SVD can be
strictly suboptimal under a \emph{parameter}-norm budget.
Replacing the transformed feasible image by the rank-constrained
Frobenius ball of radius $\kappa t$ gives a relaxation.
Imposing the unscaled radius $t$ directly in transformed coordinates
is not in general a valid relaxation or a feasible parameter construction.  Poorly excited directions
can make the inverse lift expensive.  

For an arbitrary square-integrable desired correction, the fixed-feature
orthogonal decomposition in \cref{eq:misspecified-feature-adapter-tail} still
applies for every feasible $\Delta$.  Its feature-misspecification residual
must be added to \cref{eq:rank-radius-objective}.

\subsection{A total radius over distinct parameter matrices}
\label{sec:total-parameter-radius}

Let $\Delta_1,\ldots,\Delta_E$ be updates to \emph{distinct} physical parameter
matrices, and form $d$ by concatenating their vectorized entries once each.
Then
\begin{equation}
 \|d\|_2^2=\sum_{e=1}^E\|\Delta_e\|_F^2.
 \label{eq:unique-parameter-radius}
\end{equation}
An expert reused at several tokens or depths contributes one summand, not a
fresh matrix or independent random draw at each call.  Conversely, giving each
of $E$ distinct matrices radius $t$ permits total radius $\sqrt E\,t$, not $t$.

\begin{proposition}[A common isotropic radius for distinct update blocks]
\label{prop:total-radius-allocation}
Fix targets $K_e$, ranks $r_e$ within their matrix dimensions, and positive
weights $w_e$ summing to one.  Put
$q_e=\|(K_e)_{r_e}\|_F$, $a_e=\sum_{i>r_e}\sigma_i(K_e)^2$, and
$Q_0=(\sum_e q_e^2)^{1/2}$.  Then
\begin{equation}
 \min_{\substack{\operatorname{rank}(\Delta_e)\le r_e\ \forall e\\
                  \sum_e\|\Delta_e\|_F^2\le t^2}}
       \sum_e w_e\|K_e-\Delta_e\|_F^2
 =\sum_e w_e a_e+\sum_e w_e(q_e-u_e)^2,
 \label{eq:joint-radius-value}
\end{equation}
where $u_e=q_e$ if $t\ge Q_0$, and $u_e=0$ if $t=0$.
For $0<t<Q_0$, there is a unique $\eta>0$ satisfying
\begin{equation}
 u_e=\frac{w_e q_e}{w_e+\eta},\qquad \sum_e u_e^2=t^2.
 \label{eq:radius-allocation-multiplier}
\end{equation}
An optimizer is $\Delta_e=(u_e/q_e)(K_e)_{r_e}$ when $q_e>0$ and zero otherwise.
This is a deterministic approximation identity; it assumes no independence
between parameter errors and proves no optimizer's convergence.
\end{proposition}
\begin{proof}
The inner-product inequality in \cref{thm:rank-radius-frontier} reduces each
block to $a_e+(q_e-u_e)^2$ at prescribed norm $u_e$.  An optimum need not use
$u_e>q_e$: replacing it by $q_e$ reduces both radius and objective.
The problem is therefore the scalar quadratic minimization over
$0\le u_e\le q_e$, $\sum_eu_e^2\le t^2$, and aligned truncated SVDs attain
any such choice.  The inactive and zero-radius cases follow directly.
For $0<t<Q_0$, the squared norm of
$(w_e q_e/(w_e+\eta))_e$ is continuous and strictly decreasing from $Q_0^2$
to zero as $\eta$ increases from zero to infinity.
The resulting $u$ minimizes
$\sum_e w_e(q_e-u_e)^2+\eta\sum_eu_e^2$ coordinatewise.
For any feasible $v$, this minimization and
$\sum_eu_e^2=t^2\ge\sum_ev_e^2$ imply
$\sum_e w_e(q_e-u_e)^2\le\sum_e w_e(q_e-v_e)^2$.
This proves global optimality without an assumption about numerical search.
\end{proof}

\paragraph{Several roles sharing one actual update.}
Roles are not automatically distinct variables in \cref{eq:joint-radius-value}.
For fixed tying groups $\mathcal J_e$, common isotropic geometry within each
group, and positive role weights $\pi_a$ summing to one, set
$w_e=\sum_{a\in\mathcal J_e}\pi_a$ and
$\bar K_e=w_e^{-1}\sum_{a\in\mathcal J_e}\pi_a K_a$.  Completing the square gives
\begin{equation}
 \sum_a\pi_a\|K_a-\Delta_{\alpha(a)}\|_F^2
 =\sum_e\sum_{a\in\mathcal J_e}\pi_a\|K_a-\bar K_e\|_F^2
   +\sum_e w_e\|\bar K_e-\Delta_e\|_F^2.
 \label{eq:tied-radius-bias-decomposition}
\end{equation}
Apply \cref{prop:total-radius-allocation} to the unique matrices and group
centroids in the second term.  The first term is an irreducible tying bias in
this model.  For example, targets $+1,-1$ with equal weights and one shared
scalar update have loss $1+\Delta^2\ge1$, even though separate updates of
radius one fit each target exactly.  Merely adding separate per-role optima
would give an invalid claim of joint realizability.
For two \emph{distinct} targets both equal to $1$, a total radius one instead
gives equal optimal updates $1/\sqrt2$ and average loss
$(1-1/\sqrt2)^2$, rather than zero.

For different semidefinite geometries and distinct variables, any radii $t_e$
with $\sum_e t_e^2\le t^2$ yield jointly feasible scaled lifts from
\cref{eq:psd-radius-definitions}.  Their weighted upper-bound values add.
Their individual lower-bound values add under these \emph{individual} caps;
for only a total cap, a valid lower bound minimizes the sum of individual
lower bounds over $t_e\ge0$, $\sum_e t_e^2\le t^2$.
With shared variables, evaluate the same constructed $\Delta_e$ in every
associated role; separate per-role minima are only a relaxation unless
compatibility is proved.  

\paragraph{Interface with the nonlinear parameter model.}
The concatenated $d$ in \cref{eq:unique-parameter-radius} is the same parameter
vector used in \cref{eq:joint-linear-recursion}.  Choosing radii with total at
most $t$ makes the derivative envelopes at that radius applicable, subject to
their separate state-region and route hypotheses.  For random updates with this
bound almost surely, $\E\|d\|^4\le t^4$; covariance or independence of calls
is not needed for this inequality.  Repetition still affects $G_j,c_j,b_j$,
not the count of distinct parameter entries.  Applying the radius bound to terminal risk also requires the stated route, remainder, cross-covariance, and exit controls.
For factors initialized nontrivially, the actual increment is
$BA-B_0A_0$; factor width $r$ does not by itself make this increment rank at
most $r$.  The rank and radius constraints must hold for the increment, as
emphasized by \cref{prop:factor-tangent-image}.

\subsection{Support replay recomputes the selected gain at its own state}
\label{sec:support-replay}

Fix a finite prefix (or a distribution of prefixes with finite maximum length),
not a recursively generated token sequence.  All states include the full
prefix and every value needed by attention, residual paths, and subsequent
events.  For events $j=0,\ldots,H-1$, let $\mathcal A_j$ be the finite set of
complete support patterns at that event; a pattern may specify one expert per
token.  A deterministic rule $\rho_j(s)$, including its tie convention,
selects the natural support.  Route-free events have a single possible pattern.
Let $F_{j,a}(s)$ be the complete branch extension at prescribed support $a$,
including the \emph{current-state} gain, expert inputs, and every shared path.
In a top-one residual event it has the form
\begin{equation}
 F_{j,a}(s)=B_j(s)+p_{j,a}(s)v_{j,a}(s),\qquad
 p_j(s)=\operatorname{softmax}(z_j(s)).
 \label{eq:replay-topone-branch}
\end{equation}
Here $B_j$ includes the common residual path.  Forcing $a$ uses $p_{j,a}(s)$,
not $\max_e p_{j,e}(s)$ and not a gain recorded at another state.
The branch is defined also where $a$ would not win.

On one common probability space, fix a prescribed support sequence $a_j^\circ$ (for example, recorded from a declared reference pass), an input, model
parameters, and any common external random tape.  For each realization these
objects are held the same in the following two runs:
\begin{equation}
 n_0=r_0=s_0,\qquad
 n_{j+1}=F_{j,\rho_j(n_j)}(n_j),\qquad
 r_{j+1}=F_{j,a_j^\circ}(r_j).
 \label{eq:natural-support-replay}
\end{equation}
Thus $n$ is natural execution and $r$ is \emph{support replay}.
The prescribed sequence is not revised in response to either run's terminal
loss.  It may depend on the input and correlated parameters; no independence
is assumed.  Replay is a defined counterfactual computation.  Model parameters are the same
within this pair; comparison of two models uses a separate pair for each.

\begin{theorem}[Natural execution versus support replay]
\label{thm:support-replay-bridge}
Let $\tau$ be the first $j$ with $\rho_j(r_j)\ne a_j^\circ$, setting
$\tau=H$ if there is none, and put $\Omega=\{\tau=H\}$ and
$p_{\rm diff}=\Prb(\Omega^c)$.  Then $\tau$ is also the first index at which
$\rho_j(n_j)\ne a_j^\circ$.  In particular, $n_H=r_H$ on $\Omega$.
For a common target $Y$ and the same measurable loss
$0\le\ell(s,Y)\le M<\infty$ in both runs,
\begin{equation}
 \left|R^{\rm nat}-R^{\rm rep}\right|
 :=\left|\E\ell(n_H,Y)-\E\ell(r_H,Y)\right|
 \le M p_{\rm diff}.
 \label{eq:replay-bounded-risk}
\end{equation}
This bound needs neither smoothness nor independence, and its factor $M$ is
sharp over the stated class.

For a functional refinement, suppose the prescribed branch is $L_j$-Lipschitz
between the required pairs $n_j,r_j$, and define
\begin{equation}
 \begin{split}
 f_j&=\|F_{j,\rho_j(n_j)}(n_j)-F_{j,a_j^\circ}(n_j)\|,\\
 D_H&=\sum_{j=0}^{H-1}\left(\prod_{k=j+1}^{H-1}L_k\right)f_j.
 \end{split}
 \label{eq:replay-functional-envelope}
\end{equation}
Then $\|n_H-r_H\|\le D_H$.  If also the terminal loss is $L_\ell$-Lipschitz
in this norm, then
\begin{equation}
 |R^{\rm nat}-R^{\rm rep}|
 \le\E\min\{M\mathbf1_{\Omega^c},L_\ell D_H\}.
 \label{eq:replay-functional-risk}
\end{equation}
Path-dependent finite $L_j$ are allowed, but their products remain inside
expectation.  The Lipschitz condition is on branch extensions over the compared
states, not an assumption that the discontinuous natural router is Lipschitz.
\end{theorem}
\begin{proof}
Before the first replay mismatch, induction gives $n_j=r_j$: the two rules
choose $a_j^\circ$ at the same state and apply the identical branch, including
its current-state gain.  If the first mismatch occurs, the equal pre-event
states make it a mismatch in the natural run as well.  If none occurs, equality
persists to the endpoint.  Losses therefore differ only on $\Omega^c$, and
their pointwise absolute difference is at most $M\mathbf1_{\Omega^c}$.
Taking expectations proves \cref{eq:replay-bounded-risk}.
For sharpness, use one event and a Bernoulli input of probability $p$: the
natural rule selects branch $1$ on that input and branch $0$ otherwise, replay
always selects $0$, and branch outputs/losses are respectively $M$ and zero.
The risk difference is $Mp$.
For the functional statement, insert $F_{j,a_j^\circ}(n_j)$ to get
$\|n_{j+1}-r_{j+1}\|\le f_j+L_j\|n_j-r_j\|$.
Unroll from equal initial states.  Both this norm bound times $L_\ell$ and the
bounded-loss indicator bound apply pointwise; take their minimum and expectation.
\end{proof}

\paragraph{Expert-output and gain contributions.}
At a common state in \cref{eq:replay-topone-branch}, write
$e=\rho_j(s)$, $a=a_j^\circ$, suppress $j,s$, and insert $p_ev_a$ to obtain
\begin{equation}
 f_j\le p_e\|v_e-v_a\|+|p_e-p_a|\|v_a\|.
 \label{eq:replay-gain-support-cost}
\end{equation}
The reverse insertion gives $p_a\|v_e-v_a\|+|p_e-p_a|\|v_e\|$; either is valid.
For rowwise routing and Frobenius state norm, combine per-row upper bounds by
the square root of their squared sum.  Shared paths cancel at this common state.
Equal expert outputs alone need not imply equal branch outputs: with
$v_1=v_2=1$, probabilities $(0.8,0.2)$ and prescribed expert $2$, the discrepancy
is $0.6$ despite zero expert-output diameter.  The selected masses differ.
Thus the equal-mass diameter lemma cannot be used here without a null atom or
an explicit gain term.  By contrast, if the prescribed expert equals the
natural expert at the \emph{same} state, both support and recomputed gain agree,
and $f_j=0$.

\paragraph{A first-disagreement margin bound on the correct states.}
For rowwise top-one routing, sum only over routed events. Let reference logits
$z_j^\circ$ select $a_j^\circ$ with winner gap $\gamma_j\ge0$ (minimum over
 tokens), and put $U_j=\|z_j(r_j)-z_j^\circ\|_\infty$ over all tokens and
 coordinates. With the same tie rule,
\begin{equation}
 \begin{split}
 p_{\rm diff}
 &\le\min\{1,\sum_j\Prb(\tau\ge j,\ \gamma_j\le2U_j)\}\\
 &\le\min\left\{1,\sum_j\inf_{v>0}
   \left[\Prb(\gamma_j\le2v)+\frac{\E U_j^\nu}{v^\nu}\right]\right\},
 \qquad \nu>0.
 \end{split}
 \label{eq:replay-first-margin}
\end{equation}
For the first line, a first mismatch implies the displayed margin event by
\cref{lem:topk-margin-stability}; sum over its disjoint possible locations.
For the second line, drop $\{\tau\ge j\}$, split each margin event at $U_j\le v$,
and apply Markov to $U_j>v$.  No independence across events or between margin
and perturbation is needed.  If the marginal CDF satisfies a power bound,
\cref{thm:marginal-margin-routing} gives its sharper constant instead.
Reference inputs with ties remain in the bound; they are not assumed to have
zero mass.  The moments and margins here are those of \emph{support-replay}
states.
For the down-only specialization, \cref{eq:explicit-branch-tube} is an
alternative sufficient condition for $p_{\rm diff}=0$ under its hypotheses.

\paragraph{Transferring replay-risk comparisons to natural execution.}
For two models evaluated on the same target, loss and input population, let
$b_m$ be a valid right-hand side of \cref{eq:replay-bounded-risk} or
\cref{eq:replay-functional-risk}.  Then
\begin{equation}
 R_{\rm sep}^{\rm nat}-R_{\rm reuse}^{\rm nat}
 \ge R_{\rm sep}^{\rm rep}-R_{\rm reuse}^{\rm rep}
          -b_{\rm sep}-b_{\rm reuse}.
 \label{eq:natural-replay-improvement}
\end{equation}
Consequently a replay-risk gap exceeding the two penalties certifies the
corresponding natural-risk ordering.  With only replay-risk intervals, use
its separate-model lower endpoint minus its reuse-model upper endpoint.
Two upper bounds alone do not yield this conclusion.  Each model may have its
own prescribed sequence and penalty.  The fixed-$G$ or fixed-selection assumptions of
\cref{thm:terminal-geometry-frontier,thm:endogenous-conditional-risk} remain
necessary for their exact quadratics.  

\paragraph{Bounded squared loss for normalized fixed-prefix outputs.}
If the terminal state is the final normalized logits, the frozen head and
normalizer in \cref{sec:explicit-block-constants} give a common bound
$\|s\|\le B$ on natural and replay outputs for a bounded prefix length.
For a common target with $\|Y\|\le B_Y$, the \emph{unclipped} square loss
$\ell(s,Y)=\|s-Y\|^2$ obeys $M=(B+B_Y)^2$ and
$L_\ell=2(B+B_Y)$ on this output set.  Indeed,
$|\|s-Y\|^2-\|s'-Y\|^2|
\le(\|s-Y\|+\|s'-Y\|)\|s-s'\|$.
Thus the bridge can apply to actual bounded-output square risk; it is not
limited to an artificially clipped loss.  This conclusion requires the stated
output/target bounds.

\subsection{Frozen-gain replay pays an additional functional term}
\label{sec:frozen-gain-replay}

The preparation code's \texttt{force}/\texttt{gain\_reference} mechanism fixes
both the chosen index and the supplied gain at a forced position.  The implementation uses frozen-gain replay, defined below.  For a top-one event,
fix a reference gain $q_j\in[0,1]$ and define
\[
 \overline F_{j,a_j^\circ}(s)=B_j(s)+q_jv_{j,a_j^\circ}(s),\qquad
 h_0=s_0,\quad h_{j+1}=\overline F_{j,a_j^\circ}(h_j).
\]
For rowwise routing, $q_j$ is the vector of prescribed gains.  For route-free
events let $\overline F_j=F_j$.

\begin{proposition}[A gain-aware bridge for complete-gate replay]
\label{prop:frozen-gain-replay}
Suppose the current-gain prescribed branch $F_{j,a_j^\circ}$ is $L_j$-Lipschitz
between $r_j,h_j$, and define
\begin{equation}
 c_j=\|F_{j,a_j^\circ}(h_j)-\overline F_{j,a_j^\circ}(h_j)\|,
 \qquad V_H=\sum_j\left(\prod_{k=j+1}^{H-1}L_k\right)c_j.
 \label{eq:frozen-gain-envelope}
\end{equation}
Then $\|r_H-h_H\|\le V_H$.  In the single-row form,
$c_j=|p_{j,a_j^\circ}(h_j)-q_j|\|v_{j,a_j^\circ}(h_j)\|$.
For a common loss in $[0,M]$ that is also $L_\ell$-Lipschitz,
\begin{equation}
 |R^{\rm nat}-R^{\rm frozen}|
 \le M p_{\rm diff}
      +\E\left[\mathbf1_\Omega\min\{M,L_\ell V_H\}\right].
 \label{eq:frozen-gain-risk}
\end{equation}
Here $\Omega$ and $p_{\rm diff}$ are the \emph{support-replay} event and
probability in \cref{thm:support-replay-bridge}, not agreement measured on $h$.
No independent-gain or independent-state assumption is used.  If $L_j$ are
random, all products stay inside the expectation.
\end{proposition}
\begin{proof}
Insert $F_{j,a_j^\circ}(h_j)$ to obtain
$\|r_{j+1}-h_{j+1}\|\le L_j\|r_j-h_j\|+c_j$ and unroll.
The formula for $c_j$ follows by canceling the same-state shared path and
expert output.  On $\Omega$, $n_H=r_H$, so the natural/frozen loss difference
is bounded by $\min\{M,L_\ell V_H\}$.  On its complement it is at most $M$.
Splitting the expectation proves the claim.  In particular, the right-hand
side never exceeds $M$; it does not factor a gain error from a dependent event.
\end{proof}

\paragraph{Counterexample to an index-only bound for frozen gain.}
Let $\sigma(x)=(1+e^{-x})^{-1}$ and start from $s_0=1$.
At the first residual event the winning support is always $1$, with constant
gain $g_0=\sigma(1)$, and $F_0(s;D)=s+g_0D$ for a constant expert feature.
The reference has $D=0$; the compared model changes only this output weight to
$D=1/g_0$, reaching $s_1=2$ instead of the reference state $1$.
At the second event use logits $(s,0)$, expert-$1$ output $1$, and residual map
$s\mapsto s+\sigma(s)$.  Every reference, natural, support-replay and frozen-gain
run selects expert $1$ throughout.  Nevertheless the natural/support-replay
endpoint is $2+\sigma(2)$, while reference-gain replay ends at $2+\sigma(1)$.
The common bounded Lipschitz loss $\ell(s)=\min\{1,\max\{0,s/4\}\}$ gives
risk difference $(\sigma(2)-\sigma(1))/4>0$ although $p_{\rm diff}=0$.
This two-event construction has the selected-softmax-gain semantics of
\cref{eq:replay-topone-branch}.
It disproves the index-only extension.

\paragraph{Conditions for applying the replay bounds.}
Without bounded loss, a mismatch probability $p$ does not bound squared risk:
one event may output $p^{-1}$ only on the mismatching event, giving squared
risk $p^{-1}$.  Conversely, bounded but discontinuous classification loss need
not be Lipschitz: outputs $-\epsilon$ and $+\epsilon$ can have loss difference
one for every $\epsilon>0$.  Thus the gain-to-loss refinement requires its
Lipschitz premise or a separate terminal decision-margin argument; boundedness
alone still gives the support-replay result.  The normalized fixed-prefix
branch constants in \cref{prop:explicit-block-envelopes} can supply the stated
branch Lipschitz bounds at a specified total update radius, without freezing
attention activations or selected gains.  Extra adaptive decisions must be
included in the event support/state before invoking these results.

\clearpage
\section{Assumptions and Theorem Dependencies}
\label{app:operational-assumptions}
This table summarizes the dependencies of the existing results. Each alternative retains its own assumptions.

\begingroup
\small
\setlength{\tabcolsep}{3pt}
\renewcommand{\arraystretch}{1.05}
\begin{longtable}{@{}>{\raggedright\arraybackslash}p{0.205\dimexpr\textwidth-18pt\relax}>{\raggedright\arraybackslash}p{0.305\dimexpr\textwidth-18pt\relax}>{\raggedright\arraybackslash}p{0.265\dimexpr\textwidth-18pt\relax}>{\raggedright\arraybackslash}p{0.225\dimexpr\textwidth-18pt\relax}@{}}
\toprule
Result & Key assumptions & Application conditions & Applicable extension \\
\midrule
\endfirsthead
\toprule
Result & Key assumptions & Application conditions & Applicable extension \\
\midrule
\endhead
\bottomrule
\endfoot
\multicolumn{4}{@{}l@{}}{\refstepcounter{subsection}\label{sec:assumption-guide-local}\textbf{\thesubsection\quad Local estimation and data-dependent choices}}\\[3pt]
\Cref{lem:call-replacement} &
Declared roles, interfaces and complete states; probe objects fixed independently of fitting, or covered jointly. &
Same-data learned environments need a joint uniform class or independent split. &
\Cref{app:routed-roles,app:statistical-proofs}. \\[3pt]
\Cref{thm:exact-frontier,thm:hetero-frontier} &
Fixed orthogonal designs, centered independent noise with stated moments, independent tests and fixed pooling. &
General designs and covariance need their own risk geometry. &
\Cref{thm:general-smoother-risk,thm:weighted-projection-frontier}. \\[3pt]
\Cref{prop:hilbert-bias} &
Common squared Hilbert geometry; unrestricted or appropriate realizable class for pure variance. &
Restricted classes retain distance corrections. &
\Cref{eq:restricted-hilbert-bias}. \\[3pt]
\Cref{thm:partition-frontier,thm:linear-sharing-subspace,thm:cross-depth-frontier,cor:multi-axis-additive} &
Fixed disjoint groups or linear tying spaces; balanced cells and stated unbiased covariance for dimension penalties. &
Adaptive grouping needs selection control; embeddings need realization conditions; greedy optimization needs separate analysis. &
\Cref{cor:nested-sharing-frontier}; independent selection or a uniform family bound. \\[3pt]
\Cref{eq:app-merge-statistic-law,prop:observed-dispersion}; \cref{app:classical-plugin} &
Gaussian noise and positive variance for sampling laws. The plug-in also uses known $q>0$, a fixed mean subspace and $k\ge4$. &
Degeneracies use deterministic identities; searched merges need multiplicity control.  &
Separate expectation identities and selection guarantees. \\[3pt]
\Cref{thm:finite-class} &
Bounded losses, fixed finite class and weights, independent samples, joint confidence budget. &
Use independent records and retain fitting gaps and weight mismatch. &
\Cref{eq:app-trace-concentration,prop:approximate-weighted-erm}. \\[3pt]
\Cref{prop:neff-allocation,cor:architecture-dependent-sampling} &
Declared allocation objective and continuous feasible sample counts. &
Enforce design floors and integer feasibility. &
Constrained allocation; integer allocation is a separate problem. \\[3pt]
\Cref{prop:validation-lambda,thm:finite-architecture-selection,prop:paired-real-improvement} &
Fixed fitted candidates and weights; independent validation/audit units; bounded loss or stated tail conditions. &
Outcome-dependent candidates or unknown tail constants need further analysis. &
Bounded and Gaussian quadratic-loss guarantees remain separate. \\[3pt]
\multicolumn{4}{@{}l@{}}{\refstepcounter{subsection}\label{sec:assumption-guide-composition}\textbf{\thesubsection\quad Composition, routing, and intervention}}\\[3pt]
\Cref{eq:app-path-bound}; \cref{ass:role-aggregation} &
Common coupled program and state metric; regional Lipschitz envelopes, common deterministic role weights and assembly residual. &
The terminal certificate orders bounds, not risks; random products stay inside expectations. &
\Cref{prop:additive-occupancy-transfer}; optimization and exit costs remain. \\[3pt]
\Cref{thm:terminal-geometry-frontier,prop:local-terminal-reversal} &
Unbiased finite-covariance estimates, fixed sharing and one common fixed linear terminal map. &
Retain cross-covariances; route-dependent propagation needs conditional moments or remainder control. &
\Cref{prop:gramian-amplification,prop:shared-error-coherence,prop:terminal-remainder-interval}. \\[3pt]
\Cref{app:uniform-terminal-inheritance} &
Orthogonal pooling, positive contrast rank, covariance $qI$, $q>0$; $QG^\top GQ=cQ$, $c>0$. &
An all-target condition, not a necessary condition for one target or a guarantee for adaptive sharing. &
The target-specific fixed-map comparison remains distinct. \\[3pt]
\Cref{prop:terminal-remainder-interval,thm:joint-taylor-remainder,prop:explicit-block-envelopes} &
Same-law remainder control; joint derivatives, region/route control and stated higher moments for Taylor bounds. &
Use joint derivative and remainder control; the concrete specialization fixes the prefix and perturbs down matrices only. &
Retain escape costs and justified remainder moments. \\[3pt]
\Cref{thm:endogenous-conditional-risk,prop:gaussian-threshold-moments} &
Condition on enough information to fix propagation coefficients and offsets; recompute conditional moments. &
Changing selection partitions requires candidate-specific laws. &
The half-space Gaussian law is a specialization. \\[3pt]
\Cref{lem:gate-discrepancy-revised,thm:functional-transport-gate,cor:topk-matching} &
Common input and expert outputs; equal nonnegative mass; uniform $1/k$ weights for matching. &
Dispatch changes inputs; other drop or normalization rules need their own maps. &
\Cref{thm:full-soft-dispatch}: normalized dispatch/combine maps and expert sensitivity. \\[3pt]
\Cref{prop:hierarchical-route-decomposition,thm:sequential-route-discrepancy} &
Stated hierarchy or preselected sequence; compatible spaces and reference-suffix sensitivity. &
Use separate routed events or a complete adaptive-policy comparison. &
Separate routed events or a full-policy comparison. \\[3pt]
\Cref{prop:soft-router-closed-loop,lem:topk-margin-stability,thm:no-route-cascade} &
Gate sensitivity for soft routing; aligned logits, strict margins and error envelopes for hard support. &
Sufficient, not necessary; equal support does not fix nonuniform gains. &
\Cref{thm:margin-mediated-routing,thm:marginal-margin-routing,prop:marginal-margin-sharpness} distinguish conditional and marginal control. \\[3pt]
\Cref{thm:support-replay-bridge,prop:frozen-gain-replay} &
Same prefix, parameters, target and randomness; bounded loss. Gain refinements also need branch/loss sensitivity. &
Recorded gains can incur extra cost despite identical indices. &
\Cref{prop:frozen-gain-replay}; retain dependent products and event indicators. \\[3pt]
\Cref{prop:route-label-separation,prop:aliasing,prop:average-aliasing} &
Declared intervention and extractor; distinguish worst-case and average squared-risk obstructions. &
Natural labels do not identify operations; off-policy defects need transfer. &
\Cref{eq:app-causal-rollout,eq:app-wasserstein-recursion}: coupled actions, stability and state sufficiency. \\[3pt]
\Cref{thm:counterfactual-reuse-bias,prop:no-overlap-nonidentifiability} &
Randomization, consistency, square-integrable Hilbert outcomes and unrestricted Bayes predictors; a nontrivial unrestricted effect class for non-identification. &
Local bias matches only with matched inputs, classes, losses and laws; unsupported effects are not identified. &
Conditional-mean estimation and coverage require separate analysis. \\[3pt]
\Cref{thm:ipw-route-audit,prop:optimal-intervention-allocation} &
Known positive forcing propensities, independent bounded records and fixed targets; known second moments for allocation. &
Router scores are not forcing probabilities; least-covered-action precision is not every aggregate rate. &
Independent pilot and required propensity floors; cross-fitting needs foldwise accounting. \\[3pt]
\multicolumn{4}{@{}l@{}}{\refstepcounter{subsection}\label{sec:assumption-guide-evidence}\textbf{\thesubsection\quad Interpreting diagnostics and applying a result}}\\[3pt]
\Cref{prop:rank-r-functional,thm:feature-weighted-low-rank,prop:free-core-two-role,prop:factor-tangent-image} &
Stated functional or matrix class, core, initialization, feature geometry and insertion point. &
Functional rank is not LoRA rank; approximation does not establish learned or terminal risk. &
\Cref{thm:rank-radius-frontier,prop:total-radius-allocation}: actual updates, each distinct matrix counted once. \\[3pt]
\Cref{eq:app-resource-prg}; application of certificates &
Declared evaluation population and budgets; nonnegative cost conversions; justified uniform or input-integrated bounds. &
Resource scores express preferences; uniform envelopes require justification beyond sampled maxima. &
Independent pilot/fitting/audit separation or justified joint/foldwise analysis. \\[3pt]
\end{longtable}
\endgroup

\end{document}